\documentclass[11pt]{article}
\usepackage[letterpaper,margin=1in,headheight=14pt,headsep=18pt,footskip=25pt]{geometry}
\usepackage[round,authoryear]{natbib}
\setcitestyle{authoryear,round,citesep={;},aysep={,},yysep={;}}
\usepackage{amsmath,amsfonts,bm}

\def\eqref#1{equation~\ref{#1}}

\def\plaineqref#1{\ref{#1}}

\def\1{\bm{1}}

\DeclareMathAlphabet{\mathsfit}{\encodingdefault}{\sfdefault}{m}{sl}
\SetMathAlphabet{\mathsfit}{bold}{\encodingdefault}{\sfdefault}{bx}{n}

\newcommand{\E}{\mathbb{E}}

\newcommand{\R}{\mathbb{R}}

\newcommand{\TV}{ d_{\mathrm{TV}}}

\DeclareMathOperator*{\argmin}{arg\,min}

\makeatletter
\renewcommand{\eqref}[1]{\textup{\tagform@{\ref{#1}}}}
\makeatother

\usepackage{hyperref}
\usepackage{url}
\usepackage{amsthm}
\usepackage{thm-restate}
\makeatletter
\@ifundefined{newcounteralias}{}{%
  \renewcommand\thmt@autorefsetup{%
    \@xa\def\csname\thmt@envname autorefname\@xa\endcsname
      \@xa{\thmt@thmname}%
  }%
}
\makeatother
\usepackage{amsmath}
\usepackage{amssymb}
\usepackage{amsthm}
\usepackage{mathtools}
\usepackage{bm}
\usepackage{graphicx}
\usepackage{subcaption}
\usepackage{booktabs}
\usepackage{enumitem}
\usepackage[dvipsnames]{xcolor}
\usepackage{pifont}
\usepackage{optidef}          %
\usepackage{tikz}
\usepackage{etoolbox}         %
\usepackage[capitalize,nameinlink]{cleveref}   %
\usepackage{algorithm}
\usepackage{algorithmic}
\usepackage[acronym]{glossaries}

\usepackage{dsfont}

\DeclareMathOperator{\dist}{dist}

\DeclareMathOperator{\spec}{spec}

\DeclareMathOperator{\sn}{sn}
\DeclareMathOperator{\cn}{cn}
\DeclareMathOperator{\dn}{dn}

\DeclareMathOperator{\scc}{sc}
\DeclareMathOperator{\jsc}{sc}

\DeclareMathOperator{\jcs}{cs}
\newcommand{\sech}{\operatorname{sech}}
\newcommand{\C}{\mathbb C}
\newcommand{\Z}{\mathbb Z}
\newcommand{\dif}{\mathop{}\!\mathrm{d}}
\providecommand{\am}{\operatorname{am}}

\renewcommand{\Im}{\operatorname{Im}}

\makeglossaries

\newcommand{\myacronym}[3]{%
    \newacronym{#1}{#2}{#3}%
    \expandafter\newcommand\csname #1\endcsname{\gls{#1}}%
}

\myacronym{cpu}{CPU}{Central Processing Unit}
\myacronym{ram}{RAM}{Random Access Memory}

\newcommand{\eqd}{\stackrel{d}{=}}

\newcommand{\circnum}[1]{\tikz[baseline=(char.base)]{%
    \node[draw, circle, inner sep=1pt] (char) {#1};}}
\robustify{\circnum}

\theoremstyle{plain}
\ifcsundef{theorem}{\newtheorem{theorem}{Theorem}[section]}{}
\ifcsundef{lemma}{\newtheorem{lemma}[theorem]{Lemma}}{}
\ifcsundef{proposition}{\newtheorem{proposition}[theorem]{Proposition}}{}
\ifcsundef{corollary}{\newtheorem{corollary}[theorem]{Corollary}}{}
\ifcsundef{claim}{}{}
\ifcsundef{property}{}{}
\ifcsundef{conjecture}{}{}

\theoremstyle{definition}
\ifcsundef{definition}{\newtheorem{definition}[theorem]{Definition}}{}
\ifcsundef{assumption}{\newtheorem{assumption}[theorem]{Assumption}}{}
\ifcsundef{problem}{\newtheorem{problem}[theorem]{Problem}}{}
\ifcsundef{example}{}{}

\theoremstyle{remark}
\ifcsundef{remark}{\newtheorem*{remark}{Remark}}{}

\crefname{theorem}{Theorem}{Theorems}
\crefname{lemma}{Lemma}{Lemmas}
\crefname{proposition}{Proposition}{Propositions}
\crefname{corollary}{Corollary}{Corollaries}
\crefname{claim}{Claim}{Claims}
\crefname{property}{Property}{Properties}
\crefname{conjecture}{Conjecture}{Conjectures}
\crefname{definition}{Definition}{Definitions}
\crefname{assumption}{Assumption}{Assumptions}
\crefname{problem}{Problem}{Problem}
\crefname{example}{Example}{Examples}
\crefname{equation}{Eq.}{Eqs.}
\crefname{figure}{Figure}{Figures}
\crefname{table}{Table}{Tables}
\crefname{section}{Section}{Sections}
\crefname{appendix}{Appendix}{Appendices}

\setlist[enumerate]{topsep=4pt, itemsep=1pt}
\setlist[itemize]{topsep=4pt, itemsep=1pt}

\usepackage{subcaption}
\usepackage{multirow}

\usepackage[lavender]{colepaper}
\usepackage{array}
\usepackage{etoc} %

\hypersetup{colorlinks=true,linkcolor=ColeResult,citecolor=ColeResult,
  urlcolor=ColeDefinition,pdfborder={0 0 0},
  pdftitle={Elliptic Sampling: Optimal interpolant schedules for Gaussians and Beyond},
  pdfauthor={Cole Becker, Taylan Kargin, Kimia Hassibi, Asuman Ozdaglar, Pablo Parrilo}}
\usepackage{fancyhdr}
\fancypagestyle{plain}{\fancyhf{}\fancyfoot[C]{\sffamily\small\color{ColeRemark}\thepage}}

\titleformat{\subsection}{\normalsize\sffamily\bfseries\color{ColeInk}}{\thesubsection}{.6em}{}
\titleformat{\subsubsection}{\normalsize\sffamily\bfseries\color{ColeInk}}{\thesubsubsection}{.6em}{}
\titleformat{\paragraph}[runin]{\normalsize\bfseries\color{ColeInk}}{\theparagraph}{.6em}{}
\titlespacing*{\section}{0pt}{16pt plus 3pt minus 2pt}{9pt}
\titlespacing*{\subsection}{0pt}{13pt plus 2pt minus 2pt}{7pt}
\titlespacing*{\subsubsection}{0pt}{11pt plus 2pt minus 2pt}{6pt}
\titlespacing*{\paragraph}{0pt}{8pt plus 2pt minus 2pt}{.6em}
\usepackage{needspace}
\AddToHook{cmd/section/before}{\Needspace{6\baselineskip}}
\AddToHook{cmd/subsection/before}{\Needspace{4\baselineskip}}
\makeatletter
\renewcommand{\maketitle}{%
  \thispagestyle{plain}
  \begingroup
    \renewcommand{\thefootnote}{\fnsymbol{footnote}}%
    \let\@footnotemark\H@@footnotemark
    \let\@footnotetext\H@@footnotetext
    \begingroup\centering
      {\fontsize{23}{27}\selectfont\bfseries\color{ColeInk}\@title\par}
      \vspace{12pt}
      {\normalsize\@author\par}
      \vspace{15pt}
      {\color{ColeResult!35}\hrule height .6pt}
    \endgroup
    \@thanks
  \endgroup
  \setcounter{footnote}{0}%
  \global\let\@thanks\@empty
  \vspace{13pt}}
\renewenvironment{abstract}{%
  \begin{list}{}{\setlength{\leftmargin}{13pt}\setlength{\rightmargin}{13pt}}
  \item[]\small
  {\sffamily\bfseries\color{ColeInk}Abstract}\par\vspace{4pt}
}{\end{list}\vspace{6pt}}
\makeatother

\title{Elliptic Sampling: Optimal Interpolant Schedules for Gaussians and Beyond}

\author{%
  Cole Becker\thanks{Corresponding author:
  \href{mailto:colbeck@mit.edu}{\texttt{colbeck@mit.edu}}.},
  Taylan Kargin\thanks{Equal contribution.}, Kimia Hassibi\footnotemark[2],
  Asuman Ozdaglar, Pablo A. Parrilo\\[6pt]
  {\small Laboratory for Information and Decision Systems (LIDS)\\
  Massachusetts Institute of Technology\\
  Cambridge, MA 02139}}
\date{}

\begin{document}

\maketitle

\begin{abstract}
Generative modeling based on stochastic interpolants requires specifying two key choices: an \emph{interpolant schedule} bridging the initial and target distributions, and an efficient \emph{discretization} of the resulting probability-flow ODE. In this work, we introduce a new framework for joint schedule/discretization design based on the analyticity properties of the (complexified) ODE trajectories, which in turn ensure good numerical behavior. Our formulation can be solved exactly via conformal mapping, yielding a natural family of variance-preserving \emph{elliptic schedules}, parametrized by Jacobi elliptic functions. Pairing an elliptic schedule with a tailored discretization, we obtain \emph{Zoloflow}, a new sampling algorithm closely connected to Zolotarev’s optimal rational function approximants. Crucially, both the elliptic schedule and Zoloflow explicitly incorporate the condition number $\kappa$ of the target distribution and reduce to the classical trig schedule in the limit $\kappa \rightarrow 1$. For Gaussian targets, Zoloflow achieves \emph{minimax optimality among all oblivious score-span algorithms}, attaining total-variation error $\varepsilon$ using $\Theta(\log\kappa\,\log(\sqrt d/\varepsilon))$ (exact) score queries. For the log-concave case, the elliptic schedule allows sampling with 2-Wasserstein-error $\varepsilon$ by requiring merely $\widetilde{O}(\sqrt{d} \, \mathrm{polylog}(\kappa, d, 1/\varepsilon))$ score evaluations. Complementing these theoretical results, we empirically evaluate the elliptic schedule on general (non-log-concave) target distributions, including detailed comparisons with alternative schedules on both synthetic benchmarks and existing image generation datasets.
\end{abstract}

\etocdepthtag.toc{mainbody} %

\begin{figure}[ht!]
    \centering
    \includegraphics[width=1\linewidth]{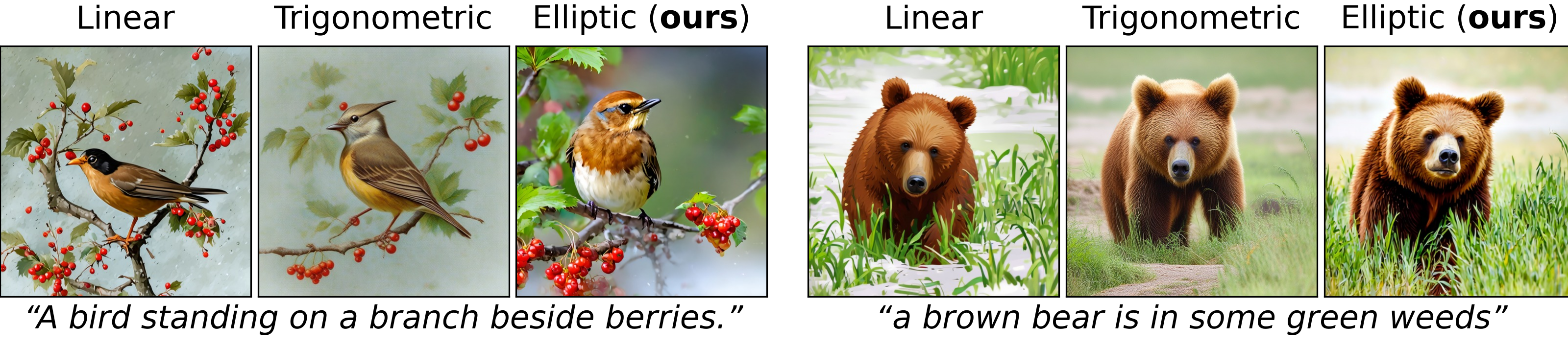}
    \vspace{-0.5cm}
    \caption{\textbf{Side-by-side comparison of images sampled via different schedules.} Images are generated using Stable Diffusion 3.5 Medium (CFG 4.5) with DPM-Solver++ with 8 neural network evaluations (8 oracle calls), starting from the same noise initializations across schedules. From left to right, the schedules are the target-rescaled linear, trigonometric, and elliptic (ours). Hyperparameters are listed in \cref{tab:sd3-fid50k-equal-budget}, and the elliptic schedule uses the heuristic 1D-sweep configuration. The text prompt is shown below the images.}
    \label{fig:front-page-schedules}
\end{figure}

\section{Introduction}
\label{sec:introduction}

Generative modeling seeks to produce novel random samples $X\sim \pi$ from a complicated target distribution $\pi$ on $\mathbb{R}^d$ using a representation learned from a finite training dataset $\mathcal{D}_{\mathrm{train}}=\{X^{(i)}\}$ of iid samples $X^{(i)}\sim \pi$. Many modern approaches learn an ordinary or stochastic differential equation (ODE or SDE) which progressively transports a simple initial reference distribution, typically Gaussian, to the target $\pi$. The stochastic interpolant (SI) framework \citep{albergo2025stochastic} unifies these flow-based (ODE) and diffusion-based (SDE) approaches and offers flexibility in shaping the transport dynamics through \emph{interpolant schedules}. 

Although different valid schedules all bridge the same initial and target distributions under exact continuous-time transport and are related through time and space reparameterizations, they can induce trajectories with substantially different regularity properties, including Lipschitz continuity and complex analyticity. These properties, in turn, govern how accurately and efficiently the resulting dynamics, most notably the probability-flow ODE (PF-ODE), can be solved numerically. Sampling performance, therefore, depends not only on the choice of interpolant schedule, but also on the \emph{discretization scheme} used to integrate the induced dynamics with a finite number of evaluations of the learned score/velocity field, typically represented by a deep neural network. Despite this intrinsic coupling, interpolant schedules and the discretization schemes are usually designed separately and chosen largely heuristically. This raises the following central question:

\begin{center}
    \textit{Given a fixed budget of evaluations of the learned score/velocity field, how should one choose both the interpolant schedule and the numerical procedure used to integrate the resulting PF-ODE?}
\end{center}

\subsection{Contributions}
We address this question by developing a principled theory of joint schedule and discretization design, which seeks the best achievable sampling accuracy over a class of target distributions under a finite budget of score/velocity-field evaluations. Our contributions are as follows:
\paragraph{1. Minimax-optimal Gaussian Sampling with Zoloflow.}
We formulate sampling from a Gaussian target with condition number at most $\kappa$ as a minimax problem over oblivious score-span algorithms with a fixed budget of exact score evaluations, and show that it reduces to a scalar rational approximation problem solved by Zolotarev's optimal approximants. The resulting algorithm attains $\TV$ error $\varepsilon$ with $\Theta(\log\kappa\,\log(\sqrt{d}/\varepsilon))$ score evaluations, which is minimax optimal. It admits a dynamical realization, \emph{Zoloflow}, which pairs a variance-preserving schedule parametrized by Jacobi elliptic functions with a modified Euler-type discretization; both depend on $\kappa$ and reduce to the trig schedule as $\kappa\to1$. 
\paragraph{2. Complex-Analytic Optimality of Elliptic Schedules.} 
We propose a complementary design criterion: the width of the complex strip around real time to which PF-ODE trajectories extend holomorphically, which governs the convergence of high-order numerical integrators. For Gaussian targets, maximizing this width is solved exactly by conformal mapping and recovers the elliptic schedule of the Zolotarev construction. For strongly log-concave targets, we establish explicit lower bounds on the width attained by the elliptic schedule.
\paragraph{3. Efficient Log-Concave Sampling with Elliptic Schedules.}
Combining these width bounds with collocation methods for analytic ODEs, we obtain a sampler for strongly log-concave targets that attains $W_2$ error $\varepsilon$ with $O(\sqrt{d}\,\log^{3.5}(\kappa d/\varepsilon))$ score evaluations. Because it relies on a generic high-order solver rather than the Gaussian-specific Zolotarev structure, this result isolates the benefit of the elliptic schedule itself.
\paragraph{4. Empirical Evaluations of Schedule and Discretization.}
On Gaussian targets, we empirically confirm the gains of Zoloflow over existing discretizations predicted by the minimax theory. Beyond Gaussians, we pair the elliptic schedule with state-of-the-art numerical solvers, and compare it against alternative schedules at matched evaluation budgets on synthetic and image-generation benchmarks. Across solvers, the elliptic schedule outperforms alternative schedules, including on non-log-concave targets beyond the scope of our theory.

Throughout the rest of the paper, we assume the score/velocity-field function evaluations are exact. All proofs and additional results are deferred to the Appendix.

\subsection{Related Work}
\paragraph{Interpolant and noise schedules.}
Stochastic interpolants \citep{albergo2025stochastic} separate the path, set by the schedule $(\alpha_t,\beta_t)$, from its numerical integration. Schedules are often chosen by hand, as with the trig schedule \citep{nichol2021improveddenoisingdiffusionprobabilistic} and the noise levels of EDM \citep{karras2022elucidatingdesignspacediffusionbased}; the continuous-time variational bound depends on the schedule only through its endpoint signal-to-noise ratios \citep{kingma2023variationaldiffusionmodels}, and \citet{noauthor_noise_2024} argues that the schedule entangles several design choices. Principled designs optimize the time grid for a given solver \citep{sabour_align_2024}, minimize an averaged Lipschitz constant of the drift \citep{chen2025lipschitz}, or adapt to high-dimensional mixtures \citep{aranguri_optimizing_2025}, and \citet{gao_Gaussian_2024} study the regularity of the resulting flows. We instead choose the schedule by the analyticity of the flow in complex time. The elliptic family reduces to the trig schedule at $\kappa=1$, contains the closed-form Gaussian schedule of Align Your Steps after rescaling (\cref{app:ays-clock}), and is asymptotically optimal for the worst-case drift-Lipschitz criterion whose Gaussian minimizer is the schedule of \citet{chen2025lipschitz} (\cref{app:lipschitz-schedules}).

\paragraph{Gaussian sampling and rational approximation.} 
Sampling from $\mathcal N(0,\Lambda^{-1})$ amounts to computing $\Lambda^{-1/2}z$, for which rational approximations, and Zolotarev's optimal ones in particular, converge far faster than polynomial ones when the spectrum is wide \citep{higham2008functions,HaleNicholas,nakatsukasa2016computing,gawlikZolotarevIterationsMatrix2018,beckermann2017singular,trefethenComputationZolotarevRational2025}. The closest work to ours is \citet{liu2026smoothedscorequeriescomplexity}, which samples Gaussians from \emph{smoothed-score} queries, our interpolant scores in variance-exploding coordinates (\cref{app:si-is-ve}). Combining these queries through a sinc-quadrature rational approximation of $x^{-1/2}$, Liu obtains $O\bigl((\log\kappa+\log(\sqrt d/\varepsilon))\log(\sqrt d/\varepsilon)\bigr)$ queries, improving the $\sqrt\kappa$ dependence of unsmoothed scores to $\log\kappa$; the accompanying lower bounds count bits received from a finite-precision oracle rather than exact queries. We improve on this in two ways. Zoloflow realizes the endpoint-normalized Zolotarev approximant, which removes the additive $\log^2(\sqrt d/\varepsilon)$ term (\cref{thm:zoloflow-Gaussian-bound}), and we prove a matching lower bound on the number of exact queries for oblivious score-span algorithms, a class that contains Liu's sampler and the standard diffusion solvers (\cref{thm:zoloflow-Gaussian-bound}). Moreover, unlike Liu's sampler, Zoloflow is a schedule plus a discretization of the probability-flow ODE, so the schedule carries over to non-Gaussian targets.

\paragraph{Fast samplers and discretizations.}
Many works accelerate sampling for a fixed schedule by improving the discretization: DDIM \citep{song2022denoisingdiffusionimplicitmodels}, the exponential integrators DEIS \citep{zhang_fast_2023} and DPM-Solver(++) \citep{lu2022dpmsolver,lu2022dpmsolverpp}, and the Heun sampler of EDM \citep{karras2022elucidatingdesignspacediffusionbased}; Align Your Steps \citep{sabour_align_2024} instead tunes the time grid for a fixed solver. We design the schedule and the discretization together. On Gaussian targets all of these solvers are oblivious score-span algorithms (\cref{lma:algs-are-ratfunc,app:additional-discretizations}), so our lower bound covers them, and on non-Gaussian targets the elliptic schedule also improves standard solvers such as DPM-Solver++ (\cref{app:logconcave-experiments}).

\paragraph{Complexity of diffusion and log-concave sampling.}
Diffusion samplers converge in polynomially many steps given accurate scores, for both the reverse SDE \citep{chen2023samplingeasylearningscore} and the probability-flow ODE \citep{chen2023probabilityflowodeprovably}, while classical samplers for strongly log-concave targets pay polynomially in the condition number \citep{Chewi26Book}. When an ODE solution is analytic, collocation reaches accuracy $\varepsilon$ in $\mathrm{polylog}(1/\varepsilon)$ steps \citep{leeAlgorithmicTheoryODEs2018}, which \citet{gatmiry_high-accuracy_2026} use to give high-accuracy guarantees for diffusion samplers. The step count depends on how far the solution extends into complex time; we choose the schedule to maximize this region, which gives a query complexity polylogarithmic in both $\kappa$ and $1/\varepsilon$ for log-concave targets (\cref{sec:logconcave}).

\paragraph{Schedules as approximation problems.}
Choosing a schedule so that a fixed budget of steps is as accurate as possible has close analogues. For gradient descent on a quadratic, stepsizes define a polynomial in the Hessian, and Chebyshev stepsizes \citep{agarwalAccelerationFractalLearning2021} attain the minimax $O(\sqrt\kappa\log(1/\varepsilon))$ rate; the silver stepsize schedule \citep{altschulerAccelerationStepsizeHedging2025,altschulerStepsizeHedgingAlternative2026} accelerates gradient descent beyond quadratics through the stepsizes alone. Score queries instead produce rational functions of the precision (\cref{lma:algs-are-ratfunc}), so Zolotarev's approximant replaces the Chebyshev polynomial and $\sqrt\kappa$ becomes $\log\kappa$. Similarly, double-exponential and sinc quadrature \citep{takahasi_double_1973,mori_discovery_2005,stenger_summary_2000}, like the sampler of \citet{liu2026smoothedscorequeriescomplexity}, change variables so that the integrand is analytic in a wide strip and then use a uniform grid; the elliptic schedule is the time change that maximizes this strip, and Zoloflow queries on a uniform grid in the new time.

\section{Preliminaries} 
\subsection{Stochastic Interpolants}
\label{sec:stochastic-interpolants}
 \emph{Stochastic interpolants} (SI) provide a general framework for constructing a continuous path between a simple reference distribution and a target \citep{albergo2025stochastic}, and are defined as the stochastic process
\begin{equation}
    x_t = \alpha_t z + \beta_t x_1,
    \qquad t \in [0,1], 
    \label{eq:stochastic-interpolant}
\end{equation}
where $z \sim \mathcal{N}(0,I_d)$ is the reference distribution, and $x_1\sim \pi$ is the target, and $(\alpha_t, \beta_t)$ is the \emph{interpolant schedule}. Stochastic interpolants induce a
deterministic probability-flow ODE with drift (velocity field) given by
\begin{equation}
\label{eq:si-velocity}
    b_t(x)\triangleq \mathbb{E}\!\left[\dot{x}_t \mid x_t=x\right]
\end{equation}
so that, under standard regularity conditions, the solution of
\begin{equation}
    \dot{X}_t = b_t(X_t),
    \qquad
    X_0 \sim \mathcal{N}(0,I_d),
    \label{eq:probability-flow-ode}
\end{equation}
has marginal law $X_t \sim p_t$ where $p_t = \mathcal{L}(x_t)$ is the law of the interpolant at time $t$. For the linear interpolant~\eqref{eq:stochastic-interpolant}, the velocity
field can be expressed using the score function $s_t(x)\triangleq\nabla\log p_t(x)$ or the conditional mean of the target sample. Indeed, using Tweedie's formula $\beta_t \mathbb{E}[x_1\mid x_t=x] = x+\alpha_t^2 s_t(x)$,
\begin{equation}
    b_t(x)
    =
    \frac{\dot{\alpha}_t}{\alpha_t}x
    +
    \beta_t
    \left(
        \frac{\dot{\beta}_t}{\beta_t}
        -
        \frac{\dot{\alpha}_t}{\alpha_t}
    \right)
    \mathbb{E}[x_1 \mid x_t=x] 
    =
    \frac{\dot{\beta}_t}{\beta_t}x
    +
    \alpha_t^2
    \left(
        \frac{\dot{\beta}_t}{\beta_t}
        -
        \frac{\dot{\alpha}_t}{\alpha_t}
    \right)
    s_t(x).
    \label{eq:si-denoiser-drift}
\end{equation}
\begin{definition}[VP interpolant schedules]
\label{def:schedule}
An interpolant schedule $(\alpha,\beta)$ is \emph{variance preserving} (VP) if $\alpha_t^2+\beta_t^2=1$ on $[0,1]$, with $\alpha$ strictly decreasing from $1$ to $0$ and $\beta$ strictly increasing from $0$ to $1$. Equivalently, $(\alpha_t,\beta_t)=(\cos\varphi(t),\sin\varphi(t))$ for a unique \emph{phase} $\varphi=\arccos\alpha$ in
\[
    \mathcal C\triangleq\bigl\{\varphi:[0,1]\to[0,\tfrac{\pi}{2}]\ \text{strictly increasing}:\ \varphi(0)=0,\ \varphi(1)=\tfrac{\pi}{2}\bigr\}.
\]
Denote $\mathcal S$ as the class of VP schedules. Choosing a schedule in $\mathcal S$ amounts to choosing a phase in $\mathcal C$. For VP schedules, $\alpha_t\dot\alpha_t+\beta_t\dot\beta_t=0$, so~\eqref{eq:si-denoiser-drift} simplifies to
\begin{equation}
    b_t(x)=\frac{\dot\beta_t}{\beta_t}\bigl(x+s_t(x)\bigr).
    \label{eq:vp-drift}
\end{equation}
\end{definition}

\subsection{Problem Setup}
\label{sec:problem_setup}
In this paper we study the following algorithm design problem. 

\begin{problem}
Given a class $\mathcal{A}_r$ of sampling algorithms that use at most $r$ score evaluations and a class $\mathsf P$ of target distributions, we seek the algorithm whose output law $\hat\pi_{\mathrm{alg}}(\pi)$ is uniformly closest to the target over the class,
\begin{equation}
    \inf_{\mathrm{alg}\in\mathcal{A}_r}\ \sup_{\pi\in\mathsf{P}}\ \mathrm{dist}\big(\hat\pi_{\mathrm{alg}}(\pi),\pi\big),
    \label{eq:minimax}
\end{equation}
for a distance $\mathrm{dist}$ between probability measures
\end{problem}
Algorithms access the target only through an exact score oracle, which returns $s_t(x)=\nabla\log p_t(x)$ for a queried pair $(t,x)$, and their cost is the number of queries. Score oracles for different schedules are interchangeable: by a change of time and spatial scale, one query for a VP schedule answers one query for any other schedule (\cref{prop:interpolant-conversion}), so the schedule is a genuine design choice.
\begin{definition}[Algorithm class $\mathcal{A}_r$: oblivious score-span algorithms]
\label{def:score span}
An algorithm $\mathrm{alg}\in\mathcal{A}_r$ is a pairing of (i) a stochastic interpolant schedule $(\alpha_t,\beta_t)$ with (ii) a discretization of the probability-flow ODE~\eqref{eq:probability-flow-ode} that makes $r$ evaluations of the exact score $s_t=\nabla\log p_t$ of the corresponding interpolant. Starting from $x_0=z\sim\mathcal N(0,I_d)$, each iterate lies in the span of the previous iterates and score evaluations,
\begin{equation*}
    x_j \in \operatorname{span}\bigl\{x_0,\dots,x_{j-1},\ s_{t_0}(x_0),\dots,s_{t_{j-1}}(x_{j-1})\bigr\}, \qquad j=1,\dots,r,
\end{equation*}
that is, for predetermined coefficients $a_{ij},b_{ij}$,
\begin{equation}
    x_j = \sum_{i=0}^{j-1} a_{ij}\, x_i + \sum_{i=0}^{j-1} b_{ij}\, s_{t_i}(x_i), \qquad j=1,\dots,r,
    \label{eq:oblivious-updates}
\end{equation}
and the output is $x_r^{\mathrm{alg}}\triangleq x_r$, with law $\hat\pi_{\mathrm{alg}}(\pi)\triangleq\mathcal{L}(x_r^{\mathrm{alg}})$. The schedule, which determines the oracle $s_t$, and the discretization, given by the query times $t_0,\dots,t_{r-1}\in(0,1)$ and the coefficients $a_{ij},b_{ij}$, are the design variables. They are fixed in advance (\emph{oblivious}): they may depend on $\kappa$, $d$, and $r$, but not on the particular target $\pi\in\mathsf P$.
\end{definition}
We note this class is quite broad, and contains explicit ODE solvers common in the diffusion literature (e.g., Euler, Heun, DDIM, Collocation, and DPM-Solver++); see the remark at the end of Appendix~\ref{app:algs-are-ratfuncs} for its relation to general linear methods and implicit solvers. For the theoretical results in this paper, we make the following assumption on the target distributions:

\begin{assumption}[Target Distributions]
\label{ass:targets}
$\mathsf P$ is one of two nested classes of centered distributions with condition number at most $\kappa\ge1$: Gaussians with precision spectrum in $[1,\kappa]$, and strongly log-concave distributions with curvature in $[1,\kappa]$,
    \begin{equation*}
        \mathsf{G}_\kappa=\big\{\mathcal N(0,\Lambda^{-1}): I_d\preceq\Lambda\preceq\kappa I_d\big\}
        \;\subset\;
        \mathsf{P}_\kappa=\big\{\pi\propto e^{-V}\ \text{centered}: I_d\preceq\nabla^2V\preceq\kappa I_d\big\}.
    \end{equation*}
\end{assumption}

\begin{remark}[Zero mean]
Centering is without loss of generality. If $x_1$ has mean $\mu$, then $x_t=\tilde x_t+\beta_t\mu$, where $\tilde x_t$ is the interpolant of the centered target, so the centered score is $\tilde s_t(y)=s_t(y+\beta_t\mu)$ and each centered query costs one query of the original score. One can therefore sample the centered target and add $\mu$ at the end; $\mu$ itself is recovered to any accuracy from a single score query near $t=0$ (\cref{app:centering}).
\end{remark}

In Section~\ref{sec:Gaussian-sampling}, we characterize the minimax query complexity of Gaussian sampling in total variation, up to universal constants, among oblivious score-span algorithms. Every algorithm in $\mathcal A_r$ outputs $q(\Lambda)z$, where $q$ has numerator and denominator degrees at most $r$. This connects sampling to relative-error rational approximation of $\lambda^{-1/2}$ on $[1,\kappa]$. Zoloflow implements an endpoint-normalized version of Zolotarev's optimal scalar approximant. Combining this construction with Gaussian TV bounds yields the matching upper and lower query-complexity bounds in Theorem~\ref{thm:zoloflow-Gaussian-bound}.
For $\mathsf P=\mathsf P_\kappa$ 
in Section~\ref{sec:logconcave} we show that, 
combined with collocation methods for analytic ODEs, this elliptic schedule samples with $W_2$ error $\varepsilon$ using $O\big(\sqrt d\,\mathrm{polylog}(\kappa,d,1/\varepsilon)\big)$ score evaluations.

\section{The Elliptic Schedule}
\label{sec:elliptic-si}
\subsection{Elliptic Stochastic Interpolant}

\paragraph{Motivation} As discussed in \cref{def:schedule}, a VP schedule is \emph{equivalent} to a phase function $\varphi \in \cal C$. Our primary criterion for choosing $\varphi$ will concern the analyticity properties it induces in the ODE solution~$X_t$, when extended to the complex plane.

For a Gaussian target $\mathcal{N}(0,\Lambda^{-1})$, the ODE~\eqref{eq:probability-flow-ode} can be solved as $X_t = (\alpha_t^2 I_d + \beta_t^2 \Lambda^{-1})^{\frac{1}{2}}X_0$, and thus the eigenmodes are given by
\[
x_\lambda(t) = \sqrt{\alpha(t)^2 + \lambda^{-1} \beta(t)^2} \, x_\lambda(0).
\]
This has branch point singularities whenever the expression inside the square root vanishes, i.e., whenever $\cos(\varphi(t))^2 + \lambda^{-1} \sin(\varphi(t))^2= 1 - (1-\lambda^{-1}) \sin^2 \varphi(t)=0$. Thus, our design problem is to find $\varphi$ such that the roots~$t$ of this equation are as far as possible from the real axis, whenever $\lambda \in [1,\kappa]$. This ensures that the ODE solution $X_t$ is analytic on a wide strip. The optimal such $\varphi$ can be nicely characterized as follows: 
\begin{tcolorbox}[colebase,colevisual=ColeResult]
\begin{restatable}[Jacobi amplitude as optimal phase]{proposition}{JacobiAM}
\label{prop:optimaiAM}
Consider the phase (angular clock) given by
\begin{equation}
\varphi_\star(t) \triangleq \am(K t \mid m),
\label{eq:thetaAm}
\end{equation}
where $m = 1-\kappa^{-1}$, $K = K(m)$, $K'=K(1-m)$, and $\am()$ is the Jacobi amplitude function and $K(\cdot)$ the complete elliptic integral of the first kind (see Appendix~\ref{app:jacobi}). 
Let $S_\gamma \triangleq \{t \in \C : |\Im t| < \gamma \}$ be a strip around the real axis of width $2\gamma$. This choice of $\varphi$ maximizes $\gamma$ (analyticity strip width), subject to the constraints
\begin{itemize}
 \item[i.]  $\varphi \text{ is holomorphic on } S_\gamma$
 \item[ii.]  $\varphi(0)=0, \, \varphi(1)=\pi/2$
 \item[iii.]  $1-(1-\lambda^{-1}) \sin^2 \varphi(t) \not = 0 \qquad 
 \forall t \in S_\gamma, \quad
 \forall \lambda \in [1,\kappa]$    
\end{itemize}
In particular, for this choice of $\varphi_\star$ we have $\gamma_\star = K'/K$.
\end{restatable}
\end{tcolorbox}
A proof is provided in Appendix~\ref{sec:proofJacobi}. 

\begin{figure}[t]
\centering
\includegraphics[width=\linewidth,height=.73\textheight,keepaspectratio]{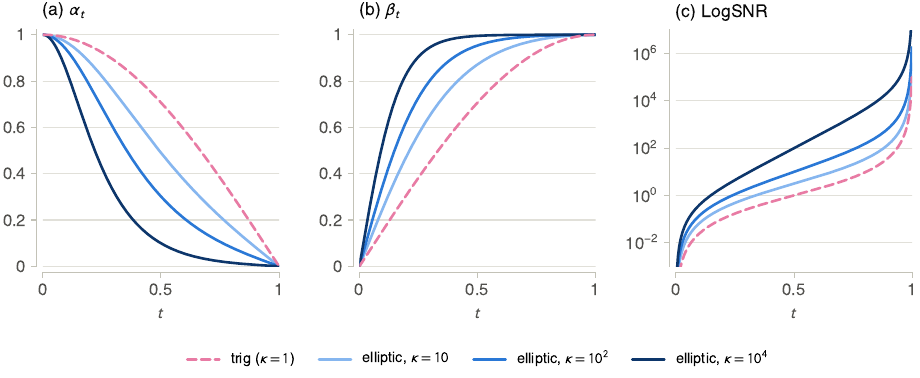}
\caption{\textbf{Elliptic schedules for several values of $\kappa$:} (a) $\alpha_t$, (b) $\beta_t$ and (c) the signal-to-noise ratio $\beta_t^2/\alpha_t^2=\jsc^2(Kt\mid m)$ on a logarithmic axis. For $\kappa=1$ the elliptic schedule is the trigonometric schedule. }
\label{fig:elliptic-schedule}
\end{figure}

\begin{figure}[t]
\centering
\includegraphics[width=0.75\linewidth,height=.73\textheight,keepaspectratio]{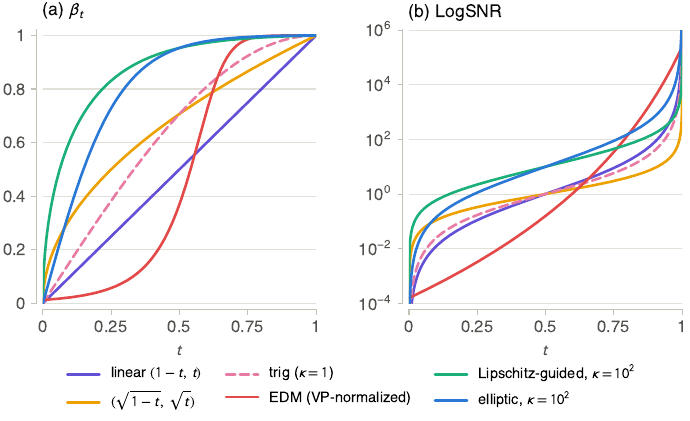}
\caption{\textbf{Common schedules next to the elliptic schedule ($\kappa=10^2$):} (a) $\beta_t$ and (b) the signal-to-noise ratio $\beta_t^2/\alpha_t^2$ on a logarithmic axis. EDM \citep{karras2022elucidatingdesignspacediffusionbased} uses the VP-normalized interpolant $(\alpha_t,\beta_t)=(\sigma_\rho(t),1)/\sqrt{1+\sigma_\rho(t)^2}$, as in \cref{app:schedule-catalog}, with $\sigma_{\max}=80$, $\sigma_{\min}=0.002$, and $\rho=7$; its SNR $\sigma_\rho(t)^{-2}$ is finite at both ends. The Lipschitz-guided schedule \citep{chen2025lipschitz} and the elliptic schedule use $\kappa=10^2$.}
\label{fig:schedule-zoo}
\end{figure}

The optimality of the Jacobi amplitude $\am()$ leads directly to the following definition:

\begin{definition}[Elliptic schedule]
    For a given $\kappa \geq 1$, let
\[
    m = 1-\kappa^{-1}, \qquad K = K(m), \qquad K' = K(1-m).
\]
The \emph{elliptic interpolant schedule} is the VP schedule corresponding to the Jacobi amplitude phase $\varphi_*(t) = \am(Kt \mid m)$:
\begin{equation}
    \alpha_t=\cos(\am(K t\mid m)),
    \qquad
    \beta_t=\sin(\am(Kt \mid m))
    \qquad t\in[0,1].
\end{equation}
These functions are well understood and known as the Jacobi elliptic functions $\cn$ and $\sn$. We thus write the elliptic schedule as
\begin{equation}
    \alpha_t=\cn(K t\mid m),
    \qquad
    \beta_t=\sn(K t\mid m),
    \qquad t\in[0,1].
\end{equation}
\label{def:ellipticschedule}
\end{definition}
The elliptic schedule is variance-preserving, and satisfies all properties in Definition~\ref{def:schedule}. The holomorphic properties of $\varphi_*$ guarantee that $\alpha_t$ and $\beta_t$ are also holomorphic on the complex strip $S_{\gamma_*}$. A unique property of the elliptic schedule is the dependence on the condition number $\kappa$ (\cref{fig:elliptic-schedule}). For $\kappa = 1$, this reduces to the trigonometric schedule from \cite{albergo2025stochastic}. \Cref{fig:schedule-zoo} places it next to common schedules.

\subsection{Elliptic schedule as a two-parameter family}
\label{sec:two-param-family}
While the construction above uses the normalized precision interval $\text{spec}(\Lambda)\subseteq [1,\kappa]$, one may extend the elliptic schedule to an arbitrary spectral interval $[\lambda_{\min},\lambda_{\max}]$ by applying the elliptic schedule with $\kappa=\lambda_{\max}/\lambda_{\min}$ to the rescaled target $\tilde x_1=\sqrt{\lambda_{\min}}\,x_1$ and dividing the sampled output by $\sqrt{\lambda_{\min}}$ (\cref{app:target-rescaling}). The elliptic schedule is thus a two-parameter family, which we can express either in endpoint coordinates $(\ell,L)=(\lambda_{\min},\lambda_{\max})$ with $\kappa \triangleq L/\ell$. 

or equivalently by center--width coordinates $(\bar\lambda,\kappa)$, related by $(\bar\lambda,\kappa)\triangleq\left(\sqrt{\ell L},\,L/\ell\right),
    [\ell,L]=[\bar\lambda/\sqrt\kappa,\,\bar\lambda\sqrt\kappa].$
\[
    (\bar\lambda,\kappa)\triangleq\left(\sqrt{\ell L},\,L/\ell\right),
    \qquad
    [\ell,L]=[\bar\lambda/\sqrt\kappa,\,\bar\lambda\sqrt\kappa].
\]

\begin{figure}[ht]
    \centering
    \includegraphics[width=0.9\linewidth,height=.73\textheight,keepaspectratio]{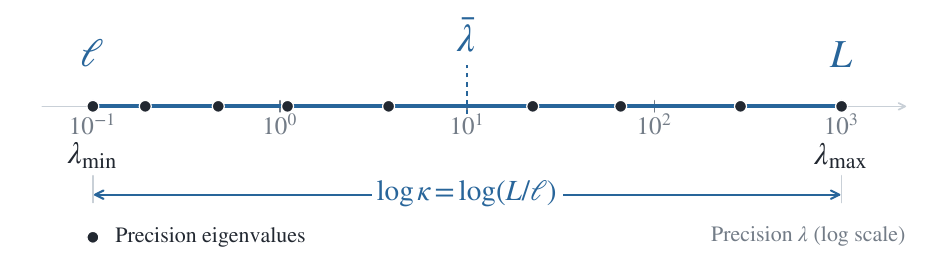}
    \caption{\textbf{Endpoint and center--width coordinates.} Black dots represent the eigenvalues $\lambda_i(\Lambda)$ of an illustrative precision matrix $\Lambda=\Sigma^{-1}$. For the matched choice $\ell=\lambda_{\min}$ and $L=\lambda_{\max}$, the geometric center $\bar\lambda=\sqrt{\ell L}$ lies at the midpoint of the interval on the logarithmic axis, and the interval has logarithmic width $\log\kappa=\log(L/\ell)$. Here $\lambda_{\min}=10^{-1}$ and $\lambda_{\max}=10^3$, so $\bar\lambda=10$ and $\kappa=10^4$.}
    \label{fig:center-width-coordinates}
\end{figure}

Here $\bar\lambda$ is the geometric center of the $[\ell,L]$ interval: on the log-precision axis, its midpoint is $\log\bar\lambda$ and its width is $\log\kappa$. Centering refers to the location of this interval, rather than to subtracting the target mean. At fixed $\kappa$, varying $\bar\lambda$ adjusts the relative source--target scale; at fixed $\bar\lambda$, varying $\kappa$ changes the interval width.\Cref{fig:center-width-coordinates} illustrates these coordinates for a matched spectral interval. For $\kappa=1$ ($\ell = L$), the elliptic schedule reduces to the trigonometric schedule, written $(\ell,L)=(\bar\lambda,\bar\lambda)$ in endpoint coordinates and $(\bar\lambda,\kappa)=(\bar\lambda,1)$ in center--width coordinates. Unless otherwise specified, we use the center--width hyperparameters $(\bar\lambda,\kappa)$ to separate scale and width. 

\paragraph{All schedules $(\alpha,\beta)$ can be centered.}
We note that any baseline schedule can accommodate a centering parameter $\bar{\lambda}$ through the same target rescaling mentioned previously\footnote{In terms of the hyperparameters $(\ell,L)$, this forces $\ell = L$.}. The idea of rescaling source or target is not new; target rescaling and shifts in log-SNR schedules are established design choices \citep{chen_importance_nodate,hoogeboom_simple_2023,esser2024scaling}. As the elliptic family gains the additional width hyperparameter $\kappa$\footnote{The Lipschitz-guided schedule of \citet{chen2025lipschitz} also depends on $\kappa$ and therefore admits a two-parameter family under target rescaling.} and reduces to a trigonometric schedule at $\kappa =1$, we thus ask whether elliptic width $\kappa$ contributes on top of a centering calibration.
\paragraph{Choosing hyperparameters $(\bar{\lambda},\kappa)$} For a Gaussian target $\mathcal N(0,\Lambda^{-1})$, matching the parameters to the spectral bounds, $(\ell,L)=(\lambda_{\min}(\Lambda),\lambda_{\max}(\Lambda))$
or equivalently in terms of positive definite covariance matrix $\Sigma$
\[
    (\bar\lambda,\kappa)=\left(\frac{1}{\sqrt{\lambda_{\min}(\Sigma)\,\lambda_{\max}(\Sigma)}},\ \frac{\lambda_{\max}(\Sigma)}{\lambda_{\min}(\Sigma)}\right).
\]
makes our proposed algorithm Zoloflow minimax optimal (\cref{sec:Gaussian-sampling}). Our experiments describe how extending this choice as a covariance heuristic for non-Gaussian targets may not be optimal. Thus in general we treat both $\bar\lambda$ and $\kappa$ as hyperparameters to tune.

\begin{figure}[!htbp]
    \centering
        \includegraphics[width=\textwidth,height=.73\textheight,keepaspectratio]{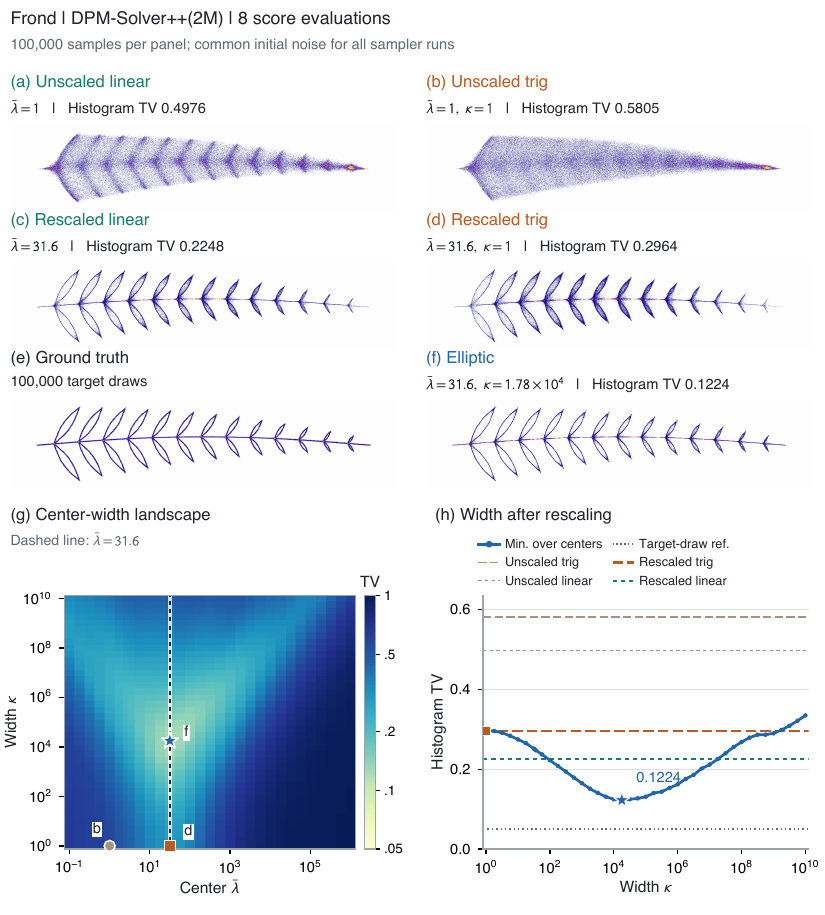}
    \caption{\textbf{Rescaling and the additional benefit of elliptic width.}
DPM-Solver++(2M) with exact scores and eight score evaluations.
The left column shows unscaled linear, rescaled linear, and ground truth;
the right shows unscaled trig, rescaled trig, and the best tested elliptic setting.
Unscaled baselines use $\bar\lambda=1$; rescaled linear (c), rescaled trig (d),
and elliptic (f) share $\bar\lambda\approx31.6$.
Comparing rescaled trig (d) with elliptic (f), at $\kappa\approx1.78\times10^4$,
isolates the additional width effect at this center; rescaled linear (c)
provides another calibrated baseline. Each sample panel contains $10^5$ draws,
with common initial noise across sampler runs.
Panel (g) shows histogram TV over the tested center--width grid (logarithmic
color scale), with markers identifying (b), (d), and (f).
Panel (h) minimizes over tested centers at each width; the lighter horizontal
lines show unscaled references. The minimizing center remains
$\bar\lambda\approx31.6$ from $\kappa=1$ through the selected minimum, so this
descent also holds at a fixed center. Histogram TV uses $100\times100$ bins
plus an outside bin and a reference of $10^6$ target draws. The grey dotted
line shows an independent target draw against that reference, not a lower
bound. Rescaled baselines are selected over 30 centers; elliptic is selected
over the same centers and 42 widths. These are best observed settings within
the tested grids.}
        \label{fig:frond-main}
\end{figure}

\Cref{fig:frond-main} compares unscaled and independently tuned rescaled baselines with the elliptic family on a non-Gaussian frond target. The shared endpoint sweeps and tuned schedule comparisons on this target and four additional line drawings are given in \Cref{app:toy2d}.

\section{Gaussian Sampling and Zoloflow}
\label{sec:Gaussian-sampling}
We now study the minimax problem \eqref{eq:minimax} over the Gaussian class $\mathsf G_\kappa$ with $\mathrm{dist}=d_{\mathrm{TV}}$
\begin{equation}
    \inf_{\mathrm{alg}\in\mathcal{A}_r}\ \sup_{\pi\in\mathsf{G}_\kappa}\ d_{\text{TV}}\big(\hat\pi_{\mathrm{alg}}(\pi),\pi\big).
    \label{eq:minimax_gauss}
\end{equation} 
We structure this section in the following way. \textbf{First}, on Gaussian targets, we show every algorithm in $\mathcal A_r$ can be understood as a rational function on its precision eigenvalues $\lambda_i \in \text{spec}(\Lambda)$. As Gaussian sampling can be realized by a matrix-vector multiplication $\Lambda^{-1/2}z$ the best algorithm will correspond to the best uniform approximation of $\lambda^{-1/2}$ on $[1,\kappa]$. \textbf{Second}, we show that the above uniform approximation is nothing other than a reformulation of an optimal rational approximation problem solved by \citet{zolotarev1877application}. \textbf{Third}, we show that the elliptic schedule, paired with a custom finite-difference discretization, attains this optimum up to constant factors. The resulting algorithm, \emph{Zoloflow} (\cref{sec:zoloflow}), therefore matches the minimax lower bound on the number of score queries up to constants (\cref{sec:Gaussian complexity}).

\subsection{Score-Oracle Algorithms for Gaussian Sampling Are Rational Functions}
Before getting into the explicit complexity results for Zoloflow, we describe the core intuition behind our results: that every algorithm $\mathrm{alg}\in \mathcal{A}_r$ (\cref{def:score span}) has a rational function representation. We formalize this in the following lemma.
\begin{lemma}[Algorithms in $\mathcal{A}_r$ are rational functions on Gaussians]
    \label{lma:algs-are-ratfunc}
    Suppose the target distribution $\pi=\mathcal{N}(0,\Lambda^{-1})$ is a zero-mean Gaussian, with precision matrix $\Lambda = U \text{diag}(\lambda_1, \dots, \lambda_d) U^T$. For any $m,n\in \mathbb{N}$ define
        \begin{equation}
        \mathcal{R}_{m,n}
        \triangleq
        \left\{
        r(x)=\frac{P(x)}{D(x)}
        \;\middle|\;
        P,D\in\mathbb{R}[x],\;
        \deg P\le m,\;
        \deg D\le n,\;
        D\not\equiv 0
        \right\}.
    \end{equation}
    as the class of all rational functions whose numerator is a polynomial of degree at most $m$ and denominator a polynomial with degree at most $n$. Then for any algorithm in $\mathcal{A}_r$ we can express its final iterate as
    \begin{equation}
        x_r^{\mathrm{alg}} = q_r^{\mathrm{alg}}(\Lambda) z = U \text{diag}(q_r^{\mathrm{alg}}(\lambda_1),\dots, q_r^{\mathrm{alg}}(\lambda_d))U^Tz
    \end{equation}
     where $q_r^{\mathrm{alg}}(\lambda) \in \mathcal{R}_{r,r}$ is a rational function that can be obtained from the algorithm coefficients $((\alpha_{t_j}), (\beta_{t_j}), (a_{ij}), (b_{ij}))$.  
    We delay a proof of this lemma to Appendix \ref{app:algs-are-ratfuncs}.
\end{lemma}
The key idea is that for Gaussian distributions, every algorithm in $\mathcal{A}_r$ (more precisely each interpolant schedule and discretization pair) admits a rational function $q(\cdot)$ which acts on the eigenvalues of the precision matrix $\Lambda$. The performance of each algorithm is nothing more than a question about the quality of the approximation of $\lambda^{-1/2}$. In particular
\begin{equation}
    q(\Lambda) \approx \Lambda^{-1/2} \implies q(\Lambda) z \approx \Lambda^{-1/2} z \eqd \mathcal{N}(0,\Lambda^{-1})
\end{equation}
A \textit{good} algorithm for Gaussians must thus give a \textit{good} rational approximation to $\lambda^{-1/2}$ uniformly over $[1,\kappa]$.

\subsection{Zolotarev's optimal rational approximation}

Let $\mathcal{R}_{p,q}$ denote the real rational functions whose numerator and denominator have degrees at most~$p$ and $q$, respectively. We would like to approximate the function $1/\sqrt{\lambda}$ on the interval $[1,\kappa]$, with an element of $\mathcal{R}_{p,q}$. For $\kappa>1$ and $r\in \mathbb{N}_0$, define the optimal relative approximation error
\begin{equation}
    E_{\kappa,r}\triangleq
    \min_{q\in\mathcal{R}_{r,r}}
    \max_{\lambda\in[1,\kappa]}
    \left|\sqrt{\lambda}\,q(\lambda)-1\right|,
    \label{eq:zolo-minimax}
\end{equation}
where the minimization is restricted to functions without poles on $[1,\kappa]$. The unique minimizer, which we denote by $q_{\kappa,r}(\lambda)$, is a reformulation of the classically known \cite{zolotarev1877application} approximation to the $\text{sign}(x)$ function, and is written in terms of Jacobi elliptic functions. We clarify below.

\paragraph{Sign function.}
Zolotarev's theorem \cite{zolotarev1877application} gives the best type-$(2r+1,2r)$ rational approximation $Z(x)$ to $\operatorname{sign}(x)$ on $[-\sqrt{\kappa},-1]\cup[1,\sqrt{\kappa}]$. Since $\operatorname{sign}(x)$ is odd, so is the optimal approximant, and thus it has the form
\[
    Z(x)=Cx\frac{P(x^2)}{Q(x^2)}
        =xq_{\kappa,r}(x^2),
    \qquad \deg P,\deg Q\leq r.
\]
The substitution $\lambda=x^2$ gives the exact error identity
\begin{equation}
    \max_{1\leq|x|\leq\sqrt{\kappa}}
    |Z(x)-\operatorname{sign}(x)|
    =
    \max_{\lambda\in[1,\kappa]}
    \left|\sqrt{\lambda}\,q_{\kappa,r}(\lambda)-1\right|
    =E_{\kappa,r}.
    \label{eq:zolo-sign-equivalence}
\end{equation}
Thus \emph{absolute-error} approximation of the sign function is equivalent to \emph{relative-error} approximation of the inverse square root, and achieves the same optimal error. Rescaling $x$ by $\sqrt{\kappa}$ gives the usual sign-approximation domain $[-1,-\sqrt{1-m}]\cup[\sqrt{1-m},1]$, where $m\triangleq1-\kappa^{-1}$; see \citet[Section~3.1]{nakatsukasa2016computing}. More generally, on any positive interval $[a,b]$, the approximant $a^{-1/2}q_{b/a,r}(\lambda/a)$ has optimal relative error $E_{b/a,r}$. Thus, the ratio $\kappa=b/a$ of the interval endpoints directly controls the optimal approximation error.

\subsection{Zoloflow}
\label{sec:zoloflow}
\begin{algorithm}[t]
\caption{Zoloflow - sample from a general distribution from score queries}
\label{alg:zoloflow}
\begin{algorithmic}[1]
\REQUIRE Step count $r$, parameter $\kappa$ (``condition number'')
\STATE Initial sample $x_0\sim\mathcal N(0,I_d)$
\STATE Set step size $h=2/(2r+1)$ and $t_j=(2j-1)/(2r+1)$ for $j=1,\ldots,r$
\STATE Set parameter $m=1-\kappa^{-1}$, $K = K(m)$, $\alpha(t)=\cn(K t\mid m)$, and $\beta(t)=\sn(Kt\mid m)$
\FOR{$j=1,\ldots,r$}
\STATE Query score: $s_{t_j}(x_{j-1})=\nabla\log p_{t_j}(x_{j-1})$
\STATE Update: $x_j\leftarrow x_{j-1}+\dfrac{\beta^2(t_j+h/2)-\beta^2(t_j)}{\beta^2(t_j)}\left(x_{j-1}+s_{t_j}(x_{j-1})\right)$
\ENDFOR
\RETURN $x_r$
\end{algorithmic}
\end{algorithm}

The elliptic schedule given in Definition~\ref{def:ellipticschedule} can in principle be paired with any numerical method (e.g., Euler, Heun, etc.) to solve the probability flow ODE~\eqref{eq:probability-flow-ode}. However, if instead we use a non-standard (but natural) discretization we obtain \emph{Zoloflow}, a sampling algorithm that is \emph{essentially optimal for Gaussians}.  

\begin{definition}[Elliptic finite-difference discretization]
\label{def:customdiscretization}
Fix a step count $r$ and an interpolant schedule $(\alpha,\beta)$. The \emph{elliptic finite-difference} update is
\begin{align}
    x_j
    &=
    x_{j-1}
    +
    h
    \left[
        \frac{
            \beta_{t_j+h/2}^2
            -
            \beta_{t_j}^2
        }{
            h\,\beta_{t_j}^2
        }
    \right]
    \left(
        x_{j-1}+ s_{t_j}(x_{j-1})
    \right),
    \qquad j=1,\ldots,r.
    \label{eq:si-finite-diff}
\end{align}
on the grid $h = \frac{2}{2r+1}$, $t_j\in \left\{\frac{1}{2r+1}, \frac{3}{2r+1}, \ldots, \frac{2r-1}{2r+1}\right\}.$
    \label{def:elliptic_finite_diff}
\end{definition}
A derivation of this approach, and a comparison with standard Euler and deterministic DDIM discretizations using the same iterate and time-grid notation are presented in Appendix~\ref{app:additional-discretizations}. 

\begin{definition}[Zoloflow]
\label{def:zoloflow}
Fix $\kappa \geq 1$ and $r\in\mathbb{N}$, a target distribution $\pi$ with $x_1\sim \pi$, and set $m\triangleq1-\kappa^{-1}$. 
Zoloflow is the $r$-step algorithm obtained by applying the elliptic finite-difference discretization (Def.~\ref{def:elliptic_finite_diff}) to the flow ODE \eqref{eq:probability-flow-ode}, using the elliptic schedule $(\alpha_t,\beta_t)$ given in Def.~\ref{def:ellipticschedule}. We use a step size $h=2/(2r+1)$ and time grid $t_j=(2j-1)/(2r+1)$, $j=1,\ldots,r$. A formal description is presented as Algorithm~$\ref{alg:zoloflow}$ above.
\end{definition}

On centered Gaussian targets with exact interpolant scores, Zoloflow exactly realizes the endpoint-normalized Zolotarev approximant:
\[
    x_r=\widehat q_{\kappa,r}(\Lambda)z,
    \qquad
    \widehat q_{\kappa,r}(\lambda)
    =\frac{q_{\kappa,r}(\lambda)}{q_{\kappa,r}(1)}.
\]
Here $q_{\kappa,r}$ is the optimal scalar relative-error approximant to $\lambda^{-1/2}$ on $[1,\kappa]$. Endpoint normalization enforces $\widehat q_{\kappa,r}(1)=1$, preserving the unit-precision mode exactly, and changes the worst-case relative error to
\[
    \widehat E_{\kappa,r}
    =\frac{2E_{\kappa,r}}{1-E_{\kappa,r}}.
\]
Thus normalization preserves the asymptotic convergence rate, although $\widehat q_{\kappa,r}$ no longer minimizes the scalar relative error. The exact realization is established in Lemma~\ref{lma:zoloflow-normalized-zolotarev}.

Although primarily motivated by Gaussian targets, both the elliptic schedule and Zoloflow can be used to sample from \emph{any} distribution. As we show in later sections, they will perform quite well from both theoretical (in the Gaussian/log-concave case) and empirical (for general distributions) viewpoints, often outperforming competing techniques.

\subsection{Score-Oracle Complexity of Gaussian Sampling} \label{sec:Gaussian complexity}
We formally define the minimax query complexity as follows:
\begin{equation}
\label{eq:r_optimal}
    r^*_{\mathrm{obl}}(d,\kappa,\varepsilon)
    \triangleq
    \min\left\{
        r\in\mathbb Z_{\geq1}:
        \begin{array}{l}
            \exists\,\mathrm{alg}\in\mathcal A_r\ \text{such that}\\[2pt]
            \displaystyle
            \sup_{I_d\preceq\Lambda\preceq\kappa I_d}
            d_{\mathrm{TV}}\!\left(
                \mathcal L(x_r^{\mathrm{alg},\Lambda}),
                \mathcal N(0,\Lambda^{-1})
            \right)
            \leq\varepsilon
        \end{array}
    \right\}.
\end{equation}
Here each algorithm, including its admissible interpolant schedule, query times, and scalar update coefficients, is chosen independently of $\Lambda$, the initialization, and the oracle responses.

\begin{theorem}[Gaussian TV query complexity among oblivious score-span algorithms]\label{thm:zoloflow-Gaussian-bound}
There exist universal constants $c,C>0$ and $\varepsilon_0\in(0,1)$ such that, for every $d\geq1$, $\kappa\geq2$, and $0<\varepsilon\leq\varepsilon_0$,
\[
    c\log\kappa\,\log\!\left(\frac{\sqrt d}{\varepsilon}\right)
    \leq r^*_{\mathrm{obl}}(d,\kappa,\varepsilon)
    \leq C\log\kappa\,\log\!\left(\frac{\sqrt d}{\varepsilon}\right).
\]
In other words,
\[
    r^*_{\mathrm{obl}}(d,\kappa,\varepsilon)
    =\Theta\!\left(\log\kappa\,\log\!\left(\frac{\sqrt d}{\varepsilon}\right)\right),
\]
with universal implicit constants throughout this parameter regime.
Zoloflow attains the upper bound: with a suitable query budget $r$, exact interpolant scores, and initialization $x_0=z\sim\mathcal N(0,I_d)$, its output law $\widehat\pi\triangleq\mathcal L(x_r)$ satisfies $d_{\mathrm{TV}}(\widehat\pi,\pi)\leq\varepsilon$ uniformly over centered Gaussian targets $\pi=\mathcal N(0,\Lambda^{-1})$ with $I_d\preceq\Lambda\preceq\kappa I_d$.
\end{theorem}

We give a proof in Appendix~\ref{app:compelxity-results-zoloflow}. 

\Cref{fig:gauss-r} compares solver--schedule pairs on a Gaussian target $\mathcal N(0, \Lambda^{-1})$ with spectrum $\spec(\Lambda) \subset [1,\kappa]$, where $\kappa=100$. For each pair $\mathrm{alg}$ it plots the worst-case relative error $E_{\kappa,r}(\mathrm{alg})=\max_{\lambda\in[1,\kappa]}|\sqrt\lambda\,q_r^{\mathrm{alg}}(\lambda)-1|$ of its rational function (\cref{lma:algs-are-ratfunc}), which controls the sampling error, $D_{\mathrm{KL}}(\widehat\pi\,\|\,\pi)\lesssim d\,E_{\kappa,r}(\mathrm{alg})^2$ by \eqref{eq:zoloflow-kl-relative-errors}. No pair can fall below the Zolotarev error $E_{\kappa,r}$ of \eqref{eq:zolo-minimax}, and Zoloflow attains $\widehat E_{\kappa,r}=2E_{\kappa,r}/(1-E_{\kappa,r})$ (\cref{app:Gaussian-experiments}).

\begin{figure}[t]
    \centering
    \includegraphics[width=\textwidth,height=.73\textheight,keepaspectratio]{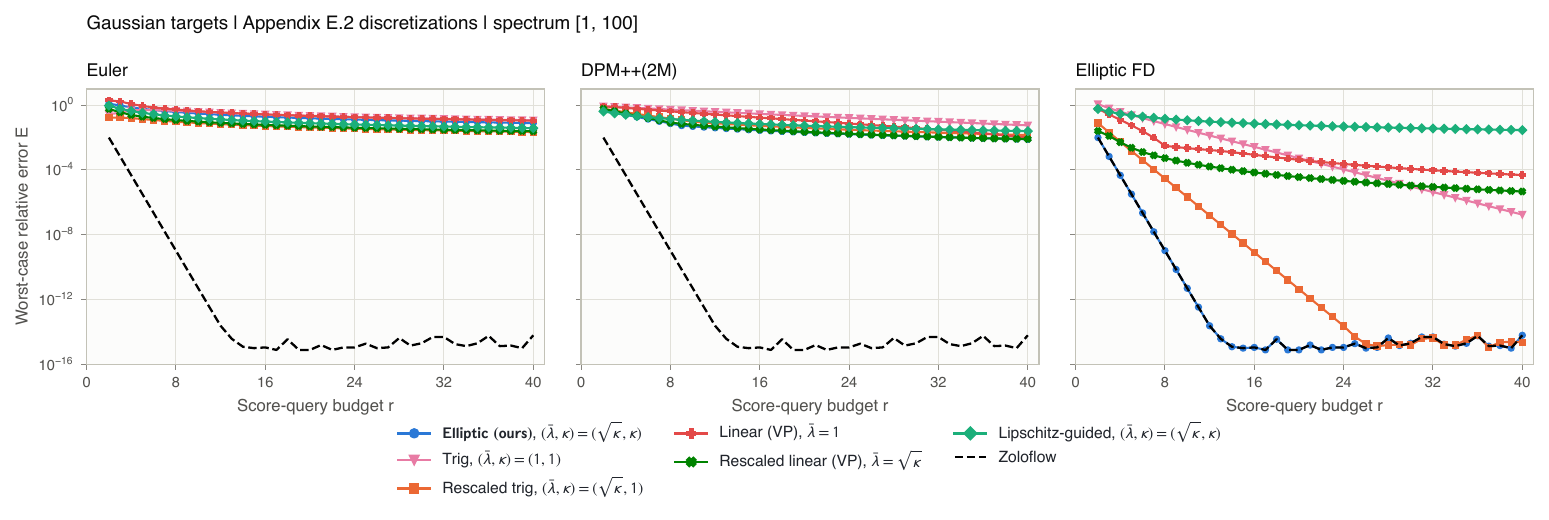}
    \caption{\textbf{Worst-case relative error $E$ against score evaluations $r$ on a Gaussian target with precision spectrum $[1,100]$.} Panels: Euler, DPM-Solver++(2M), and elliptic finite difference (FD). Colors identify schedules and rescalings; dashed black is Zoloflow (elliptic FD with the elliptic schedule). All six solvers and setup details appear in \cref{app:Gaussian-experiments}.}
    \label{fig:gauss-r}
\end{figure}

\section{Log-Concave Sampling}
\label{sec:logconcave}

We now consider the class $\mathsf{P}_{\kappa}$ of centered strongly log-concave targets (\cref{sec:problem_setup}) and measure error in the 2-Wasserstein distance, $\mathrm{dist}=W_2$ in \eqref{eq:minimax}.  Since the score is nonlinear, the rational-approximation reduction of \cref{sec:Gaussian complexity} no longer applies. We instead build on the complex-analytic criterion of \cref{prop:optimaiAM}: after a mild Gaussian smoothing, the elliptic schedule keeps all PF-ODE trajectories holomorphic and bounded on a strip around $\R$ whose width degrades only polylogarithmically in $\kappa$ and $1/\varepsilon$ (\cref{thm:holomorphic strip width}). Collocation methods for analytic ODEs \citep{leeAlgorithmicTheoryODEs2018} then yield $W_2$-error $\varepsilon$ with $O(\sqrt d\,\log^{3.5}(\kappa d/\varepsilon))$ score evaluations (\cref{thm:log concave complexity}).

\subsection{Holomorphic Width under the Elliptic Schedule}
\label{sec:lc-width}

\paragraph{Holomorphic width.}
For an ODE $\dot X_t=b_t(X_t)$ with $X_0=z$, a radius $B>0$, and the strip $S_\omega\triangleq\{t\in\C:|\Im t|<\omega\}$, define the holomorphic width
\begin{equation}
    \Omega_B(b)\triangleq\sup\bigl\{\omega>0:\ t\mapsto X_t \text{ extends holomorphically to } S_\omega \text{ for all } \|z\|\le B\bigr\}.
    \label{eq:holomorphic-width}
\end{equation}
If, in addition, all such trajectories are bounded by $M$ on $S_\omega$ and $b_t$ is $L$-Lipschitz in space, then collocation approximates $X_1$ to accuracy $\zeta$ uniformly over $\|z\|\le B$ with
\begin{equation}
    N\lesssim\max(1,L,\omega^{-1})\Bigl(1+L+\log\frac{M\max(1,L,\omega^{-1})}{\zeta}\Bigr)^{2}
    \label{eq:collocation-complexity}
\end{equation}
drift evaluations (\cref{thm:analytic ODEs}; \citealp[Thm.~2.3]{leeAlgorithmicTheoryODEs2018}). 
For VP schedules, $b_t(x)=\frac{\dot\beta_t}{\beta_t}(x+s_t(x))$, so each drift evaluation is a single score query.
\paragraph{Smoothing as early stopping.}
For Gaussian targets, the elliptic schedule makes every trajectory holomorphic on $S_\omega$ with $\omega=K(1-m)/K(m)\approx\pi/\log(16\kappa)$, independently of $d$ and $z$ (\cref{prop:optimaiAM}). On $\mathsf{P}_\kappa$, analyticity can fail at $t=1$ regardless of the schedule (\cref{ex:non-analytic log-concave example}), although it holds at every $t\in(0,1)$ for analytic schedules (\cref{app:holomorphic extension}). We therefore target a smoothed distribution: for $\widetilde\sigma\in(0,1)$, $z\sim N(0,I_d)$, and $x_1\sim\pi$ (\cref{lem:smoothing-bias}),
\begin{equation}
\begin{gathered}
\widetilde\pi\triangleq\mathcal{L}\bigl(\widetilde\sigma z+\sqrt{1-\widetilde\sigma^2}\,x_1\bigr)\in\mathsf{P}_{\widetilde\kappa},
    \quad\text{where}\quad
    \widetilde\kappa\triangleq{\kappa}(1+\widetilde\sigma^2(\kappa-1))^{-1}\le\kappa,
\end{gathered}
\label{eq:Gaussian smoothing}
\end{equation}
with smoothing error $W_2(\widetilde\pi,\pi)\le\sqrt{d/\kappa}\,(\sqrt{1+\widetilde\sigma^2(\kappa-1)}-1)$. For a VP schedule $(\alpha,\beta)$, the PF-ODE transporting $N(0,I_d)$ to $\widetilde\pi$ and its score are
  \begin{equation}\label{eq:smoothed pf-ode}
\begin{gathered}
    \dot{\widetilde X}_t=\widetilde b_t(\widetilde X_t)\triangleq\frac{\dot\beta_t}{\beta_t}\bigl(\widetilde X_t+\widetilde s_t(\widetilde X_t)\bigr),\quad\text{where}\\ %
    \widetilde s_t\triangleq s_{\tau(t)},  \quad\text{and}\quad\tau(t)\triangleq\beta^{-1}\bigl(\sqrt{1-\widetilde\sigma^2}\,\beta_t\bigr)\in[0,1),
\end{gathered}
\end{equation}
since the smoothed interpolant satisfies $\alpha_t z+\beta_t\widetilde x_1\stackrel{(d)}{=}\alpha_{\tau(t)}z+\beta_{\tau(t)}x_1$ with $\widetilde x_1\sim\widetilde\pi$. Hence each evaluation of $\widetilde b_t$ costs one query of the original score, and $\widetilde X_t=X_{\tau(t)}$, so integrating~\eqref{eq:smoothed pf-ode} over $[0,1]$ amounts to stopping the original PF-ODE early at $\tau(1)<1$ (\cref{app:log-concave holomorphic}). We seek the schedule maximizing the worst-case width 
\begin{equation}
    \omega_\star(\kappa,d,B,\widetilde\sigma)\triangleq\sup_{(\alpha,\beta)\in \mathcal{S}}\inf_{\pi\in\mathsf{P}_\kappa}\Omega_B(\widetilde b).
\end{equation}

\begin{theorem}[Holomorphic width under the elliptic schedule]
\label{thm:holomorphic strip width}
 Let $\kappa,\widehat\kappa>1$, $\widetilde\sigma\in(0,1)$, $B>0$, and $D\triangleq\sqrt d+\sqrt{B^2+d\log\kappa}$. Under the elliptic schedule (\cref{def:ellipticschedule}) with parameter $\widehat m\triangleq1-\widehat\kappa^{-1}$, for every $\pi\in\mathsf{P}_\kappa$ and $\|z\|\le B$, the trajectory of~\eqref{eq:smoothed pf-ode} is holomorphic on $S_{\omega_{\hat\kappa}}$ where
\begin{equation}
    \omega_\star(\kappa,d,B,\widetilde\sigma)\ \ge\ \omega_{\widehat\kappa}\triangleq\frac{\min\bigl(1,\widetilde\sigma\sqrt{\widehat\kappa}\bigr)}{K(\widehat m)\bigl(1+30\sqrt{1-\widetilde\sigma^2}\,D\bigr)}.
\end{equation}
and bounded on this strip:
\begin{equation}
     \sup_{t\in S_{\omega_{\hat\kappa}}}\|\widetilde X_t\|\le B+\sqrt d+1.
\end{equation}
\end{theorem}
Choosing $\widehat\kappa=\widetilde\sigma^{-2}$, where $\min(1,\widetilde\sigma\sqrt{\widehat\kappa})$ saturates, tunes the schedule to the smoothing level rather than to $\kappa$. Restricting the infimum to Gaussian targets and applying \cref{prop:optimaiAM} with the additional requirement of the phase function to be analytic then gives
\begin{equation}
    \frac{1}{D\log(4/\widetilde\sigma)}
    \ \lesssim\
    \omega_\star(\kappa,d,B,\widetilde\sigma)
    \ \le\
    \frac{K(\widetilde\kappa^{-1})}{K(1-\widetilde\kappa^{-1})}
    \ \xrightarrow{\ \widetilde\sigma\to0\ }\
    \frac{K(\kappa^{-1})}{K(1-\kappa^{-1})}\approx\frac{\pi}{\log(16\kappa)},
    \label{eq:lc-width-upper}
\end{equation}
for large $\kappa\gg 1$, so the elliptic schedule is optimal up to the factor $D\asymp\sqrt{d(1+\log\kappa)}+B$ and logarithmic terms.

\subsection{Score-Oracle Complexity of Log-Concave Sampling}
\label{sec:lc-complexity}
  With $\widehat m=1-\widetilde\sigma^2$, the smoothed drift satisfies $-K(\widehat m)I_d\preceq\nabla\widetilde b_t\preceq0$ (\cref{thm:lipschtitz}), so $L\le K(\widehat m)$ and the flow $z\mapsto\widetilde X_t$ is non-expansive. Combining this with \cref{thm:holomorphic strip width} and~\eqref{eq:collocation-complexity} gives the following result.

\begin{theorem}[Score-oracle complexity for log-concave targets]
\label{thm:log concave complexity}
Let $\pi\in\mathsf{P}_\kappa$ with $\kappa>1$, let $0<\varepsilon\le\min(1,\bar\varepsilon)$ with $\bar\varepsilon\triangleq(\kappa-1)\sqrt{d/\kappa}$, and set $\widetilde\sigma^2=\widehat\kappa^{-1}=\frac{\varepsilon}{2\bar\varepsilon}$, $ B=\sqrt d+\sqrt{2\log(32/\varepsilon^2)}$, $\zeta=\frac{\varepsilon}{4}$. Integrate~\eqref{eq:smoothed pf-ode} under the elliptic schedule with parameter $\widehat m=1-\widehat\kappa^{-1}$ by collocation
with tolerance $\zeta$, starting from\footnote{While this initialization is not truly Gaussian due to the projection step, the same query complexity bound \eqref{eq:sampling-log-cubed} can be obtained as a high-probability result when initialized with unprojected standard normal $\widetilde X_0=Z$.} $\widetilde X_0=\Pi_B(Z)\triangleq\min(1,B/\|Z\|)\,Z$ with $Z\sim N(0,I_d)$. The output $\widehat X$ satisfies $W_2(\mathcal{L}(\widehat X),\pi)\le\varepsilon$ using at most
\begin{equation}
    N\le C\log^3\!\Bigl(\frac{e\kappa(d+1)}{\varepsilon}\Bigr)\sqrt{d\log(\kappa+1)+\log\Bigl(\frac{e\kappa(d+1)}{\varepsilon}\Bigr)}
    =O\Bigl(\sqrt d\,\log^{3.5}\frac{\kappa d}{\varepsilon}\Bigr)
    \label{eq:sampling-log-cubed}
\end{equation}
exact score evaluations, where $C$ is a universal constant.
\end{theorem}
The error splits into the smoothing bias, the projection of the initialization, and the collocation error, each at most a constant fraction of $\varepsilon$. The inverse width contributes the square-root factor and one logarithm, and the accuracy term contributes the remaining $\log^2$ (\cref{app:log-concave complexity}). Unlike Zoloflow, this result uses a generic solver and thus isolates the benefit of the elliptic schedule. The $\sqrt d$ factor stems from $D$ and is absent from the Gaussian width in~\eqref{eq:lc-width-upper}. Whether it is intrinsic to $\mathsf{P}_\kappa$ remains open.

\section{Experiments}
\label{sec:experiments}

\Cref{fig:frond-main,fig:gauss-r} illustrate the elliptic schedule on a 2D target and Zoloflow on Gaussians; \cref{app:Gaussian-experiments,app:logconcave-experiments,app:toy2d} give further synthetic comparisons. Here, we test the elliptic schedule on image generation.
\paragraph{Image generation.}

\begin{figure}
    \centering
        \includegraphics[width=1\linewidth]{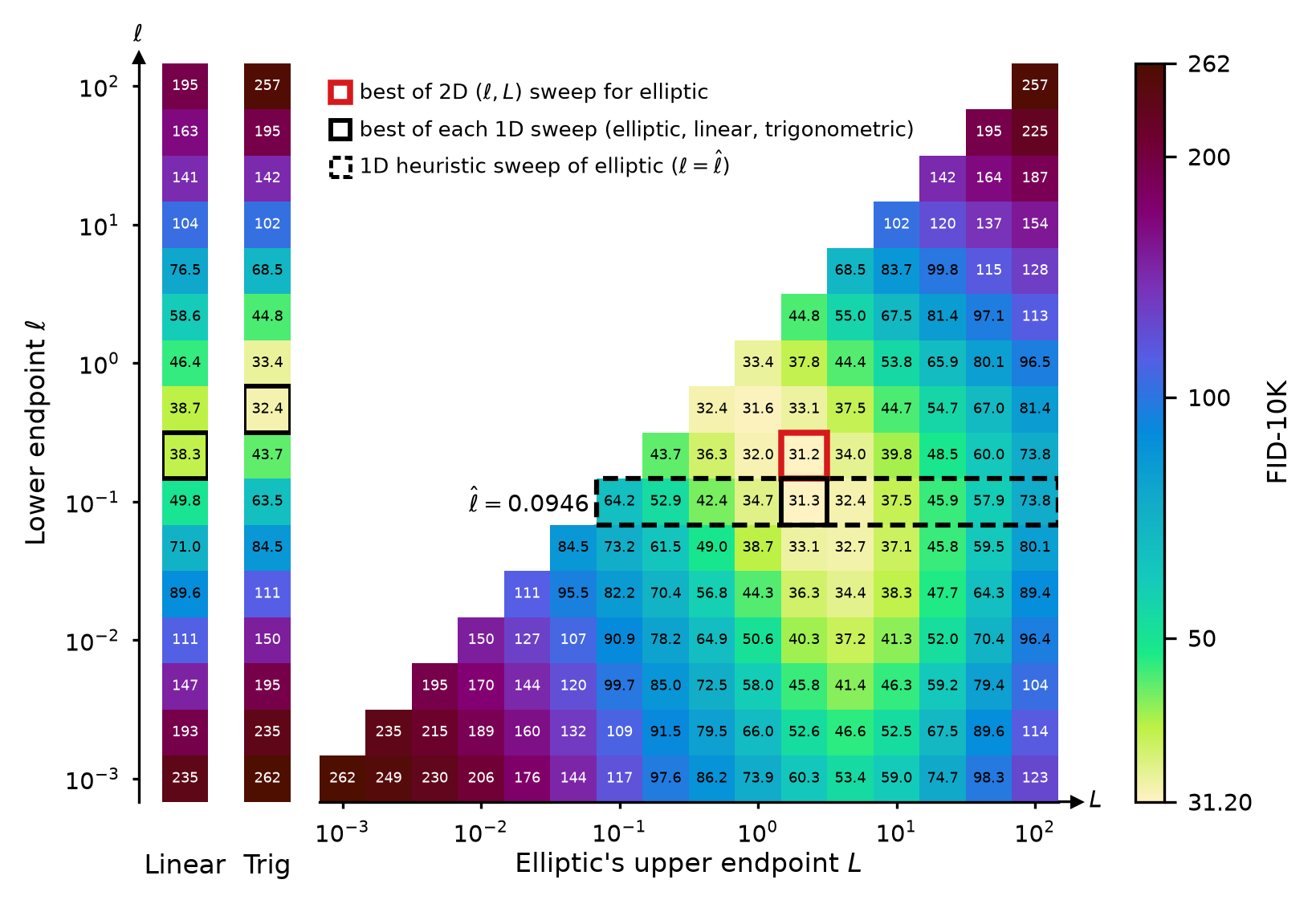}
        \caption{\textbf{FID-10K of hyperparameter sweeps of schedules with 4 oracle calls on SiT-XL/2.} \emph{Right:} the full 2D sweep over the endpoints $(\ell, L)$ of the target-rescaled elliptic schedule, with the best configuration boxed in red. \emph{Left:} 1D sweeps over $\ell$ for the linear and trigonometric schedules. For an equal-budget comparison, we also restrict the elliptic schedule to a 1D sweep over $L$ with $\ell$ fixed to our heuristic $\hat\ell$ (dashed box). The best configuration of each 1D sweep is boxed in black.}
    \label{fig:sit-heatmap-nfe4}
\end{figure}

We generate images using two pretrained models: SiT-XL/2 \citep{ma2024sit}, a 675M-parameter model trained on ImageNet to generate class-conditional images,\footnote{\url{https://github.com/willisma/SiT}.} and Stable Diffusion 3.5 Medium (SD 3.5M), a 2.5B-parameter text-to-image model.\footnote{\url{https://huggingface.co/stabilityai/stable-diffusion-3.5-medium}.} Both models learn the velocity field of the linear interpolant, which we convert to other schedules via \citet[Prop.~3.1]{chen2025lipschitz}. We set the classifier-free guidance (CFG) scales to $1.5$ for SiT-XL/2 and to $4.5$ for SD 3.5M. Since the elliptic finite-difference discretization of Zoloflow is specific to Gaussian targets, we use DPM-Solver++(2M) \citep{lu2022dpmsolverpp}, a commonly used solver described in \cref{app:catalog-of-discretizations}.

\Cref{fig:sit-heatmap-nfe4} shows FID-10K scores for SiT-XL/2 using target-rescaled linear, trigonometric, and elliptic schedules with varying hyperparameters (see \cref{sec:two-param-family,app:target-rescaling}). For the elliptic schedule, we set \(\ell=\hat\ell\) to the reciprocal of the \(99.5\)th percentile of the empirical covariance spectrum of the SiT-XL/2 training set (\Cref{fig:latent-cov-spectrum}), and we sweep over \(L\). This heuristic is near-optimal for SiT-XL/2, suggesting that a 1D sweep can suffice when target samples are available. Moreover, \cref{fig:sit-heatmap,fig:sd3-heatmap} suggest that sensitivity to \(\ell\) and \(L\) decreases with the number of oracle calls. Thus, when target samples are unavailable, a 2D sweep at a low oracle budget may identify hyperparameters that perform well at higher budgets, too. We explore this in \Cref{tab:fid50k,fig:sd3-heatmap} by fixing $\ell$ to its best value at 4 oracle calls of SD~3.5M and sweeping over \(L\) at \(8\) and \(16\) calls.

\begin{table}[t]
    \centering
    \small
    \setlength{\tabcolsep}{4.5pt}
    \renewcommand{\arraystretch}{0.95}
    \caption{\textbf{FID-50K by schedule and oracle budget on SiT-XL/2 and SD 3.5M.} Hyperparameters were tuned for SiT-XL/2 using FID-10K and for SD~3.5M using FID-5K (\cref{fig:sit-heatmap,fig:sd3-heatmap}). Linear and trigonometric schedules only have the hyperparameter $\ell$. The 1D heuristic elliptic sweep uses no more trials than linear or trigonometric. The 2D elliptic sweep has a larger tuning budget. Best per model and budget is bold, and second best is boxed.}
    \label{tab:fid50k}
    \vspace{0.4em}
    \setlength{\aboverulesep}{0pt}
    \setlength{\belowrulesep}{0pt}
    \setlength{\extrarowheight}{1.5pt}
    \setlength{\fboxsep}{1.5pt}
    \begin{tabular}{|c|l|ccc||ccc|}
    \toprule
    \rule{0pt}{3ex} & & \multicolumn{3}{c||}{SiT-XL/2} & \multicolumn{3}{c|}{SD 3.5M} \\
    \cline{3-8}
    \rule{0pt}{2.5ex}\emph{\# Oracle Calls} & \emph{Schedule} & $\ell$ & $L$ & FID-50K$\downarrow$ & $\ell$ & $L$ & FID-50K$\downarrow$ \\
    \midrule
     \multirow{4}{*}{4} & Linear & 0.22 & --- & 32.36 & 0.032 & --- & \fbox{18.85} \\
      & Trigonometric & 0.46 & 0.46 & 27.34 & 0.032 & 0.032 & 26.31 \\
      & Elliptic (\textbf{ours}) -- 1D heuristic & 0.095 & 2.15 & \textbf{25.50} & --- & --- & --- \\
      & Elliptic (\textbf{ours}) -- 2D sweep & 0.22 & 2.15 & \fbox{25.56} & 0.0032 & 1 & \textbf{16.79} \\
    \midrule
     \multirow{4}{*}{8} & Linear & 0.1 & --- & 6.07 & 0.32 & --- & 11.76 \\
      & Trigonometric & 0.46 & 0.46 & 4.52 & 0.1 & 0.1 & 15.24 \\
      & Elliptic (\textbf{ours}) -- 1D heuristic & 0.095 & 1 & \fbox{4.10} & 0.0032 & 100 & \fbox{11.56} \\
      & Elliptic (\textbf{ours}) -- 2D sweep & 0.022 & 2.15 & \textbf{3.85} & 0.01 & 100 & \textbf{10.71} \\
    \midrule
     \multirow{4}{*}{16} & Linear & 0.1 & --- & 2.80 & 1 & --- & \fbox{10.82} \\
      & Trigonometric & 0.22 & 0.22 & 2.38 & 1 & 1 & 11.76 \\
      & Elliptic (\textbf{ours}) -- 1D heuristic & 0.095 & 1 & \fbox{2.35} & 0.0032 & 10000 & 12.05 \\
      & Elliptic (\textbf{ours}) -- 2D sweep & 0.022 & 2.15 & \textbf{2.27} & 0.032 & 3160 & \textbf{9.93} \\
    \bottomrule
    \end{tabular}
    \vspace{-1em}
\end{table}

\Cref{tab:fid50k} shows the largest gains from the elliptic schedule at low oracle budgets. As the budget grows, discretization error decreases and all schedules approach the same limiting sampling performance.
\Cref{fig:front-page-schedules} displays some images generated by the best-performing rescaled linear, trigonometric, and elliptic schedules at 4 oracle calls of SD 3.5M. 
Setup details, the list of schedules, and further results are in \cref{app:imgage-generation-experiments,app:schedule-catalog}.

\section{Discussion}
We designed interpolant schedules and discretizations jointly through two complementary lenses, minimax rational approximation and complex-analytic regularity, which lead to the same elliptic schedule. For Gaussian targets, Zoloflow attains the minimax query-complexity rate among oblivious score-span algorithms using exact interpolant scores. Its discretization is tailored to Gaussian structure, and we do not identify an optimal discretization for log-concave targets. Our log-concave guarantees are upper bounds without matching lower bounds, so it remains open whether the $\sqrt{d}$ factor is necessary. Finally, for non-log-concave targets, a practical limitation is the cost of selecting the elliptic schedule parameters when training-data covariance statistics are unavailable. On SD 3.5M, the full sweep of 153 hyperparameter configurations at four oracle calls took approximately 3.1 hours on eight H200 GPUs. Once selected, a schedule can be reused across samples. The schedule parameters require heuristic selection or empirical tuning, so the value of this investment depends on how many images will be generated. 

\section{AI Usage}
Generative AI was not used for ideation. In particular, the optimal schedule and solver for Gaussian targets were developed without it. We used AI tools to help write proofs and suggest mathematical tools, to search the literature, to generate synthetic target distributions, to write and proofread code (verified by manual review and unit tests), and to edit wording. We reviewed all AI-assisted content and take full responsibility for the paper.

\subsubsection*{Acknowledgments}
Cole Becker\ is supported by the National Science Foundation Graduate Research Fellowship Program under Grant No.~DGE-2141064. Cole Becker and Taylan Kargin \ are supported by an Amazon grant.

\bibliography{Zoloflow_ArXiv}
  \bibliographystyle{plainnat}

\newpage
\appendix

\begin{center}
    \huge Appendix
\end{center}

\etocdepthtag.toc{appendix}
\etocsettagdepth{mainbody}{none}
\etocsettagdepth{appendix}{subsection}
\etocsettocstyle{}{}
\begingroup\footnotesize
\setlength{\parskip}{0pt}
\setlength{\baselineskip}{10pt}
\tableofcontents
\endgroup
\newpage

\section{Organization}
\label{app:organization}
The appendices are grouped by the main-text section they support.
\begin{enumerate}
    \item \emph{Background.} \Cref{app:jacobi} reviews the Jacobi elliptic functions and \cref{app:gaussian-diag} the eigenmode decomposition of Gaussian targets.
    \item \emph{The elliptic schedule (\cref{sec:elliptic-si}).} \Cref{sec:proofJacobi} proves that the Jacobi amplitude is the optimal phase, \cref{app:approaches-for-handling-arbitrary-spectra} handles arbitrary spectra and nonzero means, \cref{app:cov-heuristic} gives the covariance heuristic for $(\bar\lambda,\kappa)$, and \cref{app:schedules} relates the elliptic schedule to other interpolant schedules.
    \item \emph{Gaussian sampling (\cref{sec:Gaussian-sampling}).} \Cref{app:additional-discretizations} derives the elliptic finite-difference rule and catalogs standard solvers, \cref{app:zolotarev} collects Zolotarev's results, \cref{app:algs-are-ratfuncs} shows that algorithms in $\mathcal A_r$ act on Gaussians as rational functions, and \cref{app:compelxity-results-zoloflow} proves \cref{thm:zoloflow-Gaussian-bound}.
    \item \emph{Log-concave sampling (\cref{sec:logconcave}).} \Cref{app:log-concave holomorphic,app:log-concave complexity} prove \cref{thm:holomorphic strip width,thm:log concave complexity}.
    \item \emph{Experiments (\cref{sec:experiments}).} \Cref{app:experiments} gives the setup and full results.
\end{enumerate}

\section{Jacobi Elliptic Functions}
\label{app:jacobi}

The Jacobi elliptic functions are elliptic generalizations of the trigonometric and hyperbolic functions. Like all elliptic functions, they are meromorphic and doubly periodic on the complex plane. On the real line, they have a natural interpretation as sines and cosines under a nonlinear reparameterization of angle, given by the \emph{Jacobi amplitude} (inverse of the elliptic integral of the first kind). The Jacobi elliptic functions appear throughout mathematics and physics, particularly in classical mechanics (e.g., pendulum, rigid body rotations); see e.g. \cite{lawden2013elliptic} and references therein. They have been implemented in almost every computer language -- in particular, in Python they can be computed with SciPy via $\texttt{scipy.special.ellipj}$, which simultaneously provides the functions $\cn, \sn, \dn$ and the Jacobi amplitude $\am$.  

\subsection{Basic notions}

\subsubsection{Elliptic functions in terms of sine and cosine}

At their simplest level the Jacobi elliptic functions are an extension of the usual trigonometric functions, using the geometry of the ellipse. 
They are indexed by a \emph{parameter}\footnote{There are two different conventions for Jacobi elliptic functions, either using the parameter $m$ or the \emph{elliptic modulus} $k =\sqrt{m}$. We use the former notation since it is more common in modern software, and provides slightly cleaner notation.} $m\in [0,1)$. For $0\le m<1$, the \emph{incomplete elliptic integral of the first kind} is
\begin{equation}\label{eq:ell-F}
F(\varphi\mid m)\triangleq\int_0^\varphi\frac{\dif\theta}{\sqrt{1-m\sin^2\theta}},\qquad K(m)\triangleq F(\pi/2\mid m).
\end{equation}
The \emph{incomplete} integral $F:\varphi\mapsto u$ maps a variable angle $\varphi$ on a circle to elliptic time $u$. The \emph{complete} integral $K(m)$ is the time for a quarter-period; a full period takes $4K(m)$. Its complementary quarter-period is $K(1-m)$. When $m=0$, then $K(m)=\pi/2$, which is exactly the quarter period of trigonometric functions.

For every fixed $m$, since $F$ is strictly increasing for real $\varphi$, its real inverse is well-defined: the \emph{Jacobi amplitude}.
\begin{equation}\label{eq:ell-am}
u=F(\varphi\mid m)\quad\Longleftrightarrow\quad\varphi=\am(u\mid m).
\end{equation}
Here $u$ is an elliptic time coordinate, while $\varphi$ is an ordinary angular parameter. In particular, the amplitude is an angle, not the `$y$'-coordinate of a curve. The main three Jacobi elliptic functions are defined as
\begin{equation}\label{eq:ell-basic}
\boxed{\ \sn(u\mid m)=\sin\varphi,\qquad \cn(u\mid m)=\cos\varphi,\qquad \dn(u\mid m)=\sqrt{1-m\sin^2\varphi}.\ }
\end{equation}
where $\varphi \triangleq \am(u\mid m)$.

The square root is positive for real $u$. Differentiating the inverse relation gives
\begin{equation}\label{eq:ell-speed}
\frac{\dif u}{\dif\varphi}=\frac1{\sqrt{1-m\sin^2\varphi}},\qquad \frac{\dif\varphi}{\dif u}=\dn(u\mid m)\in[\sqrt{1-m},1].
\end{equation}
Thus the elliptic clock $u$ stretches angular intervals most near $\varphi=\pi/2\pmod\pi$. A trajectory moving uniformly in $u$ advances slowly through those angles. When $m=0$, there is no stretching: $u=\varphi$. Importantly, $\dn$ is also exactly the angular speed $\frac{d\varphi}{du}$ of the angle as a function of the elliptic clock $u$. Below we give some plots of $u$ and $\varphi$ for varying parameters $m$. In particular note how for larger $m$ close to the quarter period, the angle $\varphi$ moves much slower relative to the elliptic clock $u$. Intuitively, for very flat ellipses, this ensures that movement towards the end of the quarter-period (around the top of the ellipse) is not too fast.
\begin{figure}[htbp]
    \centering
    \includegraphics[width=\textwidth,height=.73\textheight,keepaspectratio]{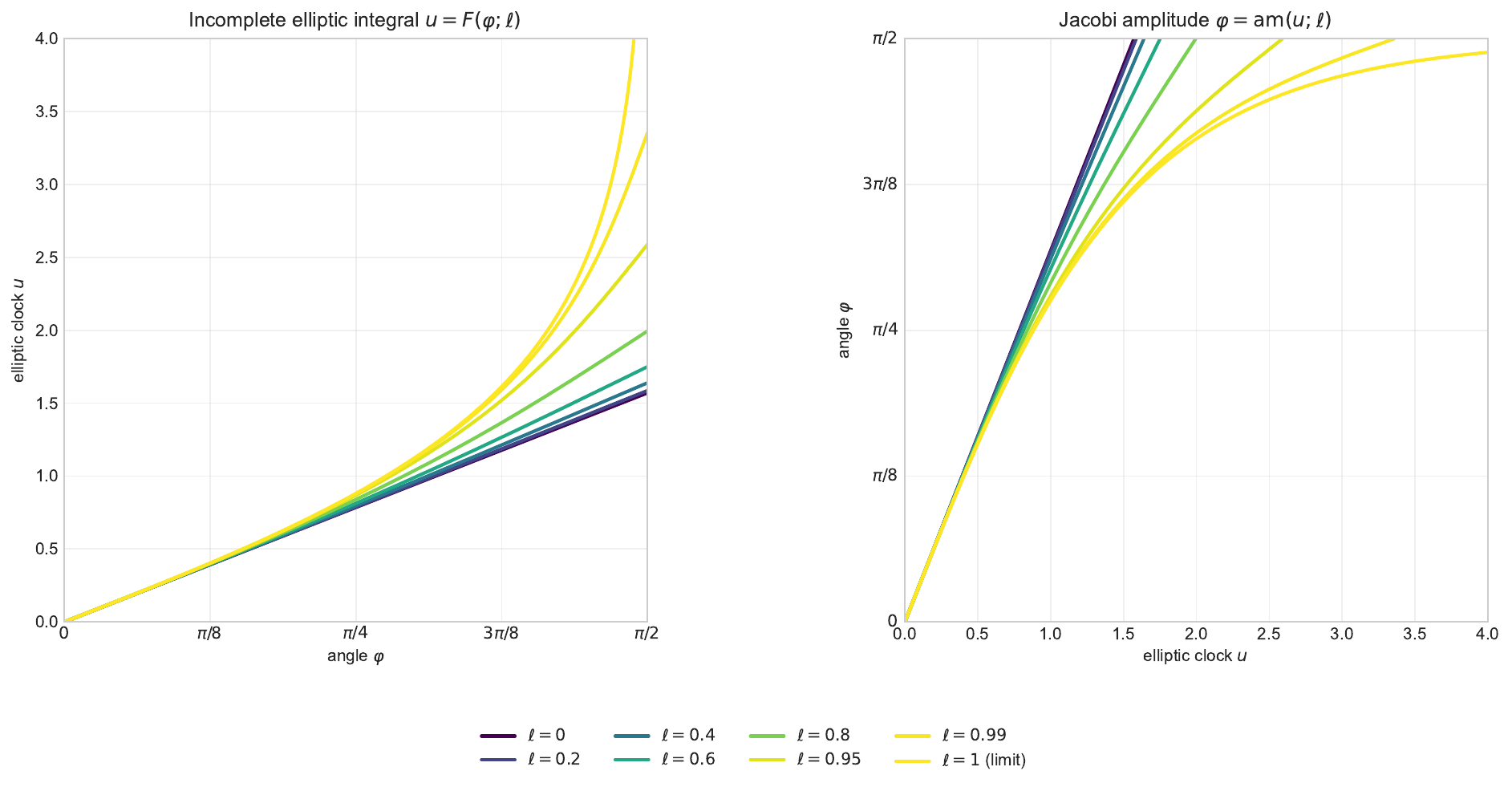}
    \caption{Angular and elliptic clocks $(\varphi, u)$ for varying parameters $m$}
    \label{fig:elliptic_clocks}
\end{figure}

\begin{figure}[htbp]
    \centering
    \includegraphics[width=\textwidth,height=.73\textheight,keepaspectratio]{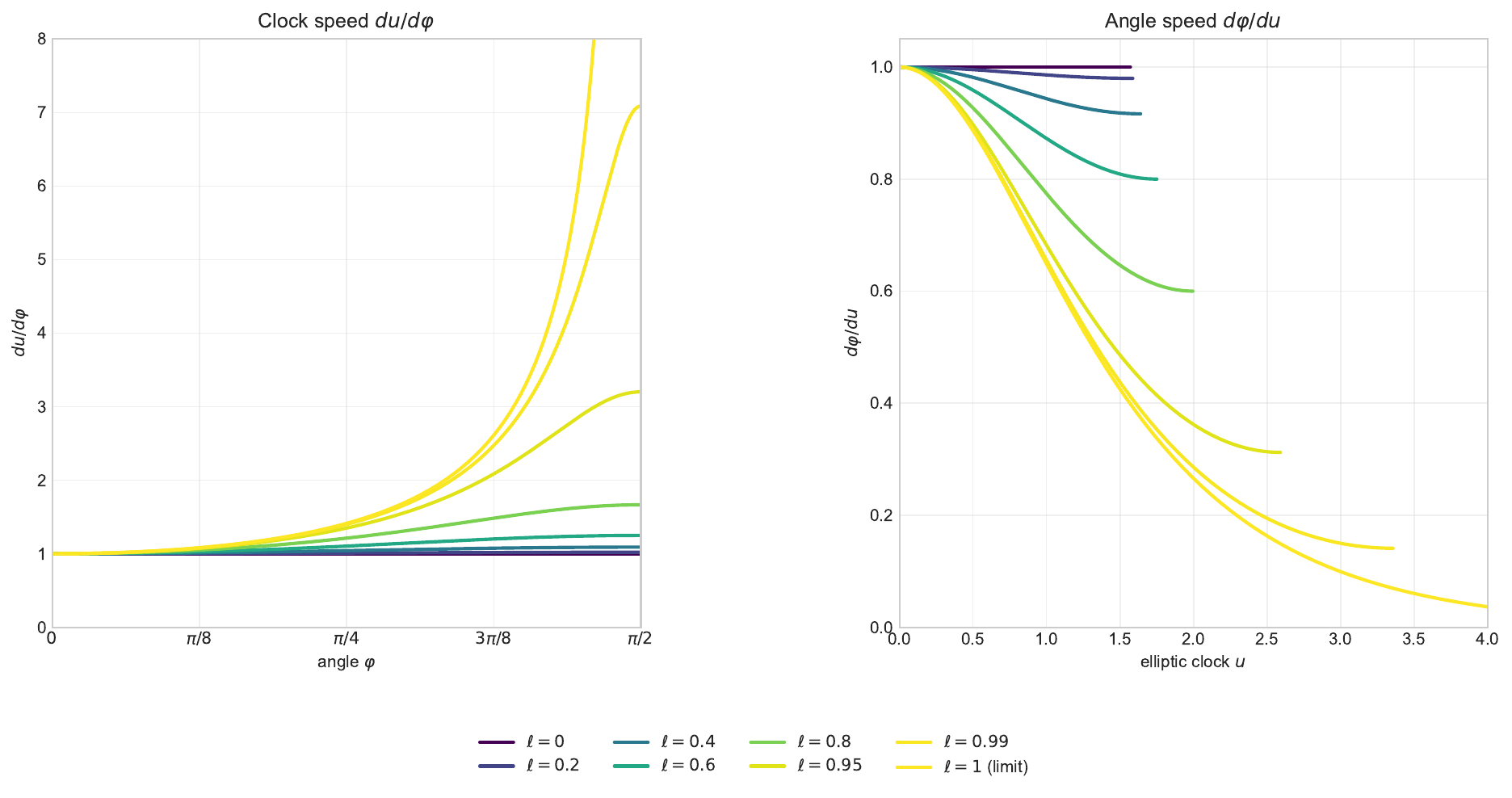}
    \caption{Angular and elliptic clock speeds $(\varphi, u)$ for varying parameters $m$}
    \label{fig:elliptic_speeds}
\end{figure}

\subsubsection{Limiting Cases $m \to 0,\ m \to 1$}
At the $m = \{0,1\}$ endpoints the Jacobi amplitude function $\am(u\mid m)$ evaluates to
\begin{equation}
    \am(u\mid0) = u, \text{ and } \am(u\mid1) = \arcsin(\tanh u)
\end{equation}
allowing us to reduce the Jacobi functions at these parameter endpoints in the corresponding table.
\begin{table}[htbp]
\centering
\renewcommand{\arraystretch}{1.5}
\begin{tabular}{@{}lcc@{}}
\toprule
Function & $m=0$ & $m \rightarrow 1$ \\
\midrule
$\varphi(u)=\operatorname{am}u$ & $u$ & $\arcsin(\tanh u)$ \\
$\sn u=\sin(\operatorname{am}u)$ & $\sin u$ & $\tanh u$ \\
$\cn u=\cos(\operatorname{am}u)$ & $\cos u$ & $\sech u$ \\
$\dn u=\sqrt{1-m\sin^2(\operatorname{am}u)}$ & $1$ & $\sech u$ \\
\bottomrule
\end{tabular}
\caption{Limiting cases of the Jacobi elliptic functions at $m=0$ and $m=1$.}
\label{tab:elliptic-limits}
\end{table}
Here $\tanh$ and $\sech$ denote the hyperbolic tangent and hyperbolic secant.

We show the standard Jacobi elliptic functions for varying active parameter $m$ along the quarter period. As the quarter period has infinite limit $\lim_{m\to 1} K(m) = \infty$, we plot these functions in both the real and normalized elliptic time $u, \text{ and } t = u/K(m)$, making explicit these endpoint conditions. With the normalized clock parameter $t = u/K(m)$, these functions $\cn(t), \ \sn(t)$ are exactly our proposed elliptic schedules $\alpha_t, \ \beta_t$. 

\begin{figure}[htbp]
    \centering
    \includegraphics[width=\textwidth,height=.73\textheight,keepaspectratio]{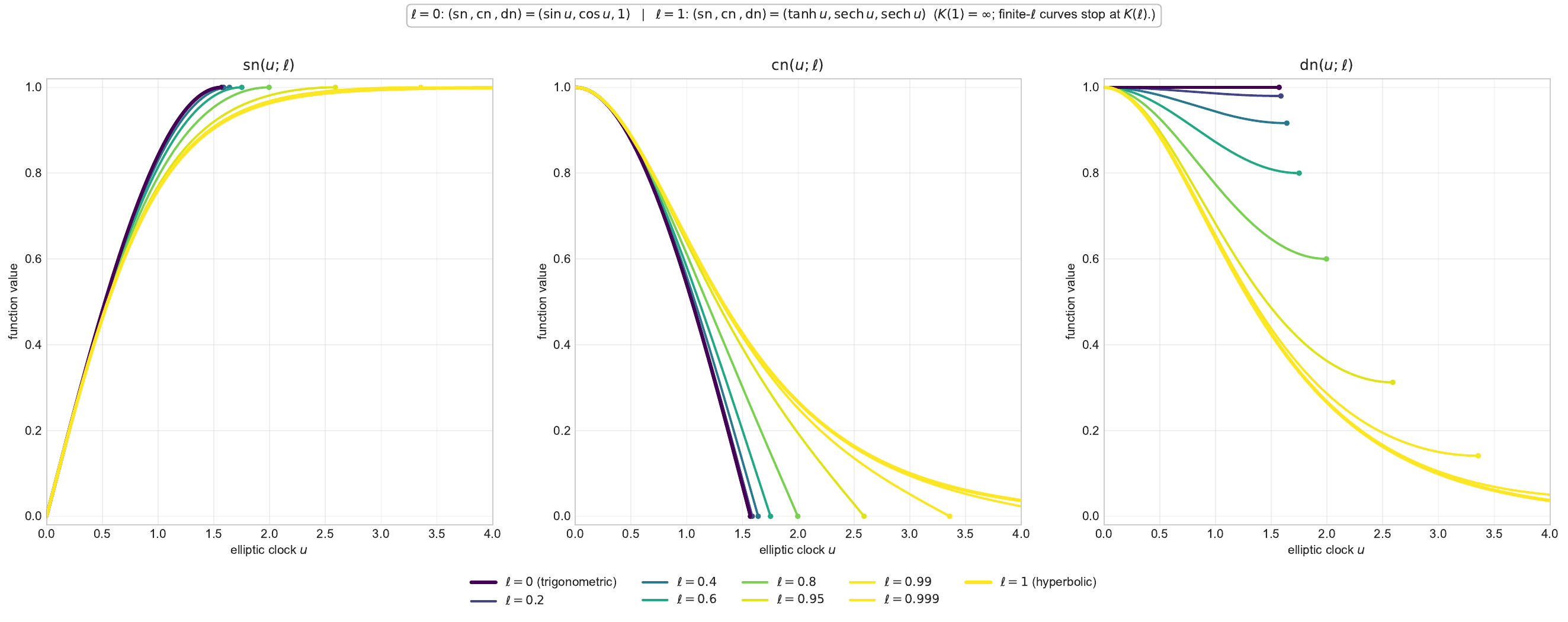}
    \caption{Elementary Jacobi Elliptic functions under raw clock $u$}
\end{figure}

\begin{figure}[htbp]
    \centering
    \includegraphics[width=\textwidth,height=.73\textheight,keepaspectratio]{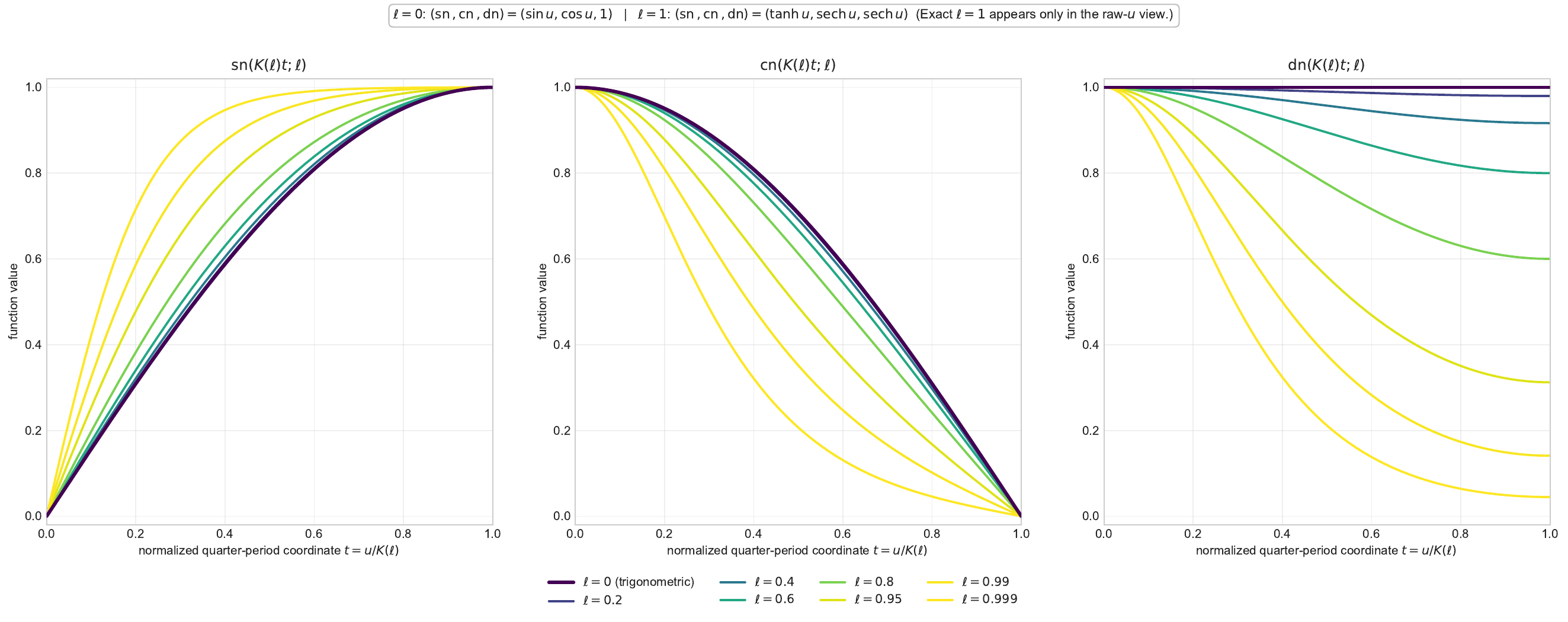}
    \caption{Elementary Jacobi Elliptic functions under normalized clock $t = u/K(m)$}
\end{figure}

\subsubsection{The elliptic family and its algebraic identities}

There are other Jacobi elliptic functions, which are ratios or reciprocals of the principal three. For example,
\begin{equation}
\jsc=\frac{\sn}{\cn}=\tan\varphi,\quad \jcs=\frac{\cn}{\sn}=\cot\varphi,\quad
\operatorname{sd}=\frac{\sn}{\dn},\quad \operatorname{cd}=\frac{\cn}{\dn},\quad \operatorname{ns}=\frac1{\sn}.
\end{equation}
There are twelve functions in total. The mnemonic is to regard the letters $s,c,d,n$ as $\sn,\cn,\dn,1$ and to interpret two distinct letters as a quotient. Their basic identities are
\begin{equation}\label{eq:ell-identities}
\boxed{\ \sn^2+\cn^2=1,\qquad \dn^2+m\sn^2=1,\qquad \dn^2=(1-m)+m\cn^2.\ }
\end{equation}
Arguments $(u\mid m)$ are suppressed here and below. These are meromorphic identities, valid wherever the displayed expressions are finite.

\subsection{Elliptic functions as differential equations}
It is often common, and perhaps more natural to define the Jacobi elliptic functions through a system of first-order differential equations. Indeed $\sn, \cn, \dn$ may be defined as the unique solutions to the following initial value problem.
\begin{equation}
\begin{aligned}
    \dot{\sn} &= \cn\,\dn, \quad & \sn(0) = 0\\
    \dot{\cn} &= -\sn\,\dn, \quad & \cn(0) = 1\\
    \dot{\dn} &= -m\sn\,\cn, \quad & \dn(0) = 1
\end{aligned}
\end{equation}
where the arguments $(u\mid m)$ are suppressed and dots denote $u$-derivatives. A nice consequence of this interpretation is one gets for free the algebraic identities \eqref{eq:ell-identities}. 

\subsection{Complex structure: poles, periods, and a torus} \label{app:complex elliptic}

\subsubsection{Complex extensions of elliptic functions and plots}

With active quarter-period $K(m)$ and complementary quarter-period $K(1-m)$, generating periods and representative poles are
\begin{center}
\begin{tabular}{@{}lcc@{}}\toprule
Function & Generators of its period lattice & Example pole\\\midrule
$\sn(u\mid m)$ & $4K(m),\ 2iK(1-m)$ & $iK(1-m)$\\
$\cn(u\mid m)$ & $4K(m),\ 2K(m)+2iK(1-m)$ & $iK(1-m)$\\
$\dn(u\mid m)$ & $2K(m),\ 4iK(1-m)$ & $iK(1-m)$
\\\bottomrule
\end{tabular}
\end{center}
For example,
\begin{equation}
\sn(u+4K(m)\mid m)=\sn(u\mid m),\qquad \sn(u+2iK(1-m)\mid m)=\sn(u\mid m).
\end{equation}

\begin{figure}[htbp]
    \centering
    \includegraphics[width=\textwidth,height=.73\textheight,keepaspectratio]{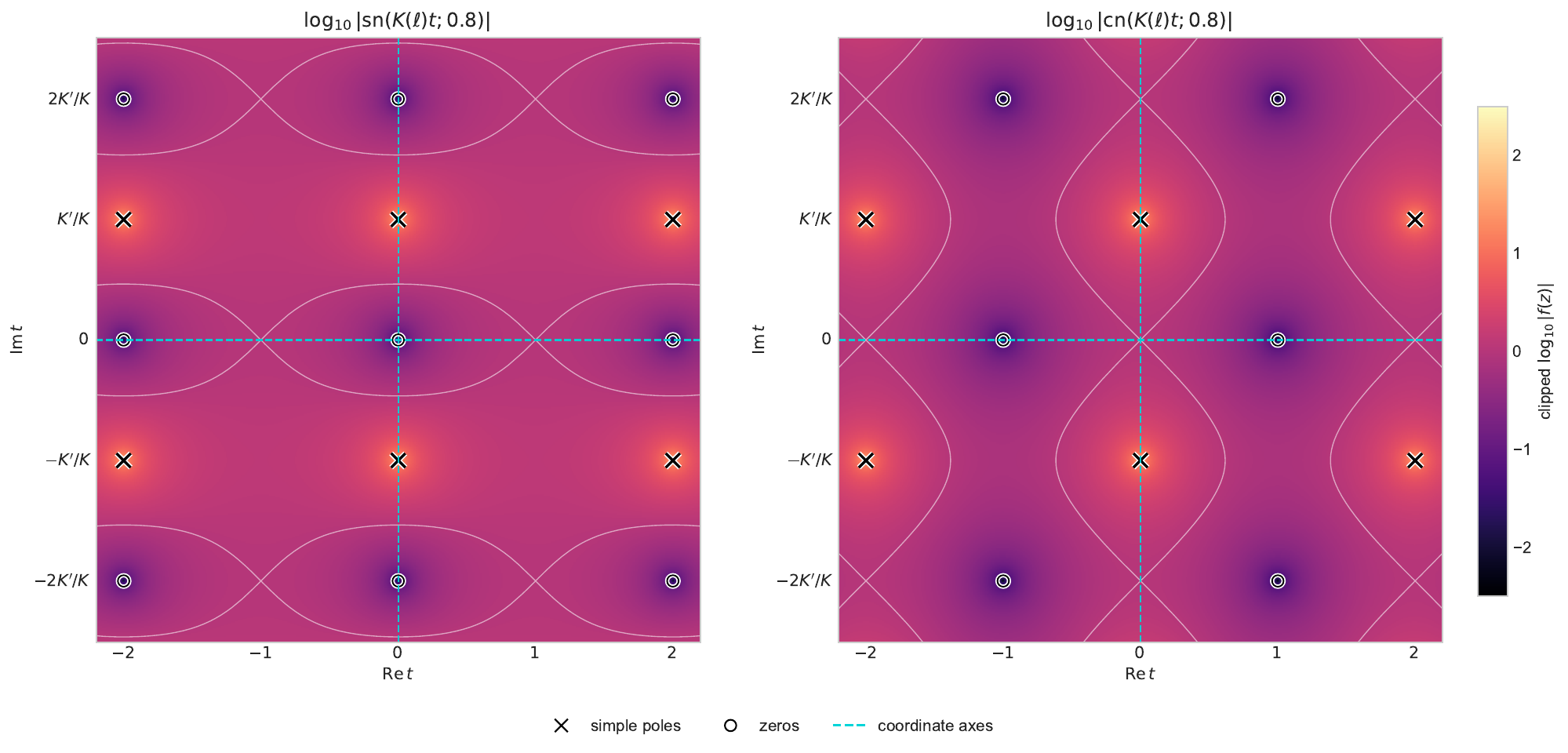}
    \caption{Plots of holomorphic extensions to $|\sn|$ and $|\cn|$ in normalized time $t = \frac{u}{K(m)}$, with parameter $m=0.64$ (modulus $\ell=\sqrt m=0.8$ in the plot labels). Zeros are labeled with ``$\circ$" and poles are labeled with ``$\times$." The white lines represent
    $|f(z)|=1$ level sets. Lighter indicates $|f(z)|$ large and darker indicates $|f(z)|$ small.}
    \label{fig:holomorphic_sn_cn}
\end{figure}

The principal three share the simple-pole locations
\begin{equation}\label{eq:ell-poles}
u=2jK(m)+(2n+1)iK(1-m),\qquad j,n\in\Z.
\end{equation}
Consequently, all three are holomorphic in the horizontal strip $|\operatorname{Im}u|<K(1-m)$. The functions have different period lattices: in particular, $2K(m)$ is a \emph{sign-changing shift}, not a period, of $\sn$ and $\cn$.
\paragraph{In normalized $t = \frac{u}{K(m)}$ coordinates.} As in the paper, it may be more natural to visualize what happens. In Figure~\ref{fig:holomorphic_sn_cn}, we plot $\sn(K(m)t\mid m), \ \cn(K(m)t\mid m)$ for complex valued $t\in \mathbb{C}$ for $m=0.8^2=0.64$. Again, the simple-pole locations are shared, and now both $\sn(K(m)t\mid m)$ and $\cn(K(m)t\mid m)$ are holomorphic on the strip $|\Im{t} |< \frac{K(1-m)}{K(m)}$ sharing poles at
\begin{equation}
    t=2j+(2n+1)i\frac{K(1-m)}{K(m)},\qquad j,n\in\Z.
\end{equation}

\textbf{Paper notation}
\begin{equation}\label{eq:ell-paper}
\begin{aligned}
    m&\triangleq1-\kappa^{-1}, \\
    1-m&=\kappa^{-1},\\
\alpha_t=\cn(K(m)t\mid m),&\quad \beta_t=\sn(K(m)t\mid m).
\end{aligned}
\end{equation}
\textbf{The active parameter in these functions is $m=1-\kappa^{-1}$.} Thus the associated ellipse has eccentricity $\sqrt m$ and semiaxis ratio $\sqrt{1-m}$. Equivalently, the schedule is a cosine--sine (trig) schedule with a nonlinear angular clock:
\begin{equation}
\varphi_t=\am(K(m)t\mid m),\qquad (\alpha_t,\beta_t)=(\cos\varphi_t,\sin\varphi_t),\qquad
\dot\varphi_t=K(m)\dn(K(m)t\mid m).
\end{equation}
It follows directly that $(\alpha_0,\beta_0)=(1,0)$, $(\alpha_1,\beta_1)=(0,1)$, and $\alpha_t^2+\beta_t^2=1$. For $\kappa=1$, $m=0$ and the schedule is exactly $(\cos(\pi t/2),\sin(\pi t/2))$.

Its signal-to-noise ratio and the positive Zolotarev shift parameters are generated by the same function:
\begin{equation}
\frac{\beta_t^2}{\alpha_t^2}=\jsc^2(K(m)t\mid m),\qquad
\gamma_i=\jsc^2\!\left(\frac{iK(m)}{2r+1}\mid m\right),\quad i=1,\ldots,2r.
\end{equation}
In the inverse-square-root product, $-\gamma_{2j}$ and $-\gamma_{2j-1}$ are the zero and pole locations, respectively. The $\gamma_i$ are \emph{positive shifts}, not the locations themselves~\cite{nakatsukasa2016computing}.

\section{Gaussian Diagonalization and Eigenmodes}
\label{app:gaussian-diag}

When both the initial and target distributions are centered Gaussians, the analysis of the SDEs and ODEs simplifies significantly, since the dynamics can be decoupled by working on an eigenbasis. We describe the details below. 

\begin{definition}[Gaussian eigenmodes]

\label{def:eigenmodes}
Let $\pi=\mathcal N(0,\Lambda^{-1})$, where $\Lambda \succ 0$, and fix an orthogonal eigendecomposition
\[
\Lambda=U\operatorname{diag}(\lambda_1,\ldots,\lambda_d)U^\top,
\qquad U=[u_1,\ldots,u_d], \qquad U^\top U=I_d.
\]
For each $k\in\{1,\ldots,d\}$, the \emph{$k$-th eigenmode} of $\pi$ is the line $\operatorname{span}\{u_k\}$, whose precision is $\lambda_k$.
Any $x\in\mathbb R^d$ can be decomposed into eigenmodes as
\[
x^{(k)}\triangleq u_ku_k^\top x,
\qquad x=\sum_{k=1}^d x^{(k)}.
\]
\end{definition}

Given a scalar function $f$ defined on the spectrum $\operatorname{spec}(\Lambda)$, there is a natural matrix extension given by $f(\Lambda) \triangleq U\operatorname{diag}\!\bigl(f(\lambda_1),\ldots,f(\lambda_d)\bigr)U^\top$. As a linear map, $f(\Lambda)$ acts modewise, since
\[
f(\Lambda)x =\sum_{k=1}^d f(\lambda_k)x^{(k)}.
\]
\emph{Working along an eigenmode} means projecting onto its eigendirection, so that multiplication by $f(\Lambda)$ becomes scalar multiplication by $f(\lambda_k)$. If the specific mode $\lambda_k$ is understood from the context, we may write $\lambda=\lambda_k$, $x_\lambda=x^{(k)}$ and $\lambda$-eigenmode for short.

\paragraph{Gaussian scores and drifts.}
For a centered Gaussian target $\pi=\mathcal N(0,\Lambda^{-1})$, the interpolant \eqref{eq:stochastic-interpolant} has law $p_t=\mathcal N(0,\Sigma_t)$, where $\Sigma_t=\alpha_t^2I_d+\beta_t^2\Lambda^{-1}$. Its score and probability-flow drift are
\begin{equation}
\label{eq:Gaussian-score-drift}
    s_t(x)=-\Sigma_t^{-1}x,
    \qquad
    b_t(x)=\bigl(\alpha_t\dot\alpha_t I_d+\beta_t\dot\beta_t\Lambda^{-1}\bigr)\Sigma_t^{-1}x,
\end{equation}
and since the matrices commute, both maps preserve the eigenmodes. In the notation of Definition~\ref{def:eigenmodes},
\begin{equation}
\label{eq:Gaussian-eigenmode-score-drift}
    s_t(x_\lambda)=-\frac{\lambda}{\lambda\alpha_t^2+\beta_t^2}\,x_\lambda,
    \qquad
    b_t(x_\lambda)=g_\lambda(t)x_\lambda,
\end{equation}
where the \emph{eigenmode velocity coefficient} is
\begin{equation}
    g_\lambda(t)\triangleq
    \frac{\lambda\alpha_t\dot\alpha_t+\beta_t\dot\beta_t}{\lambda\alpha_t^2+\beta_t^2}
    =\frac12\frac{d}{dt}\log\!\left(\lambda\alpha_t^2+\beta_t^2\right).
\label{eq:Gaussian-eigenmode-coefficient}
\end{equation}
Consequently, each component of the probability-flow solution evolves independently according to $\dot X_\lambda(t)=g_\lambda(t)X_\lambda(t)$. For variance-preserving schedules, these expressions simplify to $g_\lambda(t)=-(\lambda-1)\beta_t\dot\beta_t/(\lambda\alpha_t^2+\beta_t^2)$ and 
$b_t(x_\lambda) = (\dot \beta_t/\beta_t) (x_\lambda + s_t(x_\lambda))$.

\paragraph{Gaussian sampling and the inverse square root.}
A natural approach to sampling from $\mathcal N(0,\Lambda^{-1})$ is to consider $\Lambda^{-\frac{1}{2}}z$, where $z \sim \mathcal N(0,I_d)$ is a standard Gaussian. By the arguments above, this leads to consideration of the scalar function $\lambda \mapsto 1/\sqrt{\lambda}$, where $\lambda$ is in $\operatorname{spec}(\Lambda)$. The inverse square root has a singularity (branch point) at the origin, but it can be well-approximated (e.g., by polynomials or rational functions) on compact intervals on the positive real axis. We are particularly interested in optimal rational approximations (see e.g. \cite{higham2008functions}); see Section~\ref{app:zolotarev}.

\section{Optimality properties of the elliptic schedule}
\label{sec:proofJacobi}
Here we provide a proof of the optimality of the Jacobi amplitude, stated as Proposition~\ref{prop:optimaiAM}. The basic idea is to construct a conformal map from the singularity-free region to the strip.
\begin{tcolorbox}[colebase,colevisual=ColeResult]
\JacobiAM*
\end{tcolorbox}

\begin{proof}
We first rewrite the nonvanishing constraint as an avoidance condition in the complex \(\varphi\)-plane.  For \(\lambda\in[1,\kappa]\),
$1-\bigl(1-\lambda^{-1}\bigr)\sin^2 \varphi=0$
if and only if
$ \sin^2 \varphi = {\lambda}/(\lambda-1)$.
As \(\lambda\) ranges from \(1\) to \(\kappa\), this ranges over the interval $\left[{\kappa}/(\kappa-1),\infty\right)
= [m^{-1},\infty).$
Thus the third condition is equivalent to
$\sin^2\varphi(t)\notin[m^{-1},\infty), \forall t\in S_\gamma$.
Consider the domain $\Omega_m$ given by the connected component containing $[0,\pi/2]$ of 
$\left\{ z\in\C:\sin^2z\notin[m^{-1},\infty) \right\}$.
A standard conformal representation of $\Omega_m$ is given by the
incomplete elliptic integral
\[
F(z\mid m)
=
\int_0^z
\frac{d\zeta}{\sqrt{1-m\sin^2\zeta}}.
\]
Namely, $
    \Psi_m(z)
    \triangleq
    {F(z\mid m)}/{K(m)}
$ is a biholomorphism
$
    \Psi_m:\Omega_m
    \longrightarrow
    S_{\gamma_\star}$, with inverse $\Psi_m^{-1}(w) = \am(K(m)w\mid m)$.
The boundary interpolation conditions are satisfied, since
$ \Psi_m(0)=0$ and $\Psi_m(\pi/2)=1$,  since $F(0\mid m)=0$ and $F(\pi/2\mid m)=K(m)$.

We now prove optimality, using Schwarz-Pick.  Let \(\varphi\) be any other phase satisfying the three conditions of the proposition on \(S_\gamma\).  Since \(S_\gamma\) is
connected, \(\varphi(0)=0\), its image must lie in the component \(\Omega_m\).
Hence $ h\triangleq \Psi_m\circ\varphi $
defines a holomorphic map $
h:S_\gamma\longrightarrow S_{\gamma_\star}.$
The endpoint constraints give $h(0)=0$ and $h(1)=1$.
We now apply the Schwarz--Pick theorem to the two horizontal strips.  Recall that $
z\longmapsto
\tanh\!\left(\frac{\pi z}{4a}\right)$
maps \(S_a=\{z:|\Im z|<a\}\) biholomorphically onto the unit disk.
Consequently, the hyperbolic distance between two points \(x,y\in\R\) on
the center line of \(S_a\) is
$d_{S_a}(x,y)=\frac{\pi}{2a}|x-y|.$
In particular, $d_{S_\gamma}(0,1)={\pi}/{(2\gamma)}$
and $d_{S_{\gamma_\star}}(0,1)
={\pi}/{(2\gamma_\star)}.$
Since holomorphic maps decrease hyperbolic distance, Schwarz--Pick gives $
d_{S_{\gamma_\star}}\bigl(h(0),h(1)\bigr)
\leq
d_{S_\gamma}(0,1).$
Using \(h(0)=0\) and \(h(1)=1\), we obtain
\[
\frac{\pi}{2\gamma_\star}
\leq
\frac{\pi}{2\gamma},
\]
and hence
\[
\gamma\leq\gamma_\star
=
\frac{K(1-m)}{K(m)},
\]
with equality if and only if $\varphi(t)=\am(K(m) t|m)$. Thus no admissible phase can be holomorphic on a wider horizontal strip.

Finally, we verify that the bound is attained.  The inverse of the extremal conformal map is $\varphi_\star(t)
=
\am(Kt\mid m).
$
It is holomorphic on $S_{\gamma_\star}$, 
satisfies the boundary conditions $\varphi_\star(0)=0$, 
$\varphi_\star(1)= \am(K\mid m)=\frac{\pi}{2}$, and $1 - (1-\lambda^{-1} )\sin^2 \varphi_\star(t)$ does not vanish on the strip $S_{\gamma_\star}$ for all $\lambda \in [1,\kappa]$.
Therefore the Jacobi amplitude \(\varphi_\star\) is admissible and attains the maximal strip width
$\gamma_\star = {K(1-m)}/{K(m)}$.
\end{proof}

\section{Probability-Flow Discretizations}
\label{app:additional-discretizations}

\subsection{Derivation of elliptic finite-difference discretization}
\label{sec:derivationcustomdiscretization}

Recall that if $(\alpha_t,\beta_t)$ satisfy $\alpha_t^2 + \beta_t^2 = 1$ (a \emph{variance preserving} schedule), the drift terms in~\eqref{eq:Gaussian-score-drift} can be simplified as
 \begin{equation}
     b_t(x) = \frac{\dot \beta_t }{\beta_t}\left(x + s_t(x)\right).
     \label{eq:ODElogform}
 \end{equation}
Since 
\[
\frac{\dot \beta_t }{\beta_t} = \frac{d}{dt} (\log \beta_t ) = 
\frac{1}{2} \frac{d}{dt} (\log \beta_t^2 )  = 
\frac{1}{2 \beta_t^2} \frac{d}{dt} (\beta_t^2),
\]
we can rewrite the drift term to obtain
\begin{equation}
    b_t(x) = \frac{\dot \beta_t }{\beta_t}\left(x + s_t(x)\right)
     = \frac{1}{2 \beta_t^2} \frac{d}{dt} (\beta_t^2) \left(x + s_t(x)\right) 
     \approx \frac{1}{2 \beta_t^2} \frac{(\beta_{t+(h/2)}^2 - \beta_{t}^2)}{(h/2)} \left(x + s_t(x)\right) 
    \label{eq:ODElogform2}
\end{equation}
where in the last step we used a half-step finite-difference scheme to approximate the derivative. This yields the discretization described in Definition~\ref{def:customdiscretization}.

\subsection{Catalog of Discretizations}
\label{app:catalog-of-discretizations}
\paragraph{Notation.}
For a budget of $r$ score queries,
\begin{align}
    h&\triangleq\frac{2}{2r+1}
    &&\text{grid spacing,} \notag\\
    t_j&\triangleq\frac{2j-1}{2r+1},\quad j\in[r+1]
    &&\text{query times $t_1,\ldots,t_r$ and endpoint $t_{r+1}=1$,}
    \label{eq:shared-query-grid}\\
    x^0&\sim\mathcal N(0,I)
    &&\text{initial state at $t_1$; $x^j$ is the state after $j$ updates,}
    \notag\\
    \hat x_0(x,t)&\triangleq-\alpha_t\,s_t(x),
    \quad
    \hat x_1(x,t)\triangleq\frac{x+\alpha_t^2\,s_t(x)}{\beta_t}
    &&\text{noise and data predictions (one score query each),}
    \label{eq:predictions}\\
    \rho_t&\triangleq\log\frac{\beta_t}{\alpha_t}
    &&\text{half-log-SNR.} \notag
\end{align}

\paragraph{One query per step.}
These methods query once at each of $t_1,\ldots,t_r$ and return $x^r$.
Unless stated otherwise, $j=1,\ldots,r$.
\begin{itemize}
    \item \textbf{Elliptic Finite-Difference}:
    \begin{equation}
        x^j=x^{j-1}
        +\frac{\beta_{t_j+h/2}^2-\beta_{t_j}^2}{\beta_{t_j}^2}
        \bigl(x^{j-1}+s_{t_j}(x^{j-1})\bigr).
        \label{eq:si-zoloflow-fd-appendix}
    \end{equation}

    \item \textbf{Euler:}
    \begin{equation}
        x^j=x^{j-1}+h\,b_{t_j}(x^{j-1}).
        \label{eq:si-euler}
    \end{equation}

    \item \textbf{DDIM}~\citep{song2022denoisingdiffusionimplicitmodels}:
    \begin{equation}
        x^j=\alpha_{t_{j+1}}\hat x_0(x^{j-1},t_j)
        +\beta_{t_{j+1}}\hat x_1(x^{j-1},t_j).
        \label{eq:si-ddim}
    \end{equation}

    \item \textbf{DPM-Solver++(2M)}~\citep{lu2022dpmsolverpp}: let
    $d_j\triangleq\hat x_1(x^{j-1},t_j)$ be the data prediction at step $j$
    and $\delta_j\triangleq\rho_{t_{j+1}}-\rho_{t_j}$ be the step in
    half-log-SNR, where $\rho_t \triangleq \log (\beta_t / \alpha_t)$. For $j=1,\ldots,r-1$,
    \begin{equation}
        x^j=\frac{\alpha_{t_{j+1}}}{\alpha_{t_j}}x^{j-1}
        +\beta_{t_{j+1}}\bigl(1-e^{-\delta_j}\bigr)D_j,
        \qquad
        D_j=\begin{cases}
            d_1, & j=1,\\[2pt]
            \bigl(1+\frac{\delta_j}{2\delta_{j-1}}\bigr)d_j
            -\frac{\delta_j}{2\delta_{j-1}}d_{j-1}, & j\ge2,
        \end{cases}
        \label{eq:si-dpmpp-2m}
    \end{equation}
    and the last step is the DDIM readout $x^r=d_r$, since
    $\delta_r=+\infty$.
\end{itemize}

\paragraph{Multistage methods.}
An $m$-stage method (with $r>m$) takes $M\triangleq\lfloor(r-1)/m\rfloor$
macro-steps on the uniform macrogrid
\begin{equation}
    \tau_n\triangleq t_1+n\Delta,
    \qquad
    \Delta\triangleq\frac{t_r-t_1}{M},
    \qquad n=0,\ldots,M,
    \label{eq:multistage-macrogrid}
\end{equation}
so that $\tau_M=t_r$, with $\tau_n=t_{mn+1}$ when $m\mid r-1$. It then covers
the last interval $[t_r,1]$ with one Euler step,
\begin{equation}
    x^{M+1}=x^M+h\,b_{t_r}(x^M),
    \label{eq:si-closing-step}
\end{equation}
and returns $x^{M+1}$. This uses $mM+1\le r$ queries and never queries the
model at $t=1$, where the drift may blow up. Within a macro-step, write
$x=x^n$, $x'=x^{n+1}$, $\tau=\tau_n$, and $\tau'=\tau_{n+1}$.
\begin{itemize}
    \item \textbf{Heun} ($m=2$):
    \begin{equation}
        k_1=b_\tau(x),
        \qquad
        k_2=b_{\tau'}(x+\Delta k_1),
        \qquad
        x'=x+\frac{\Delta}{2}(k_1+k_2).
        \label{eq:si-heun}
    \end{equation}

    \item \textbf{RK4} ($m=4$):
    \begin{equation}
    \begin{aligned}
        k_1&=b_\tau(x),
        &k_2&=b_{\tau+\Delta/2}\bigl(x+\tfrac{\Delta}{2}k_1\bigr),\\
        k_3&=b_{\tau+\Delta/2}\bigl(x+\tfrac{\Delta}{2}k_2\bigr),
        &k_4&=b_{\tau'}(x+\Delta k_3),\\
        x'&=x+\tfrac{\Delta}{6}(k_1+2k_2+2k_3+k_4).
    \end{aligned}
    \label{eq:si-rk4}
    \end{equation}
\end{itemize}

\section{Zolotarev rational function approximation results}
\label{app:zolotarev}

Let $\mathcal{R}_{p,q}$ denote the real rational functions whose numerator and denominator have degrees at most~$p$ and $q$, respectively. We would like to approximate the function $1/\sqrt{\lambda}$ on the interval $[1,\kappa]$, with an element of $\mathcal{R}_{p,q}$. For $\kappa>1$ and $r\in \mathbb{N}_0$, define the optimal relative approximation error
\begin{equation}
    E_{\kappa,r}\triangleq
    \min_{q\in\mathcal{R}_{r,r}}
    \max_{\lambda\in[1,\kappa]}
    \left|\sqrt{\lambda}\,q(\lambda)-1\right|,
    \label{eq:zolo-minimax app}
\end{equation}
where the minimization is restricted to functions without poles on $[1,\kappa]$. The unique minimizer, denoted by $q_{\kappa,r}$, is the \emph{Zolotarev inverse-square-root approximant}. It is classically known \cite{zolotarev1877application} that there is an explicit expression for $q_{\kappa,r}$ in terms of Jacobi elliptic functions. To present this, we first discuss optimal rational approximations to the sign function.

\paragraph{Sign function.}
Zolotarev's theorem gives the best type-$(2r+1,2r)$ rational approximation $Z(x)$ to $\operatorname{sign}(x)$ on $[-\sqrt{\kappa},-1]\cup[1,\sqrt{\kappa}]$. Since $\operatorname{sign}(x)$ is odd, so is the optimal approximant, and thus it has the form
\[
    Z(x)=Cx\frac{P(x^2)}{Q(x^2)}
        =xq_{\kappa,r}(x^2),
    \qquad \deg P,\deg Q\leq r.
\]
The substitution $\lambda=x^2$ gives the exact error identity
\begin{equation}
    \max_{1\leq|x|\leq\sqrt{\kappa}}
    |Z(x)-\operatorname{sign}(x)|
    =
    \max_{\lambda\in[1,\kappa]}
    \left|\sqrt{\lambda}\,q_{\kappa,r}(\lambda)-1\right|
    =E_{\kappa,r}.
    \label{eq:zolo-sign-equivalence app}
\end{equation}
Thus \emph{absolute-error} approximation of the sign function is equivalent to \emph{relative-error} approximation of the inverse square root, and achieves the same optimal error. Rescaling $x$ by $\sqrt{\kappa}$ gives the usual sign-approximation domain $[-1,-\sqrt{1-m}]\cup[\sqrt{1-m},1]$, where $m\triangleq1-\kappa^{-1}$; see \citet[Section~3.1]{nakatsukasa2016computing}. More generally, on any positive interval $[a,b]$, the approximant $a^{-1/2}q_{b/a,r}(\lambda/a)$ has optimal relative error $E_{b/a,r}$. Thus, the ratio $\kappa=b/a$ of the interval endpoints directly controls the optimal approximation error.

\paragraph{Zolotarev approximant and the elliptic grid.}
The Zolotarev optimal approximant $q_{\kappa,r}$ can be explicitly described using Jacobi elliptic functions. To this end, define the \emph{elliptic parameter}
\begin{equation}
    m\triangleq1-\kappa^{-1}
\end{equation}
and notice that $m \in [0,1)$. 
Let $K\triangleq K(m)$, where $K(\cdot)$ is the complete elliptic integral of the first kind. Define 
\begin{equation}
    u_i=\frac{i}{2r+1},\qquad
    \gamma_i=\jsc^2(K u_i\mid m)
    =\frac{\sn^2(K u_i\mid m)}{\cn^2(K u_i\mid m)},
    \qquad i=1,\ldots,2r.
    \label{eq:zolo-grid}
\end{equation}
Notice that the $u_i$ describe a uniform grid on $[0,1]$ (excluding the endpoints), and the $\gamma_i$ evaluate an elliptic function on this grid. Background on Jacobi elliptic functions is given in Appendix~\ref{app:jacobi}. 

The Zolotarev approximant is
\begin{equation}
    q_{\kappa,r}(\lambda)
    =C_{\kappa,r}\prod_{j=1}^r
    \frac{\lambda+\gamma_{2j}}{\lambda+\gamma_{2j-1}},
    \qquad
    q_{\kappa,r}(1)+\sqrt{\kappa}\,q_{\kappa,r}(\kappa)=2,
    \label{eq:zolo-product}
\end{equation}
where the endpoint condition uniquely determines $C_{\kappa,r}>0$ \citep[Section~3.1]{nakatsukasa2016computing}. The zero and pole shifts are $\xi_j\triangleq\gamma_{2j}$ and $\zeta_j\triangleq\gamma_{2j-1}$, respectively; the actual zeros and poles are on the negative real axis, given by $-\xi_j$ and $-\zeta_j$. Thus they are generated by evaluating $\jsc^2$ at alternating even and odd points of the same uniform $u$-grid. In particular, it is the ``elliptic times'' $Ku_i$, not the shifts $\gamma_i$, that are uniformly spaced.

The relative error equioscillates between $\pm E_{\kappa,r}$, with endpoint values
\begin{equation}
    q_{\kappa,r}(1)=1-E_{\kappa,r},
    \qquad
    \sqrt{\kappa}\,q_{\kappa,r}(\kappa)=1+E_{\kappa,r}.
    \label{eq:zolo-endpoint-errors}
\end{equation}

\paragraph{Endpoint normalization.}
For our sampling construction, we use a slightly modified approximant that is asymptotically equivalent. It differs only by a constant factor, since we will require the approximation to be exact at $\lambda=1$. Concretely, we use 
\begin{equation}
    \widehat q_{\kappa,r}(\lambda)
    \triangleq\frac{q_{\kappa,r}(\lambda)}{q_{\kappa,r}(1)}
    =
    \prod_{j=1}^r
    \frac{1+\gamma_{2j-1}}{1+\gamma_{2j}}
    \frac{\lambda+\gamma_{2j}}{\lambda+\gamma_{2j-1}}.
    \label{eq:zolo-normalized}
\end{equation}
This removes the implicit constant $C_{\kappa,r}$ and enforces $\widehat q_{\kappa,r}(1)=1$, so the unit-precision eigenmode is preserved exactly. Its relative error is
\begin{equation}
    \widehat E_{\kappa,r}
    \triangleq
    \max_{\lambda\in[1,\kappa]}
    \left|\sqrt{\lambda}\,\widehat q_{\kappa,r}(\lambda)-1\right|
    =\frac{2E_{\kappa,r}}{1-E_{\kappa,r}}.
    \label{eq:zolo-normalized-error}
\end{equation}
Indeed, $\sqrt{\lambda}\,\widehat q_{\kappa,r}(\lambda)$ ranges from $1$ to $(1+E_{\kappa,r})/(1-E_{\kappa,r})$, and when the error is small, $\widehat E_{\kappa,r} \approx 2E_{\kappa,r}$. Although endpoint normalization sacrifices exact minimax optimality in \eqref{eq:zolo-minimax app}, it preserves the asymptotic convergence rate and makes all expressions considerably simpler.

\paragraph{Approximation rates.}
Set $\rho_\kappa\triangleq\exp\!\left(\pi K(1-m)/K(m)\right)>1$. The minimax error $E_{\kappa,r}$ satisfies \citep[Appendix~A]{beckermann2017singular}

\begin{equation}
    \frac{4\rho_\kappa^{-(2r+1)}}
    {(1+\rho_\kappa^{-2(2r+1)})^4}
    \leq E_{\kappa,r}
    \leq 4\rho_\kappa^{-(2r+1)}.
    \label{eq:zolo-error-bounds}
\end{equation}
Consequently, for fixed $\kappa>1$ as $r\to\infty$,
\begin{equation}
    E_{\kappa,r}\sim4\rho_\kappa^{-(2r+1)},
    \qquad
    \widehat E_{\kappa,r}\sim8\rho_\kappa^{-(2r+1)}.
    \label{eq:zolo-error-asymptotics}
\end{equation}
Moreover, the standard elliptic-integral asymptotics give
\[
    \log\rho_\kappa=\frac{\pi K(1-m)}{K(m)}
    \sim\frac{\pi^2}{\log(16\kappa)}
    \qquad(\kappa\to\infty).
\]
Thus, for $\kappa\geq2$ and sufficiently small $\varepsilon>0$, the minimum degree required to achieve either error at most $\varepsilon$ is $\Theta(\log\kappa\,\log(1/\varepsilon))$\footnote{Ignoring constant factors, see Appendix.}. When $\kappa=1$, the approximation interval is a singleton and the constant function $1$ achieves zero error. These results and their proofs are given in detail in Section~\ref{app:zolotarev-proofs}.

\subsection{Proofs}\label{app:zolotarev-proofs}
\begin{proposition}[Equivalence of sign and inverse-square-root minimax errors]
\label{lma:sign-invsqrt-equivalence}
Fix $\kappa>1$ and an integer $r\geq0$, and let
$m\triangleq1-\kappa^{-1}$ and $\delta_\kappa\triangleq\sqrt{1-m}=\kappa^{-1/2}$. Define
\begin{align}
    E_{p,q}(\delta_\kappa)&\triangleq
    \min_{R\in\mathcal R_{p,q}}\max_{x\in[-1,-\delta_\kappa]\cup[\delta_\kappa,1]}
    |R(x)-\operatorname{sign}(x)|,\\
    E_{\kappa,r}&\triangleq
    \min_{q\in\mathcal R_{r,r}}\max_{\lambda\in[1,\kappa]}
    |\sqrt{\lambda}\,q(\lambda)-1|,
\end{align}
where the minima are restricted to real rational functions without poles on the respective approximation domains. Then
\[
    E_{2r+1,2r+1}(\delta_\kappa)=E_{2r+1,2r}(\delta_\kappa)=E_{\kappa,r}.
\]
In particular, allowing denominator degree $2r+1$ rather than $2r$ does not improve the optimal sign approximation error.
\end{proposition}

\begin{proof}
Zolotarev's explicit solution of the sign minimax problem
\citep[proof of Theorem~3.3, equation~(3.6)]{beckermann2017singular}
states that a global minimizer over $\mathcal R_{2r+1,2r+1}$ has the form
$R(x)=xP(x^2)/Q(x^2)$ with $\deg P,\deg Q\leq r$.
Consequently, restricting to this form does not change either sign minimax error. By oddness,
\[
    E_{2r+1,2r+1}(\delta_\kappa)=E_{2r+1,2r}(\delta_\kappa)
    =\min_{\deg P,\deg Q\leq r}\max_{x\in[\delta_\kappa,1]}
    \left|x\frac{P(x^2)}{Q(x^2)}-1\right|,
\]
where $Q$ has no zeros on $[\delta_\kappa^2,1]$. Now, for any admissible $P, Q$ with degree $\leq r$, set
\[
    \lambda\triangleq\frac{x^2}{\delta_\kappa^2},
    \qquad
    q(\lambda)\triangleq\delta_\kappa\,\frac{P(\delta_\kappa^2\lambda)}{Q(\delta_\kappa^2\lambda)}.
\]
Since $\delta_\kappa^{-2}=\kappa$, this gives
\[
    \max_{x\in[\delta_\kappa,1]}
    \left|x\frac{P(x^2)}{Q(x^2)}-1\right|
    = \max_{\frac{x^2}{\delta_\kappa^2}\in [1, 1/\delta_\kappa^2]} \left|x \frac{P(x^2)}{Q(x^2)} - 1\right| =
    \max_{\lambda\in[1,\kappa]}
    \left|\sqrt{\lambda}\,q(\lambda)-1\right|.
\]
As $P, Q$ range over the admissible polynomials of degree at most $r$, the rescaled rational function $q(\lambda) = \delta_\kappa\,\frac{P(\delta_\kappa^2 \lambda)}{Q(\delta_\kappa^2 \lambda)}$ ranges over all admissible functions in $\mathcal{R}_{r,r}$. Hence
\begin{equation}
     \min_{\deg P,\deg Q\leq r}\max_{\lambda\in [1,\kappa]}\left|\sqrt{\lambda}\delta_\kappa \frac{P(\delta_\kappa^2 \lambda)}{Q(\delta_\kappa^2 \lambda)}- 1\right| = \min_{q \in \mathcal{R}_{r,r}}\max_{\lambda \in [1,\kappa]} \left|\sqrt{\lambda} q(\lambda) - 1\right|,
\end{equation}
where both minima are over the admissible functions described above. Therefore,
\[
    E_{2r+1,2r+1}(\delta_\kappa)=E_{2r+1,2r}(\delta_\kappa)=E_{\kappa,r}.
\]
\end{proof}

\begin{proposition}[Rates of $E_{\kappa, r}$]
    for all $\kappa > 1, r \geq 0$, with $m\triangleq1-\kappa^{-1}$,
    \begin{equation}
        \frac{1}{4} \exp\left(-\frac{\pi^2(2r+1)}{\ln(\kappa)}\right) \leq E_{\kappa, r} \leq 4\exp\left(-\frac{\pi^2(2r+1)}{\ln(16\kappa)}\right)
    \end{equation}
    more precisely
    \begin{equation}
        \frac{1}{4} \exp\left(-\pi(2r+1)\frac{K(1-m)}{K(m)}\right) \leq E_{\kappa, r} \leq 4\exp\left(-\pi (2r+1)\frac{K(1-m)}{K(m)}\right)
    \end{equation}
    so that for all $\kappa >1, r \geq 0$
    \begin{equation}
        E_{\kappa, r} = \Theta\left(\exp\left(-\pi(2r+1)\frac{K(1-m)}{K(m)}\right)\right)
    \end{equation}
\end{proposition}

\begin{proof}
Fix $\kappa>1$ and $r\in\mathbb Z_{\geq0}$, and write
\[
    N\triangleq2r+1,\qquad
    m\triangleq1-\kappa^{-1},\qquad
    \delta_\kappa\triangleq\sqrt{1-m}=\kappa^{-1/2}.
\]
Following \citet[equation~(3.2)]{beckermann2017singular}, define
\[
    \mu(m)\triangleq\frac{\pi K(m)}{2K(1-m)},
    \qquad
    \rho\triangleq\exp\!\left(\frac{\pi^2}{2\mu(m)}\right)
    =\exp\!\left(\pi\frac{K(1-m)}{K(m)}\right)>1.
\]
By Lemma~\ref{lma:sign-invsqrt-equivalence},
$E_{\kappa,r}=E_{N,N}(\delta_\kappa)$. Applying the sign-approximation bounds of
\citet[Appendix~A, equation~(A.5)]{beckermann2017singular}
with $a=\delta_\kappa$ and $b=1$ therefore gives
\begin{equation}
\label{eq:zolo-error-rho-bounds}
    \frac{4\rho^{-N}}{(1+\rho^{-2N})^4}
    \leq E_{\kappa,r}
    \leq \frac{4\rho^{-N}}{(1+\rho^{-2N})^2}
    \leq 4\rho^{-N}.
\end{equation}
Since $0<\rho^{-2N}<1$, we have $(1+\rho^{-2N})^4\leq16$.
Consequently,
\begin{equation}
\label{eq:zolo-error-elliptic-bounds}
    \frac14\exp\!\left(-\pi N\frac{K(1-m)}{K(m)}\right)
    \leq E_{\kappa,r}
    \leq 4\exp\!\left(-\pi N\frac{K(1-m)}{K(m)}\right).
\end{equation}
This proves the elliptic bounds and the $\Theta$-statement, with constants
independent of both $\kappa$ and $r$.

To obtain the elementary bounds, we use the elliptic-integral inequalities
\citep[equation~(19.9.5)]{DLMF}
\[
    \ln\!\left(\frac{(1+m^{1/4})^2}{\sqrt{1-m}}\right)
    <\mu(m)
    <\ln\!\left(\frac{2(1+\sqrt m)}{\sqrt{1-m}}\right).
\]
Because $0<m<1$, these imply
\[
    \ln\!\left(1/\sqrt{1-m}\right)<\mu(m)<\ln\!\left(4/\sqrt{1-m}\right).
\]
Taking reciprocals and using $1-m=\kappa^{-1}$ yields
\begin{equation}
\label{eq:elliptic-ratio-log-bounds}
    \frac{\pi}{\ln(16\kappa)}
    <\frac{K(1-m)}{K(m)}=\frac{\pi}{2\mu(m)}
    <\frac{\pi}{\ln\kappa}.
\end{equation}
Thus
\[
    \exp\!\left(-\frac{\pi^2N}{\ln\kappa}\right)
    \leq \exp\!\left(-\pi N\frac{K(1-m)}{K(m)}\right)
    \leq \exp\!\left(-\frac{\pi^2N}{\ln(16\kappa)}\right).
\]
Combining this with \eqref{eq:zolo-error-elliptic-bounds} and substituting
$N=2r+1$ proves
\[
    \frac14\exp\!\left(-\frac{\pi^2(2r+1)}{\ln\kappa}\right)
    \leq E_{\kappa,r}
    \leq 4\exp\!\left(-\frac{\pi^2(2r+1)}{\ln(16\kappa)}\right).
\]
\end{proof}

\begin{corollary}[Asymptotic Sharpness]
    From \eqref{eq:zolo-error-rho-bounds} we get that for any fixed $\kappa > 1$, with $m\triangleq1-\kappa^{-1}$, the elliptic upper bound is asymptotically exact, i.e.
    \begin{equation}
        E_{\kappa, r} \sim 4\rho^{-(2r+1)},\quad r\to\infty
    \end{equation}
    moreover from \eqref{eq:elliptic-ratio-log-bounds} we have
    \begin{equation}
        \ln \rho = \pi \frac{K(1-m)}{K(m)} \sim \frac{\pi^2}{\ln 16\kappa}, \quad \kappa \to \infty
    \end{equation}
\end{corollary}

It will also be helpful to analyze the error associated with the normalized Zolotarev rational approximation to $\lambda^{-1/2}.$ For 
\begin{equation}
    \hat{q}_{\kappa, r}(\lambda)\triangleq \frac{\bar{q}_{\kappa, r}(\lambda)}{\bar{q}_{\kappa,r}(1)} 
\end{equation}
we define
\begin{equation}
    \hat{E}_{\kappa, r} \triangleq \max_{\lambda\in [1,\kappa]}\left|\lambda^{1/2} \hat{q}_{\kappa, r}(\lambda) - 1\right| = \frac{2 E_{\kappa, r}}{1-E_{\kappa, r}}
\end{equation}

\begin{corollary}[Rates of $\hat{E}_{\kappa, r}$]
    \label{cor:e_hat_rates}
    Let $m\triangleq1-\kappa^{-1}$.
    Whenever
    \begin{equation}
        \pi (2r+1)\frac{K(1-m)}{K(m)} \geq \ln 8 \approx 2.08 \footnote{a simple sufficient condition is $r\geq 2\ln 16\kappa$ by directly applying the lower bound on $K(1-m)/K(m)$}
    \end{equation}
    the previous results tell us that 
    \begin{equation}
        \frac{1}{2} \exp\left(-\pi(2r+1)\frac{K(1-m)}{K(m)}\right) \leq \hat{E}_{\kappa, r} \leq 16\exp\left(-\pi (2r+1)\frac{K(1-m)}{K(m)}\right)
    \end{equation}
    so that
    \begin{equation}
        \hat{E}_{\kappa, r} = \Theta \left(\exp\left[-\pi(2r+1) \frac{K(1-m)}{K(m)}\right]\right)
    \end{equation}
    as a consequence, using the bounds on $\frac{K(1-m)}{K(m)}$ we get that
    \begin{equation}
        \frac{1}{2} \exp\left(-\frac{\pi^2(2r+1)}{\ln(\kappa)}\right) \leq \hat{E}_{\kappa, r} \leq 16\exp\left(-\frac{\pi^2(2r+1)}{\ln(16\kappa)}\right)
    \end{equation}
    The two lower bounds remain valid for every $\kappa>1$ and $r\in\mathbb Z_{\geq0}$, without the displayed condition.
    
\end{corollary}

\begin{proof}
Fix $\kappa>1$ and $r\in\mathbb Z_{\geq0}$, and set
\[
    N\triangleq2r+1,\qquad
    \eta\triangleq\rho^{-N}=\exp\!\left(-\pi N\frac{K(1-m)}{K(m)}\right).
\]
By \eqref{eq:zolo-error-elliptic-bounds},
\[
    \frac14\eta\leq E_{\kappa,r}\leq4\eta.
\]
First, $0\leq E_{\kappa,r}<1$ for all such parameters. Indeed, the constant approximant $q(\lambda)=2/(1+\sqrt{\kappa})$ is admissible for every $r\geq0$, so
\[
    E_{\kappa,r}\leq
    \max_{\lambda\in[1,\kappa]}
    \left|\frac{2\sqrt{\lambda}}{1+\sqrt{\kappa}}-1\right|
    =\frac{\sqrt{\kappa}-1}{\sqrt{\kappa}+1}<1.
\]
Consequently,
\[
    \widehat E_{\kappa,r}
    =\frac{2E_{\kappa,r}}{1-E_{\kappa,r}}
    \geq2E_{\kappa,r}
    \geq\frac12\eta,
\]
which proves the elliptic lower bound without any additional condition.

For the upper bound, suppose that $\pi N K(1-m)/K(m)\geq\ln8$. Then $\eta\leq1/8$, and hence $E_{\kappa,r}\leq4\eta\leq1/2$. Therefore,
\[
    \widehat E_{\kappa,r}
    =\frac{2E_{\kappa,r}}{1-E_{\kappa,r}}
    \leq4E_{\kappa,r}
    \leq16\eta.
\]
Together, these inequalities establish the elliptic bounds and the $\Theta$-statement, with universal constants throughout the stated regime.

Finally, \eqref{eq:elliptic-ratio-log-bounds} gives
\[
    \exp\!\left(-\frac{\pi^2N}{\ln\kappa}\right)
    \leq\eta
    \leq\exp\!\left(-\frac{\pi^2N}{\ln(16\kappa)}\right).
\]
Combining these inequalities with the preceding lower and upper bounds proves the elementary bounds. The elementary lower bound is again unconditional, whereas the elementary upper bound holds under $\pi N K(1-m)/K(m)\geq\ln8$. In particular, a sufficient condition for both upper bounds is
\[
    \frac{\pi^2(2r+1)}{\ln(16\kappa)}\geq\ln8.
\]
\end{proof}

\begin{corollary}[Accuracy requirement for endpoint normalization]
\label{cor:normalized-accuracy}
For every $\kappa>1$, $r\in\mathbb Z_{\geq0}$, and $\varepsilon>0$,
\begin{equation}
\label{eq:normalized-accuracy-equivalence}
    \widehat E_{\kappa,r}\leq\varepsilon
    \quad\Longleftrightarrow\quad
    E_{\kappa,r}\leq\frac{\varepsilon}{2+\varepsilon}.
\end{equation}
Consequently, $\widehat E_{\kappa,r}\leq\varepsilon$ is guaranteed whenever
\begin{equation}
\label{eq:normalized-accuracy-step-bound}
    r\geq
    \frac{\ln(16\kappa)}{2\pi^2}
    \ln\!\left(\frac{4(2+\varepsilon)}{\varepsilon}\right)
    -\frac12,
\end{equation}
with $r$ chosen as a nonnegative integer.
\end{corollary}

\begin{proof}
Since $0\leq E_{\kappa,r}<1$, multiplying by the positive denominator yields
\[
    \frac{2E_{\kappa,r}}{1-E_{\kappa,r}}\leq\varepsilon
    \quad\Longleftrightarrow\quad
    (2+\varepsilon)E_{\kappa,r}\leq\varepsilon,
\]
which proves \eqref{eq:normalized-accuracy-equivalence}.
Using the elementary upper bound on $E_{\kappa,r}$, it therefore suffices that
\[
    4\exp\!\left(-\frac{\pi^2(2r+1)}{\ln(16\kappa)}\right)
    \leq\frac{\varepsilon}{2+\varepsilon}.
\]
Taking logarithms and rearranging gives
\eqref{eq:normalized-accuracy-step-bound}.
\end{proof}

\section{Score-span Algorithms on Gaussians are Rational Functions}
\label{app:algs-are-ratfuncs}

\begin{proof}[Proof of Lemma~\ref{lma:algs-are-ratfunc}]
Fix a centered Gaussian target $\pi=\mathcal N(0,\Lambda^{-1})$ and an algorithm $\mathrm{alg}\in\mathcal A_r$. Recall that its iterates satisfy
\begin{equation}
\label{eq:oblivious-updates-2}
    x_j=\sum_{i=0}^{j-1}a_{ij}x_i+\sum_{i=0}^{j-1}b_{ij}s_{t_i}(x_i),
    \qquad j=1,\ldots,r.
\end{equation}
For a precision eigenvalue $\lambda\in\operatorname{spec}(\Lambda)$, let $x_{\lambda,j} \triangleq u_\lambda u_\lambda^T x_j$ denote the corresponding eigenmode component of $x_j$. By the Gaussian score formula,
\begin{equation}
    \label{eq:Gaussian-score-formula}
    s_{t_i}(x_{\lambda,i})
    =
    -\frac{\lambda}{\alpha_{t_i}^2\lambda+\beta_{t_i}^2}\,x_{\lambda,i}.
\end{equation}
Therefore, projecting \eqref{eq:oblivious-updates-2} onto this eigenmode gives
\begin{equation}
\label{eq:eigenmode-oblivious-update}
    x_{\lambda,j}
    =
    \sum_{i=0}^{j-1}
    \left(
        a_{ij}
        -
        b_{ij}\frac{\lambda}
        {\alpha_{t_i}^2\lambda+\beta_{t_i}^2}
    \right)x_{\lambda,i}.
\end{equation}
Indeed 
\begin{align}
        x_j &= \sum_{i=0}^{j-1} \left[a_{ij} I_d  - b_{ij}\left(\alpha_{t_i}^2 I_d + \beta_{t_i}(\Lambda^{-1})\right)^{-1}\right]x_i \\
        &= \sum_{i=0}^{j-1} U \left[a_{ij} I_d  - b_{ij}\left(\alpha_{t_i}^2 I_d + \beta_{t_i}^2(\text{diag}(1/\lambda_{1}, \dots , 1/\lambda_{d})\right)^{-1}\right] U^T x_i \\
        &= \sum_{i=1}^{j-1} U \text{diag}\left(a_{ij} - \frac{b_{ij}\lambda_1}{\alpha_{t_i}^2 \lambda_1 + \beta_{t_i}^2}, \dots, a_{ij} - \frac{b_{ij}\lambda_d}{\alpha_{t_i}^2 \lambda_d + \beta_{t_i}^2}\right) U^T x_i \\
        &= \sum_{k=1}^d  \ \sum_{i=1}^{j-1} \left(a_{ij} - \frac{b_{ij}\lambda_1}{\alpha_{t_i}^2 \lambda_1 + \beta_{t_i}^2}\right) u_k u_k^T x_i
    \end{align}
    so along any $\lambda$-eigenmode $x_{\lambda,j}$
    \begin{align}
        x_{\lambda,j} &= \sum_{i=1}^{j-1}\left(a_{ij} - \frac{b_{ij}\lambda}{\alpha^2_{t_i}\lambda + \beta^2_{t_i}}\right) x_{\lambda,i}
    \end{align}

We prove inductively that for every $j=0,\ldots,r$ there exists
$q_j(\lambda) = \frac{P_j(\lambda)}{D_j(\lambda)}\in\mathcal R_{j,j}$ such that
\begin{equation}
    x_{\lambda,j}=q_j(\lambda)x_{\lambda,0}.
\end{equation}
For $j=0$, take $q_0(\lambda)=1$.

For the induction step, define
\begin{equation}
    d_i(\lambda)\triangleq
    \alpha_{t_i}^2\lambda+\beta_{t_i}^2,
    \qquad
    D_j(\lambda)\triangleq\prod_{\ell=0}^{j-1}d_\ell(\lambda),
    \qquad D_0(\lambda)\triangleq1.
\end{equation}
Suppose for each $i<j$ that
\begin{equation}
    q_i(\lambda)=\frac{P_i(\lambda)}{D_i(\lambda)},
    \qquad \deg P_i\leq i.
\end{equation}
Substituting into \eqref{eq:eigenmode-oblivious-update} yields
\begin{align}
    q_j(\lambda)
    &=
    \sum_{i=0}^{j-1}
    \frac{
        \left(a_{ij}d_i(\lambda)-b_{ij}\lambda\right)
        P_i(\lambda)
    }{
        D_{i+1}(\lambda)
    } \\
    &=
    \frac{P_j(\lambda)}{D_j(\lambda)},
\end{align}
where
\begin{equation}
\label{eq:Pj-recursion}
    P_j(\lambda)
    \triangleq
    \sum_{i=0}^{j-1}
    \left(a_{ij}d_i(\lambda)-b_{ij}\lambda\right)
    P_i(\lambda)
    \prod_{\ell=i+1}^{j-1}d_\ell(\lambda).
\end{equation}
Each summand in \eqref{eq:Pj-recursion} has degree at most
\[
    1+i+(j-i-1)=j,
\]
and hence $\deg P_j\leq j$. Since $\deg D_j=j$, it follows that
$q_j\in\mathcal R_{j,j}$.

Taking $j=r$, there is therefore a rational function
$q_r^{\mathrm{alg}}\in\mathcal R_{r,r}$ such that
\[
    x_{\lambda,r}
    =
    q_r^{\mathrm{alg}}(\lambda)x_{\lambda,0}
\]
for every eigenmode $\lambda$. Since the same scalar function acts on every eigenmode,
\begin{equation}
    x_r^{\mathrm{alg}}
    =
    q_r^{\mathrm{alg}}(\Lambda)z
    =
    U\operatorname{diag}\!\left(
        q_r^{\mathrm{alg}}(\lambda_1),\ldots,
        q_r^{\mathrm{alg}}(\lambda_d)
    \right)U^\top z.
\end{equation}
Thus every algorithm in $\mathcal A_r$ acts on a centered Gaussian target through a type-$(r,r)$ rational function of its precision matrix.
\end{proof} 

\section{Zoloflow Complexity Results on Gaussians}
\label{app:compelxity-results-zoloflow}

\begin{lemma}[Zoloflow realizes the endpoint-normalized Zolotarev approximation]
\label{lma:zoloflow-normalized-zolotarev}
Fix $\kappa>1$, an integer $r\geq1$, and a centered Gaussian target
$\pi=\mathcal N(0,\Lambda^{-1})$ with $I_d \preceq \Lambda \preceq \kappa I_d$. Let $m$, $u_i$ and $\gamma_i$ be as in \eqref{eq:zolo-grid}, and write $\cn_i\triangleq\cn(u_i\mid m)$ and $\sn_i\triangleq\sn(u_i\mid m)$, so that $\gamma_i=\sn_i^2/\cn_i^2$. Let $q_{\kappa,r}$ be the Zolotarev approximant \eqref{eq:zolo-product} and $\widehat q_{\kappa,r}$ its endpoint normalization \eqref{eq:zolo-normalized}.

Run Zoloflow (Algorithm~\ref{alg:zoloflow}) with parameters $(\kappa,r)$,
exact interpolant scores, and initialization $x_0=z\sim\mathcal N(0,I_d)$.
For each precision eigenmode $\lambda$, let $x_{\lambda,j}$ denote the
corresponding component of $x_j$. Then, for $j=1,\ldots,r$,
\begin{align}
\label{eq:zoloflow-eigenmode-factor}
    x_{\lambda,j}
    &=\frac{\lambda\cn_{2j}^2+\sn_{2j}^2}
    {\lambda\cn_{2j-1}^2+\sn_{2j-1}^2}\,x_{\lambda,j-1}\\
    &=\frac{1+\gamma_{2j-1}}{1+\gamma_{2j}}
    \frac{\lambda+\gamma_{2j}}{\lambda+\gamma_{2j-1}}\,
    x_{\lambda,j-1}.
    \notag
\end{align}
Consequently,
\begin{equation}
\label{eq:zoloflow-Gaussian-output}
    x_{\lambda,r}=\widehat q_{\kappa,r}(\lambda)x_{\lambda,0},
    \qquad
    x_r=\widehat q_{\kappa,r}(\Lambda)z.
\end{equation}
Thus Zoloflow implements the endpoint-normalized approximant exactly,
without any additional scalar normalization of its initialization or output.
\end{lemma}

\begin{proof}
Recall that Zoloflow uses $h=2/(2r+1)$ and $t_j=(2j-1)/(2r+1)$, so
\[
    K(m)t_j=u_{2j-1},\qquad K(m)(t_j+h/2)=u_{2j}.
\]
For a fixed precision eigenmode $\lambda>0$, define
\[
    D_i(\lambda)\triangleq\lambda\cn_i^2+\sn_i^2
    =\lambda+(1-\lambda)\sn_i^2,
\]
where the second equality follows from $\cn_i^2+\sn_i^2=1$.
The Gaussian score preserves this eigenmode and satisfies
\[
    s_{t_j}(x_{\lambda,j-1})
    =-\frac{\lambda}{D_{2j-1}(\lambda)}\,x_{\lambda,j-1}.
\]
Consequently, projecting the update in Algorithm~\ref{alg:zoloflow} onto this eigenmode as described in \ref{app:algs-are-ratfuncs}
\begin{align*}
    x_{\lambda,j}
    &=\left[
        1+\frac{\sn_{2j}^2-\sn_{2j-1}^2}{\sn_{2j-1}^2}
        \left(1-\frac{\lambda}{D_{2j-1}(\lambda)}\right)
    \right]x_{\lambda,j-1}\\
    &=\left[
        1+\frac{(1-\lambda)(\sn_{2j}^2-\sn_{2j-1}^2)}
        {D_{2j-1}(\lambda)}
    \right]x_{\lambda,j-1}\\
    &=\frac{D_{2j}(\lambda)}{D_{2j-1}(\lambda)}\,x_{\lambda,j-1}.
\end{align*}
Here we used
$D_{2j-1}(\lambda)-\lambda=(1-\lambda)\sn_{2j-1}^2$
in the second equality.

Next, since $\gamma_i=\sn_i^2/\cn_i^2$ and $\cn_i^2+\sn_i^2=1$,
\[
    \cn_i^2=\frac{1}{1+\gamma_i},
    \qquad
    D_i(\lambda)=\cn_i^2(\lambda+\gamma_i)
    =\frac{\lambda+\gamma_i}{1+\gamma_i}.
\]
Thus each step satisfies
\[
    x_{\lambda,j}
    =\frac{1+\gamma_{2j-1}}{1+\gamma_{2j}}
    \frac{\lambda+\gamma_{2j}}{\lambda+\gamma_{2j-1}}\,
    x_{\lambda,j-1}.
\]
All divisions are valid because $0<u_i<K(m)$ and $\lambda>0$.

Iterating from $x_0=z$ yields
\begin{align*}
    x_{\lambda,r}
    &=\left[
        \prod_{j=1}^r
        \frac{1+\gamma_{2j-1}}{1+\gamma_{2j}}
        \frac{\lambda+\gamma_{2j}}{\lambda+\gamma_{2j-1}}
    \right]x_{\lambda,0}\\
    &=\frac{q_{\kappa,r}(\lambda)}
    {q_{\kappa,r}(1)}\,x_{\lambda,0}
    =\widehat q_{\kappa,r}(\lambda)x_{\lambda,0},
\end{align*}
where the constant $C_{\kappa,r}$ cancels in the ratio.
Finally, writing $\Lambda=U\operatorname{diag}(\lambda_1,\ldots,\lambda_d)U^\top$
and reassembling the eigenmode components gives
\[
    x_r
    =U\operatorname{diag}\!\left(
        \widehat q_{\kappa,r}(\lambda_1),\ldots,
        \widehat q_{\kappa,r}(\lambda_d)
    \right)U^\top z
    =\widehat q_{\kappa,r}(\Lambda)z.
\]
Hence the step factors already incorporate the endpoint normalization,
and no additional scaling of the initialization or output is required.
\end{proof}

We are now ready to prove the upper bound in \cref{thm:zoloflow-Gaussian-bound}.

\begin{proof}[Proof of Theorem~\ref{thm:zoloflow-Gaussian-bound}]
Write $\widehat q\triangleq\widehat q_{\kappa,r}$.
By Lemma~\ref{lma:zoloflow-normalized-zolotarev},
\[
    x_r=\widehat q(\Lambda)z,\qquad z\sim\mathcal N(0,I_d).
\]
Since $\widehat q(\Lambda)$ is symmetric, the output law is
\[
    \widehat\pi=\mathcal N\!\left(0,\widehat q(\Lambda)^2\right).
\]

\paragraph{KL divergence in terms of relative errors.}
Let $\lambda_1,\ldots,\lambda_d\in[1,\kappa]$ be the eigenvalues of $\Lambda$, and define
\[
    e_k\triangleq\sqrt{\lambda_k}\,\widehat q(\lambda_k)-1,
    \qquad k=1,\ldots,d.
\]
The endpoint-normalization identity gives
\[
    0\leq e_k\leq\widehat E_{\kappa,r}.
\]
In particular, $\widehat q(\lambda_k)>0$, so $\widehat q(\Lambda)^2\succ0$.
Moreover, the eigenvalues of $\Lambda\widehat q(\Lambda)^2$ are
\[
    \lambda_k\widehat q(\lambda_k)^2=(1+e_k)^2.
\]
Taking the expectation of the Gaussian log-density ratio therefore yields
\begin{align}
\label{eq:zoloflow-kl-relative-errors}
    D_{\mathrm{KL}}(\widehat\pi\|\pi)
    &=\frac12\left(
        \operatorname{tr}\!\left(\Lambda\widehat q(\Lambda)^2\right)
        -d-\ln\det\!\left(\Lambda\widehat q(\Lambda)^2\right)
    \right) \notag\\
    &=\frac12\sum_{k=1}^d
        \left((1+e_k)^2-1-2\ln(1+e_k)\right) \notag\\
    &=\sum_{k=1}^d
        \left(e_k+\frac{e_k^2}{2}-\ln(1+e_k)\right).
\end{align}
For every $u\geq0$,
\[
    \ln(1+u)\geq u-\frac{u^2}{2}.
\]
Indeed, the difference between the two sides vanishes at zero and has derivative $u^2/(1+u)\geq0$. Applying this inequality to each $e_k$ gives
\begin{equation}
\label{eq:zoloflow-kl-upper-bound}
    D_{\mathrm{KL}}(\widehat\pi\|\pi)
    \leq\sum_{k=1}^d e_k^2
    \leq d\,\widehat E_{\kappa,r}^{\,2}.
\end{equation}

\paragraph{From KL divergence to total variation.}
By Pinsker's inequality,
\begin{equation}
\label{eq:zoloflow-pinsker}
    d_{\mathrm{TV}}(\widehat\pi,\pi)
    \leq\sqrt{\frac12D_{\mathrm{KL}}(\widehat\pi\|\pi)}
    \leq\left(\frac12\sum_{k=1}^d e_k^2\right)^{1/2}
    \leq\sqrt{\frac d2}\,\widehat E_{\kappa,r}.
\end{equation}

\paragraph{Applying the rational-approximation rate.}
Setting $m\triangleq1-\kappa^{-1}$ and $A_{\kappa,r}\triangleq
    \frac{\pi^2(2r+1)}{\ln(16\kappa)}.$ we have 
from \eqref{eq:elliptic-ratio-log-bounds},
\[
     A_{\kappa,r} \leq \pi(2r+1)\frac{K(1-m)}{K(m)}.
\]
Consequently, whenever $A_{\kappa,r}\geq\ln8$ (sufficient if $r\geq 2\ln 16\kappa$), the condition of Corollary~\ref{cor:e_hat_rates} is satisfied, and
\[
    \widehat E_{\kappa,r}
    \leq16\exp\!\left(-\pi(2r+1)\frac{K(1-m)}{K(m)}\right)
    \leq16\exp\left(-\frac{\pi^2(2r+1)}{\ln(16\kappa)}\right)
\]
Combining this with \eqref{eq:zoloflow-pinsker} yields
\begin{equation}
\label{eq:zoloflow-tv-exponential-bound}
    d_{\mathrm{TV}}(\widehat\pi,\pi)
    \leq8\sqrt{2d}\,
    \exp\!\left(-\frac{\pi^2(2r+1)}{\ln(16\kappa)}\right). 
\end{equation}
The query condition \eqref{eq:zoloflow-tv-query-threshold} is equivalent to
\[
    A_{\kappa,r}\geq
    \ln\!\left(\frac{8\sqrt{2d}}{\varepsilon}\right).
\]
Because $d\geq1$ and $\varepsilon<1$, the right-hand side exceeds $\ln8$. Thus the preceding bound applies and gives
$d_{\mathrm{TV}}(\widehat\pi,\pi)\leq\varepsilon$.

Finally, choose the smallest positive integer satisfying \begin{equation}
\label{eq:zoloflow-tv-query-threshold}
    r\geq
    \frac{\ln(16\kappa)}{2\pi^2}
    \ln\!\left(\frac{8\sqrt{2d}}{\varepsilon}\right)
    -\frac12.
\end{equation}
Each Zoloflow step uses one score query, so the number of queries is
\[
    O\!\left(
        \ln(16\kappa)\,
        \ln\!\left(\frac{8\sqrt{2d}}{\varepsilon}\right)
    \right).
\]
For $\kappa\geq2$ and $0<\varepsilon\leq1/2$, this is
$O(\ln\kappa\,\ln(\sqrt d/\varepsilon))$, with a universal implicit constant.
\end{proof}

Next we restate the lower bound. Recall the minimax query complexity
\begin{equation}
\tag{\ref{eq:r_optimal}}
    r^*_{\mathrm{obl}}(d,\kappa,\varepsilon)
    \triangleq
    \min\left\{
        r\in\mathbb Z_{\geq1}:
        \begin{array}{l}
            \exists\,\mathrm{alg}\in\mathcal A_r\ \text{such that}\\[2pt]
            \displaystyle
            \sup_{I_d\preceq\Lambda\preceq\kappa I_d}
            d_{\mathrm{TV}}\!\left(
                \mathcal L(x_r^{\mathrm{alg},\Lambda}),
                \mathcal N(0,\Lambda^{-1})
            \right)
            \leq\varepsilon
        \end{array}
    \right\}.
\end{equation}
Here each algorithm, including its admissible interpolant schedule,
query times, and scalar update coefficients, is chosen independently
of $\Lambda$, the initialization, and the oracle responses.

\begin{proof}[Proof of Lower Bound in Theorem~\ref{thm:zoloflow-Gaussian-bound}]
\label{app:Gaussian_lower_bound}
Fix $d\geq1$, $\kappa\geq2$, an integer $r\geq1$, and
$\mathrm{alg}\in\mathcal A_r$. Write $\delta\triangleq\varepsilon$.

\paragraph{Rational representation.}
By Lemma~\ref{lma:algs-are-ratfunc}, there is a real rational function
$q\triangleq q_r^{\mathrm{alg}}\in\mathcal R_{r,r}$ such that
\[
    x_r^{\mathrm{alg},\Lambda}=q(\Lambda)z,
    \qquad z\sim\mathcal N(0,I_d).
\]
Because the schedule, query times, and scalar coefficients are fixed
independently of the target, the same function $q$ applies to every
$I_d\preceq\Lambda\preceq\kappa I_d$. Moreover, $q$ has no poles on
$[1,\kappa]$: the proof of the rationality lemma gives a denominator
$\prod_{i=0}^{r-1}(\alpha_{t_i}^2\lambda+\beta_{t_i}^2)$, which is
strictly positive for $\lambda>0$.
In particular, $q$ is continuous on $[1,\kappa]$, and
\[
    \mathcal L(x_r^{\mathrm{alg},\Lambda})
    =\mathcal N\!\left(0,q(\Lambda)^2\right).
\]

\paragraph{A worst-case Gaussian target.}
We first prove
\begin{equation}
\label{eq:obl-tv-rational-lower-bound}
    \sup_{I_d\preceq\Lambda\preceq\kappa I_d}
    d_{\mathrm{TV}}\!\left(
        \mathcal L(x_r^{\mathrm{alg},\Lambda}),
        \mathcal N(0,\Lambda^{-1})
    \right)
    \geq\frac1{100}\min\{1,\sqrt d\,E_{\kappa,r}\}.
\end{equation}

If $q(\lambda_0)=0$ for some $\lambda_0\in[1,\kappa]$, choose
$\Lambda=\lambda_0I_d$. The algorithm then outputs zero almost surely,
whereas $\mathcal N(0,\lambda_0^{-1}I_d)$ assigns zero probability to
the origin. Their total variation distance is therefore $1$, so
\eqref{eq:obl-tv-rational-lower-bound} holds in this case.

Otherwise, continuity implies that $q$ has constant sign on $[1,\kappa]$.
Thus there is a single $\sigma\in\{-1,1\}$ such that
$\sigma q(\lambda)=|q(\lambda)|$ throughout the interval.
Since $\sigma q\in\mathcal R_{r,r}$ is admissible in the definition of
$E_{\kappa,r}$,
\[
    \max_{\lambda\in[1,\kappa]}
    \left|\sqrt{\lambda}\,|q(\lambda)|-1\right|
    =
    \max_{\lambda\in[1,\kappa]}
    \left|\sqrt{\lambda}\,\sigma q(\lambda)-1\right|
    \geq E_{\kappa,r}.
\]
Choose $\lambda_*\in[1,\kappa]$ attaining this maximum and set
$\Lambda_*\triangleq\lambda_*I_d$. For this target,
\[
    \mathcal L(x_r^{\mathrm{alg},\Lambda_*})
    =\mathcal N\!\left(0,q(\lambda_*)^2I_d\right),
    \qquad
    \mathcal N(0,\Lambda_*^{-1})
    =\mathcal N(0,\lambda_*^{-1}I_d).
\]
Applying the same-mean Gaussian TV lower bound of
\citet[Theorem~1.1]{devroye2018total}, whose relative covariance
eigenvalues here are all $\lambda_*q(\lambda_*)^2$, gives
\begin{equation}
\label{eq:obl-isotropic-tv-lower-bound}
    d_{\mathrm{TV}}\!\left(
        \mathcal L(x_r^{\mathrm{alg},\Lambda_*}),
        \mathcal N(0,\Lambda_*^{-1})
    \right)
    \geq
    \frac1{100}\min\!\left\{
        1,\sqrt d\,\left|\lambda_*q(\lambda_*)^2-1\right|
    \right\}.
\end{equation}
Furthermore,
\begin{align*}
    \left|\lambda_*q(\lambda_*)^2-1\right|
    &=
    \left(\sqrt{\lambda_*}|q(\lambda_*)|+1\right)
    \left|\sqrt{\lambda_*}|q(\lambda_*)|-1\right|\\
    &\geq
    \left|\sqrt{\lambda_*}|q(\lambda_*)|-1\right|
    \geq E_{\kappa,r}.
\end{align*}
Substituting into \eqref{eq:obl-isotropic-tv-lower-bound} proves
\eqref{eq:obl-tv-rational-lower-bound}.

\paragraph{From TV accuracy to query complexity.}
Suppose now that $\mathrm{alg}$ satisfies the uniform accuracy guarantee
\[
    \sup_{I_d\preceq\Lambda\preceq\kappa I_d}
    d_{\mathrm{TV}}\!\left(
        \mathcal L(x_r^{\mathrm{alg},\Lambda}),
        \mathcal N(0,\Lambda^{-1})
    \right)
    \leq\delta.
\]
For $\delta<1/100$, \eqref{eq:obl-tv-rational-lower-bound} implies
\[
    \min\{1,\sqrt d\,E_{\kappa,r}\}\leq100\delta<1,
    \qquad\text{and hence}\qquad
    E_{\kappa,r}\leq\frac{100\delta}{\sqrt d}.
\]
Combining this with the previously established elementary lower bound,
obtained from \eqref{eq:zolo-error-elliptic-bounds} and
\eqref{eq:elliptic-ratio-log-bounds}, yields
\[
    \frac14\exp\!\left(-\frac{\pi^2(2r+1)}{\ln\kappa}\right)
    \leq E_{\kappa,r}
    \leq\frac{100\delta}{\sqrt d}.
\]
Taking logarithms gives the non-asymptotic necessary condition
\begin{equation}
\label{eq:obl-query-lower-bound-explicit}
    2r+1
    \geq
    \frac{\ln\kappa}{\pi^2}
    \ln\!\left(\frac{\sqrt d}{400\delta}\right).
\end{equation}

Choose the universal constant $\delta_0\triangleq400^{-2}$.
For $0<\delta\leq\delta_0$ and $d\geq1$, we have
$\sqrt d/\delta\geq400^2$, so
\[
    \ln\!\left(\frac{\sqrt d}{400\delta}\right)
    =
    \ln\!\left(\frac{\sqrt d}{\delta}\right)-\ln400
    \geq\frac12\ln\!\left(\frac{\sqrt d}{\delta}\right).
\]
Since $r\geq1$ implies $2r+1\leq3r$,
\eqref{eq:obl-query-lower-bound-explicit} therefore gives
\[
    r\geq
    \frac1{6\pi^2}\,
    \ln\kappa\,
    \ln\!\left(\frac{\sqrt d}{\delta}\right).
\]
This holds for every algorithm and query budget satisfying the uniform
accuracy guarantee. Taking the minimum over such budgets proves
the lower bound of \ref{thm:zoloflow-Gaussian-bound}, for example with
$c=1/(6\pi^2)$ and $\delta_0=400^{-2}$.

Finally, Zoloflow belongs to $\mathcal A_r$, and
Theorem~ \ref{thm:zoloflow-Gaussian-bound} supplies a matching
upper bound with a universal constant throughout this parameter regime.
This proves the lower bound
\end{proof}

\section{Rescaling and Centering: Approaches for handling $\mathcal N(\mu, \Lambda^{-1})$, where spec$(\Lambda) \subset [\lambda_{\min} , \lambda_{\max}]$ and $\mu \neq 0$}
\label{app:approaches-for-handling-arbitrary-spectra}

\label{app:spectrum_rescaling}

Zoloflow is engineered to be optimal for centered Gaussian targets $x_1 \sim \mathcal{N}(0,\Lambda^{-1})$ that are normalized such that $\operatorname{spec}(\Lambda) \subset [1, \kappa]$. In this section, we explore how to modify Zoloflow to achieve the same optimal relative error $\hat{E}_{\kappa, r}$ when we instead have $\mu \neq 0$ and $\text{spec}(\Lambda) \subset [ \lambda_{min}, \lambda_{max}]$ for $0 < \lambda_{\min} < \lambda_{\max}$. We begin with the following two different techniques to accommodate the arbitrary endpoints.
\begin{enumerate}
    \item \textbf{Target rescaling} works by (1) building a modified stochastic interpolant by rescaling the target $x_1$ so that the spectrum is normalized to $[1,\kappa]$, (2) running the new interpolant ODE where we know Zoloflow is optimal, and then (3) rescaling the output back to the original spectrum.
    \item \textbf{Gauge rescaling} works by rescaling the interpolant schedules themselves by a \emph{gauge} $g(t)$, a time-dependent spatial rescaling that is amortized over the interpolant time.
\end{enumerate}
We discuss both of these options in \Cref{app:target-rescaling,sec:gauge-rescaling} for centered distributions. Then in \cref{app:centering}, we argue that generalizing to $\mu\neq 0$ is quite simple.

\subsection{Target rescaling}
\label{app:target-rescaling}

\begin{definition}[Target Rescaled Stochastic Interpolant]
For any VP interpolant pair $(\alpha, \beta)$ consider the rescaled target
\begin{equation}
    x_1^{(\lambda_{\min})}\triangleq \sqrt{\lambda_{\min}}\,x_1,
    \
    \Lambda^{(\lambda_{\min})}\triangleq \frac{\Lambda}{\lambda_{\min}}, \\\implies x_1^{(\lambda_{\min})}\sim\mathcal N\!\left(0,(\Lambda^{(\lambda_{\min})})^{-1}\right), \  I_d\preceq\Lambda^{(\lambda_{\min})}\preceq\kappa I_d.
\end{equation}
We define the \textit{target rescaled Stochastic Interpolant} transporting $z$ to $x_1^{(\lambda_{\min})}$ as
\begin{equation}
    \label{eq:flow_right_endpoint}
     y_t = \alpha_t z + \beta_t x_1^{(\lambda_{\min})} = \alpha_t z + \beta_t \left[\lambda_{\min}^{1/2} x_1\right].
\end{equation}
\end{definition}
Throughout this subsection, the scale $\lambda_{\min}>0$ is fixed, and we suppress the dependence of $y_t$, $\widetilde p_t$, and $\widetilde s_t$ on $\lambda_{\min}$. Intuitively, this rescaling maps the left endpoint of the target precision spectrum to $I_d$, which again reduces the distance our initial Gaussian $z$ must travel and recovers our original setup. The associated probability flow ODE is given by
\begin{equation}
    \dot y_t = \frac{\dot\beta_t}{\beta_t}\left(y + \tilde{s}_{t}(y)\right), \quad y_0\sim \mathcal{N}(0,I), \quad \tilde{s}_t \triangleq \nabla \log p_{y_t}(y).
\end{equation}
\paragraph{Equivalent schedule interpretation and reuse of the original score.}
Target rescaling can equivalently be viewed as retaining the original target $x_1$ and using coefficients
\[
    \alpha_t^*\triangleq\alpha_t,\qquad
    \beta_t^*\triangleq\sqrt{\lambda_{\min}}\,\beta_t,
    \qquad y_t=x_t^*.
\]
Both interpretations define the same interpolant and hence the same probability-flow ODE. Although the starred coefficients are generally not VP or endpoint-normalized, the general case of Proposition~\ref{prop:interpolant-conversion} still applies. Under its assumptions, with $\sigma(t)=\alpha_t/\beta_t$, the time change and spatial scaling are
\[
    \tau(t)\triangleq\sigma^{-1}\!\left(\frac{\sigma(t)}{\sqrt{\lambda_{\min}}}\right),
    \qquad c_t=A_{\lambda_{\min}}(t)\triangleq\sqrt{\alpha_t^2+\lambda_{\min}\beta_t^2}.
\]
as $(\alpha_{\tau(t)},\beta_{\tau(t)}) = \left(\frac{\alpha_t}{A_{\lambda_{\min}}(t)}, \frac{\beta_t \lambda_{\min}^{1/2}}{A_{\lambda_{\min}}(t)}\right)$ gives $c_t = \beta_t^* / \beta_{\tau(t)} = A_{\lambda_{\min}}(t).$
Consequently, the rescaled score and velocity satisfy
\begin{align}
    \widetilde s_t(y)
    &=\frac{1}{A_{\lambda_{\min}}(t)}
      s_{\tau(t)}\!\left(\frac{y}{A_{\lambda_{\min}}(t)}\right),\\
    \widetilde b_t(y)
    &=\frac{\dot\beta_t}{\beta_t}\bigl(y+\widetilde s_t(y)\bigr)
    =\frac{\dot\beta_t}{\beta_t}
      \left[y+\frac{1}{A_{\lambda_{\min}}(t)}
      s_{\tau(t)}\!\left(\frac{y}{A_{\lambda_{\min}}(t)}\right)\right].
\end{align}
Thus the target rescaled flow can be implemented using the original score oracle.

The following proposition states that Zoloflow on the rescaled interpolant also recovers the optimal Gaussian sampling rate. We articulate this method for any VP pair $(\alpha, \beta)$ through the following algorithm
\begin{algorithm}[t]
\caption{Sampling with target rescaling}
\label{alg:target-rescaling}
\begin{algorithmic}[1]
\REQUIRE Spectral endpoints $\lambda_{\min}>0$ and $\lambda_{\max}\geq\lambda_{\min}$, step count $r$, discretization method $\mathcal D$, VP schedule $(\alpha_t,\beta_t)$
\REQUIRE Score oracle $\widetilde s_t=\nabla\log p_{y_t}$ for $y_t=\alpha_t z+\sqrt{\lambda_{\min}}\,\beta_t x_1$
\STATE Sample $y_0=z\sim\mathcal N(0,I_d)$
\STATE Define $b_t(y)\triangleq \dfrac{\dot\beta_t}{\beta_t}\bigl(y+\widetilde s_t(y)\bigr)$
\STATE Compute $y_r$ using $r$ steps of $\mathcal D$ for $\dot y_t=b_t(y_t)$ with schedule $(\alpha_t,\beta_t)$, initialized at $y_0$
\STATE $x_r\leftarrow \lambda_{\min}^{-1/2}y_r$
\RETURN $x_r$
\end{algorithmic}
\end{algorithm}

\begin{proposition}[Zoloflow optimality on the target-rescaled interpolant]
\label{prop:zoloflow_rescaled}
Let $\pi=\mathcal N(0,\Lambda^{-1})$ with $\operatorname{spec}(\Lambda)\subseteq[\lambda_{\min},\lambda_{\max}]$, set $\kappa\triangleq\lambda_{\max}/\lambda_{\min}$, and take the elliptic schedule $(\alpha_t,\beta_t)=(\cn(K(m)t\mid m),\sn(K(m)t\mid m))$ with $m=1-\kappa^{-1}$. Run Zoloflow, i.e.\ the finite-difference discretization of Definition~\ref{def:elliptic_finite_diff} with $r$ steps and exact scores, on the target-rescaled flow~\eqref{eq:flow_right_endpoint} from $y_0=z\sim\mathcal N(0,I_d)$, and output $x_r\gets\lambda_{\min}^{-1/2}y_r$. Then
\begin{equation}
    x_r=\lambda_{\min}^{-1/2}\,\hat q_{\kappa,r}\!\left(\frac{\Lambda}{\lambda_{\min}}\right)z = Q_{\lambda_{\min},\kappa,r}(\Lambda)\,z,
    \qquad Q_{\lambda_{\min},\kappa,r}(\lambda)\triangleq\lambda_{\min}^{-1/2}\,\hat q_{\kappa,r}(\lambda/\lambda_{\min}),
\end{equation}
and
\begin{equation}
    \max_{\lambda\in[\lambda_{\min},\lambda_{\max}]}\left|\lambda^{1/2}Q_{\lambda_{\min},\kappa,r}(\lambda)-1\right|
    =\frac{2E_{\kappa,r}}{1-E_{\kappa,r}}=\hat E_{\kappa,r},
\end{equation}
the same relative error that $\hat q_{\kappa,r}$ attains on $[1,\kappa]$. Moreover, in terms of the reference score $s_t$, the flow~\eqref{eq:flow_right_endpoint} reads
\begin{equation}
    \dot y_t=\frac{\dot\beta_t}{\beta_t}\left(y_t+\frac{1}{A_{\lambda_{\min}}(t)}\,s_{\tau(t)}\!\left(\frac{y_t}{A_{\lambda_{\min}}(t)}\right)\right),
    \qquad A_{\lambda_{\min}}(t)=\sqrt{\alpha_t^2+\lambda_{\min}\beta_t^2},
\end{equation}
where, writing $(\alpha_t,\beta_t)=(\cos\varphi(t),\sin\varphi(t))$, $\tau(t)=\varphi^{-1}\!\big(\arctan(\sqrt{\lambda_{\min}}\tan\varphi(t))\big)$. For the elliptic schedule,
\begin{equation}
    \tau(t)=\frac{1}{K(m)}\,F\!\left(\arctan\!\big[\sqrt{\lambda_{\min}}\,\scc(K(m)t\mid m)\big]\;\middle|\;m\right).
\end{equation}
\end{proposition}

\begin{proof}
\emph{Reduction to a single eigenmode.} Write $\alpha_i\triangleq\alpha(i/(2r+1))$ and $\beta_i\triangleq\beta(i/(2r+1))$ for $i=1,\dots,2r$, so that the query times and half-steps of Definition~\ref{def:elliptic_finite_diff} are $t_j=\tfrac{2j-1}{2r+1}$ and $t_j+\tfrac h2=\tfrac{2j}{2r+1}$, and $\gamma_i=\beta_i^2/\alpha_i^2$ zeros and poles. The rescaled target $\sqrt{\lambda_{\min}}\,x_1$ is $\mathcal N\big(0,(\Lambda/\lambda_{\min})^{-1}\big)$. Fix an eigenvalue $\lambda$ of $\Lambda$ and set $\mu\triangleq\lambda/\lambda_{\min}\in[1,\kappa]$. By the Gaussian score formula, the rescaled score acts on this eigenmode as
\begin{equation}
    \tilde s_t(y_\lambda)=-\frac{\mu}{\mu\alpha_t^2+\beta_t^2}\,y_\lambda .
\end{equation}
The same expression follows from~\eqref{eq:Gaussian-score-formula}: with $s_\tau(x_\lambda)=-\lambda x_\lambda/(\lambda\alpha_\tau^2+\beta_\tau^2)$ and $(\alpha_{\tau(t)},\beta_{\tau(t)})=(\alpha_t,\sqrt{\lambda_{\min}}\beta_t)/A_{\lambda_{\min}}(t)$, one gets $A_{\lambda_{\min}}(t)^2(\lambda\alpha_{\tau(t)}^2+\beta_{\tau(t)}^2)=\lambda\alpha_t^2+\lambda_{\min}\beta_t^2=\lambda_{\min}(\mu\alpha_t^2+\beta_t^2)$.

\emph{One step.} Let $D_i(\mu)\triangleq\mu\alpha_i^2+\beta_i^2$. Since $\alpha_i^2+\beta_i^2=1$, we have $D_i(\mu)-\mu=(1-\mu)\beta_i^2$, hence $y_\lambda+\tilde s_{t_j}(y_\lambda)=\frac{(1-\mu)\beta_{2j-1}^2}{d_{2j-1}(\mu)}\,y_\lambda$. The finite-difference update with coefficient $(\beta_{2j}^2-\beta_{2j-1}^2)/\beta_{2j-1}^2$ therefore gives
\begin{equation}
    y_{\lambda,j}
    =\left[1+\frac{(1-\mu)(\beta_{2j}^2-\beta_{2j-1}^2)}{d_{2j-1}(\mu)}\right]y_{\lambda,j-1}
    =\frac{\mu+(1-\mu)\beta_{2j}^2}{d_{2j-1}(\mu)}\,y_{\lambda,j-1}
    =\frac{\mu\,\alpha_{2j}^2+\beta_{2j}^2}{\mu\,\alpha_{2j-1}^2+\beta_{2j-1}^2}\,y_{\lambda,j-1}.
\end{equation}
This identity holds for any VP schedule; the elliptic choice enters only in the next step.

\emph{All steps.} Since $D_i(\mu)=\alpha_i^2(\mu+\gamma_i)=(\mu+\gamma_i)/(1+\gamma_i)$,
\begin{equation}
    y_{\lambda,r}=\prod_{j=1}^r\frac{1+\gamma_{2j-1}}{1+\gamma_{2j}}\cdot\frac{\mu+\gamma_{2j}}{\mu+\gamma_{2j-1}}\;y_{\lambda,0}
    =\hat q_{\kappa,r}(\mu)\,y_{\lambda,0}
\end{equation}
by~\eqref{eq:zolo-normalized}, exactly as in Lemma~\ref{lma:zoloflow-normalized-zolotarev} with $\Lambda$ replaced by $\Lambda/\lambda_{\min}$. Reassembling the eigenmodes gives $y_r=\hat q_{\kappa,r}(\Lambda/\lambda_{\min})z$, and multiplying by $\lambda_{\min}^{-1/2}$ gives the claimed form of $x_r$.

\emph{Error.} For $\lambda\in[\lambda_{\min},\lambda_{\max}]$, $\mu=\lambda/\lambda_{\min}$ ranges over $[1,\kappa]$, and $\lambda^{1/2}Q_{\lambda_{\min},\kappa,r}(\lambda)=\mu^{1/2}\hat q_{\kappa,r}(\mu)$. Hence the maximum relative error over $[\lambda_{\min},\lambda_{\max}]$ equals that of $\hat q_{\kappa,r}$ over $[1,\kappa]$, which is $\hat E_{\kappa,r}=2E_{\kappa,r}/(1-E_{\kappa,r})$ by~\eqref{eq:zolo-normalized-error}.

\emph{Reference-score form.} Substituting the rescaled score formula \ref{prop:interpolant-conversion} into~\eqref{eq:flow_right_endpoint} gives the displayed ODE. The time change is defined by equality of noise-to-signal ratios, $\alpha_{\tau}/\beta_{\tau}=\alpha_t/(\sqrt{\lambda_{\min}}\beta_t)$, i.e.\ $\cot\varphi(\tau)=\cot\varphi(t)/\sqrt{\lambda_{\min}}$, i.e.\ $\tan\varphi(\tau)=\sqrt{\lambda_{\min}}\tan\varphi(t)$. For the elliptic schedule $\varphi(t)=\operatorname{am}(K(m)t\mid m)$, so $\tan\varphi(t)=\scc(K(m)t\mid m)$, and inverting the amplitude with the incomplete elliptic integral $F(\cdot\mid m)$ gives $K(m)\tau(t)=F\big(\arctan[\sqrt{\lambda_{\min}}\,\scc(K(m)t\mid m)]\mid m\big)$.
\end{proof}

\begin{corollary}[Same complexity as Zoloflow]
\label{cor:zoloflow_rescaled_complexity}
In the setting of Proposition~\ref{prop:zoloflow_rescaled}, let $\hat\pi\triangleq\mathcal L(x_r)$. Then
\begin{equation}
    d_{\mathrm{TV}}(\hat\pi,\pi)\le\sqrt{\tfrac d2}\,\hat E_{\kappa,r},
\end{equation}
and $d_{\mathrm{TV}}(\hat\pi,\pi)\le\varepsilon$ whenever $r\ge\frac{\ln(16\kappa)}{2\pi^2}\ln\!\big(\frac{8\sqrt{2d}}{\varepsilon}\big)-\frac12$. In particular $r=O\big(\ln\kappa\,\ln(\sqrt d/\varepsilon)\big)$ with $\kappa=\lambda_{\max}/\lambda_{\min}$, independently of the scale $\lambda_{\min}$.
\end{corollary}

\begin{proof}
Total variation is invariant under the common bijection $x\mapsto\sqrt{\lambda_{\min}}\,x$, which maps $\hat\pi$ to $\mathcal L(y_r)$ and $\pi$ to $\mathcal N\big(0,(\Lambda/\lambda_{\min})^{-1}\big)$. By Proposition~\ref{prop:zoloflow_rescaled}, $y_r=\hat q_{\kappa,r}(\Lambda/\lambda_{\min})z$ is exactly the output of Zoloflow with parameters $(\kappa,r)$ on a centered Gaussian target whose precision satisfies $I_d\preceq\Lambda/\lambda_{\min}\preceq\kappa I_d$.
Theorem~\ref{thm:zoloflow-Gaussian-bound} 
therefore applies verbatim and gives both claims.
\end{proof}

\subsection{Gauge rescaling}
\label{sec:gauge-rescaling}
Fix a VP interpolant schedule $(\alpha,\beta)$ and a scale parameter $\lambda_{\min}>0$.

\begin{definition}[Gauge]
A gauge is a strictly positive function $g\in C^1([0,1])$ satisfying $g_0=1$ and $g_1=\lambda_{\min}^{-1/2}$. The corresponding gauge-rescaled schedule is
\begin{equation}
\label{eq:gauge-rescaled-schedule}
    \alpha_t^* \triangleq g_t\alpha_t,
    \qquad
    \beta_t^* \triangleq \sqrt{\lambda_{\min}}\,g_t\beta_t,
    \qquad t\in[0,1].
\end{equation}
These endpoint conditions ensure $(\alpha_0^*,\beta_0^*)=(1,0)$ and $(\alpha_1^*,\beta_1^*)=(0,1)$.
\end{definition}

For independent $z\sim\mathcal N(0,I_d)$ and $x_1\sim\pi$, the gauge-rescaled interpolant satisfies
\begin{equation}
    x_t^*=\alpha_t^*z+\beta_t^*x_1
    =g_t\underbrace{\left(\alpha_tz+\sqrt{\lambda_{\min}}\beta_tx_1\right)}_{y_t}.
\end{equation}
Thus it is a time-dependent spatial rescaling of the target interpolant $y_t$, with $x_0^*=z$ and $x_1^*=x_1$. Every gauge preserves the shifted noise-to-signal ratio $\alpha_t^*/\beta_t^*=\alpha_t/(\sqrt{\lambda_{\min}}\beta_t)$ for $t\in(0,1)$.

Let $p_t^*$ denote the density of $x_t^*$ and define $s_t^*(x)\triangleq\nabla_x\log p_t^*(x)$. Using $\alpha_t^2+\beta_t^2=1$, the associated probability-flow ODE becomes
\begin{equation}
\label{eq:gauge-probability-flow}
    \dot X_t=\frac{\dot g_t}{g_t}X_t
    +\frac{\dot\beta_t}{\beta_t}\left(X_t+g_t^2s_t^*(X_t)\right),
    \qquad t\in(0,1),\qquad X_0=z.
\end{equation}
Equivalently, one can use \ref{prop:interpolant-conversion} to write the ODE in terms of the original score. We consider two choices of gauge.

\subsubsection{VP gauge}
\label{sec:scaled-Gaussian-reference}
The choice
\begin{equation}
    g_t^{\mathrm{VP}}\triangleq \frac{1}{A_{\lambda_{\min}}(t)} = \frac{1}{\sqrt{\alpha_t^2+\lambda_{\min}\beta_t^2}}
\end{equation}
gives the variance-preserving schedule
\begin{equation}
    \label{eq:VPgauge-schedule}
    \alpha_t^*=\frac{\alpha_t}{A_{\lambda_{\min}}(t)},
    \qquad
    \beta_t^*=\frac{\sqrt{\lambda_{\min}}\beta_t}{A_{\lambda_{\min}}(t)}.
\end{equation}
The associated probability-flow ODE simplifies to
\begin{equation}
    \dot X_t
    =\frac{\dot\beta_t}{\beta_t\left(\alpha_t^2+\lambda_{\min}\beta_t^2\right)}
      \left(X_t+s_t^*(X_t)\right),
    \qquad t\in(0,1),\qquad X_0=z.
\end{equation}

\begin{proposition}[Gaussian approximation under the VP gauge]
\label{prop:vp-gauge}
Let $\pi=\mathcal N(0,\Lambda^{-1})$ with $\lambda_{\min}I_d\preceq\Lambda\preceq\lambda_{\max}I_d$, set $\kappa\triangleq\lambda_{\max}/\lambda_{\min}$, and let $(\alpha_t^*,\beta_t^*)$ be the VP-gauge schedule \eqref{eq:VPgauge-schedule} of the elliptic base schedule with parameter $\kappa$. Run the finite-difference discretization of Definition~\ref{def:elliptic_finite_diff} with the schedule $(\alpha^*,\beta^*)$ itself with the exact scores $s_t^*$ of the interpolant $x_t^*=\alpha_t^*z+\beta_t^*x_1$ for $r$ steps. Then $x_r=Q_{\lambda_{\min},\kappa,r}(\Lambda)z$, where
\begin{equation}
    Q_{\lambda_{\min},\kappa,r}(\lambda)
    \triangleq\frac{q_{\kappa,r}(\lambda/\lambda_{\min})}{q_{\kappa,r}(1/\lambda_{\min})}.
\end{equation}
Here $q_{\kappa,r}$ denotes the Zolotarev inverse-square-root approximant \eqref{eq:zolo-product} on $[1,\kappa]$, with relative error $E_{\kappa,r}$. If $1\in[\lambda_{\min},\lambda_{\max}]$, equivalently $1/\lambda_{\min}\in[1,\kappa]$, then
\begin{equation}
    \max_{\lambda\in[\lambda_{\min},\lambda_{\max}]}
    \left|\sqrt{\lambda}\,Q_{\lambda_{\min},\kappa,r}(\lambda)-1\right|
    \leq\frac{2E_{\kappa,r}}{1-E_{\kappa,r}}.
\end{equation}
\end{proposition}

\begin{proof}
Write $\alpha_i\triangleq\alpha(i/(2r+1))$, $\beta_i\triangleq\beta(i/(2r+1))$ and $D_i(\mu)\triangleq\mu\alpha_i^2+\beta_i^2$ for $i=1,\dots,2r$, so that the query time $t_j$ and the half-step $t_j+h/2$ of the $j$-th update correspond to $i=2j-1$ and $i=2j$; define $\alpha_i^*$, $\beta_i^*$ and $D_i^*(\lambda)\triangleq\lambda\alpha_i^{*2}+\beta_i^{*2}$ in the same way for the VP-gauge schedule.

\emph{One step.} Fix an eigenvalue $\lambda$ of $\Lambda$ and let $x_{\lambda,j}$ denote the corresponding component of $x_j$. The interpolant $x_t^*$ has law $\mathcal N(0,\alpha_t^{*2}I_d+\beta_t^{*2}\Lambda^{-1})$, so its score acts on this eigenmode as $s_t^*(x_\lambda)=-\lambda x_\lambda/(\lambda\alpha_t^{*2}+\beta_t^{*2})$. Since $\alpha_t^{*2}+\beta_t^{*2}=1$, the one-step identity in the proof of Proposition~\ref{prop:zoloflow_rescaled}, which uses only the VP property, gives
\begin{equation}
    x_{\lambda,j}=\frac{D_{2j}^*(\lambda)}{D_{2j-1}^*(\lambda)}\,x_{\lambda,j-1},
    \qquad j=1,\dots,r.
\end{equation}

\emph{The gauge.} Put $\mu\triangleq\lambda/\lambda_{\min}$ and $\mu_0\triangleq1/\lambda_{\min}$. By \eqref{eq:VPgauge-schedule},
\begin{equation}
    D_i^*(\lambda)
    =\frac{\lambda\alpha_i^2+\lambda_{\min}\beta_i^2}{\alpha_i^2+\lambda_{\min}\beta_i^2}
    =\frac{\lambda_{\min}\,(\mu\alpha_i^2+\beta_i^2)}{\lambda_{\min}\,(\mu_0\alpha_i^2+\beta_i^2)}
    =\frac{D_i(\mu)}{D_i(\mu_0)}.
\end{equation}

\emph{All steps.} Multiplying the one-step factors,
\begin{equation}
    x_{\lambda,r}
    =\left.\prod_{j=1}^r\frac{D_{2j}(\mu)}{D_{2j-1}(\mu)}\middle/\prod_{j=1}^r\frac{D_{2j}(\mu_0)}{D_{2j-1}(\mu_0)}\right.\;x_{\lambda,0}
    =\frac{\widehat q_{\kappa,r}(\mu)}{\widehat q_{\kappa,r}(\mu_0)}\,x_{\lambda,0}
    =\frac{q_{\kappa,r}(\mu)}{q_{\kappa,r}(\mu_0)}\,x_{\lambda,0}.
\end{equation}
The second equality is the identity $\prod_{j}D_{2j}(\mu)/D_{2j-1}(\mu)=\widehat q_{\kappa,r}(\mu)$ of Lemma~\ref{lma:zoloflow-normalized-zolotarev} for the elliptic base schedule, which holds for every $\mu>0$; the third cancels the common factor $1/q_{\kappa,r}(1)$ in \eqref{eq:zolo-normalized}. Reassembling the eigenmodes gives $x_r=Q_{\lambda_{\min},\kappa,r}(\Lambda)z$.

\emph{Error.} Let $e(\mu)\triangleq\sqrt\mu\,q_{\kappa,r}(\mu)-1$, so that $|e(\mu)|\le E_{\kappa,r}$ for $\mu\in[1,\kappa]$ by \eqref{eq:zolo-minimax app}. For $\lambda\in[\lambda_{\min},\lambda_{\max}]$ we have $\mu\in[1,\kappa]$, and $1\in[\lambda_{\min},\lambda_{\max}]$ is exactly $\mu_0\in[1,\kappa]$. Since $\sqrt\lambda=\sqrt\mu/\sqrt{\mu_0}$,
\begin{equation}
    \sqrt\lambda\,Q_{\lambda_{\min},\kappa,r}(\lambda)
    =\frac{\sqrt\mu\,q_{\kappa,r}(\mu)}{\sqrt{\mu_0}\,q_{\kappa,r}(\mu_0)}
    =\frac{1+e(\mu)}{1+e(\mu_0)}
    \in\left[\frac{1-E_{\kappa,r}}{1+E_{\kappa,r}},\,\frac{1+E_{\kappa,r}}{1-E_{\kappa,r}}\right],
\end{equation}
where $E_{\kappa,r}<1$ (see the proof of Corollary~\ref{cor:e_hat_rates}). Both endpoints of this interval lie within $2E_{\kappa,r}/(1-E_{\kappa,r})$ of $1$, which proves the bound.
\end{proof}

\subsubsection{Exponential gauge}
\label{sec:exponential-rescaling}
The choice $g_t^{\mathrm{exp}}\triangleq\lambda_{\min}^{-t/2}$ gives the generally non-VP schedule
\begin{equation}
    \alpha_t^*=\lambda_{\min}^{-t/2}\alpha_t,
    \qquad
    \beta_t^*=\lambda_{\min}^{(1-t)/2}\beta_t.
\end{equation}
Its probability-flow ODE is
\begin{equation}
    \dot X_t=-\frac{\log\lambda_{\min}}{2}X_t
    +\frac{\dot\beta_t}{\beta_t}\left(X_t+\lambda_{\min}^{-t}s_t^*(X_t)\right),
    \qquad t\in(0,1),\qquad X_0=z.
\end{equation}
\paragraph{Generalized Zoloflow.}
We use an exponential integrator-like discretization of the exponential-gauge ODE: the score-dependent term is approximated by the elliptic finite-difference rule, while the linear term $-\tfrac12\log(\lambda_{\min})X_t$ is integrated exactly. Distributing this linear evolution equally across $r$ updates gives the factor
\begin{equation}
    \exp\!\left(-\frac{\log\lambda_{\min}}{2r}\right)
    =\lambda_{\min}^{-1/(2r)}
\end{equation}
per update.

\begin{definition}[Generalized Zoloflow]
\label{def:generalized-zoloflow}
Fix $\lambda_{\min}>0$, $\kappa>1$, and an integer $r\geq1$. Let $(\alpha_t,\beta_t)=(\cn(K(m)t\mid m),\sn(K(m)t\mid m))$, where $m\triangleq1-\kappa^{-1}$, and let $s_t^*$ denote the score under the exponential-gauge schedule. Define
\begin{equation}
    h\triangleq\frac{2}{2r+1},
    \qquad
    t_j\triangleq\frac{2j-1}{2r+1},
    \qquad
    \omega_j\triangleq\frac{\beta_{t_j+h/2}^2-\beta_{t_j}^2}{\beta_{t_j}^2},
    \qquad j=1,\ldots,r.
\end{equation}
Starting from $x_0=z\sim\mathcal N(0,I_d)$, Generalized Zoloflow performs the updates
\begin{equation}
\label{eq:generalized-zoloflow-update}
    x_j=\lambda_{\min}^{-1/(2r)}
    \left[x_{j-1}+\omega_j\left(x_{j-1}+\lambda_{\min}^{-t_j}s_{t_j}^*(x_{j-1})\right)\right],
    \qquad j=1,\ldots,r,
\end{equation}
and returns $x_r$. The finite-difference weights use the base schedule $\beta$, rather than the transformed schedule $\beta^*$.
\end{definition}

For a centered Gaussian target $\pi=\mathcal N(0,\Lambda^{-1})$ with $\lambda_{\min}I_d\preceq\Lambda\preceq\lambda_{\min}\kappa I_d$, exact scores satisfy, along a precision eigenmode $\lambda$,
\begin{equation}
    \lambda_{\min}^{-t}s_t^*(x_\lambda)
    =-\frac{\lambda/\lambda_{\min}}{(\lambda/\lambda_{\min})\alpha_t^2+\beta_t^2}\,x_\lambda.
\end{equation}
Consequently, the eigenmode update is
\begin{equation}
    x_{\lambda,j}
    =\lambda_{\min}^{-1/(2r)}
    \frac{(\lambda/\lambda_{\min})\alpha_{t_j+h/2}^2+\beta_{t_j+h/2}^2}
         {(\lambda/\lambda_{\min})\alpha_{t_j}^2+\beta_{t_j}^2}
    x_{\lambda,j-1}.
\end{equation}
Defining the positive zero and pole shifts
\begin{equation}
    \xi_j\triangleq\frac{\beta_{t_j+h/2}^2}{\alpha_{t_j+h/2}^2},
    \qquad
    \zeta_j\triangleq\frac{\beta_{t_j}^2}{\alpha_{t_j}^2},
\end{equation}
this becomes
\begin{equation}
\label{eq:generalized-zoloflow-eigenmode}
    x_{\lambda,j}
    =\lambda_{\min}^{-1/(2r)}
    \frac{\lambda/\lambda_{\min}+\xi_j}{\lambda/\lambda_{\min}+\zeta_j}
    \frac{1+\zeta_j}{1+\xi_j}
    x_{\lambda,j-1}.
\end{equation}
Thus the output is
\begin{equation}
\label{eq:generalized-zoloflow-output}
    x_r=\lambda_{\min}^{-1/2}
    \widehat q_{\kappa,r}\!\left(\frac{\Lambda}{\lambda_{\min}}\right)z,
\end{equation}
where $\widehat q_{\kappa,r}$ is the endpoint-normalized Zolotarev approximant on $[1,\kappa]$.

\subsection{Centering}
\label{app:centering}
\Cref{ass:targets} takes the target to be centered; this is without loss of generality. Let $\tilde x_1\sim\pi$ be centered, and suppose the data instead have mean $\mu$, i.e.\ $x_1=\tilde x_1+\mu$. Write 
\begin{align}
    \tilde x_t=\alpha_t z+\beta_t\tilde x_1, \quad &\text{(centered)} \\ 
    x_t=\alpha_t z+\beta_t x_1, \quad &\text{(not centered)}
\end{align}
for the two interpolants, with respective $\tilde s_t=\nabla\log\tilde p_t$ and $s_t=\nabla\log p_t$. Then $\tilde x_t = x_t - \beta_t\mu$. Additionally we can write the velocities $\tilde{b}_t(\tilde{x}) = \mathbb{E}[\dot \tilde{x}_t \mid \tilde{x}_t = \tilde{x}]$, and $b_t(x) = \mathbb{E}[\dot x_t\mid x_t = x]$. 

\begin{lemma}[Centered score]
\label{lem:centered-score}
For all $t\in(0,1)$ and $y\in\R^d$, \begin{equation}
    \tilde s_t(y)=s_t(y+\beta_t\mu).
\end{equation} Additionally, one can also write the corresponding velocity fields in terms of one another
\begin{equation}
    \tilde{b}_t(y) = b_t(y + \beta_t \mu) - \dot \beta_t \mu.
\end{equation}
\end{lemma}
\begin{proof}
Since $x_t$ is a deterministic translate of $\tilde x_t$, $p_t(x)=\tilde p_t(x-\beta_t\mu)$. Take $\nabla\log$ of both sides and substitute $x=y+\beta_t\mu$. The velocity expression follows the same logic.
\end{proof}

Each evaluation of $\tilde s_t$ therefore costs exactly one evaluation of $s_t$; note that the offset is $\beta_t\mu$, not $\mu$. Given the oracle $s_t$, one samples the law of $x_1$ in two steps.
\begin{enumerate}
    \item \textbf{Estimate the mean.} By Tweedie's formula,
    \begin{equation}
        \hat\mu_t(x)\triangleq\frac{x+\alpha_t^2 s_t(x)}{\beta_t}=\mathbb{E}[x_1\mid x_t=x].
        \label{eq:mean-estimate}
    \end{equation}
    As $t\to0$ the likelihood of $x_t$ given $x_1$ flattens, so the posterior mean tends to the prior mean: $\hat\mu_t(x)\to\mu$ for every fixed $x$ for targets with finite moments (e.g., translates of targets in $\mathsf P_\kappa$). A single score query thus recovers $\mu$ to any desired accuracy.
    \item \textbf{Sample the centered target, then translate.} Run any sampler for $\pi$ with the centered oracle $\tilde s_t(\cdot)=s_t(\cdot+\beta_t\hat\mu)$, producing $\hat{\tilde x}_1$, and output $\hat x_1=\hat{\tilde x}_1+\hat\mu$.
\end{enumerate}

\section{Parameter-selection heuristic for general distributions}
\label{app:cov-heuristic}

As described in \cref{sec:two-param-family}, implementing the elliptic schedule requires choosing two parameters:
\begin{enumerate}
    \item a center $\bar\lambda>0$, the scale set by target rescaling (\cref{app:approaches-for-handling-arbitrary-spectra}), which any schedule admits;
    \item a width $\kappa\ge1$, the condition number of \cref{def:ellipticschedule} (then set $m=1-\kappa^{-1}$), which is specific to the elliptic family.
\end{enumerate}
The paper mostly works with $(\bar\lambda,\kappa)$. Here we use the equivalent spectral endpoint parameterization $(\ell,L)=(\bar\lambda/\sqrt\kappa,\,\bar\lambda\sqrt\kappa)$, because the heuristic sets the endpoints directly. For Gaussians, optimal parameter choices for $(\ell,L)$ depend on curvature endpoints $-\nabla^2 \log \pi$ or equivalently the smallest and largest eigenvalues of the precision matrix $(\lambda_{\min}(\Lambda), \lambda_{\max}(\Lambda)),$ where
\begin{equation}
    \lambda_{\min} I_d \preceq -\nabla^2 \log \pi(x) = \Lambda \preceq \lambda_{\max}I_d.
\end{equation}
However, this equivalence decouples for general $\pi$, and so it's not obvious how to pick these parameters. For a general distribution $\pi$ with precision matrix $\Lambda_\pi = \Sigma^{-1}_\pi$ the two immediate extensions of these choices are 
\begin{enumerate}
    \item \textbf{Curvature endpoints}: \begin{equation}
        (\ell,L) = \left(\lambda_{\min}(-\nabla^2\log \pi),\lambda_{\max}(-\nabla^2 \log \pi)\right)
    \end{equation}
    \item \textbf{Precision matrix endpoints}  
    \begin{equation}
        (\ell, L) = \left(\lambda_{\min}(\Lambda), \lambda_{\max}(\Lambda)\right)
    \end{equation}
\end{enumerate}
The immediate issue with the curvature endpoints is generally $\ell < 0$ can occur which is prohibited. Nevertheless even in the log-concave setting we provide evidence that this choice is too conservative in \ref{app:logconcave-experiments}. We propose the precision matrix endpoints as our heuristic:

\paragraph{The heuristic.}
Let $x_1\sim\pi$ have mean $\mu$ and covariance $\Sigma=\mathbb E[(x_1-\mu)(x_1-\mu)^\top]$. We use the endpoints of the inverse covariance spectrum to define a reference setting for the target-rescaled elliptic schedule (\cref{app:target-rescaling}):
\begin{equation}
    \ell\;=\;\lambda_{\min}(\Sigma^{-1})\;=\;\frac{1}{\lambda_{\max}(\Sigma)},
    \qquad
    L\;=\;\lambda_{\max}(\Sigma^{-1})\;=\;\frac{1}{\lambda_{\min}(\Sigma)},
    \qquad 
    \kappa\;=\;\frac L \ell.
    \label{eq:cov-heuristic}
\end{equation}
In center--width form, $(\bar\lambda,\kappa)=\bigl(1/\sqrt{\lambda_{\min}(\Sigma)\lambda_{\max}(\Sigma)},\,\lambda_{\max}(\Sigma)/\lambda_{\min}(\Sigma)\bigr)$, as stated in \cref{sec:two-param-family}.
The heuristic has three properties that motivate it:
\begin{enumerate}
    \item \emph{It is exact in the Gaussian case.} If $x_1\sim\mathcal N(m,\Lambda^{-1})$ then $\Sigma^{-1}=\Lambda$, so \eqref{eq:cov-heuristic} returns the endpoints of the precision spectrum, which is the optimal choice of \cref{sec:Gaussian-sampling}.
    \item \emph{It matches isotropic covariance by rescaling.} If $\Sigma=\sigma^2 I$, then $\ell=L=\sigma^{-2}$ and the schedule is trig rescaled to the data variance, coinciding with the closed-form isotropic Gaussian schedule of \citet{sabour_align_2024}. The heuristic assigns width one in this case; this need not be the best width for a non-Gaussian target with isotropic covariance.
    \item \emph{It is bounded by the curvature.} For strongly log-concave $\pi$, we have 
    \begin{equation}
        \lambda_{\min}\le\lambda_{\min}(\Sigma^{-1}) \leq \lambda_{\max}(\Sigma^{-1})\le\lambda_{\max},
    \end{equation}
so the interval \eqref{eq:cov-heuristic} always lies inside $[\lambda_{\min},\lambda_{\max}]$ and is never wider than the curvature recipe.
\end{enumerate}

For general non-Gaussian targets, these properties motivate an initial parameter choice for a given solver and query budget. The covariance center and width can then be adjusted separately, or equivalently one can tune the endpoints $(\ell,L)$. The five toy experiments in \Cref{app:toy2d} compare independently tuned elliptic, trigonometric, and linear schedules on a shared endpoint grid. The covariance-derived minimum precision is marked as a reference row, without fixing the upper endpoint.

In practice, we may not have access to the exact covariance $\Sigma$, so we recommend computing the sample covariance $\hat{\Sigma}$, taking the 0.5th and 99.5th percentiles of the spectrum $\lambda_{\min}^{(0.5)}(\hat{\Sigma})$ and $\lambda_{\max}^{(99.5)}(\hat{\Sigma})$ to calculate 
\begin{equation}
    (\ell, L) = \left(\frac{1}{\lambda_{\max}^{(99.5)}}, \frac{1}{\lambda_{\min}^{(0.5)}}\right),
\end{equation}
and using these values of $\ell$ and $L$ for the target-rescaled elliptic schedule. Alternatively, it can serve to guide a more in-depth hyperparameter sweep. 

\section{Interpolant schedules}
\label{app:schedules}

\subsection{Interpolant schedule catalog}
\label{app:schedule-catalog}
We collect the schedules used in the paper. Their dependence on target scale and spectral width is discussed in \cref{sec:two-param-family}; the following subsections explain their conversions and the elliptic schedule's Lipschitz properties. 
\begin{itemize}
    \item \textbf{Linear Schedule:} \begin{equation}\alpha^{lin}(t) = {1 - t}, \quad \beta^{lin}(t) = t.\end{equation}
    \item \textbf{Square Root Schedule:} \begin{equation}\label{eq:square-root-schedule}\alpha^{sqrt}(t) = \sqrt{1 - t}, \quad \beta^{sqrt}(t) = \sqrt t.\end{equation}
    \item \textbf{Trigonometric Schedule:} \begin{equation}\label{eq:trigonometric-schedule}\alpha^{trig}(t) = \cos\left(\frac 1 2 \pi t\right), \quad \beta^{trig}(t) = \sin\left(\frac 1 2 \pi t\right).\end{equation}
    \item \textbf{Elliptic Schedule:} For a given $\kappa \geq 1$, with $m = 1 - \kappa^{-1}$ and $K = K(m)$, \begin{equation}\label{eq:appendix-elliptic-schedule}\alpha^{ell}(t) = \cn(K t \ |\ m), \quad \beta^{ell}(t) = \sn(K t \ |\ m).\end{equation}
    \item \textbf{Lipschitz-Guided Schedule:} For a given $\kappa > 1$, \begin{equation}\label{eq:drift-lipschitz-schedule}\alpha^{lip}(t) = \sqrt{\frac{\kappa^{1-t}-1}{\kappa-1}}, \quad \beta^{lip}(t) = \sqrt{\frac{\kappa-\kappa^{1-t}}{\kappa-1}}.\end{equation} This schedule was introduced in \cite{chen2025lipschitz}.

    \item \textbf{EDM Schedule (VP-normalized):} For given $\sigma_{\max} > \sigma_{\min} > 0$ and $\rho > 0$, let
    \begin{equation}
      \sigma_\rho(t) = \left((1-t)\sigma_{\max}^{1/\rho} + t\sigma_{\min}^{1/\rho}\right)^{\rho}.
    \end{equation}
    The VP normalization of the EDM noise schedule \citep{karras2022elucidating} is
    \begin{equation}
      \alpha^{edm}(t) = \frac{\sigma_\rho(t)}{\sqrt{1+\sigma_\rho(t)^2}}, \quad
      \beta^{edm}(t) = \frac{1}{\sqrt{1+\sigma_\rho(t)^2}}.
    \end{equation}
    \begin{sloppypar}
    Thus $(\alpha^{edm}(t))^2+(\beta^{edm}(t))^2=1$, and the signal-to-noise ratio is
    $\operatorname{SNR}(t)=\allowbreak(\beta^{edm}(t)/\alpha^{edm}(t))^2=\allowbreak\sigma_\rho(t)^{-2}$.
    With finite positive noise endpoints, this traces a truncated VP path: its endpoint SNRs are $\sigma_{\max}^{-2}$ and $\sigma_{\min}^{-2}$.
    \end{sloppypar}
\end{itemize}

\subsection{Conversions between interpolant schedules}

\begin{proposition}[Conversion between interpolant schedules]
\label{prop:interpolant-conversion}
Let $z\sim\mathcal N(0,I_d)$ and $x_1\sim\pi$ be independent, with $\mathbb E\|x_1\|<\infty$. Let $\alpha,\beta,\alpha^*,\beta^*\in C^1((0,1))$ be positive coefficient functions.
For each schedule, define the interpolant and its associated density, score, velocity, and noise-to-signal ratio as follows:
\begin{center}
\begingroup
\normalfont
\renewcommand{\arraystretch}{1.35}
\setlength{\tabcolsep}{5pt}
\begin{tabular}{@{}lll@{}}
\toprule
Interpolant & $x_t=\alpha_t z+\beta_t x_1$ & $x_t^*=\alpha_t^*z+\beta_t^*x_1$ \\
Density & $p_t$ & $p_t^*$ \\
Score & $s_t=\nabla\log p_t$ & $s_t^*=\nabla\log p_t^*$ \\
Velocity & $b_t(x)=\mathbb E[\dot x_t\mid x_t=x]$ & $b_t^*(x)=\mathbb E[\dot x_t^*\mid x_t^*=x]$ \\
Noise-to-signal ratio & $\sigma(t)\triangleq\alpha_t/\beta_t$ & $\sigma^*(t)\triangleq\alpha_t^*/\beta_t^*$ \\
\bottomrule
\end{tabular}
\endgroup
\end{center}

Assume that
\[
    \dot\sigma(t)\neq0 \quad\text{for all }t\in(0,1),
    \qquad \sigma^*((0,1))\subseteq\sigma((0,1)).
\]
Then the time change and spatial scaling
\[
    \tau(t)\triangleq\sigma^{-1}(\sigma^*(t)),
    \qquad c_t\triangleq\frac{\beta_t^*}{\beta_{\tau(t)}}
    =\frac{\alpha_t^*}{\alpha_{\tau(t)}}
\]
are well-defined and $C^1$. For $t\in(0,1)$, the conversions are given below, where $X_u$ denotes any solution of $\dot X_u=b_u(X_u)$. The last column applies when both schedules are variance-preserving (VP).
\begin{equation*}
\boxed{
\begin{array}{c|c|c|c}
    & \text{Object} & \text{General schedules} & \text{Both VP }(c_t=1) \\[4pt]
    \hline
    \textbf{Interpolant} & x_t^* & c_t x_{\tau(t)} & x_{\tau(t)} \\[6pt]
    \textbf{ODE solution} &X_t^* & c_t X_{\tau(t)} & X_{\tau(t)} \\[6pt]
    \textbf{Score} &s_t^*(x) & c_t^{-1}s_{\tau(t)}(x/c_t) & s_{\tau(t)}(x) \\[8pt]
    \begin{gathered}
        \textbf{Velocity field}\\
        \text{(from velocity)}
    \end{gathered}
    & b_t^*(x)
    & \dfrac{\dot c_t}{c_t}x+c_t\dot\tau(t)b_{\tau(t)}(x/c_t)
    & \dot\tau(t)b_{\tau(t)}(x) \\[12pt]
    \begin{gathered}
        \textbf{Velocity field}\\
        \text{(from score)}
    \end{gathered}
    & b_t^*(x)
    & \begin{aligned}
        &\dfrac{\dot\beta_t^*}{\beta_t^*}x
        +\dfrac{(\alpha_t^*)^2}{c_t}
        \left(\dfrac{\dot\beta_t^*}{\beta_t^*}
        -\dfrac{\dot\alpha_t^*}{\alpha_t^*}\right)
        s_{\tau(t)}(x/c_t)
    \end{aligned}
    & \dfrac{\dot\beta_t^*}{\beta_t^*}
      \bigl(x+s_{\tau(t)}(x)\bigr)
\end{array}
}
\end{equation*}
The derivatives appearing in the table are
\[
    \frac{\dot c_t}{c_t}
    =\frac{\dot\alpha_t^*}{\alpha_t^*}
    -\dot\tau(t)\frac{\dot\alpha_{\tau(t)}}{\alpha_{\tau(t)}},
    \qquad
    \dot\tau(t)=\frac{(\sigma^*)'(t)}{\sigma'(\tau(t))}.
\]

To explain these conversions, first note that the time change matches the noise-to-signal ratios, and hence the SNRs, of the two schedules:
    \begin{equation}
        \sigma(\tau(t)) = \frac{\alpha_{\tau(t)}}{\beta_{\tau(t)}} = \frac{\alpha_t^*}{\beta_t^*} = \sigma^*(t).
    \end{equation}
The spatial rescaling then gives the interpolant, score, and velocity identities
\begin{align}
    x_t^* &= c_t x_{\tau(t)},\\
    s_t^*(x) &= \frac{1}{c_t}s_{\tau(t)}(x/c_t),\\
    b_t^*(x) &= \frac{\dot c_t}{c_t}x
    +c_t\dot\tau(t)b_{\tau(t)}(x/c_t).
\end{align}
Consequently, any solution $\dot X_u=b_u(X_u)$ on $(0,1)$ yields a solution $X_t^*\triangleq c_tX_{\tau(t)}$ of $\dot X_t^*=b_t^*(X_t^*)$. 
Finally, if both coefficient pairs are variance-preserving, then $c_t=1$, so
\begin{equation}
    s_t^*(x)=s_{\tau(t)}(x),
    \qquad b_t^*(x)=\dot\tau(t)b_{\tau(t)}(x).
\end{equation}
\end{proposition} 

Proposition~\ref{prop:interpolant-conversion} shows that, under its assumptions, two linear stochastic interpolants with the same target $x_1\sim\pi$ and Gaussian reference $z\sim\mathcal N(0,I_d)$ are related by a change of time and a rescaling of space. The time change $\tau$ is chosen so that the two schedules have the same noise-to-signal ratio $\alpha/\beta$ at corresponding times, and the scores and velocity fields then transform accordingly. When both schedules are variance-preserving, the spatial rescaling is trivial ($c_t=1$) and only the time change remains.

\subsection{Bijection between VP interpolant schedules and noise schedules}
\label{app:si-is-ve}

\textbf{VP interpolant schedules, noise schedules, and SNR schedules are in one-to-one correspondence.}
For positive coefficients on $t\in(0,1)$, the forward map is
\[
    \sigma_t=\frac{\alpha_t}{\beta_t},
    \qquad
    \operatorname{SNR}(t)=\frac{\beta_t^2}{\alpha_t^2}=\sigma_t^{-2}.
\]
Conversely, specifying either the noise level $\sigma_t$ or the SNR uniquely determines the VP coefficients:
\begin{equation}
    (\alpha_t,\beta_t)
    =\frac{(\sigma_t,1)}{\sqrt{1+\sigma_t^2}}
    =\frac{\bigl(1,\sqrt{\operatorname{SNR}(t)}\bigr)}
    {\sqrt{1+\operatorname{SNR}(t)}}.
    \label{eq:ve-to-vp-schedule}
\end{equation}
Thus a noise schedule specifies the time-varying SNR; its squared noise level $\sigma_t^2$ is the inverse SNR. Every positive interpolant schedule defines an SNR, and the VP constraint selects a unique interpolant schedule for each SNR. Without that constraint, a common rescaling of $\alpha_t$ and $\beta_t$ leaves the SNR unchanged.

In our noise-to-data orientation, full VP schedules correspond to strictly decreasing noise clocks with $\sigma_t\to\infty$ as $t\downarrow0$ and $\sigma_t\to0$ as $t\uparrow1$; equivalently, SNR increases from $0$ to $\infty$. The VP coefficients extend to the endpoints by these limits. Finite positive noise endpoints instead describe a truncated VP path.

\paragraph{Why the correspondence holds.}
Let $x_t=\alpha_tz+\beta_tx_1$ follow a VP interpolant. Dividing by the data coefficient gives the corresponding VE-corrupted variable:
\begin{equation}
    y_t\triangleq\frac{x_t}{\beta_t}
    =x_1+\sigma_tz,
    \qquad \sigma_t=\frac{\alpha_t}{\beta_t}.
    \label{eq:vp-to-ve-noise-clock}
\end{equation}
Conversely, normalizing the coefficients $(\sigma_t,1)$ of $x_1+\sigma_tz$ to have squared sum one gives \eqref{eq:ve-to-vp-schedule}. These operations are inverses, establishing the bijection. The VP and VE states are related by $x_t=\beta_ty_t$, so their probability-flow ODEs transform accordingly: if $\dot X_t=b_t(X_t)$, then $Y_t=X_t/\beta_t$ satisfies
\begin{equation}
    \dot Y_t
    =\frac{1}{\beta_t}b_t(\beta_tY_t)
     -\frac{\dot\beta_t}{\beta_t}Y_t.
    \label{eq:vp-to-ve-flow-dilation}
\end{equation}

An increasing VE grid $\sigma_{\min}=\sigma_0<\cdots<\sigma_n=\sigma_{\max}$ therefore corresponds to the reverse-ordered interpolant times $t_i=\sigma^{-1}(\sigma_i)$. The clock fixes the correspondence between query times; the numerical update rule is an additional choice. The following AYS example makes this conversion explicit.

\subsubsection{Gaussian AYS as a special case of the elliptic schedule}
\label{app:ays-clock}
For isotropic Gaussian targets, the closed-form AYS noise grid is recovered by the VP form of the target-rescaled elliptic schedule at $\kappa=1$. This correspondence concerns the analytic Gaussian example in \citet[Theorem~3.1]{sabour_align_2024}, which motivates their later KL-divergence upper bound (KLUB) framework; the identification is with this analytic result, not with schedules obtained by numerical KLUB optimization. We show the correspondence by expressing the Gaussian AYS schedule as a VP interpolant and identifying its coefficients with those of the rescaled trigonometric schedule.

\paragraph{AYS in VP form.}
For $x_1\sim\mathcal N(0,c^2I_d)$ with $c>0$, this analytic schedule spaces the angle $\arctan(\sigma/c)$ uniformly:
\begin{equation}
    \sigma_i^\star
    =c\tan\!\left[
      \left(1-\frac{i}{n}\right)
      \arctan\!\left(\frac{\sigma_{\min}}{c}\right)
      +\frac{i}{n}
      \arctan\!\left(\frac{\sigma_{\max}}{c}\right)
    \right],
    \qquad i=0,\ldots,n.
    \label{eq:ays-Gaussian-grid}
\end{equation}
A continuous clock generating this grid in our noise-to-data orientation is, with $\theta(u)=\pi u/2$,
\begin{equation}
    \sigma^{\mathrm{AYS}}(u)=c\cot\theta(u),
    \qquad
    \arctan\!\left(\frac{\sigma^{\mathrm{AYS}}(u)}{c}\right)
    =\frac{\pi}{2}(1-u),
    \qquad u\in(0,1).
    \label{eq:ays-shifted-clock}
\end{equation}
Explicitly, define the interpolant-time grid by
\[
    u_i\triangleq1-\frac{2}{\pi}\arctan\!\left(\frac{\sigma_i^\star}{c}\right)
    =\left(1-\frac{i}{n}\right)u_0+\frac{i}{n}u_n,
    \qquad i=0,\ldots,n.
\]
Then $\sigma^{\mathrm{AYS}}(u_i)=\sigma_i^\star$ exactly. For $0<\sigma_{\min}<\sigma_{\max}<\infty$, the times are equally spaced on $[u_n,u_0]\subset(0,1)$, with $u_0>\cdots>u_n$. Sampling traverses $i=n,n-1,\ldots,0$, so $u$ increases as noise decreases. The grid is uniform in $u$; the noise levels themselves are generally nonuniform.

The conversion \eqref{eq:ve-to-vp-schedule} gives the VP coefficients
\begin{equation}\label{eq:ays-vp-interpolant}
    (\alpha_u^{\mathrm{AYS}},\beta_u^{\mathrm{AYS}})
    =\frac{(c\cos\theta(u),\sin\theta(u))}
    {\sqrt{c^2\cos^2\theta(u)+\sin^2\theta(u)}},
\end{equation}
which define the interpolant $x_u^{\mathrm{AYS}}=\alpha_u^{\mathrm{AYS}}z+\beta_u^{\mathrm{AYS}}x_1$, with $z\sim\mathcal N(0,I_d)$ independent of $x_1$.

\paragraph{Identification with elliptic at $\kappa=1$.}
At $\kappa=1$, the elliptic coefficients are $(\cos\theta(u),\sin\theta(u))$. Rescaling the target to $x_1/c$ (\cref{app:target-rescaling}) makes the coefficients relative to the original target $x_1$ equal to $(\cos\theta(u),c^{-1}\sin\theta(u))$. Their VP normalization is exactly \eqref{eq:ays-vp-interpolant}.
The uniform grid $\{u_i\}_{i=0}^n$ therefore realizes the Gaussian AYS schedule in this rescaled elliptic time. This identifies the schedule and query grid. The cited AYS optimality result concerns Euler/DDIM; Zoloflow uses the elliptic finite-difference update.

\subsection{Asymptotic drift-Lipschitz optimality of the elliptic schedule}
\label{app:lipschitz-schedules}

The Lipschitz-guided schedule of \citet{chen2025lipschitz} controls the spatial sensitivity of the velocity field. On the Gaussian class with precision spectra in $[1,\kappa]$, it attains the optimal worst-case drift-Lipschitz bound. The elliptic schedule, although chosen for analyticity, approaches the same optimum as $\kappa\to\infty$, both for the worst-case Lipschitz constant and for their averaged squared criterion. We make this connection precise below.

\subsubsection{The drift-Lipschitz criterion and its optimal value}

Consider $\pi=\mathcal N(0,\Lambda^{-1})$ with $I_d\preceq\Lambda\preceq\kappa I_d$, where $\kappa>1$. We optimize over the class $\widetilde{\mathcal S}$ of monotone VP schedules with $\alpha_0=\beta_1=1$, $\alpha_1=\beta_0=0$, positive coefficients on $(0,1)$, and
\begin{equation}
    \alpha,\beta\in C([0,1])\cap C^1((0,1)),
    \qquad \alpha^2,\beta^2\in C^1([0,1]).
    \label{eq:endpoint-relaxed-vp-class}
\end{equation}
This permits the square-root endpoint behavior of the Lipschitz-guided schedule.
For a precision eigenvalue $\lambda\in[1,\kappa]$, the normalized probability-flow solution and its drift multiplier are
\begin{equation}\label{eq:lip-solution-multiplier}
    x_\lambda(t)=\sqrt{\alpha_t^2+\beta_t^2/\lambda},
    \qquad
    g_\lambda(t)=\frac{d}{dt}\log x_\lambda(t),
    \qquad \dot x_\lambda(t)=g_\lambda(t)x_\lambda(t).
\end{equation}
Since the Gaussian velocity field is linear, its spatial Lipschitz constant is $\max_{\lambda\in\operatorname{spec}(\Lambda)}|g_\lambda(t)|$. Define the worst-case criterion
\begin{equation}\label{eq:drift-lipschitz-problem}
    L^b[\alpha,\beta]
    \triangleq\sup_{t\in(0,1)}\max_{\lambda\in[1,\kappa]}|g_\lambda(t)|.
\end{equation}
Writing $q_t=\beta_t^2$, the VP constraint gives
\[
    |g_\lambda(t)|
    =\frac{(1-\lambda^{-1})\dot q_t}
    {2\bigl(1-(1-\lambda^{-1})q_t\bigr)}.
\]
Because $\dot q_t\geq0$, this expression increases with $\lambda$, so the worst case is $\lambda=\kappa$. The endpoints $x_\kappa(0)=1$ and $x_\kappa(1)=\kappa^{-1/2}$ therefore imply
\begin{equation}\label{eq:lip-optimal-value}
    L^b[\alpha,\beta]
    =\sup_t|g_\kappa(t)|
    \geq\int_0^1|g_\kappa(t)|\,dt
    =\tfrac12\log\kappa
    \triangleq d_\kappa.
\end{equation}
The bound is attained by $x_\kappa(t)=\kappa^{-t/2}$, for which $g_\kappa(t)=-d_\kappa$ is constant. Thus the optimal value is $d_\kappa$.

\subsubsection{The Lipschitz-guided optimizer and its connection to prior work}

Solving $\alpha_t^2+\beta_t^2=1$ and $\alpha_t^2+\beta_t^2/\kappa=\kappa^{-t}$ gives exactly the Lipschitz-guided schedule in \eqref{eq:drift-lipschitz-schedule}. Equality in \eqref{eq:lip-optimal-value} forces this constant logarithmic speed, so the optimizer is unique for $\kappa>1$.
It coincides with the Gaussian optimizer of \citet[Proposition~3.9 and Equation~(3.11)]{chen2025lipschitz}, taking their smallest covariance eigenvalue to be $\kappa^{-1}$.

Their criterion is the averaged squared spatial Lipschitz constant,
\begin{equation}\label{eq:chen-averaged-lipschitz}
    A_2[\alpha,\beta;\pi]
    \triangleq\int_0^1\mathbb E_{X\sim p_t}
    \bigl[\|\nabla_x b_t(X)\|_{\mathrm{op}}^2\bigr]dt.
\end{equation}
For Gaussian targets the Jacobian is independent of $X$. Taking the worst case over $I_d\preceq\Lambda\preceq\kappa I_d$ and applying Cauchy--Schwarz yields
\begin{equation}\label{eq:averaged-lipschitz-bound}
    A_2^{\mathrm{wc}}[\alpha,\beta]
    \triangleq\sup_{I_d\preceq\Lambda\preceq\kappa I_d}
    A_2[\alpha,\beta;\mathcal N(0,\Lambda^{-1})]
    =\int_0^1|g_\kappa(t)|^2dt
    \geq d_\kappa^2.
\end{equation}
The Lipschitz-guided schedule attains equality. It therefore minimizes both $L^b$ and $A_2^{\mathrm{wc}}$.

\subsubsection{The elliptic schedule approaches the same optimum}

For the elliptic schedule, with $m=1-\kappa^{-1}$, the worst-mode solution is $x_\kappa^{\mathrm{ell}}(t)=\dn(K(m)t\mid m)$. Hence
\begin{equation}
    g_\kappa^{\mathrm{ell}}(t)
    =-K(m)m\frac{\sn(K(m)t\mid m)\cn(K(m)t\mid m)}
    {\dn(K(m)t\mid m)}.
\end{equation}
Maximizing its magnitude, attained when $\sn^2(K(m)t\mid m)=1/(1+\sqrt{1-m})$, gives
\begin{equation}\label{eq:schedule-drift-comparison}
    L_{\mathrm{ell}}^b=(1-\kappa^{-1/2})K(m),
    \qquad L_{\mathrm{opt}}^b=d_\kappa=\tfrac12\log\kappa.
\end{equation}
Since $K(m)\sim\tfrac12\log(16\kappa)$ as $\kappa\to\infty$,
\begin{equation}\label{eq:elliptic-lipschitz-ratio}
    \frac{L_{\mathrm{ell}}^b}{L_{\mathrm{opt}}^b}
    =\frac{2(1-\kappa^{-1/2})K(m)}{\log\kappa}
    \longrightarrow1.
\end{equation}
The averaged squared criterion has the same conclusion, because
\begin{equation}
    1\leq\frac{A_2^{\mathrm{wc}}[\alpha^{\mathrm{ell}},\beta^{\mathrm{ell}}]}
    {d_\kappa^2}
    \leq\left(\frac{L_{\mathrm{ell}}^b}{d_\kappa}\right)^2
    \longrightarrow1.
\end{equation}
Thus elliptic asymptotically achieves both Gaussian Lipschitz benchmarks attained exactly by the Lipschitz-guided schedule. These are regularity guarantees for the continuous flow; sampling error also depends on the discretization.

At $\kappa=1$, every VP schedule has zero drift for the standard Gaussian target. The elliptic family tends to the trigonometric schedule, while the Lipschitz-guided family tends to the square-root schedule; neither is uniquely selected by the degenerate Lipschitz criterion.

\section{Holomorphic Extension of PF-ODE Trajectories for Log-Concave Targets}\label{app:log-concave holomorphic}

Throughout this section of the appendix, we assume that, for $\kappa>1$, the target $\pi(x)\propto e^{-V(x)}$ belongs to the class $\mathsf{P}_{\kappa}$ of centered log-concave distributions over $\R^d$ with $V \in \mathcal{C}^2(\R^d)$ and
\begin{equation}
    I_d \preceq \nabla^2 V(x) \preceq \kappa I_d, \quad x\in \R^d.
\end{equation}

For a smoothing level $\widetilde\sigma\in(0,1)$, the smoothed target is $\widetilde x_1=\widetilde\sigma z^\prime + \sqrt{1-\widetilde\sigma^2}\,x_1$, where $z^\prime$ is an independent standard Gaussian. Its law is $\widetilde\pi$, and tildes on the drift, score, and trajectory refer to this smoothed target. The target curvature parameter $\kappa$ and the elliptic schedule parameter $\widehat\kappa$ retain their original meanings.

In the rest of this appendix section, we will use
\begin{equation}\label{eq:canonical interpolant}
    x^\star_u = \sqrt{1-u^2}\,z + u\, x_1,\qquad u\in[-1,1]
\end{equation}
as a \emph{canonical VP interpolant}. Write $p^\star_u = \mathcal{L}(x^\star_u)$ for its marginal density, $s^\star_u(x) = \nabla \log p_u^\star(x)$ for its score and $X^\star_u$ for its PF-ODE trajectory satisfying
\begin{equation}
    \frac{dX_u^{\star,z}}{du}=b_u^\star(X_u^{\star,z}),\quad X_0^{\star,z} = z, \quad u\in (-1,1)
    \label{eq:lc-canonical-ode}
\end{equation}
Here, $b_u^\star(x)$ is the drift which satisfies
\begin{equation}\label{eq:canonical drift and score}
    \begin{aligned}
    b_u^\star(x) &= \frac{x + s^\star_u(x)}{u},\quad u\neq 0, \\
                 &=\frac{\E[x_1\mid x_u^\star=x]-ux}{1-u^2},\quad u\in (-1,1).
\end{aligned}
\end{equation}
Existence and uniqueness of the trajectories $X^\star_u$ for $u\in (-1,1)$ are established in \cref{lem:lc-canonical-real-bounds}

All vector norms are Euclidean, including the Hermitian Euclidean norm on $\C^d$. In holomorphic formulas, quadratic forms use transpose rather than conjugate transpose. We use the domains
\begin{equation}
    S_\omega=\{t\in\C \mid |\operatorname{Im}t|<\omega\},\qquad \Omega_\rho=\{t\in\C \mid \dist(t,[0,1])<\rho \}.
    \label{eq:lc-analytic-domains}
\end{equation}

\subsection{Proof of Theorem~\ref{thm:holomorphic strip width}: Holomorphic Width}
\begin{proof}[Proof of \cref{thm:holomorphic strip width}]
The proof follows three steps:
\paragraph{i. Representing any SI flow map in terms of a canonical VP-SI flow map.}
With independent standard Gaussians $z,z^\prime$, the smoothed target is $\widetilde x_1=\widetilde\sigma z^\prime + \sqrt{1-\widetilde\sigma^2}\,x_1$. Under a VP schedule $(\alpha,\beta)\in \mathcal{S}$, its stochastic interpolant satisfies
\begin{equation}
    \widetilde{x}_t{=} \alpha_t {z} + \beta_t\widetilde{x}_1 \stackrel{(d)}{=} \sqrt{\alpha_t^2 + \widetilde{\sigma}^2 \beta_t^2 }\, z + \sqrt{1-\widetilde{\sigma}^2} \, \beta_t\, x_1 = x^\star_{u(t)} \text{ where } u(t) \triangleq \sqrt{1-\widetilde{\sigma}^2}\, \beta_t,
\end{equation}
since $\sqrt{\alpha_t^2 + \widetilde{\sigma}^2 \beta_t^2 } = \sqrt{1 - (1-\widetilde{\sigma}^2) \beta_t^2 } = \sqrt{1-u(t)^2}$. 

Moreover, the drift for the smoothed target becomes 
\begin{equation}
    \widetilde{b}_t(x) = \mathbb{E}[ \dot{\widetilde{x}}_t \mid  \widetilde{x}_t=x] = \mathbb{E}\left[ \dot{u}(t) \frac{d}{du} x^\star_{u(t)} \mid  x^\star_{u(t)}=x\right] = \dot{u}(t) b^\star_{u(t)}(x).
\end{equation}
Chain rule and uniqueness of the PF-ODE trajectory imply
\begin{equation}\label{eq:lc-smoothed-canonical-trajectory}
   \widetilde{X}_t^z = X^{\star,z}_{u(t)}, \qquad t\in[0,1],
\end{equation}
for any initialization $z\in \R^d$. The holomorphic extension of the PF-ODE trajectory $t\mapsto \widetilde X^{z}_t$ of the smoothed target can be constructed by composing the holomorphic extensions of the time reparameterization $t\mapsto u(t)= \sqrt{1-\widetilde{\sigma}^2}\, \beta_t$, which is determined solely by the schedule, and the canonical VP trajectory $u\mapsto X^{\star,z}_u$, which is schedule independent and determined solely by the underlying distribution.

\paragraph{ii. Schedule-independent holomorphic extension of canonical VP trajectory.} Let $D = \sqrt{d} + \sqrt{B^2 + d\log \kappa}$ as in the theorem. By
\cref{thm:analytic extension canonical SI} and \cref{cor:lc-canonical-domain}, every canonical trajectory with initialization in $\|z\|\leq B$ extends holomorphically to 
\begin{equation}
    \mathcal{D}_{X^\star}(\kappa, d, B)  \triangleq \bigcup_{u_\circ \in (-1,1)} \left\{u\in \C \mid |u-u_\circ| < \frac{1-u_\circ^2}{30 D}\right\},
\end{equation}
and they are uniformly bounded on this domain:
\begin{equation}
    \sup_{u \in \mathcal{D}_{X^\star}(\kappa, d, B)} \| X^{\star,z}_u\| \leq B + \sqrt{d} + 1.
\end{equation}

\paragraph{iii. Elliptic schedule holomorphically maps a strip $S_{\omega_{\hat\kappa}}$ to the domain $\mathcal{D}_{X^\star}(\kappa, d, B)$.} Set $\widehat m = 1-\widehat\kappa^{-1}$ and
\begin{equation}
    r \triangleq \min(1,\widetilde\sigma\sqrt{\widehat\kappa}),\qquad
    \omega_{\widehat\kappa}\triangleq\frac{r}{K(\widehat m)(1+30\sqrt{1-\widetilde\sigma^2}\,D)}.
    \label{eq:lc-proof-strip-parameters}
\end{equation}
Consider the elliptic schedule $\alpha_t = \cn(\widehat{K} t \mid \widehat m)$ and $\beta_t = \sn(\widehat{K} t \mid \widehat m)$ where $\widehat{K}=K(\widehat{m})$ and $\widehat{K}^\prime = K(1-\widehat{m})$. As discussed in \cref{app:complex elliptic}, the mapping $t\mapsto u(t) = \sqrt{1-\widetilde\sigma^2}\, \beta_t$ extends holomorphically to the strip
\begin{equation}
    S_{\frac{\widehat{K}^\prime}{\widehat{K}}} =   \left\{t \in \C \mid |\Im(t)| < \frac{K(1-\widehat m)}{K(\widehat m)} \right\}.
\end{equation}
Note that $\omega_{\widehat\kappa} < \frac{1}{K(\widehat m)}< \frac{K(1-\widehat m)}{K(\widehat m)} $, thus $S_{\omega_{\hat\kappa}} \subset S_{\frac{\widehat{K}^\prime}{\widehat{K}}}$ and the function $t\mapsto u(t)$ is holomorphic on the smaller strip $S_{\omega_{\hat\kappa}}$. We will show that $u(S_{\omega_{\hat\kappa}}) \subset \mathcal{D}_{X^\star}(\kappa, d, B)$, which implies the trajectories $t\mapsto \widetilde X^{z}_t = X^{\star,z}_{u(t)}$ with $\|z\|\leq B$ extend holomorphically to the strip $S_{\omega_{\hat\kappa}} $ under this elliptic schedule. 

Fix real $t_\circ \in \R$ and complex $t \in \C$ such that 
\begin{equation}
    \Delta \triangleq \frac{K(\widehat m)}{r}|t-t_\circ| < 1.
\end{equation}
Applying \cref{thm:local elliptic bounds} with parameters $m=\widehat m$ and $\widetilde m=1-\widetilde\sigma^2$ gives the following bounds along the segment $[0,1]\ni \theta \mapsto t_\theta = t_\circ+\theta(t-t_\circ)$
\begin{equation}
    |\cn(K(\widehat m)t_\theta\mid\widehat m)|\leq\frac{\sqrt{1-u(t_\circ)^2}}{1-\theta\Delta},
    \qquad
    |\dn(K(\widehat m)t_\theta\mid\widehat m)|\leq\frac{\sqrt{1-u(t_\circ)^2}}{r(1-\theta\Delta)}.
\end{equation}
Using the derivative identity 
\begin{equation}
    \dot u(t)=\sqrt{1-\widetilde\sigma^2}\,K(\widehat m)\cn(K(\widehat m)t\mid\widehat m)\dn(K(\widehat m)t\mid\widehat m)
\end{equation}
and integrating along this segment, we obtain
\begin{align}
    &|u(t) - u(t_\circ)| = \sqrt{1-\widetilde{\sigma}^2} K(\widehat m) \left|\int_{0}^{1}\cn(K(\widehat m) t_\theta \mid \widehat m) \dn(K(m)t_\theta \mid \widehat m) (t-t_\circ) d\theta\right|,\\
    &\quad\leq  \sqrt{1-\widetilde{\sigma}^2} K(\widehat m) |t-t_\circ| \int_{0}^{1} |\cn(K(\widehat m)t_\theta \mid \widehat m)|\, |\dn(K(\widehat m) t_\theta \mid \widehat m)|  d\theta,\\
    &\quad\leq (1-u(t_\circ)^2) \sqrt{1-\widetilde\sigma^2}\int_0^1\frac{\Delta}{(1-\theta\Delta)^2}\,d\theta.
\end{align}
Thus, we get the bound
\begin{equation} \label{eq:lc-elliptic-image-displacement}
     \frac{|u(t)-u(t_\circ)|}{1-u(t_\circ)^2}\leq \frac{\sqrt{1-\widetilde\sigma^2}\,\Delta}{1-\Delta}.
\end{equation}
In particular, for any $ t \in S_{\omega_{\hat\kappa}}$, setting $t_\circ = \Re(t)$ gives
\begin{equation}
    \Delta  = \frac{K(\widehat m)}{r}|t-t_\circ| =  \frac{K(\widehat m)}{r}|\Im(t)|< \frac{1}{1+30 \sqrt{1-\widetilde\sigma^2} \,D} %
\end{equation}
Since $|u(t_\circ)|\leq \sqrt{1-\widetilde\sigma^2}<1$ and $\Delta \mapsto \frac{\Delta}{1-\Delta}$ is increasing on $\Delta \in (0,1)$, we get
\begin{equation}
     \frac{|u(t)-u(t_\circ)|}{1-u(t_\circ)^2}\leq \frac{\sqrt{1-\widetilde\sigma^2}\,\Delta}{1-\Delta}< \frac{1}{30 D},
\end{equation}
implying $u(t)$ belongs to $\mathcal{D}_{X^\star}(\kappa, d, B)$ whenever $ t \in S_{\omega_{\hat\kappa}}$. Thus, \eqref{eq:lc-smoothed-canonical-trajectory} extends holomorphically to $S_{\omega_{\hat\kappa}}$ by composition with the holomorphic extension of the canonical VP trajectory $\mathcal{D}_{X^\star}(\kappa, d, B) \ni u\mapsto X^{\star,z}_u$ and its norm is at most
\begin{equation}
    \sup_{t \in S_{\omega_{\hat\kappa}}} \| \widetilde X^z_t\| = \sup_{t \in S_{\omega_{\hat\kappa}}} \| X^{\star,z}_{u(t)}\|  \leq \sup_{u \in \mathcal{D}_{X^\star}(\kappa, d, B)} \| X^{\star,z}_{u}\|\leq B + \sqrt{d}+1.
\end{equation}
The constants are uniform over $\pi \in \mathsf{P}_{\kappa}$ and $\|z\|\leq B$, proving the stated lower bound on $\omega_\star$.
\end{proof}

\subsection{Holomorphic Extension of the Canonical Interpolant Trajectories}
\label{app:holomorphic extension}

\begin{lemma}[Holomorphic extension of the canonical trajectories]\label{thm:analytic extension canonical SI}
    Fix $B>0$ and define 
    \begin{equation}
        D=\sqrt d+\sqrt{B^2+d\log\kappa}.
    \end{equation}
    For every real $u_\circ \in (-1,1)$ and $\|z\|\leq B$, the canonical interpolant trajectories $u\mapsto X_u^{\star,z}$ defined in \eqref{eq:lc-canonical-ode} extend holomorphically to a disk around $u_\circ$, namely,
    \begin{equation}\label{eq:lc-canonical-local-radius}
        \mathbb{B}_{\rho_\circ}(u_\circ)\triangleq \left\{u\in \C \mid |u-u_\circ| < \rho_\circ\right\}, \quad \text{where}\quad \rho_\circ \triangleq \frac{1-u_\circ^2}{30D}.
    \end{equation}
    Moreover, for centered target $\mathbb{E}[x_1]=0$, the holomorphic extensions are bounded, satisfying
    \begin{equation}\label{eq:lc-canonical-local-magnitude}
        \sup_{\substack{u \in  \mathbb{B}_{\rho_\circ}(u_\circ),\\\|z\|\leq B }}\|X_u^{\star,z} - X_{u_\circ}^{\star,z}\|< 1,\quad \text{and}\quad   \sup_{\substack{u \in  \mathbb{B}_{\rho_\circ}(u_\circ),\\\|z\|\leq B }}\|X_u^{\star,z}\|\leq B + \sqrt{d}+1.
    \end{equation}
    
\end{lemma}
\begin{proof}
\cref{lem:lc-canonical-real-bounds} establishes existence, uniqueness, and real bounds of $u\mapsto X_{u}^{\star,z}$ on $u\in (-1,1)$. 

\paragraph{Conversion to SNR coordinate.}
Fix a real $u_\circ \in (-1,1)$ and $\|z\|\leq B$ and let $X_\circ^\star \triangleq X_{u_\circ}^{\star,z}$ be the value of the trajectory at $u_\circ$ for initialization $z$. Given this real time and trajectory value pair $(X_\circ^\star, u_\circ)$, we aim to show that the initial value problem
\begin{equation}
    \frac{dX_u^{\star}}{du}=b_u^\star(X_u^{\star}),\quad \text{where} \quad X_{u_\circ}^{\star} = X_\circ^\star,
\end{equation}
has a unique bounded holomorphic solution on the disk $\mathbb{B}_{\rho_\circ}(u_\circ)$ defined in \eqref{eq:lc-canonical-local-radius}. To do so, we introduce the change of variables $h:(X^\star,u) \mapsto (Y,s)$ given by%
\begin{equation}
    Y \triangleq \frac{X^\star - u \overline{X}_{1\mid u_\circ}^\star}{\sqrt{1-u^2}}, \qquad s \triangleq \frac{u}{\sqrt{1-u^2}}.
\end{equation}
Here, $\overline{X}_{1\mid u_\circ}^\star$ is the unique mode of the $(1-u_\circ^2)^{-1}$-strongly log-concave posterior density
\begin{equation}
     p_{1\mid u_\circ}(x_1\mid x) \propto \exp\left(-\frac{\|x-u_\circ x_1\|^2}{2(1-u_\circ^2)} - V(x_1)\right),%
\end{equation}
of the target $x_1$ conditioned on $x_{u_\circ} = X_\circ^\star$, i.e.,
\begin{equation}\label{eq: mode def}
    \overline{X}_{1\mid u_\circ}^\star \triangleq \argmin_{x_1\in \R^d} \left\{\frac{\|X_\circ^\star-u_\circ x_1\|^2}{2(1-u_\circ^2)} + V(x_1) \right\} = \operatorname{prox}_{\frac{1-u_\circ^2}{u_\circ^2}V}\left(\frac{X_\circ^\star}{u_\circ}\right).
\end{equation}

Under this change of coordinates, the canonical interpolant becomes
\begin{equation}
    x^\star_u = \sqrt{1-u^2}z + u x_1 \stackrel{(d)}{=} \sqrt{1-u^2} y_{\frac{u}{\sqrt{1-u^2}}} +  u\overline{X}_{1\mid u_\circ}^\star, \quad \text{where}\quad  y_s = z + s (x_1 - \overline{X}_{1\mid u_\circ}^\star).
\end{equation}
The initial-value problem under these coordinates transforms to
\begin{align}
    \frac{d Y_s}{ds} &= m_s(Y_s), \qquad Y_{s_\circ} = Y_\circ \triangleq \frac{X_\circ^\star - u_\circ \overline{X}_{1\mid u_\circ}^\star}{\sqrt{1-u_\circ^2}}, \quad s_\circ \triangleq \frac{u_\circ}{\sqrt{1-u_\circ^2}}\label{eq:snr ivp}
\end{align}
where the drift $(y,s)\mapsto m_s(y)$ is the conditional mean
\begin{align}
     m_s(y) &\triangleq \mathbb{E}[x_1 -\overline{X}_{1\mid u_\circ}^\star  \mid  y_s  = y ], \qquad s \in \R, y \in \R^d. 
\end{align}

\paragraph{Bounding the real-valued initialization.}
Define the following mode-centered and rescaled conditional random sample
\begin{equation}
    \xi_\circ \triangleq \frac{x_{1\mid u_\circ}^\star - \overline{X}_{1\mid u_\circ}^\star}{\sqrt{1-u_\circ^2}},\quad \text{where}\quad x_{1\mid u_\circ}^\star \sim p_{1\mid u_\circ}(\cdot \mid X_\circ^\star),
\end{equation}
which has a $1-$strongly log-concave density with mode at zero. Notice that the translated real-valued initial condition $Y_\circ$ can be expressed in terms of $\xi_\circ$ and the VE score using Tweedie's identity as
\begin{align}
    Y_\circ = \frac{X_\circ^\star - u_\circ \overline{X}_{1\mid u_\circ}^\star}{\sqrt{1-u_\circ^2}} &= \frac{X_\circ^\star - u_\circ \E[x_1\mid x_{u_\circ}^\star = X_\circ^\star]}{\sqrt{1-u_\circ^2}} + u_\circ\frac{ \E[x_1\mid x_{u_\circ}^\star = X_\circ^\star]-\overline{X}_{1\mid u_\circ}^\star}{\sqrt{1-u_\circ^2}},\\
    &=-{\frac{1}{s_\circ}} s_{\frac{1}{s_\circ^2}}^{VE}\left(\frac{X_\circ^\star}{u_\circ}\right) + u_\circ \E[\xi_\circ].
\end{align}
Since $\sqrt{\E[\|\xi_\circ\|^2]}\leq \sqrt{d}$ by \cref{thm:second moment bound} and $|s_\circ|^{-1}\,\| s_{{s_\circ^{-2}}}^{VE}({X_\circ^\star/ u_\circ}) \| \leq \sqrt{B^2 + d\log \kappa}$ by \cref{thm:VE score bound}, we get
\begin{align}
    \|Y_\circ\| &\leq \left\| {\frac{1}{s_\circ}}s_{\frac{1}{s_\circ^2}}^{VE}\left(\frac{X_\circ^\star}{u_\circ}\right) \right\|+ |u_\circ| \|\E[\xi_\circ]\|,\\
    &\leq \sqrt{B^2 + d\log \kappa}+  \sqrt{\E[\|\xi_\circ\|^2]}, \\
    &\leq \sqrt{B^2 + d\log \kappa}+  \sqrt{d} = D.\label{eq:Y_circ bound}
\end{align}

\paragraph{Holomorphic extension of the drift.}
The drift $(y,s)\mapsto m_s(y)$ can be extended to a complex-valued domain meromorphically as a normalized expectation of $\xi_\circ$ under a complex-exponential tilting as follows:
\begin{gather}
     m_s(y)  = \sqrt{1-u_\circ^2}\,\frac{\E[\xi_\circ \exp(  \xi_\circ^\top \chi_\circ(y,s)+\eta_\circ(s)\|\xi_\circ\|^2)]}{\E[\exp(  \xi_\circ^\top \chi_\circ(y,s)+\eta_\circ(s)\|\xi_\circ\|^2)]},\label{eq:ratio of complex tilted expectations}\\
     \text{where}\quad \chi_\circ(y,s)\triangleq\sqrt{1-u_\circ^2}\left(s y -s_\circ Y_\circ\right),\quad\text{and}\quad \eta_\circ(s)\triangleq -\frac{1-u_\circ^2}{2}(s^2-s_\circ^2).
\end{gather}
Note that although both the numerator and denominator are expectations of entire functions in $(y,s)$, the expectations with respect to $1-$strongly log-concave random variable $\xi_\circ$ take a finite value and are holomorphic when $\Re(\eta_\circ(s))<\frac{1}{2}$ (or equivalently $\Re(s^2)>-1$). 

Moreover, the ratio remains holomorphic on this region except at the zeros of the denominator in \eqref{eq:ratio of complex tilted expectations}. In particular, by \cref{thm:complex tilting}, $m_s(y)$ is holomorphic and bounded as $\|m_s(y)\| \leq 2\sqrt{d(1-u_\circ^2)}$ on the domain
\begin{equation}
    \mathcal{D}_{m} \triangleq  \left\{(y,s) \in \C^d\times \C \mid \|\chi_\circ(y,s)\|< \frac12,\quad |\eta_\circ(s)| < \frac{1}{16\sqrt{d}} \right\}.
\end{equation}
We verify that the product domain $\mathbb{B}_{r_{Y}}(Y_\circ)\times \mathbb{B}_{r_{s}}(s_\circ)$ with $r_{s} \triangleq (24 \sqrt{1-u_\circ^2} D)^{-1}$ and $r_{Y} \triangleq 2\sqrt{d(1-u_\circ^2)} r_{s}$ remains inside $\mathcal{D}_{m}$, namely, $m_s(y)$ is holomorphic on $\mathbb{B}_{r_{Y}}(Y_\circ)\times \mathbb{B}_{r_{s}}(s_\circ)$. Note that, on $\mathbb{B}_{r_{Y}}(Y_\circ)\times \mathbb{B}_{r_{s}}(s_\circ)$, we have $|s|= |s-s_\circ + s_\circ| \leq r_{s} + |s_\circ|$, and thus
\begin{align}
    \|\chi_\circ(y,s)\| &= \sqrt{1-u_\circ^2}\|sy - s Y_\circ + s Y_\circ - s_\circ Y_\circ\|,\\
    &\leq  \sqrt{1-u_\circ^2} \left( |s| \|y - Y_\circ\| + \| Y_\circ\||s-s_\circ|\right),\\
    &\leq  \sqrt{1-u_\circ^2}( r_{s} + |s_\circ|) r_{Y} + \sqrt{1-u_\circ^2}  D r_{s},\\
    &\leq \left(\frac{1}{24D} + 1\right) \frac{2\sqrt{d(1-u_\circ^2)} }{24 \sqrt{1-u_\circ^2} D} + \frac{1}{24},\\
    &= \left(\frac{1}{24D} + 1\right) \frac{\sqrt{d} }{12  D} + \frac{1}{24},\\
    &\leq  \frac{25}{24} \frac{1 }{12 } + \frac{1}{24} \leq \frac12,
\end{align}
where the third inequality follows from \eqref{eq:Y_circ bound}, and the penultimate inequality follows from $D\geq\sqrt{d}\geq 1$. Similarly,  we get
\begin{align}
   |\eta_\circ(s)| &= \frac{1-u_\circ^2}{2}|s-s_\circ|\, |s+s_\circ|,\\
   &\leq \frac{\sqrt{1-u_\circ^2}}{48 D} (|s| + |s_\circ|),\\
   &\leq \frac{\sqrt{1-u_\circ^2}}{48 D} (r_s + 2|s_\circ|),\\
   &\leq\frac{1}{24D}\left(\frac{1}{48D} + 1\right),\\
   &\leq \frac{1}{24\sqrt{d}} \frac{49}{48}\leq \frac{1}{16\sqrt{d}}.
\end{align}
Hence, $m_s(y)$ is holomorphic and bounded by $2\sqrt{d(1-u_\circ^2)}$ on  $\mathbb{B}_{r_{Y}}(Y_\circ)\times \mathbb{B}_{r_{s}}(s_\circ)$ with $r_{s} = (24 \sqrt{1-u_\circ^2} D)^{-1}$ and $r_{Y} = 2\sqrt{d(1-u_\circ^2)} r_{s}$.

\paragraph{Holomorphic extension of trajectories $s\mapsto Y_s$ and $u\mapsto X_u$.} 
By \cref{lem:lc-holomorphic-ivp}, the initial value problem \eqref{eq:snr ivp} has a holomorphic solution $s\mapsto Y_s$ on $\mathbb{B}_{r_s}(s_\circ)$ with 
\begin{equation}\label{eq:ys-ycirc}
    \|Y_s - Y_\circ\|\leq  2\sqrt{d(1-u_\circ^2)} |s-s_\circ| \leq r_Y \leq \frac{1}{12}.
\end{equation}
We want to show that the trajectories $u\mapsto X_u^\star = \sqrt{1-u^2}\, Y_{\frac{u}{\sqrt{1-u^2}}} + u \overline{X}_{1\mid u_\circ}^\star$ extend holomorphically to $ \mathbb{B}_{\rho_\circ}(u_\circ)$ with $\rho_\circ = \frac{1-u_\circ^2}{30D} \leq \frac{1-u_\circ^2}{30}$. Note that on $u \in \mathbb{B}_{\rho_\circ}(u_\circ)$, we have 
\begin{align}
    |(1-u^2) - (1-u_\circ^2)| =|u-u_\circ|\,|u+u_\circ|&\leq \rho_\circ(2+\rho_\circ) ,\\
    &\leq \frac{1-u_\circ^2}{30D}\left(2+\frac{1}{30}\right) = \frac{61(1-u_\circ^2)}{900 D}< \frac{1-u_\circ^2}{14}.\label{eq:1-u2-1-ucirc2}
\end{align}
This implies that $1-u^2$ remains entirely on the right-half plane with 
\begin{equation}
     0<\frac{13}{14}(1-u_\circ^2)\leq |1-u^2|\leq \frac{15}{14}(1-u_\circ^2). \label{eq:1-u2 bound}
\end{equation}
Thus, both maps $u\mapsto \sqrt{1-u^2}$ and $u\mapsto \frac{u}{\sqrt{1-u^2}}$ are holomorphic on the disk $ \mathbb{B}_{\rho_\circ}(u_\circ)$ as the principal square-root is holomorphic on the right half plane. Moreover, both functions are bounded on $ \mathbb{B}_{\rho_\circ}(u_\circ)$, which can be shown by integrating their derivatives $-u(1-u^2)^{-1/2}$ and $(1-u^2)^{-3/2}$, respectively, along the line segment $u(\theta)\triangleq u_\circ + \theta (u-u_\circ)$ for $\theta\in[0,1]$, which yields
\begin{align}
    |\sqrt{1-u^2} - \sqrt{1-u_\circ^2}| &\leq |u-u_\circ| \left|\int_{0}^{1} -u(\theta)(1-u(\theta)^2)^{-1/2} d\theta \right|,\\
    &\leq |u-u_\circ| \sup_{v \in \mathbb{B}_{\rho_\circ}(u_\circ)}\left|v(1-v^2)^{-1/2}\right|,\\
    &\leq \rho_\circ  \left(\sup_{v \in \mathbb{B}_{\rho_\circ}(u_\circ)}|v| \right)\, \left(\inf_{v \in \mathbb{B}_{\rho_\circ}(u_\circ)}|1-v^2|\right)^{-1/2},\\
    &\leq \rho_\circ (1 + \rho_\circ) \, \left(\frac{13}{14}(1-u_\circ^2)\right)^{-1/2} < \frac{\sqrt{1-u_\circ^2}}{14 D},\label{eq: sqrt 1-u2 diff}
\end{align}
and similarly
\begin{align}
    \left|\frac{u}{\sqrt{1-u^2}} - \frac{u_\circ}{\sqrt{1-u_\circ^2}}\right| &\leq |u-u_\circ| \left|\int_{0}^{1}  (1-u(\theta)^2)^{-3/2}  d\theta\right|,\\
    &\leq |u-u_\circ| \sup_{v \in \mathbb{B}_{\rho_\circ}(u_\circ)}\left|(1-v^2)^{-3/2}\right|,\\
    &\leq \rho_\circ \left(\frac{13}{14}(1-u_\circ^2)\right)^{-3/2} <\frac{1}{24 \sqrt{1-u_\circ^2} D}= r_s.
\end{align}
The last bound implies that the image of the disk $\mathbb{B}_{\rho_\circ}(u_\circ)$ under $u\mapsto \frac{u}{\sqrt{1-u^2}}$ lies inside the disk $\mathbb{B}_{r_s}(s_\circ)$ and the mapping $u\mapsto Y_{\frac{u}{\sqrt{1-u^2}}}$ is holomorphic on $\mathbb{B}_{\rho_\circ}(u_\circ)$ as a composition. Hence, the canonical trajectory $u\mapsto X_u^\star = \sqrt{1-u^2}\, Y_{\frac{u}{\sqrt{1-u^2}}} + u \overline{X}_{1\mid u_\circ}^\star$ is holomorphic on $\mathbb{B}_{\rho_\circ}(u_\circ)$.

Finally, to obtain the bound on the canonical trajectories, observe that
\begin{align}
    X_u^\star - X_\circ^\star &= \left(\sqrt{1-u^2}\, Y_{\frac{u}{\sqrt{1-u^2}}} - \sqrt{1-u_\circ^2} Y_\circ\right) + (u-u_\circ) \overline{X}_{1\mid u_\circ}^\star,\\
    &= \sqrt{1-u^2}(Y_{\frac{u}{\sqrt{1-u^2}}} - Y_\circ) +\left(\sqrt{1-u^2} -  \sqrt{1-u_\circ^2} \right) Y_\circ + (u-u_\circ) \overline{X}_{1\mid u_\circ}^\star.\label{eq:xu - xcirc}
\end{align}
The first two terms in \eqref{eq:xu - xcirc} are bounded as
\begin{equation}
    \|\sqrt{1-u^2}(Y_{\frac{u}{\sqrt{1-u^2}}} - Y_\circ) \| \leq \sqrt{\frac{15}{14}(1-u_\circ^2)} \frac{1}{12} <\frac{\sqrt{1-u_\circ^2}}{11},\label{eq:term1 bdd}
\end{equation}
which follows from \eqref{eq:ys-ycirc} and \eqref{eq:1-u2 bound}, and
\begin{align}
    \left\|\left(\sqrt{1-u^2} -  \sqrt{1-u_\circ^2} \right) Y_\circ \right\| =  |\sqrt{1-u^2} -  \sqrt{1-u_\circ^2} | \,\|Y_\circ\|<\frac{\sqrt{1-u_\circ^2}}{14}, \label{eq:term2 bdd}
\end{align}
which follows from \eqref{eq: sqrt 1-u2 diff} and \eqref{eq:Y_circ bound}. 

To bound the third term, notice that the first-order optimality conditions for the conditional mode $\overline{X}_{1\mid u_\circ}$ in \eqref{eq: mode def} imply
\begin{equation}
    \nabla V(\overline{X}_{1\mid u_\circ}^\star) = \frac{u_\circ}{1-u_\circ^2}(X_\circ^\star-u_\circ \overline{X}_{1\mid u_\circ}^\star) = s_\circ Y_\circ.
\end{equation}
Defining the unconditional mode $\overline{X}_1 \triangleq \argmin_{x} V(x)$ of target $\pi$, 1-strong log-concavity and Cauchy-Schwarz imply
\begin{align}
    \|\overline{X}_{1\mid u_\circ}^\star - \overline{X}_{1}\|^2 &\leq (\overline{X}_{1\mid u_\circ}^\star - \overline{X}_{1})^\top (  \nabla V(\overline{X}_{1\mid u_\circ}^\star)-   \nabla V(\overline{X}_{1}) ),\\
    &= (\overline{X}_{1\mid u_\circ}^\star - \overline{X}_{1})^\top  \nabla V(\overline{X}_{1\mid u_\circ}^\star),\\
    &\leq \|\overline{X}_{1\mid u_\circ}^\star - \overline{X}_{1}\| \| \nabla V(\overline{X}_{1\mid u_\circ}^\star)\|.
\end{align}
Thus, \eqref{eq:Y_circ bound} yields
\begin{align}
    \|\overline{X}_{1\mid u_\circ}^\star - \overline{X}_{1}\| &\leq \| \nabla V(\overline{X}_{1\mid u_\circ}^\star)\| = |s_\circ| \|Y_\circ\| \leq |s_\circ| D.
\end{align}
For a centered target, we have 
\begin{equation}
    \|\overline{X}_{1}\| = \|\E[x_1 -\overline{X}_{1} ]\| \leq \E[\|x_1 -\overline{X}_{1} \|^2]^{1/2} \leq \sqrt{d}  
\end{equation}
where the last inequality follows from \cref{thm:second moment bound}. Thus, 
\begin{equation}
    \|\overline{X}_{1\mid u_\circ}^\star\| \leq \sqrt{d} + |s_\circ| D    
\end{equation}
and the third term is bounded by
\begin{equation}\label{eq:term3 bdd}
    \|(u-u_\circ) \overline{X}_{1\mid u_\circ}^\star\| = |u-u_\circ| \,\|\overline{X}_{1\mid u_\circ}^\star\| \leq \rho_\circ (\sqrt{d} + |s_\circ| D) < \frac{\sqrt{1-u_\circ^2}}{15}.
\end{equation}
Combining all three bounded terms \eqref{eq:term1 bdd}, \eqref{eq:term2 bdd}, and \eqref{eq:term3 bdd} yields 
\begin{equation}
    \|X_u^\star - X_\circ^\star\| \leq \left(\frac{1}{11}+\frac{1}{14}+\frac{1}{15}\right)\sqrt{1-u_\circ^2} < 1.
\end{equation}
Finally, since $\|X^\star_\circ\| \leq B +\sqrt{d}$ by equation \eqref{eq:lc-real-canonical-norm} in \cref{lem:lc-canonical-real-bounds}, we obtain
\begin{equation}
    \|X_u^\star \| \leq B +\sqrt{d}+1.
\end{equation}
This completes the proof.

\end{proof}

\begin{corollary}[A common canonical holomorphic domain]
\label{cor:lc-canonical-domain}
Under the assumptions of \cref{thm:analytic extension canonical SI}, the local extensions agree on overlaps and define a holomorphic trajectory on
\begin{equation}
     \mathcal{D}_{X^\star}(\kappa, d, B) \triangleq\bigcup_{u_\circ\in(-1,1)}
    \left\{u\in\C:|u-u_\circ|<\frac{1-u_\circ^2}{30D}\right\}.
    \label{eq:lc-glued-canonical-domain}
\end{equation}
Moreover, if $\mathbb E[x_1]=0$, then $\sup_{u\in \mathcal{D}_{X^\star},\,\|z\|\leq B}\|X_u^{\star,z}\|\leq B+\sqrt d+1$.
\end{corollary}

\begin{proof}
Each disk has radius smaller than the distance of its real center to $\{-1,1\}$. Two overlapping disks with real centers have a connected intersection meeting $(-1,1)$ in a nonempty real interval. Their local extensions agree with the same real trajectory there, so the identity theorem makes them agree throughout the intersection. They therefore glue to one holomorphic function. The magnitude bound follows from the bound on each disk.
\end{proof}

\subsubsection{Real Canonical Trajectories and Posterior Estimates}
\begin{lemma}[Existence, non-expansiveness, and real trajectory bounds]
\label{lem:lc-canonical-real-bounds}
The following holds for the canonical ODE \eqref{eq:lc-canonical-ode},
\begin{itemize}
    \item[i.] Existence-uniqueness: For every initialization $z\in \R^d$, a unique solution $u\mapsto X_u^{\star,z}$ exists for $u\in(-1,1)$.
    \item[ii.] Non-expansiveness: For every $u\in(-1,1)$, the flow map $z\mapsto X_u^{\star,z}$ is nonexpansive, and $X_u^{\star,z}$ has density $p_u^\star$ whenever $z\sim N(0,I_d)$.
    \item[iii.] Boundedness: For centered targets, i.e., $\mathbb E [x_1]=0$, the trajectories are bounded as
\begin{equation}
    \|X_u^{\star,z}\|\leq\sqrt{\|z\|^2+d}\leq\|z\|+\sqrt d.
    \label{eq:lc-real-canonical-norm}
\end{equation}
\end{itemize}
\end{lemma}

\begin{proof}
Define the posterior expectation
\begin{equation}
    \widehat{X}_1^\star(x,u) \triangleq \E[x_1 \mid x_u^\star = x],
\end{equation}
where $x_{u}^\star = \sqrt{1-u^2}z + ux_1 $. The posterior identity in \cref{lem:lc-posterior} and its upper covariance bound give
\begin{equation}
    \nabla_x\widehat{X}_1^\star(x,u)=\frac{u}{1-u^2}\operatorname{Cov}(x_1\mid x_u^\star=x),
    \qquad
    0\preceq\operatorname{Cov}(x_1\mid x_u^\star=x)\preceq(1-u^2)I_d.
\end{equation}
Therefore
\begin{equation}
    \nabla_x b_u^\star(x)
    =\frac{u}{1-u^2}
      \left(\frac{\operatorname{Cov}(x_1\mid x_u^\star=x)}{1-u^2}-I_d\right).
    \label{eq:lc-canonical-drift-jacobian}
\end{equation}
The drift is globally Lipschitz in $x$, uniformly on each compact subinterval of $(-1,1)$, and continuous in $u$ there. At $u=0$, it equals $\mathbb E[x_1]$. Standard real ODE existence and uniqueness, together with this linear-growth bound, yield a unique trajectory throughout $(-1,1)$. Differentiating the stochastic interpolant shows that its conditional velocity is exactly $b_u^\star$, so its densities satisfy the corresponding continuity equation. Uniqueness for a globally Lipschitz velocity field on each compact time interval then gives $\mathcal L(X_u^{\star,Z})=p_u^\star$.

For $u\geq0$, \eqref{eq:lc-canonical-drift-jacobian} is negative semidefinite. Hence the squared distance between two trajectories cannot increase as $u$ increases from zero. For negative times, symmetry of the Gaussian reference and uniqueness give
\begin{equation}
    \widehat{X}_1^\star(x,-u)=\widehat{X}_1^\star(-x,u),\qquad
    X_{-u}^{\star,z}=-X_u^{\star,-z},\qquad 0\leq u<1,
    \label{eq:lc-canonical-reflection}
\end{equation}
which proves nonexpansiveness there as well. If the target is centered, then $\mathbb E [X_u^{\star,Z}]=0$, and
\begin{equation}
    \|X_u^{\star,z}\|
    =\left\|\mathbb E[X_u^{\star,z}-X_u^{\star,Z}]\right\|
    \leq\mathbb E\|z-Z\|
    \leq\sqrt{\|z\|^2+d}.
\end{equation}

\end{proof}

\begin{lemma}[VE score is bounded for log-concave targets]\label{thm:VE score bound}
    For any $z\in \R^d$, let $[0,\infty)\ni\tau\mapsto \Phi^{VE}_\tau(z)$ denote the reverse flow map for the VE PF-ODE such that 
    \begin{equation}
        \frac{d \Phi^{VE}_\tau(z)}{d\tau} =-\frac12 s^{VE}_\tau( \Phi^{VE}_\tau(z)), \quad\text{and}\quad  \lim_{\tau\to\infty} \frac{\Phi^{VE}_\tau(z)}{\sqrt{\tau}} = z.
    \end{equation}
    Then, for all $\tau>0$, we have that 
    \begin{equation}\label{eq:VE score norm bound}
        \|s^{VE}_\tau( \Phi^{VE}_\tau(z))\| \leq \sqrt{\frac{\|z\|^2 + d\log(\kappa)}{\tau}}
    \end{equation}
\end{lemma}
\begin{proof}
Since $p \propto \exp(-V(x))$ is a log-concave distribution, the noisy distribution $p_\tau^{VE} = p \ast N(0,\tau I_d)$ is also log-concave. In fact, Tweedie's formula and posterior identities yield quantitative bounds
\begin{equation}\label{eq:VE curvature bounds}
     \frac{1}{\tau +1 }I_d \preceq \nabla^2 V_\tau \preceq  \frac{\kappa}{\kappa \tau + 1} I_d
\end{equation}
for all $\tau\geq 0$, where $V_\tau(x) \triangleq -\log  p^{VE}_\tau(x)$. Since $\nabla V_\tau$ is Lipschitz with constant $(\tau+\kappa^{-1})^{-1}$, we get the gradient estimate
\begin{equation}
     \|\nabla V_\tau( x)\|^2 \leq \frac{2}{\tau + \kappa^{-1}} \left(V_\tau(x) - \min V_\tau \right), 
\end{equation}
which can be used to get a bound on $s^{VE}_\tau(\Phi^{VE}_\tau(z)) = -\nabla V_\tau(\Phi^{VE}_\tau(z))$ as 
\begin{equation}\label{eq:score norm bound via gradient}
     \|s^{VE}_\tau( \Phi^{VE}_\tau(z))\|^2 \leq \frac{2}{\tau + \kappa^{-1}} \left(-\log  p^{VE}_\tau( \Phi^{VE}_\tau(z)) + \log \max_{x} p^{VE}_\tau(x)\right).
\end{equation}
Therefore, we need a lower bound on $\log  p^{VE}_\tau( \Phi^{VE}_\tau(z))$ and an upper bound on $\max_{x} p^{VE}_\tau(x)$. 

\paragraph{i. Lower bound on $\log  p^{VE}_\tau( \Phi^{VE}_\tau(z))$.} We can utilize its evolution in $\tau$, which is governed by the following ODE
\begin{align}
    \frac{d\,}{d\tau}\log p^{VE}_\tau(\Phi^{VE}_\tau(z))  &= \frac{1}{p^{VE}_\tau(\Phi^{VE}_\tau(z) )} \left(\nabla_x p^{VE}_\tau(\Phi^{VE}_\tau(z) )^\top \frac{d\Phi^{VE}_\tau(z)}{d\tau} + \partial_\tau p^{VE}_\tau(\Phi^{VE}_\tau(z) ) \right), \label{eq:logp eq 1}\\
    &= \frac12\left(- {\|s^{VE}_\tau(\Phi^{VE}_\tau(z) )\|^2} + \nabla_x^\top  s^{VE}_\tau(\Phi^{VE}_\tau(z) ) +  {\|s^{VE}_\tau(\Phi^{VE}_\tau(z) )\|^2} \right), \label{eq:logp eq 2}\\
    &= \frac12 \operatorname{tr}(\nabla  s^{VE}_\tau(\Phi^{VE}_\tau(z) ) ), \label{eq:logp eq 3}
\end{align}
where \eqref{eq:logp eq 1} follows from chain rule and \plaineqref{eq:logp eq 2} follows from the continuity equation 
\begin{equation}
    \partial_\tau p^{VE}_\tau(x) + \nabla \cdot \left(-\frac12 s^{VE}_\tau(x)p^{VE}_\tau(x) \right) = \partial_\tau p^{VE}_\tau(x)-\frac12 \Delta p^{VE}_\tau(x) =0.
\end{equation}
Since $\nabla  s^{VE}_\tau(\Phi^{VE}_\tau(z)) = -\nabla^2 V_\tau(\Phi^{VE}_\tau(z))$, the lower curvature bound in \plaineqref{eq:VE curvature bounds} implies 
\begin{equation}
     \frac{d\,}{d\tau}\log p^{VE}_\tau(\Phi^{VE}_\tau(z)) \leq -\frac{d}{2(\tau +1)},
\end{equation}
By Gronwall's inequality, integrating both sides yields
\begin{equation}\label{eq:logpt lower bound T}
    \log p^{VE}_\tau(\Phi^{VE}_\tau(z)) \geq \log p^{VE}_T(\Phi^{VE}_T(z)) +\frac{d}{2}\log \frac{T+1}{\tau+1},
\end{equation}
for any $0\leq \tau \leq T$. In particular, this inequality also holds as $T\longrightarrow \infty$. To get the limiting term, note that
\begin{equation}
    (2\pi T)^{d/2} p^{VE}_T(\Phi^{VE}_T(z)) = \mathbb{E}_{X\sim p}\left[ \exp\left(-\frac12 \left\|\frac{\Phi^{VE}_T(z) - X}{\sqrt{T}}\right\|^2\right) \right] \longrightarrow \exp(-\|z\|^2/2)
\end{equation}
as $T\longrightarrow \infty$ by the entrance condition $\lim_{T\to\infty} T^{-1/2}\Phi^{VE}_T(z)  = z$. Therefore,   
\begin{equation}
    \log p^{VE}_T(\Phi^{VE}_T(z)) + \frac{d}{2}\log (T)   \longrightarrow -\frac{\|z\|^2}{2} - \frac{d}{2}\log(2\pi)  \quad \text{as} \quad T\longrightarrow\infty.
\end{equation}
Together with \eqref{eq:logpt lower bound T}, this gives the lower bound
\begin{equation}\label{eq:logpt lower}
    \log p^{VE}_\tau(\Phi^{VE}_\tau(z)) \geq -\frac{\|z\|^2 + d\log(2\pi(\tau+1))}{2}.
\end{equation}

\paragraph{ii. Upper bound on $\max_{x} p^{VE}_\tau(x)$.} Note that the upper Hessian bound in \plaineqref{eq:VE curvature bounds} implies that
\begin{equation}
    V_\tau(x) \leq V_\tau (x_\tau^\star) + \frac{1}{2(\tau+\kappa^{-1})}\|x-x_\tau^\star\|^2, 
\end{equation}
where $x_\tau^\star = \argmin_{x^\prime} V_\tau(x^\prime)$. Negating and exponentiating both sides yield
\begin{equation}
    p^{VE}_\tau(x) \geq \max_{x^\prime} p^{VE}_\tau(x^\prime) \; \exp \left( -\frac{\|x-x_\tau^\star\|^2}{2(\tau+\kappa^{-1})}\right).
\end{equation}
Integrating both sides gives the upper bound
\begin{equation}\label{eq:logpt upper}
    \max_{x^\prime} p^{VE}_\tau(x^\prime) \leq  \; \left(2\pi (\tau+\kappa^{-1})\right)^{-d/2}.
\end{equation}
Substituting the lower bound \plaineqref{eq:logpt lower} and upper bound \plaineqref{eq:logpt upper} into \plaineqref{eq:score norm bound via gradient} yields
\begin{equation}
    \|s^{VE}_\tau( \Phi^{VE}_\tau(z))\|^2 \leq \frac{1}{\tau + \kappa^{-1}} \left(\|z\|^2 + d\log\left(\frac{\tau+1}{\tau+\kappa^{-1}}\right) \right).
\end{equation}
Applying the inequalities $\frac{1}{\tau+\kappa^{-1}}\leq \frac{1}{\tau}$ and $\frac{\tau+1}{\tau+\kappa^{-1}} \leq \kappa $ gives the inequality \plaineqref{eq:VE score norm bound}.
\end{proof}

\subsubsection{Complex Tilting and Local Complex ODE Existence}

\begin{lemma}[Complex-valued tilting]\label{thm:complex tilting}
    Let $X \sim {p}({x})\propto \exp(-{V}({x}))$ be such that 
    $\nabla^2 {V}\succeq I_d $
    and $\nabla V(0) = 0$, i.e., the mode is at $0$. Then, for complex valued $\chi \in \C^d$ and $\eta \in \C$ satisfying 
    \begin{equation}
        \|\chi\|<\frac12 \quad  \text{and} \quad |\eta| < \frac{1}{16\sqrt{d}},
    \end{equation}
    we have that
    \begin{equation} \label{eq:bounded ratio}
        \mathbb{E} \left[ e^{X^\top \chi +  \eta\|X\|^2} \right] \neq 0 \quad \text{and}\quad   \left\|\frac{\mathbb{E} \left[X e^{X^\top \chi+  \eta\|X\|^2} \right]}{\mathbb{E} \left[e^{X^\top \chi + \eta\|X\|^2} \right]}\right\| \leq 2\sqrt{d}.
    \end{equation}
\end{lemma}
\begin{proof}
    Let us split the exponent into its real and imaginary parts
    $x^\top \chi + \eta \|x\|^2 = R(x\mid \chi,\eta) + i\Theta(x\mid \chi,\eta)$ where
    \begin{equation}
        R(x\mid \chi,\eta) = x^\top \Re(\chi)+ \Re(\eta) \|x\|^2 \quad\text{and}\quad \Theta(x\mid \chi,\eta) = x^\top \Im(\chi)+ \Im(\eta) \|x\|^2.
    \end{equation}
    After exponentiation, the real part can be absorbed into the distribution $p(x)$ yielding a tilted distribution
    \begin{equation}
        \widetilde{p}(x\mid \chi,\eta) \triangleq \frac{1}{Z(\chi,\eta)} e^{R(x\mid \chi,\eta)} p(x),\quad Z(\chi,\eta) \triangleq \mathbb{E}_{p}[e^{R(X\mid \chi,\eta)}] > 0.
    \end{equation}
    Note that the curvature of this tilted distribution is bounded below by
    \begin{equation}
        -\nabla_x^2\log  \widetilde{p}(x\mid \chi,\eta)\succeq (1-2\Re(\eta))I_d \succeq (1-2|\eta|)I_d \succeq \frac{7}{8}I_d.
    \end{equation}
    Thus, the tilted distribution remains strongly log-concave.
    
    The expressions in \plaineqref{eq:bounded ratio} can be written as expectations under the tilted distribution $\widetilde{X}\sim \widetilde{p}(\cdot\mid \chi,\eta)$ as
    \begin{gather}
        \mathbb{E}_{p} \left[ e^{X^\top \chi +  \eta\|X\|^2} \right] = Z(\chi,\eta)  \mathbb{E}_{\widetilde{p}(\cdot\mid \chi,\eta)}  \left[  e^{i \Theta(\widetilde{X}\mid \chi,\eta)} \right],\\
        \frac{\mathbb{E}_{p} \left[X e^{X^\top \chi+  \eta\|X\|^2} \right]}{\mathbb{E}_{p} \left[e^{X^\top \chi + \eta\|X\|^2} \right]} =  \frac{\mathbb{E}_{\widetilde{p}(\cdot\mid \chi,\eta)} \left[\widetilde{X} e^{i \Theta(\widetilde{X}\mid \chi,\eta)} \right]}{\mathbb{E}_{\widetilde{p}(\cdot\mid \chi,\eta)} \left[ e^{i \Theta(\widetilde{X}\mid \chi,\eta)} \right]}.
    \end{gather}
    Since $Z(\chi,\eta)>0$, we only need to show that the magnitude of the purely complex-phase integral $\mathbb{E}_{\widetilde{p}(\cdot\mid \chi,\eta)}  \left[  e^{i \Theta(\widetilde{X}\mid \chi,\eta)} \right]$ is bounded below by a non-zero constant and control the magnitude of the numerator $\mathbb{E}_{\widetilde{p}(\cdot\mid \chi,\eta)} \left[\widetilde{X} e^{i \Theta(\widetilde{X}\mid \chi,\eta)} \right]$ from above.
    
    To show that $\mathbb{E}_{\widetilde{p}} [  e^{i \Theta(\widetilde{X}\mid \chi,\eta)}]$ is non-vanishing, notice that its magnitude remains the same under constant phase shifts, thus
    \begin{align}
        \left|\mathbb{E}_{\widetilde{p}}  \left[  e^{i \Theta(\widetilde{X}\mid \chi,\eta)} \right]\right| &= \left|\mathbb{E}_{\widetilde{p}}  \left[  e^{i (\Theta(\widetilde{X}\mid \chi,\eta) - \mathbb{E}_{\widetilde{p}}[\Theta(\widetilde{X}\mid \chi,\eta)])} \right]\right|,\\
        &\geq \mathbb{E}_{\widetilde{p}}  \left[  \cos( \Theta(\widetilde{X}\mid \chi,\eta) - \mathbb{E}_{\widetilde{p}}[\Theta(\widetilde{X}\mid \chi,\eta)]) \right],\\
        &\geq 1- \frac12 \operatorname{Var}_{\widetilde{p}}(\Theta(\widetilde{X}\mid \chi,\eta)),
    \end{align}
    where the last inequality follows from $\cos(\theta)\geq 1-\theta^2/2$ for $\theta\in \R$. 

    On the other hand, the magnitude of the numerator can be bounded above by the second moment $\widetilde{\mathsf{M}}_2(\chi,\eta) \triangleq \mathbb{E}_{\widetilde{p}}[\|\widetilde{X}\|^2 ]^{1/2}$ of the tilted distribution $\widetilde{p}$ as
    \begin{equation}
        \left\|\mathbb{E}_{\widetilde{p}} \left[\widetilde{X} e^{i \Theta(\widetilde{X}\mid \chi,\eta)} \right]\right\| \leq \mathbb{E}_{\widetilde{p}}[\|\widetilde{X}\| ] \leq \mathbb{E}_{\widetilde{p}}[\|\widetilde{X}\|^2 ]^{1/2}=\widetilde{\mathsf{M}}_2(\chi,\eta).
    \end{equation}
    Hence, $\operatorname{Var}_{\widetilde{p}}(\Theta(\widetilde{X}\mid \chi,\eta))< 2$ ensures that the first integral in \plaineqref{eq:bounded ratio} is non-vanishing and the bound on the ratio-of-expectations in \plaineqref{eq:bounded ratio} is 
    \begin{equation}
        \left\|\frac{\mathbb{E}_{p} \left[X e^{X^\top \chi+  \eta\|X\|^2} \right]}{\mathbb{E}_{p} \left[e^{X^\top \chi + \eta\|X\|^2} \right]}\right\| \leq \frac{\mathbb{E}_{\widetilde{p}}[\|\widetilde{X}\|^2 ]^{1/2}}{1- \frac12 \operatorname{Var}_{\widetilde{p}}(\Theta(\widetilde{X}\mid \chi,\eta))}.
    \end{equation}

    Since $\widetilde{p}$ remains $7/8$-strongly log-concave, Brascamp-Lieb variance inequality in \cref{thm:brascamp lieb} implies
    \begin{align}
        \operatorname{Var}_{\widetilde{p}}(\Theta(\widetilde{X}\mid \chi,\eta)) &\leq \frac{8}{7}\mathbb{E}_{\widetilde{p}}[\|\nabla \Theta(\widetilde{X}\mid \chi,\eta)\|^2],\\
        &= \frac{8}{7}\mathbb{E}_{\widetilde{p}}[\|\Im(\chi) + 2\Im(\eta) \widetilde{X}\|^2],\\
        &\leq \frac{8}{7}\left(\|\Im(\chi)\| +2|\Im(\eta)| \mathbb{E}_{\widetilde{p}}[\|\widetilde{X}\|^2 ]^{1/2}\right)^2 \leq \frac{2}{7}\left(1 + \frac{{\widetilde{\mathsf{M}}_2(\chi,\eta)}}{4\sqrt{d}}\right)^2.\label{eq:variance bound}
    \end{align}
    Thus, both the numerator and the denominator are controlled by the second moment: 
    \begin{equation}\label{eq:ratio bound in M2}
         \left\|\frac{\mathbb{E}_{p} \left[X e^{X^\top \chi+  \eta\|X\|^2} \right]}{\mathbb{E}_{p} \left[e^{X^\top \chi + \eta\|X\|^2} \right]}\right\| \leq \frac{\widetilde{\mathsf{M}}_2(\chi,\eta)}{1- \frac{1}{7} \left(1 + \frac{{\widetilde{\mathsf{M}}_2(\chi,\eta)}}{4\sqrt{d}}\right)^2}.
    \end{equation}

    To get an upper bound on the second moment $\widetilde{\mathsf{M}}_2(\chi,\eta)$, note that the tilted distribution $\widetilde{p}(\widetilde{x}\mid \chi,\eta)$ is $\frac{7}{8}$-strongly log-concave and $\|\nabla_x \log\widetilde{p}(0\mid \chi,\eta)\| = \|\Re(\chi)\| < 1/2$. Thus, \cref{thm:second moment bound} implies 
    \begin{equation}
        \widetilde{\mathsf{M}}_2(\chi,\eta) \leq \frac{2}{7}(1+\sqrt{1+14 d}) \leq \sqrt{2d}.
    \end{equation}
    Substituting this bound on $ \widetilde{\mathsf{M}}_2$ in \plaineqref{eq:variance bound} and \plaineqref{eq:ratio bound in M2} yields $ \operatorname{Var}_{\widetilde{p}}(\Theta(\widetilde{X}\mid \chi,\eta)) < \frac{4}{7} <2$, ensuring the first expectation in \plaineqref{eq:bounded ratio} is non-vanishing and the upper bound on the ratio of expectations holds.
\end{proof}

\begin{lemma}[Local Bounds on Jacobi Elliptic Functions]\label{thm:local elliptic bounds}
    Let $m,\widetilde{m}\in (0,1)$ and define $r \triangleq \min\left(1,\sqrt{\frac{1-\widetilde{m}}{1-m}}\right)$. For any real $u_\circ \in \R$ and $u\in \C$ such that $|u-u_\circ|< r$, the following holds:
    \begin{equation}
        \max\left(|\sn(u\mid m)|, \frac{|\cn(u\mid m)|}{\sqrt{1-\widetilde{m} \sn^2(u_\circ\mid m)}},\, \frac{r|\dn(u\mid m)|}{\sqrt{1-\widetilde{m} \sn^2(u_\circ\mid m)}} \right) \leq \frac{1}{1-r^{-1}|u-u_\circ|}.
    \end{equation}

\end{lemma}
\begin{proof}
    For notational simplicity, let $s_\circ \triangleq  \sqrt{\widetilde{m}}\sn(u_\circ \mid m)$ and define the following functions under the coordinate transformation $v \triangleq \frac{u-u_\circ}{r}$:
    \begin{gather}
        f_1(v) \triangleq \sn(u_\circ + r v\mid m), \quad f_2(v) \triangleq \frac{\cn(u_\circ + r v\mid m)}{\sqrt{1-s_\circ^2}}, \quad f_3(v) \triangleq \frac{r \dn(u_\circ + r v\mid m)}{\sqrt{1-s_\circ^2}},\\
        \text{and}\quad h(\rho) \triangleq \max_{i=1,2,3}\,\max_{|v|\leq \rho}|f_i(v)|\quad \text{for}\quad \rho\geq 0
    \end{gather}
    We aim to show that $h(\rho) \leq \frac{1}{1-\rho}$ whenever $\rho<1$. 
    
    First, all functions in the triplet $(f_1,f_2,f_3)$ are holomorphic on the open unit ball $\mathbb{D} \triangleq \{v\in \C \mid |v|<1\}$, since $|u-u_\circ|< r \leq  1$ implies $|\Im(u)|< 1 < \pi/2 \leq K(1-m)$ where all three Jacobi elliptic functions $\sn(u\mid m)$, $\cn(u\mid m)$, and $\dn(u\mid m)$ are holomorphic. 
    
    Second, we show that $h(0)\leq 1$. Clearly $|f_1(0)| = |\sn(u_\circ\mid m)|\leq 1$ for any real $u_\circ \in \R$. Similarly, $|f_2(0)|\leq 1$ since $\cn^2(u_\circ \mid m) \leq \cn^2(u_\circ \mid m) +  (1-\widetilde{m})  \sn^2(u_\circ \mid m) = 1-s_\circ^2$. To show $|f_3(0)|\leq 1$, one can use the standard Jacobi elliptic function identities:
    \begin{align}
        \dn^2(u_\circ \mid m) &=1 - m \sn^2(u_\circ\mid m),\\
        &= \cn^2(u_\circ\mid m) + (1-m)\sn^2(u_\circ\mid m),\\
        &\leq \max\left(1,\, \frac{1-m}{1-\widetilde{m}}\right)\left[\cn^2(u_\circ\mid m) +  (1-\widetilde{m})\sn^2(u_\circ \mid m)\right ],\\
        &=\frac{1-s_\circ^2}{r^2}.
    \end{align}

    Third, we aim to derive a bound on each $|f_i(v)|$ in terms of $|v|$ for $v\in \mathbb{D}$. Standard derivative identities of Jacobi elliptic functions imply that the triplet $(f_1,f_2,f_3)$ satisfies the following system of differential equations:
    \begin{align}
        \frac{d}{dv} f_1(v) &= c_1  f_2(v) f_3(v),\\
        \frac{d}{dv} f_2(v) &= c_2  f_1(v) f_3(v),\\
        \frac{d}{dv} f_3(v) &= c_3  f_1(v) f_2(v),
    \end{align}
    where the coefficients are $c_1 \triangleq 1-s_\circ^2$, $c_2 \triangleq -1$, and $c_3 \triangleq -r^2 m$, with all three of them satisfying $|c_i|\leq 1$. Integrating these ODEs along the straight line from the origin to any $v \in \mathbb{D}$ and using the bounds $|f_i(0)|\leq 1$ and $|c_i|\leq 1$ give
    \begin{align}
        |f_i(v)| &= \left|f_i(0) + c_i \int_{0}^{1} f_j(\theta v)\, f_k(\theta v) \,v d\theta\right|,\\
        &\leq |f_i(0)| + |c_i| \int_{0}^{1} |f_j(\theta v)|\, |f_k(\theta v)| |v| d\theta,\\
        &\leq 1+ \int_{0}^{1} h^2(\theta |v|) |v| d\theta.
    \end{align}
    Taking the maximum of both sides over $i=1,2,3$ and $|v|\leq \rho <1$ gives the following integral inequality for the monotonic function $[0,1)\ni \rho \mapsto h(\rho)$:    
    \begin{equation}
        h(\rho) \leq H(\rho) \triangleq 1+ \int_{0}^{\rho} h^2(\rho^\prime) d\rho^\prime.
    \end{equation}
    Taking the derivative of $H(\rho)$ gives
    \begin{equation}
        \frac{d}{d\rho}H(\rho) = h^2(\rho) \leq H^2(\rho) \implies \frac{d}{d\rho}\left(-\frac{1}{H(\rho)}\right) \leq 1
    \end{equation}
    Noting $H(0)=1$ and applying the fundamental theorem of calculus, we get the following bound
    \begin{equation}
        -\frac{1}{H(\rho)} = -1 + \int_{0}^{\rho}\frac{d}{d\rho^\prime}\left(-\frac{1}{H(\rho^\prime)}\right) d\rho^\prime  \leq -1 + \rho.
    \end{equation}
    Thus, $h(\rho) \leq H(\rho)\leq \frac{1}{1-\rho}$.
\end{proof}

\subsection{Why smoothing is needed}
\label{app:lc-endpoint}

The preceding guarantees use a positive smoothing level to keep the canonical time away from its terminal values. The following example shows that this restriction reflects a real obstruction: curvature bounds alone do not ensure terminal analyticity, even with Gaussian initialization and a freely chosen VP clock.

\begin{proposition}[Almost-sure terminal nonanalyticity]
\label{ex:non-analytic log-concave example}
For every $\kappa>1$ and $d\ge1$, there is a centered density $\pi_d\propto e^{-V_d}$ on $\mathbb R^d$, with $V_d\in C^2$ and
\begin{equation}
    I_d\preceq\nabla^2V_d(x)\preceq\kappa I_d
    \qquad(x\in\mathbb R^d),
\end{equation}
having the following property. Consider any fixed deterministic continuous VP schedule $(\alpha,\beta)$, continuously differentiable on $(0,1)$, with
\begin{equation}
    (\alpha_0,\beta_0)=(1,0),\qquad
    (\alpha_1,\beta_1)=(0,1),\qquad
    \alpha_t,\beta_t>0\quad(0<t<1).
\end{equation}
The schedule is chosen independently of the sampled initialization. For its probability-flow ODE, write $X_t(z)$ for the trajectory initialized at $X_0(z)=z$, understood by continuous extension at $0$. Then
\begin{equation}
    \mathbb P_{Z\sim N(0,I_d)}
    \bigl(t\mapsto X_t(Z)\text{ is not real analytic at }t=1\bigr)=1.
\end{equation}
In fact, for each such schedule, at most $7^d$ initializations can have a trajectory analytic at $t=1$. The target is independent of the schedule. Here analyticity at $1$ means agreement, for real $t<1$ sufficiently close to $1$, with a real-analytic function on a neighborhood of $1$.
\end{proposition}

\begin{proof}
We first construct a one-dimensional target and analyze its quantiles. Let
\begin{equation}
    \lambda=\frac{\kappa+1}{2},\qquad
    P(x)=(1-x^2)^4,\qquad
    \psi(x)=P(x)\mathbf1_{\{|x|\le1\}},
\end{equation}
and define
\begin{equation}
    p(x)=\sqrt{\frac{\lambda}{2\pi}}e^{-\lambda x^2/2}
         +\eta\psi'(x),
    \qquad \eta>0.
\end{equation}
The function $\psi$ is $C^3$, and $\psi'$ is supported on $[-1,1]$ and has integral zero. On this compact interval, $p$ converges in $C^2$ to the strictly positive Gaussian density as $\eta\downarrow0$. Thus, for sufficiently small fixed $\eta>0$, $p$ is a strictly positive probability density, $V=-\log p\in C^2$, and
\begin{equation}
    1<m\triangleq\frac{\lambda+1}{2}
    \le V''(x)\le\kappa
    \qquad(x\in\mathbb R).
    \label{eq:p05-strong-curvature}
\end{equation}
Outside $[-1,1]$, $V''=\lambda$ exactly. If $W\sim p$, its mean is
\begin{equation}
    \mu=\mathbb E W
       =-\eta\int_{-1}^1(1-x^2)^4\,dx
       =-\frac{256}{315}\eta.
\end{equation}
Our centered scalar target is the law of $W-\mu$; translation preserves the curvature bounds.

\paragraph{Canonical quantile curves.}
Write $\Phi$ for the standard normal CDF and $F$ for the CDF of $W$:
\begin{equation}
    F(x)=\Phi(\sqrt\lambda x)+\eta\psi(x).
\end{equation}
For $q>0$, let $F_q$ be the CDF of $W+\sqrt q\,G$, where $G\sim N(0,1)$ is independent of $W$, and put
\begin{equation}
    H_q f(x)\triangleq\mathbb E f(x+\sqrt q\,G).
\end{equation}
Then
\begin{equation}
    F_q(x)=\Phi\!\left(\frac{x}{\sqrt{q+1/\lambda}}\right)
           +\eta H_q\psi(x).
    \label{eq:p05-heat-cdf}
\end{equation}
For a terminal point $y\in\mathbb R$, define
\begin{equation}
    U_y(q)=F_q^{-1}(F(y)),\qquad
    C_y(q)=\frac{U_y(q)-\mu}{\sqrt{1+q}}.
\end{equation}
These curves extend continuously to $q=0$, with $U_y(0)=y$. For a VP schedule, set $q(t)=\alpha_t^2/\beta_t^2$ on $(0,1)$. Its centered interpolant has the law of $(W+\sqrt{q(t)}G-\mu)/\sqrt{1+q(t)}$. Consequently its scalar PF-ODE trajectory initialized at $z$ is
\begin{equation}
    X_t(z)=C_y(q(t)),\qquad y=F^{-1}(\Phi(z)).
    \label{eq:p05-quantile-trajectory}
\end{equation}
Indeed, the scalar continuity equation implies that the marginal CDF is constant along a PF-ODE trajectory, and its initial quantile is $\Phi(z)$. Positivity of the densities makes the quantile representation unique. Notice that $q(t)>0$ and $q(t)\to0$ as $t\uparrow1$; no analyticity of $q$ is assumed.

\paragraph{An analytic reference and its nonzero remainder.}
Fix $y\ne\pm1$ and let $d_y=\operatorname{dist}(y,\{-1,1\})>0$. We construct a function $A_y$, real analytic near $q=0$, such that
\begin{equation}
    C_y(q)=A_y(q)+r_y(q),\qquad
    r_y(q)\ne0,\qquad
    \lim_{q\downarrow0}q\log|r_y(q)|=-\frac{d_y^2}{2}.
    \label{eq:p05-flat-remainder}
\end{equation}

If $|y|>1$, set $a_y(q)=y\sqrt{1+\lambda q}$. Since $F(y)=\Phi(\sqrt\lambda y)$, \eqref{eq:p05-heat-cdf} gives
\begin{equation}
    F_q(a_y(q))-F(y)=\eta H_q\psi(a_y(q))>0.
\end{equation}
If $|y|<1$, introduce instead
\begin{equation}
    B(q,x)=\Phi\!\left(\frac{x}{\sqrt{q+1/\lambda}}\right)
           +\eta H_qP(x).
\end{equation}
This is real analytic near $(0,y)$: in particular,
\begin{equation}
    H_qP(x)=\sum_{j=0}^4\frac{q^j}{2^j j!}P^{(2j)}(x)
\end{equation}
is a polynomial. Since $B(0,y)=F(y)$ and $\partial_x B(0,y)=p(y)>0$, the analytic implicit-function theorem gives an analytic $a_y(q)$ satisfying $a_y(0)=y$ and $B(q,a_y(q))=F(y)$. Because $P$ is nonnegative on the entire real line,
\begin{equation}
    F_q(a_y(q))-F(y)
    =-\eta H_q\!\left(P\mathbf1_{\{|x|>1\}}\right)(a_y(q))<0.
\end{equation}
The inequalities in both cases are strict for every sufficiently small $q>0$, since Gaussian kernels are strictly positive.

In either case the positive tail integral $J_y(q)$ on the corresponding right-hand side satisfies
\begin{equation}
    \lim_{q\downarrow0}q\log J_y(q)=-\frac{d_y^2}{2}.
    \label{eq:p05-tail-rate}
\end{equation}
For completeness, these are integrals of the form
\begin{equation}
    \frac{1}{\sqrt{2\pi q}}
    \int_E w(x)\exp\!\left(-\frac{(x-a_y(q))^2}{2q}\right)\,dx,
\end{equation}
where $E=(-1,1)$ and $w=\psi$ in the exterior case, or $E=(-\infty,-1)\cup(1,\infty)$ and $w=P$ in the interior case. The weight is positive on $E$, grows at most polynomially, and $\operatorname{dist}(a_y(q),E)\to d_y$. For an upper bound, for any $\theta\in(0,1)$ split the Gaussian exponent into factors with coefficients $1-\theta$ and $\theta$. The first is bounded by $\exp(-(1-\theta)\operatorname{dist}(a_y(q),E)^2/(2q))$; the integral of the second against $w/\sqrt{2\pi q}$ is bounded uniformly for small $q$, because $a_y(q)$ stays bounded. For a lower bound, integrate over a fixed small closed interval in $E$ on which $w>0$ and whose points have distance at most $d_y+\epsilon$ from $y$. Letting first $q\downarrow0$, then $\theta,\epsilon\downarrow0$, proves \eqref{eq:p05-tail-rate}.

Set $A_y(q)=(a_y(q)-\mu)/\sqrt{1+q}$. The mean value theorem applied to $F_q$ between $a_y(q)$ and $U_y(q)$ now proves \eqref{eq:p05-flat-remainder}. Indeed, both points tend to $y$, and Gaussian convolution of $p$ converges locally uniformly to $p$. The intervening density therefore tends to $p(y)>0$, so division by that density does not change the logarithmic rate. The factor $(1+q)^{-1/2}$ also does not change it. Thus $r_y$ is nonzero and $r_y(q)=o(q^N)$ for every $N>0$.

\paragraph{Only one reference curve has zero endpoint derivative.}
Differentiating the defining reference equation at $q=0$ gives $a_y'(0)=-p'(y)/(2p(y))$, in both cases. Therefore
\begin{equation}
    A_y'(0)
    =-\frac12\left(\frac{p'(y)}{p(y)}+y-\mu\right)
    =\frac12\bigl(V'(y)-y+\mu\bigr).
    \label{eq:p05-reference-slope}
\end{equation}
The last expression, defined also at $y=\pm1$, has derivative $(V''(y)-1)/2\ge(m-1)/2>0$ and tends to opposite infinities at the two ends of the real line. Hence it vanishes at exactly one point, denoted $y_c$.

\paragraph{An arbitrary time change cannot regularize two different tail rates.}
Fix a schedule. Suppose that $C_{y_1}(q(t))$ and $C_{y_2}(q(t))$ are both analytic at $t=1$, with $y_i\notin\{-1,1,y_c\}$. We show that $d_{y_1}=d_{y_2}$. Abbreviate $A_i=A_{y_i}$ and $r_i=r_{y_i}$. Since $A_1'(0)\ne0$, its local inverse is analytic. Thus
\begin{equation}
    v(t)=A_1^{-1}\bigl(C_{y_1}(q(t))\bigr)
\end{equation}
is analytic near $1$, satisfies $v(1)=0$, and has the expansion
\begin{equation}
    v(t)
    =q(t)+\frac{r_1(q(t))}{A_1'(q(t))}
       +O(r_1(q(t))^2).
\end{equation}
In particular $v(t)/q(t)\to1$, so $v$ is not identically zero. Its finite order of vanishing implies
\begin{equation}
    q(t)\asymp(1-t)^n
\end{equation}
for some integer $n\ge1$ as $t\uparrow1$.

The function
\begin{equation}
    D(t)=C_{y_2}(q(t))-A_2(v(t))
\end{equation}
is analytic near $1$. With $q=q(t)$, Taylor expansion gives
\begin{equation}
    D(t)
    =r_2(q)-\frac{A_2'(q)}{A_1'(q)}r_1(q)
       +O(r_1(q)^2).
\end{equation}
If $d_{y_1}\ne d_{y_2}$, \eqref{eq:p05-flat-remainder} implies that one of $|r_1|,|r_2|$ dominates the other exponentially. The ratio of derivatives tends to a nonzero constant, so $D(t)$ is nonzero for all $t<1$ sufficiently close to $1$. On the other hand, both remainders are smaller than every power of $q$, and $q(t)\asymp(1-t)^n$. Thus $D(t)$ is smaller than every power of $1-t$. An analytic function with that property vanishes identically near $1$, a contradiction. Therefore $d_{y_1}=d_{y_2}$.

For each $r>0$, the set $\{y:\operatorname{dist}(y,\{-1,1\})=r\}$ contains at most four points. Hence the terminal points with analytic trajectories lie in $\{-1,1,y_c\}$ together with at most four other points. There are at most seven such terminal points and, because $z\mapsto F^{-1}(\Phi(z))$ is a bijection, at most seven initializations. Every finite set has zero standard Gaussian probability.

\paragraph{Extension to dimension $d$.}
Take $\pi_d$ to be the product of $d$ copies of the centered scalar target. Its potential has diagonal Hessian with entries satisfying \eqref{eq:p05-strong-curvature}, and its PF-ODE acts coordinatewise. For each schedule, a vector trajectory is analytic only if all its scalar coordinates are analytic. Thus there are at most $7^d$ such initializations, proving the claim.
\end{proof}

\begin{remark}[Quantifiers and analytic schedules]
The target above is fixed independently of both the schedule and any failure probability $\delta>0$. For each schedule, let $\mathcal E_{\alpha,\beta}$ be its finite set of possibly analytic initializations. Taking $\mathcal Z_\delta=\mathbb R^d\setminus\mathcal E_{\alpha,\beta}$ gives Gaussian probability one, hence greater than $1-\delta$ for every $\delta>0$.

The exceptional set may depend on the schedule. If $\alpha^2$ and $\beta^2$ are analytic near $t=1$, a common full-measure set works for all these schedules:
\begin{equation}
    \mathcal Z
    =\mathbb R^d\setminus\{-\sqrt\lambda,\sqrt\lambda\}^d.
\end{equation}
Indeed, $q(t)$ is then analytic and has a finite-order zero at $1$. For every $y\ne\pm1$, $A_y(q(t))$ is analytic, while $r_y(q(t))$ is nonzero and smaller than every power of $1-t$, so $C_y(q(t))$ is not analytic. The two excluded scalar terminal points correspond to $z=\pm\sqrt\lambda$, since $F(\pm1)=\Phi(\pm\sqrt\lambda)$.

The condition of positive noise immediately before $1$ is essential: a schedule that reaches $(\alpha,\beta)=(0,1)$ early and remains there has constant, hence analytic, terminal trajectories.
\end{remark}

\section{Score-Oracle Complexity for Log-Concave Targets}\label{app:log-concave complexity}

\subsection{Proof of Theorem~\ref{thm:log concave complexity}: The Score-Oracle Complexity Bound}
\label{app:lc-complexity-proof}

We prove \cref{thm:log concave complexity} for $\kappa>1$ in the admissible accuracy regime $0<\varepsilon\le\min(1,\bar\varepsilon)$, where 
\begin{equation}
\bar\varepsilon=(\kappa-1)\sqrt{\frac{d}{\kappa}}.    
\end{equation}
The algorithm is the finite-query collocation method in \cref{alg:collocation method}, applied to the smoothed probability-flow ODE with a projected Gaussian initialization. The supporting smoothing, drift, integration, and initialization estimates are proved below. When $\kappa=1$, the standing assumptions force $\pi=N(0,I_d)$, so returning the Gaussian initialization already gives an exact sample without any score queries.

\begin{proof}[Proof of \cref{thm:log concave complexity} in the stated admissible regime]
Use the theorem parameters: %
\begin{equation}
\begin{gathered}
    \widetilde\sigma^2=\widehat\kappa^{-1}=\frac{\varepsilon}{2\bar\varepsilon},\qquad 
     \widehat m =1-\widehat\kappa^{-1}, \qquad
     B=\sqrt{d}+\sqrt{2\log(32/\varepsilon^2)},\qquad 
    \zeta\triangleq \varepsilon/4,
    \end{gathered}
\end{equation}
The restriction on $\varepsilon \leq \overline{\varepsilon} $ ensures $0<\widetilde\sigma^2\leq 1/2$, $\widehat\kappa\geq 2$, and a real positive radius $B>0$. %

\paragraph{Uniform numerical approximation.}
With $D=\sqrt d+\sqrt{B^2+d\log\kappa}$, \cref{thm:holomorphic strip width} supplies the uniform trajectory bound and strip half-width
\begin{equation}
    M\triangleq B+\sqrt d+1,
    \qquad
    \rho \triangleq \frac{1}{K(\widehat m)\bigl(1+30\sqrt{1-\widetilde\sigma^2}\,D\bigr)}.
    \label{eq:lc-complexity-analytic-data}
\end{equation}
The stadium $\Omega_\rho=\{t\in\C:\operatorname{dist}(t,[0,1])<\rho\}$ is contained in the strip $S_\rho$, so the same holomorphic extension and bound $M$ hold on $\Omega_\rho$. By \cref{thm:lipschtitz}, the smoothed drift is spatially $L$-Lipschitz with $L=K(\widehat m)$ and the endpoint map $z\mapsto\widetilde X_1^z$ is non-expansive. In particular 
\begin{equation}
    R\triangleq \max\{L,\rho^{-1}\}=\rho^{-1}\ge1 \quad \text{and}\quad \zeta\leq M.
\end{equation}
Applying \cref{thm:analytic ODEs} on the initialization set $\{z:\|z\|\le B\}$ gives a numerical endpoint $\widehat X_1^z$ satisfying
\begin{equation}
    \sup_{\|z\|\le B}\|\widehat X_1^z-\widetilde X_1^z\|\le\zeta.
    \label{eq:lc-complexity-uniform-error}
\end{equation}
All segment counts, polynomial degrees, and Picard iteration counts are fixed by the displayed bounds and are uniform over the target class and the initialization ball.

\paragraph{Error allocation.}
By \cref{lem:smoothing-bias} and the inequality $\sqrt{1+u}-1\le u/2$ for $u\ge0$,
\begin{align}
    W_2(\widetilde\pi,\pi)
    &\le\sqrt{\frac d\kappa}\left(\sqrt{1+\widetilde\sigma^2(\kappa-1)}-1\right)\notag\le\frac{\widetilde\sigma^2}{2}(\kappa-1)\sqrt{\frac d\kappa}
    =\frac\varepsilon4.
    \label{eq:lc-complexity-smoothing-budget}
\end{align}
Let $\Pi_B:  x \mapsto \min(1, B/\|x\|)\, x$ denote Euclidean projection onto the closed ball of radius $B$ and output $\widehat X=\widehat X_1(\Pi_BZ)$ for $Z\sim\mathcal N(0,I_d)$. The chosen radius is the radius $B_{\delta}$ in \cref{thm:gaussian tail} for $\delta=\varepsilon/4$, because $4\log(4\sqrt2/\varepsilon)=2\log(32/\varepsilon^2)$. Since $\widetilde X_1(Z)$ has law $\widetilde\pi$ and $z\mapsto\widetilde X_1^z$ is non-expansive, \cref{thm:gaussian tail} and \eqref{eq:lc-complexity-uniform-error} give
\begin{equation}
    W_2\bigl(\mathcal L(\widehat X),\widetilde\pi\bigr)
    \le\left(\mathbb E\|\widehat X_1(\Pi_BZ)-\widetilde X_1\|^2\right)^{1/2}
    \le\zeta+\delta=\frac\varepsilon2.
    \label{eq:lc-complexity-projection-budget}
\end{equation}
The three errors---smoothing, projected initialization, and numerical integration---are thus each at most $\varepsilon/4$. The triangle inequality for $W_2$ yields $W_2(\mathcal L(\widehat X),\pi)\le3\varepsilon/4\le\varepsilon$.

\paragraph{Counting score queries.}
For the original VP schedule, let $s_t$ denote the interpolant score with target $\pi$. The score of the smoothed interpolant is $\widetilde s_t=s_{\tau(t)}$, where
\begin{equation}
    \tau(t)=\beta^{-1}\bigl(\sqrt{1-\widetilde\sigma^2}\,\beta_t\bigr),
    \qquad
    \widetilde b_t(x)=\frac{\dot\beta_t}{\beta_t}\bigl(x+s_{\tau(t)}(x)\bigr).
    \label{eq:lc-complexity-score-query}
\end{equation}
For every stage time $t\in(0,1)$ in \cref{alg:collocation method}, the transformed time satisfies $\tau(t)\in(0,1)$, and the displayed prefactor is finite. Each drift evaluation therefore requires exactly one original-score evaluation. The quadrature nodes are interior, so the algorithm does not query the apparent singularity at $t=0$. Its full score-query count, including every Picard stage, is the drift-query count in \cref{thm:analytic ODEs}.

We next substitute the parameters into that bound. Since $\bar\varepsilon\le\sqrt{\kappa d}$ and $0<\widetilde\sigma\le1/\sqrt2$, the complete elliptic integral satisfies
\begin{equation}
    K(\widehat m)=K(1-\widetilde\sigma^2) \leq \frac \pi 2 + \log\left(\frac{1}{\widetilde\sigma}\right) \leq \frac{\pi}{\log 16} \log\left(\frac{4}{\widetilde\sigma}\right)
    \le C H,
    \label{eq:lc-complexity-elliptic-bound}
\end{equation}
where
\begin{equation}
    C \triangleq \frac{\pi}{\log 4} \quad \text{and}\quad H\triangleq \log\left(\frac{e \kappa (d+1)}{\varepsilon}\right).
\end{equation}
The first inequality follows directly from the definition of the complete elliptic integral of the first kind:
\begin{equation}
    K(1-s^2)=\int_0^{\pi/2}\frac{du}{\sqrt{\sin^2u+s^2\cos^2u}}.
\end{equation}
On $[0,\pi/4]$, use $\sin u\ge2u/\pi$ and $\cos u\ge1/\sqrt2$; on $[\pi/4,\pi/2]$, use $\sin u\ge1/\sqrt2$. The resulting elementary integral is bounded by $\frac{\pi}{\log 16}\log(4/s)$. The third inequality in \eqref{eq:lc-complexity-elliptic-bound} follows from $\widetilde\sigma^{-2}=2\bar\varepsilon/\varepsilon\le2\sqrt{\kappa d}/\varepsilon$.

Because $H\ge1$, $\log(1/\varepsilon)\le H$, and $\log(\kappa+1)\ge\log2$, the initialization radius obeys
\begin{align}
    B^2&\le2d+4\log(32/\varepsilon^2)\le C(d+H),\notag
\end{align}
which implies
\begin{align}
    D&\le C\sqrt{d\log(\kappa+1)+H},\qquad M\le C\sqrt{d+H},\qquad R\le C H\sqrt{d\log(\kappa+1)+H}.\label{eq:lc-complexity-data-bounds}
\end{align}
Consequently
\begin{equation}
    1+L+\log(MR/\zeta)\le C H.
    \label{eq:lc-complexity-log-bound}
\end{equation}
Indeed, after substituting $\zeta=\varepsilon/4$ and \eqref{eq:lc-complexity-data-bounds}, the logarithmic term is at most a constant plus $\log H+\log(1/\varepsilon)+\frac12\log(d+H)+\frac12\log(d\log(\kappa+1)+H)$. Each term is bounded by a universal multiple of $H$, using $\log d\le H$, $\log(\kappa+1)\le C H$, and $\log H\le H$. Combining \cref{thm:analytic ODEs} with \cref{eq:lc-complexity-data-bounds,eq:lc-complexity-log-bound} gives
\begin{equation}
    N\le C H^3\sqrt{d\log(\kappa+1)+H}
    \le C\sqrt d\,H^{3.5},
    \label{eq:lc-complexity-final-bound}
\end{equation}
where the last inequality uses $d\ge1$ and $d\log(\kappa+1)+H\le C dH$. This is the claimed exact score-query bound.
\end{proof}

\subsubsection{Finite-query Integration of Analytic Trajectories}
\label{app:lc-collocation}

We give a finite-query realization of the collocation bound used above. The construction follows the collocation and Picard-iteration framework of \citet[Theorem~2.3]{leeAlgorithmicTheoryODEs2018}. The proof below supplies the derivative approximation, integrated interpolation estimate, and error propagation needed for the present hypotheses. It uses only drift evaluations at real states and real times.

\begin{theorem}[ODEs with Analytic and Bounded Trajectories]\label{thm:analytic ODEs}
    Consider the following ODE
    \begin{equation}
        \frac{dX_t^z}{dt}= b_t(X_t^z), \quad X_0^z = z \in \mathcal{Z}, \quad t\in [0,1],
    \end{equation}
    where the initialization belongs to a set $\mathcal{Z}\subset \R^d$. Suppose the following assumptions hold:
    \begin{itemize}
        \item[i.] There exists $\rho>0$ such that every trajectory $t\mapsto X_t^z$ extends holomorphically to $\Omega_\rho = \{t\in\C\mid \dist(t,[0,1])<\rho\}$ for any initialization $z\in\mathcal{Z}$.
        \item[ii.] There exists $M>0$ such that the trajectories are uniformly bounded on $\Omega_\rho$ by $M$ for any initialization $z\in\mathcal{Z}$, i.e., $\sup_{\substack{t\in \Omega_\rho,\\ z\in \mathcal{Z}}}\|X_t^z\|\leq M$.
        \item[iii.] There exists $L>0$ such that $\|b_t(x)-b_t(y)\|\leq L\|x-y\|$ for all $x,y\in \R^d$ and $t\in [0,1]$.
    \end{itemize}
    Let $R\triangleq \max(1, L,\rho^{-1})$. Then, there exists an algorithm taking an accuracy level $\zeta>0$ and producing a terminal estimate $\widehat{X}_1^z$ such that
    \begin{equation}
        \sup_{z\in\mathcal{Z}}\|\widehat{X}_1^z - X_1^z\|\leq \zeta,
    \end{equation} 
    using at most 
    \begin{equation}\label{eq:lc-analytic-ode-cost}
        N(\rho,M,L,\zeta) \leq C R \left[1+L+\log\left(\max\left(1,\frac{MR}{\zeta}\right)\right)\right]^2,
    \end{equation}
    exact drift evaluations $(x,t)\mapsto b_t(x)$, where $C>0$ is a universal constant. In particular, for $\zeta\in (0,M)$, \cref{alg:collocation method} achieves the desired error and complexity bounds, whereas for $\zeta\geq M$, the zero output suffices without any drift evaluations.
\end{theorem}
For $0<\zeta<M$, define the number of segments $Q$, their length $h$, the auxiliary tolerances $\eta,\delta$, the number of nodes $n$, and the number of Picard iterations $J$ by
\begin{align}
    Q&=\left\lceil\max\{1,2L,4/\rho\}\right\rceil,
    &h&=Q^{-1},\notag\\
    \eta&=\frac{\zeta e^{-2L}}8,
    &\delta&=\frac{\zeta e^{-2L}}{4Q},\notag\\
    n&=\max\left\{1,\left\lceil\log_4\left(\frac{2M}{\rho\eta}\right)\right\rceil\right\},
    &J&=\max\left\{1,\left\lceil\log_2\left(\frac{8M}{\delta}\right)\right\rceil\right\}.
    \label{eq:lc-collocation-parameters}
\end{align}
Let $c_1,\ldots,c_n\in(0,1)$ be the $n$ Gauss--Legendre nodes on $[0,1]$, let $w_1,\ldots,w_n>0$ be their quadrature weights with $\sum_iw_i=1$, and let $\ell_i$ be the associated Lagrange basis polynomials. Precompute
\begin{equation}
    G_{\ell i}=\int_0^{c_\ell}\ell_i(u)\,du,
    \qquad 1\le\ell,i\le n.
    \label{eq:lc-collocation-integration-matrix}
\end{equation}
Gaussian quadrature exactness gives $\int_0^1\ell_i(u)\,du=w_i$. All parameters, nodes, and integration weights are independent of the individual trajectory and of the oracle replies.

\begin{algorithm}[t]
\caption{Finite-query collocation for an analytic trajectory}
\label{alg:collocation method}
\begin{algorithmic}[1]
\REQUIRE Drift oracle $b$, initialization $z$, bounds $\rho,M,L$ as in \cref{thm:analytic ODEs}, tolerance $0<\zeta<M$
\STATE Choose $Q,h,\eta,\delta,n,J$ by \eqref{eq:lc-collocation-parameters} and precompute $c_i,w_i,G_{\ell i}$ by \eqref{eq:lc-collocation-integration-matrix}.
\STATE Set $v_0=z$.
\FOR{$j=0,\ldots,Q-1$}
    \STATE Set $Y_i^{(0)}=v_j$ for $i=1,\ldots,n$.
    \FOR{$k=0,\ldots,J-1$}
        \STATE Query $F_i^{(k)}=b_{jh+hc_i}(Y_i^{(k)})$ for $i=1,\ldots,n$.
        \STATE Set $Y_\ell^{(k+1)}=v_j+h\sum_{i=1}^nG_{\ell i}F_i^{(k)}$ for $\ell=1,\ldots,n$.
        \STATE Set $E^{(k+1)}=v_j+h\sum_{i=1}^nw_iF_i^{(k)}$.
    \ENDFOR
    \STATE Set $v_{j+1}=E^{(J)}$.
\ENDFOR
\RETURN $\widehat X_1^z=v_Q$.
\end{algorithmic}
\end{algorithm}

\begin{proof}[Proof of \cref{thm:analytic ODEs}]
If $\zeta\ge M$, the bound $\|X_1(z)\|\le M$ justifies the zero output. Assume henceforth $0<\zeta<M$, use \eqref{eq:lc-collocation-parameters}, and fix an initialization $z\in\mathcal Z$. All estimates below are uniform in $z$. For a real vector-valued function on an interval $I$, write $\|f\|_{\infty,I}=\sup_{t\in I}\|f(t)\|$.

\paragraph{Approximation of the derivative.}
On the segment $I_j=[jh,(j+1)h]$, let $m_j=(j+1/2)h$. The disk $|t-m_j|<\rho$ lies in $\Omega_\rho$, so Cauchy's formula applied to the bounded vector-valued trajectory gives Taylor coefficients $a_k$ satisfying
\begin{equation}
    X_t^{z}=\sum_{k=0}^{\infty}a_k(t-m_j)^k,
    \qquad \|a_k\|\le M\rho^{-k}.
\end{equation}
The coefficient bound follows first on every circle of radius less than $\rho$, then by taking its radius to $\rho$. Since $h\le\rho/4$, the degree-$(n-1)$ polynomial $q_j(t)=\sum_{k=1}^n k a_k(t-m_j)^{k-1}$ satisfies
\begin{align}
    \|\dot X_t^{z}-q_j(t)\|
    &\le\frac M\rho\sum_{k=n+1}^{\infty}k\left(\frac{|t-m_j|}{\rho}\right)^{k-1}\notag\\
    &\le\frac M\rho\,8^{-n}\left(\frac{n+1}{1-1/8}+\frac{1/8}{(1-1/8)^2}\right)
    \le\frac{2M}{\rho}\,4^{-n}\le\eta,
    \qquad t\in I_j.
    \label{eq:lc-collocation-derivative-approximation}
\end{align}
The penultimate inequality uses $n+1\le2^n$ for $n\ge1$. The coefficients of $q_j$ are real because the trajectory is real on the real interval. These polynomials are used only to analyze approximation error; the algorithm does not compute them or query derivatives.

Let $\mathcal I_j$ denote interpolation at the nodes $jh+hc_i$, and let $\mathcal J_j$ denote integration from $jh$. By \cref{lem:lc-integrated-gauss-interpolation}, the operator $\mathcal J_j\mathcal I_j$ has norm at most $h$ from the maximum Euclidean nodal norm to the uniform Euclidean norm. Because $\mathcal I_jq_j=q_j$, \eqref{eq:lc-collocation-derivative-approximation} implies
\begin{equation}
    \|\mathcal J_j\mathcal I_j\dot X-\mathcal J_j\dot X\|_{\infty,I_j}
    \le2h\eta.
    \label{eq:lc-collocation-consistency}
\end{equation}
One term is bounded using integrated interpolation, and the ordinary integral of $\dot X-q_j$ contributes the other $h\eta$.

\paragraph{Contraction and comparison with the original trajectory.}
For a numerical initial state $v\in\R^d$, define an operator on continuous real curves on $I_j$ by
\begin{equation}
    \mathcal T_v(y)(t)=v+\mathcal J_j\mathcal I_j\bigl[b_{\cdot}(y(\cdot))\bigr](t).
    \label{eq:lc-collocation-operator}
\end{equation}
The interpolation operator uses only the values at the fixed nodes, so $\mathcal T_v$ maps continuous curves to polynomial curves. The spatial Lipschitz assumption and \cref{lem:lc-integrated-gauss-interpolation} give
\begin{equation}
    \|\mathcal T_v(y)-\mathcal T_v(\widetilde y)\|_{\infty,I_j}
    \le hL\|y-\widetilde y\|_{\infty,I_j},
    \qquad hL\le\frac12.
    \label{eq:lc-collocation-contraction}
\end{equation}
Thus $\mathcal T_v$ has a unique fixed point $u_v$. Starting from the constant curve $y^{(0)}(t)=v_j$, the nodal values and endpoint of the Picard curves $y^{(k+1)}=\mathcal T_{v_j}(y^{(k)})$ are exactly those computed by \cref{alg:collocation method}.

Let $e_j=\|v_j-X_{jh}(z)\|$ and $c_h=(1-hL)^{-1}\le2$. Comparing the fixed point with the restriction of the original exact trajectory to $I_j$, using its integral equation and \cref{eq:lc-collocation-consistency,eq:lc-collocation-contraction}, yields
\begin{equation}
    \|u_{v_j}-X\|_{\infty,I_j}
    \le hL\|u_{v_j}-X\|_{\infty,I_j}+e_j+2h\eta,
    \qquad
    \|u_{v_j}-X\|_{\infty,I_j}\le c_h(e_j+2h\eta).
    \label{eq:lc-collocation-fixed-point-error}
\end{equation}
This comparison requires analyticity only of the original trajectory $X_t^{z}$. It does not assume an analytic extension of an exact ODE restarted from an approximate numerical state.

\paragraph{Finite iteration count and propagation.}
Cauchy's estimate gives $\|\dot X_t^{z}\|\le M/\rho$ for every real $t\in[0,1]$. Suppose temporarily that $e_j\le\zeta<M$. Since $h/\rho\le1/4$, $h\le1$, and $\eta\le M/8$, \eqref{eq:lc-collocation-fixed-point-error} gives
\begin{align}
    \|u_{v_j}-v_j\|_{\infty,I_j}
    &\le\|u_{v_j}-X\|_{\infty,I_j}+\|X-v_j\|_{\infty,I_j}\notag\\
    &\le2e_j+4h\eta+e_j+Mh/\rho
    \le3M+M/2+M/4<4M.
\end{align}
The $J$ Picard iterations therefore satisfy
\begin{equation}
    \|y^{(J)}-u_{v_j}\|_{\infty,I_j}
    \le2^{-J}4M\le\delta,
    \qquad
    e_{j+1}\le c_h e_j+2c_h h\eta+\delta.
    \label{eq:lc-collocation-error-recurrence}
\end{equation}
We verify the temporary assumption by induction, starting from $e_0=0$. If the preceding recurrences have been justified and $e_j\le\zeta$, the calculation above establishes the recurrence for segment $j$. Since $-\log(1-hL)\le2hL$, we have $c_h^Q\le e^{2L}$. Unrolling only these already justified recurrences then gives
\begin{equation}
    e_{j+1}
    \le c_h^{j+1}\bigl(2c_h(j+1)h\eta+(j+1)\delta\bigr)
    \le e^{2L}(4\eta+Q\delta)
    =\frac{3\zeta}{4}<\zeta.
    \label{eq:lc-collocation-global-error}
\end{equation}
This closes the induction and proves $e_Q\le\zeta$ without assuming an a priori bound on numerical iterates.

\paragraph{Cost.}
Each of the $Q$ segments uses $J$ iterations of $n$ drift evaluations. The final endpoint uses the same queried values as the final stage update, so there is no additional endpoint query and the count is exactly $QnJ$. Since $R\ge1$ and $0<\zeta<M$, the parameter choices imply
\begin{equation}
    Q\le5R,
    \qquad
    n,J\le C\left(1+L+\log\frac{MR}{\zeta}\right).
\end{equation}
Their product proves \eqref{eq:lc-analytic-ode-cost}. All query states are real and all query times $jh+hc_i$ belong to $(0,1)$.
\end{proof}

\begin{remark}[Arithmetic cost]
The direct dense stage updates require $O(dQJn^2)$ arithmetic operations in addition to the drift evaluations and the precomputation of quadrature data. The theorem is an exact real-oracle query bound. A finite-precision implementation would require additional control of quadrature-coefficient, arithmetic, and oracle errors.
\end{remark}

\begin{lemma}[Integrated Gauss--Legendre interpolation]
\label{lem:lc-integrated-gauss-interpolation}
Let $c_i,w_i,\ell_i$ be the $n$ Gauss--Legendre nodes, weights, and Lagrange basis on $[0,1]$. For arbitrary $a_1,\ldots,a_n\in\R^d$,
\begin{equation}
    \sup_{0\le\theta\le1}\left\|\int_0^\theta\sum_{i=1}^n\ell_i(u)a_i\,du\right\|
    \le\max_{1\le i\le n}\|a_i\|.
    \label{eq:lc-integrated-gauss-bound}
\end{equation}
On an interval of length $h$, the corresponding bound is multiplied by $h$.
\end{lemma}

\begin{proof}
Set $P(u)=\sum_i\ell_i(u)a_i$. The scalar function $\|P(u)\|^2$ is a polynomial of degree at most $2n-2$, so Gaussian quadrature exactness and positivity of the weights imply
\begin{equation}
    \int_0^1\|P(u)\|^2\,du
    =\sum_{i=1}^nw_i\|a_i\|^2
    \le\max_i\|a_i\|^2.
\end{equation}
For every $\theta\in[0,1]$, the integral triangle inequality followed by Cauchy--Schwarz gives
\begin{equation}
    \left\|\int_0^\theta P(u)\,du\right\|
    \le\sqrt\theta\left(\int_0^\theta\|P(u)\|^2\,du\right)^{1/2}
    \le\max_i\|a_i\|.
\end{equation}
This proves \eqref{eq:lc-integrated-gauss-bound}; an affine change of variables gives the factor $h$ on a segment of length $h$.
\end{proof}

\subsubsection{Gaussian initialization tails}

\begin{lemma}[Gaussian initialization tails]\label{thm:gaussian tail}
    For any $B>0$, let $\Pi_{B}: z \mapsto \min(1, B/\|z\|) z$ denote the Euclidean projection onto the ball $\{z\in \R^d \mid \|z\|\leq B\}$. Suppose $T:\R^d \to \R^d$ is a globally $L$-Lipschitz map
    and $\widehat{T}:\R^d \to \R^d$ is measurable such that
    \begin{equation}
        \sup_{\|z\|\leq B_{\delta}}\|\widehat{T}(z)-{T}(z)\|\leq\zeta,\quad \text{where}\quad B_\delta \triangleq \sqrt{d} + \sqrt{4\log\left(\frac{\sqrt{2}}{\delta}\right)},
    \end{equation}
    for some $\zeta>0$ and $\delta\in (0,\sqrt{2})$. Then, 
    \begin{equation}\label{eq:lc-gaussian-initialization-bound}
        \left(\mathbb{E}_{Z\sim N(0,I_d)}\left[\|\widehat{T}(\Pi_{B_\delta}Z)-{T}(Z)\|^2\right]\right)^{1/2} \leq \zeta +L\delta.
    \end{equation}
\end{lemma}

\begin{proof}
The Gaussian moment-generating function is $\mathbb E e^{\theta\|Z\|^2}=(1-2\theta)^{-d/2}$ for $0\le\theta<1/2$. Exponential Markov inequality and optimization in $\theta$ therefore give, for $r\ge\sqrt d$,
\begin{equation}
    \mathbb P(\|Z\|\ge r)
    \le\exp\left[-\frac12\left(r^2-d-d\log\frac{r^2}{d}\right)\right].
\end{equation}
For $r=\sqrt d+u$, use $\log(1+u/\sqrt d)\le u/\sqrt d$ to obtain $\mathbb P(\|Z\|\ge\sqrt d+u)\le e^{-u^2/2}$ for $u\ge0$. Hence, for every $a\ge0$, the tail-integral formula and $(a+v)^2\ge a^2+v^2$ give
\begin{align}
    \mathbb E(\|Z\|-\sqrt d-a)_+^2
    &=\int_0^\infty2v\,\mathbb P(\|Z\|>\sqrt d+a+v)\,dv\notag\\
    &\le2e^{-a^2/2}\int_0^\infty v e^{-v^2/2}\,dv
    =2e^{-a^2/2}.
\end{align}
Taking $a=\sqrt{4\log(\sqrt2/\delta)}$ and using $\|Z-\Pi_{B_\delta}Z\|=(\|Z\|-B_\delta)_+$ yields $\mathbb E\|Z-\Pi_{B_\delta}Z\|^2\le\delta^2$. The triangle inequality in $L^2$, the assumed uniform approximation, and the Lipschitz bound now imply
\begin{align}
    \|\widehat T(\Pi_{B_\delta}Z)-T(Z)\|_{L^2}
    &\le\|\widehat T(\Pi_{B_\delta}Z)-T(\Pi_{B_\delta}Z)\|_{L^2}
    +\|T(\Pi_{B_\delta}Z)-T(Z)\|_{L^2}\notag\\
    &\le\zeta+L\|\Pi_{B_\delta}Z-Z\|_{L^2}
    \le\zeta+L\delta.
\end{align}
This is \eqref{eq:lc-gaussian-initialization-bound}.
\end{proof}

\subsubsection{Smoothing Bias and Drift Lipschitz Constant Bounds}
\label{app:lc-smoothing-real-drift}

\begin{lemma}[Sharp Smoothing Bias]\label{lem:smoothing-bias}
    For $\widetilde\sigma\in(0,1)$, let $\widetilde\pi$ be the density of $\widetilde x_1=\sqrt{1-\widetilde\sigma^2}\,x_1+\widetilde\sigma z$, where $z\sim N(0,I_d)$ and $x_1\sim \pi$ and set
    \begin{equation}
        \widetilde\kappa\triangleq \frac{\kappa}{1+\widetilde\sigma^2(\kappa-1)} < \kappa.
    \end{equation}
    Then,  $I_d\preceq-\nabla^2\log\widetilde\pi(x)\preceq\widetilde\kappa I_d$ for every $x\in\R^d$, and
    \begin{align}
        W_2(\widetilde{\pi}, \pi) &\leq \sqrt{\frac{d}{\kappa}}(\sqrt{\frac{\kappa}{\widetilde\kappa}}-1), \label{eq:lc-sharp-smoothing-w2}
    \end{align}
    where equality is attained by $\pi = N(0,\kappa^{-1} I_d)$. 
\end{lemma}
\begin{proof}
The curvature claim follows from \cref{lem:lc-posterior} with signal coefficient $\sqrt{1-\widetilde\sigma^2}$ and noise standard deviation $\widetilde\sigma$. Introduce the forward Ornstein--Uhlenbeck marginals
\begin{equation}
q_r=\mathcal L\bigl(e^{-r}X_1+\sqrt{1-e^{-2r}}Z\bigr),\qquad W_r=-\log q_r,\qquad r\geq0.
\end{equation}
Here and below the same symbol denotes a law and its density when no confusion is possible. For this proof only, denote the curvature upper bound along this Ornstein--Uhlenbeck curve by $k_r$ and the ratio of the target bound to this bound by $R_r$:
\begin{equation}
k_r=\frac{\kappa}{e^{-2r}+\kappa(1-e^{-2r})},\qquad R_r=\frac{\kappa}{k_r}=\kappa-(\kappa-1)e^{-2r}.
\end{equation}
At $T=-\frac12\log(1-\widetilde\sigma^2)$, these quantities satisfy $q_T=\widetilde\pi$, $k_T=\widetilde\kappa$, and $R_T=\kappa/\widetilde\kappa$. By \cref{lem:lc-posterior}, $I_d\preceq\nabla^2W_r\preceq k_r I_d$. The Ornstein--Uhlenbeck equation is the continuity equation
\begin{equation}
\partial_rq_r+\nabla\cdot(q_rv_r)=0,\qquad v_r(x)=\nabla W_r(x)-x,\qquad 0\preceq Dv_r(x)\preceq(k_r-1)I_d.
\label{eq:lc-ou-velocity}
\end{equation}
Let $T_r(y)$ be the solution of $\partial_rT_r(y)=v_r(T_r(y))$ with $T_0(y)=y$. On each finite time interval these continuously differentiable vector fields are globally Lipschitz in space, so $y\mapsto T_r(y)$ is a global $\mathcal{C}^1$ diffeomorphism. If $J_r(y)=D_yT_r(y)$, differentiating $q_r(T_r(y))\det J_r(y)$ using the continuity and variational equations shows that it equals $\pi(y)$. The identity at $r=0$ follows by continuity. Consequently, $(T_r)_\#\pi=q_r$. Strong convexity gives Gaussian tails, and the upper Hessian bounds give at most linear growth of all scores and velocities used below; these facts justify the expectations and integration by parts.

Because $q_r$ is centered and $\mathbb E_{q_r}\nabla W_r=0$, the vector $v_r$ has mean zero. Applying \cref{thm:brascamp lieb} to each of its components yields
\begin{align}
\mathbb E_{q_r}\|v_r\|^2
&\leq\mathbb E_{q_r}\operatorname{tr}\bigl[(\nabla^2W_r-I_d)(\nabla^2W_r)^{-1}(\nabla^2W_r-I_d)\bigr]\\
&\leq d\,\frac{(k_r-1)^2}{k_r},
\label{eq:lc-ou-speed}
\end{align}
where the last inequality uses that $(\lambda-1)^2/\lambda$ is increasing for $\lambda\geq1$. Minkowski's integral inequality and $(T_r)_\#\pi=q_r$ now give
\begin{align}
\bigl(\mathbb E\|T_T(X_1)-X_1\|^2\bigr)^{1/2}
&\leq\int_0^T\bigl(\mathbb E_{q_r}\|v_r\|^2\bigr)^{1/2}\,dr\\
&\leq\sqrt d\int_0^T\frac{k_r-1}{\sqrt{k_r}}\,dr
=\sqrt{\frac d\kappa}\,(\sqrt{R_T}-1).
\label{eq:lc-ou-displacement}
\end{align}
For the last equality, $R_r'=2(\kappa-R_r)$ and hence $(k_r-1)/\sqrt{k_r}=R_r'/(2\sqrt{\kappa R_r})$. Since $(X_1,T_T(X_1))$ is a coupling of $\pi$ and $q_T$, \eqref{eq:lc-ou-displacement} proves the Wasserstein bound.

When $\pi=N(0,\kappa^{-1}I_d)$, the smoothed density is $N(0,\widetilde\kappa^{-1}I_d)$, and the Gaussian Wasserstein formula gives equality in the bound.
\end{proof}

\begin{lemma}[Lipschitz Constant of Drift under Elliptic Schedule]\label{thm:lipschtitz}
    Fix $\widetilde{\sigma}\in (0,1)$, set $\widehat{m} = 1- \widetilde{\sigma}^{2}$ and consider the elliptic schedule $\alpha_t = \cn(K(\widehat m)t\mid \widehat m)$, $\beta_t = \sn(K(\widehat m)t\mid \widehat m)$ applied to the smoothed target $\widetilde\pi$ in \cref{lem:smoothing-bias}. Write $\widetilde p_t$ for the density of the random interpolant $\widetilde x_t=\alpha_tz+\beta_t\widetilde x_1$ with $\widetilde x_1 \sim \widetilde \pi$ and $\widetilde s_t\triangleq\nabla \log\widetilde p_t$. Its PF-ODE drift
    \begin{equation}
    \widetilde b_t(x)=\frac{\dot\beta_t}{\beta_t}\bigl(x+\widetilde s_t(x)\bigr),
    \end{equation}
    satisfies
    \begin{equation} \label{eq:lc-elliptic-drift-jacobian}
        -K(\widehat{m}) I_d \preceq \nabla \widetilde{b}_t(x)\preceq0, \quad x\in \R^d, t\in [0,1].
    \end{equation}
    Thus, its Lipschitz constant is at most $K(\widehat m)$, i.e., $\operatorname{Lip}(\widetilde{b}_t(\bullet)) \leq K(\widehat{m})=K(1- \widetilde{\sigma}^{2})$ for all $t\in[0,1]$ and the resulting trajectories $X_t^z$ with initialization $X_0^z = z$ are non-expansive, i.e.,
    \begin{equation}\label{eq:lc-elliptic-nonexpansive}
        \|\widetilde{X}_t^{z_1}-\widetilde{X}_{t}^{z_2}\|\leq  \|\widetilde{X}_s^{z_1}-\widetilde{X}_{s}^{z_2}\|\leq \|z_1-z_2\|,
    \end{equation}
    for all $0\leq s\leq t\leq 1$ and $z_1,z_2\in \R^d$.
\end{lemma}
\begin{proof}
Let $\widetilde V_t \triangleq -\log \widetilde p_t$. For $0<t<1$, combining the independent Gaussian terms gives
\begin{equation}
\widetilde x_t\stackrel d=\sqrt{1-\widehat m\beta_t^2}\,z+\sqrt{\widehat m}\,\beta_t x_1.
\end{equation}
Apply \cref{lem:lc-posterior} with signal coefficient $\sqrt{\widehat m}\,\beta_t$ and noise standard deviation $\sqrt{1-\widehat m\beta_t^2}$. The interpolated potential satisfies
\begin{equation}
I_d\preceq\nabla^2\widetilde V_t\preceq\frac1{1-(1-\kappa^{-1})\widehat m\beta_t^2}I_d\preceq\frac1{1-\widehat m\beta_t^2}I_d.
\end{equation}
The derivative identity $\dot\beta_t=K(\widehat m)\alpha_t\sqrt{1-\widehat m\beta_t^2}$ therefore yields
\begin{equation}
-K(\widehat m)\frac{\widehat m\alpha_t\beta_t}{\sqrt{1-\widehat m\beta_t^2}}I_d\preceq \nabla \widetilde b_t(x)\preceq0.
\end{equation}
On this interval, $\alpha_t,\beta_t\geq0$, $\widehat m\alpha_t^2\leq1-\widehat m\beta_t^2$, and $\sqrt{\widehat m}\,\beta_t\leq1$. Hence the scalar ratio in the lower bound lies in $[0,1]$, proving \eqref{eq:lc-elliptic-drift-jacobian} for interior times.

At $t=0$, the apparent singularity is removable. The score identity in \cref{lem:lc-posterior}, expressed using the canonical denoiser, gives
\begin{equation}
x-\nabla\widetilde V_t(x)=\frac{\sqrt{\widehat m}\,\beta_t\widehat{X}_1^\star(x,\sqrt{\widehat m}\,\beta_t)-\widehat m\beta_t^2x}{1-\widehat m\beta_t^2}.
\end{equation}
For fixed $x$, differentiation of the canonical posterior integrals near $s=0$ is justified by Gaussian tail domination. Since $\widehat{X}_1^\star(x,0)=\mathbb E X_1=0$, we have $\widehat{X}_1^\star(x,s)=O(s)$, locally uniformly in $x$. The numerator is therefore $O(\beta_t^2)$, while $\dot\beta_t/\beta_t=O(\beta_t^{-1})$, so the drift tends to zero. At $t=1$, $\dot\beta_t=0$ and the positive terminal smoothing keeps the score finite and continuous, so the drift again tends to zero. Defining $\widetilde b_0=\widetilde b_1=0$ gives the asserted extension, and its endpoint Jacobians obey the same bound.

For two solutions, let $h_t=\widetilde X_t^{z_1}-\widetilde X_t^{z_2}$. The fundamental theorem of calculus in space and the upper bound in \eqref{eq:lc-elliptic-drift-jacobian} imply
\begin{align}
\frac{d}{dt}\|h_t\|^2
&=2h_t^\top\bigl(\widetilde b_t(\widetilde X_t^{z_1})-\widetilde b_t(\widetilde X_t^{z_2})\bigr)\\
&=2\int_0^1h_t^\top \nabla \widetilde b_t(\widetilde X_t^{z_2}+u h_t)h_t\,du\leq0.
\end{align}
Integration over any subinterval of $[0,1]$ proves \eqref{eq:lc-elliptic-nonexpansive}.
\end{proof}

\subsection{Additional Technical Theorems}\label{app:tehcnical theorems}

\begin{lemma}[Cauchy's Estimate]
    Let $f:\mathbb{B}_\gamma(\tau_0) \mapsto \C^d$ be holomorphic and bounded on the ball $\mathbb{B}_\gamma(\tau_0) = \{\tau\in \C \mid |\tau-\tau_0|< \gamma\}$. Then,
    \begin{equation}
        \|f^{(n)}(\tau_0)\|\leq \frac{n!}{\gamma^n}\, \sup_{\tau \in \mathbb{B}_\gamma(\tau_0)} \|f(\tau)\|.
    \end{equation}
    \label{lem:Cauchy}
\end{lemma}

\begin{lemma}[{Existence of Holomorphic Solutions of ODEs \cite[Thm 4.1]{teschl2012ordinary}}]\label{lem:lc-holomorphic-ivp}
Given $(x_0,\tau_0) \in \mathbb{C}^d\times \mathbb{C} $, let $\Omega \triangleq \{(x,\tau) \in \mathbb{C}^d\times\mathbb{C} \mid \|x-x_0\|<\delta,\, |\tau-\tau_0|<\varepsilon \}$ be an open rectangle and suppose $F : \Omega \to \mathbb{C}^d$ is analytic and bounded with 
\begin{equation}
    M \triangleq \sup_{(x,\tau) \in \Omega} \|F(x,\tau)\| < \infty.
\end{equation}
Then the initial value problem 
\begin{equation}
    \frac{dx_\tau}{d\tau} = F(x_\tau,\tau), \quad x_{\tau_0} = x_0,
\end{equation}
has a unique analytic solution defined in the disc $\{\tau \in \mathbb{C} \mid |\tau-\tau_0|<\varepsilon_0\}$, where
\begin{equation}
    \varepsilon_0 \triangleq \min\left(\varepsilon, \frac{\delta}{M}\right).
\end{equation}
\end{lemma}

\begin{lemma}[Brascamp-Lieb variance inequality]%
\label{thm:brascamp lieb}
    For $p(x) \propto \exp(-V(x))$ with $V \in \mathcal{C}^2$ and $\nabla^2V(x)\succ 0$ and any $f \in \mathcal{C}^1$, we have that
    \begin{equation}
        \operatorname{Var}_p(f) \leq \mathbb{E}_{X\sim p}\left[ \nabla f(X)^\top (\nabla^2 V(X))^{-1} \nabla f(X)\right].
    \end{equation}
\end{lemma}

\begin{lemma}[Gaussian posterior identities and curvature]\label{lem:lc-posterior}
Let $x_1\sim\pi\propto e^{-V}$, where $V\in\mathcal{C}^2(\R^d)$ and
$\lambda_{\min}I_d\preceq\nabla^2V\preceq\lambda_{\max}I_d$ for
$0<\lambda_{\min}\leq\lambda_{\max}<\infty$.
For coefficients $\alpha_t>0$ and $\beta_t\in\R$, let
$x_t=\alpha_t z+\beta_t x_1$, where $z\sim N(0,I_d)$ is independent of $x_1$.
Write $p_t$ for the density of $x_t$, $V_t\triangleq-\log p_t$, and
\begin{equation}
    \widehat{X}_1(x,t)\triangleq\E[x_1\mid x_t=x],
    \qquad
    \Sigma_{1\mid t}(x)\triangleq\operatorname{Cov}(x_1\mid x_t=x).
\end{equation}
Then
\begin{align}
    \nabla_x\widehat{X}_1(x,t)
    &=\frac{\beta_t}{\alpha_t^2}\Sigma_{1\mid t}(x),
    \label{eq:lc-posterior-mean-derivative}\\
    s_t(x)=\nabla\log p_t(x)
    &=-\frac{x-\beta_t\widehat{X}_1(x,t)}{\alpha_t^2},
    \label{eq:lc-posterior-score}\\
    \nabla^2V_t(x)
    &=\alpha_t^{-2}I_d-\beta_t^2\alpha_t^{-4}\Sigma_{1\mid t}(x).
    \label{eq:lc-posterior-hessian}
\end{align}
The posterior covariance satisfies
\begin{equation}
    \left(\E[\nabla^2V(x_1)\mid x_t=x]
        +\frac{\beta_t^2}{\alpha_t^2}I_d\right)^{-1}
    \preceq\Sigma_{1\mid t}(x)
    \preceq
    \E\left[\left(\nabla^2V(x_1)
        +\frac{\beta_t^2}{\alpha_t^2}I_d\right)^{-1}
        \,\middle|\,x_t=x\right].
    \label{eq:lc-posterior-conditional-bounds}
\end{equation}
In particular,
\begin{gather}
    \frac{1}{\lambda_{\max}+\beta_t^2/\alpha_t^2}I_d
    \preceq\Sigma_{1\mid t}(x)
    \preceq\frac{1}{\lambda_{\min}+\beta_t^2/\alpha_t^2}I_d,
    \label{eq:lc-posterior-covariance-bounds}\\
    \frac{\lambda_{\min}}{\beta_t^2+\alpha_t^2\lambda_{\min}}I_d
    \preceq\nabla^2V_t(x)
    \preceq\frac{\lambda_{\max}}{\beta_t^2+\alpha_t^2\lambda_{\max}}I_d.
    \label{eq:lc-convolution-curvature}
\end{gather}
For $\pi\in\mathsf P_\kappa$ and the canonical interpolant
$x_u^\star=\sqrt{1-u^2}\,z+u x_1$, $|u|<1$, write
$V_u^\star\triangleq-\log p_u^\star$,
$\widehat{X}_1^\star(x,u)\triangleq\E[x_1\mid x_u^\star=x]$, and
$\Sigma_{1\mid u}^\star(x)\triangleq\operatorname{Cov}(x_1\mid x_u^\star=x)$.
The preceding formulas specialize to
\begin{gather}
    \nabla_x\widehat{X}_1^\star(x,u)
    =\frac{u}{1-u^2}\Sigma_{1\mid u}^\star(x),
    \qquad
    \nabla V_u^\star(x)
    =\frac{x-u\widehat{X}_1^\star(x,u)}{1-u^2},
    \label{eq:lc-canonical-posterior-identities}\\
    \frac{1-u^2}{\kappa(1-u^2)+u^2}I_d
    \preceq\Sigma_{1\mid u}^\star(x)
    \preceq(1-u^2)I_d,
    \label{eq:lc-canonical-covariance-bounds}\\
    I_d\preceq\nabla^2V_u^\star(x)
    \preceq\frac{1}{1-(1-\kappa^{-1})u^2}I_d.
    \label{eq:lc-canonical-curvature}
\end{gather}
\end{lemma}

\begin{proof}
The conditional density of $x_1$ given $x_t=x$ is proportional to
$\exp(-V_{1\mid t}(x_1\mid x))$, where
\begin{equation}
    V_{1\mid t}(x_1\mid x)
    =V(x_1)+\frac{\|x-\beta_t x_1\|^2}{2\alpha_t^2},
    \qquad
    \nabla_{x_1}^2V_{1\mid t}(x_1\mid x)
    =\nabla^2V(x_1)+\frac{\beta_t^2}{\alpha_t^2}I_d.
\end{equation}
Gaussian tail domination permits differentiation of the posterior integrals.
Differentiating the quotient defining each component of $\widehat{X}_1(x,t)$
gives \eqref{eq:lc-posterior-mean-derivative}; differentiating the Gaussian
convolution defining $p_t$ gives \eqref{eq:lc-posterior-score}.
Differentiating this score identity once more gives
\eqref{eq:lc-posterior-hessian}. These identities also hold when $\beta_t=0$.

Apply \cref{thm:brascamp lieb} to the posterior and the function
$x_1\mapsto v^\top x_1$, for each $v\in\R^d$, to obtain the upper inequality
in \eqref{eq:lc-posterior-conditional-bounds}.
For the lower inequality, all expectations below are conditional on $x_t=x$,
and all derivatives of $V_{1\mid t}$ are with respect to $x_1$.
By \cref{thm:stein},
\begin{gather}
    \E[\nabla V_{1\mid t}(x_1\mid x)]=0,
    \qquad
    \E[(x_1-\widehat{X}_1(x,t))
        \nabla V_{1\mid t}(x_1\mid x)^\top]=I_d,\\
    \E[\nabla V_{1\mid t}(x_1\mid x)
        \nabla V_{1\mid t}(x_1\mid x)^\top]
    =\E[\nabla^2V_{1\mid t}(x_1\mid x)].
\end{gather}
The covariance matrix of the pair
$(x_1,\nabla V_{1\mid t}(x_1\mid x))$ is positive semidefinite.
Its lower-right block is positive definite, so its Schur complement gives
$\Sigma_{1\mid t}(x)\succeq(\E[\nabla^2V_{1\mid t}(x_1\mid x)])^{-1}$,
proving the lower inequality.
The curvature assumptions give \eqref{eq:lc-posterior-covariance-bounds};
substitution into \eqref{eq:lc-posterior-hessian} proves
\eqref{eq:lc-convolution-curvature}.
Finally, take $\alpha_t=\sqrt{1-u^2}$, $\beta_t=u$,
$\lambda_{\min}=1$, and $\lambda_{\max}=\kappa$ to obtain the canonical formulas.
\end{proof}

\begin{lemma}[Stein's Identity]
\label{thm:stein}
    Let $X\sim \pi(x)$ be a random variable with $\mathcal{C}^1(\mathcal{X})$ probability density over a domain $\mathcal{X}\subseteq \R^d$. For a smooth function $f\in \mathcal{C}^1(\mathcal{X})$ with finite expectations $\E_\pi[|f(X)|]<\infty$ and $\E_\pi[\|\nabla f(X)\|]<\infty$, it holds that
    \begin{equation}
        \mathbb{E}_{\pi}[\nabla f(X)] = -\mathbb{E}_{\pi}[f(X)\,\nabla \log \pi(X) ] \quad \text{whenever} \quad \int_{\mathcal{X}} \nabla (f(x)\pi(x)) dx = 0.
    \end{equation}
    In particular, for $\mathcal{X}=\R^d$, this holds as long as $f(x)\pi(x)\longrightarrow 0$ as $\|x\|\longrightarrow \infty$.
\end{lemma}

\begin{lemma}[Second-moment bound under strong log-concavity]\label{thm:second moment bound}
    Let $X \sim p(x)\propto \exp(-V(x))$ with $V\in\mathcal{C}^2$ be such that 
    $\nabla^2 V\succeq \lambda_{\min} I_d $ and $\|\nabla V(0)\| \leq M$ for  $\lambda_{\min}>0$ and $M\geq 0$ . Then,
    \begin{equation}
        \mathbb{E}_{p}[\|X\|^2]^{1/2} \leq \frac{M+ \sqrt{M^2 + 4d\lambda_{\min}}}{2\lambda_{\min}} \leq \frac{M+  \sqrt{M^2 + 4\lambda_{\min}}}{2\lambda_{\min}}\sqrt{d}. %
    \end{equation}
\end{lemma}
\begin{proof}
    Using the fundamental theorem of calculus, one gets
    \begin{align}
        \nabla V(x) &= \nabla V(0) + \int_{0}^{1} \nabla^2 V(\theta x) x  d\theta.
   \end{align}
    Taking the inner product of both sides with $x$ yields
   \begin{align}
      x^\top \nabla V(x) &= x^\top \nabla V(0) + \int_{0}^{1} x^\top \nabla^2 V(\theta x) x  d\theta,\\
       &\geq - \|x\| \| \nabla V(0) \| + \lambda_{\min}\|x\|^2,\\
       &\geq  \lambda_{\min}\|x\|^2 - M \|x\| .
   \end{align}
   where the second inequality follows from Cauchy-Schwarz and the strong log-concavity.  Note that, for any $\rho \in (0,1)$, it holds that 
   \begin{align}
        x^\top \nabla V(x)\geq \lambda_{\min}\|x\|^2 - M \|x\| \geq \rho \lambda_{\min} \|x\|^2 - \frac{M^2}{4\lambda_{\min}(1-\rho)}.
   \end{align}
   Taking the expectation of both sides under $p(x)$ and noting $\mathbb{E}_{p}[X^\top\nabla V(X)] = d$ by Stein's identity in \cref{thm:stein} since $\lim_{\|x\|\to\infty} x p(x)=0$ due to strong log-concavity, we get
   \begin{align}
      \mathbb{E}_{p}[\|X\|^2] \leq \frac{1}{\rho}\left(\frac{d}{\lambda_{\min}} +  \frac{M^2}{4\lambda_{\min}^2(1-\rho)}  \right), \forall \rho\in (0,1).
   \end{align}
   Taking the infimum over $\rho\in (0,1)$ yields
   \begin{align}
      \mathbb{E}_{p}[\|X\|^2] \leq \left(\frac{M+ \sqrt{M^2 + 4d\lambda_{\min}}}{2\lambda_{\min}} \right)^2.
   \end{align}
   
\end{proof}

\section{Experiments}
\label{app:experiments}

\subsection{Gaussian experiments}
\label{app:Gaussian-experiments}

\paragraph{Summary.}
On a Gaussian target, every solver we test acts on each precision mode $\lambda$ as a rational function $M(\lambda)$ (\cref{app:algs-are-ratfuncs}), so we evaluate its worst-case relative error over the spectrum, $E=\max_{\lambda}\bigl|\sqrt\lambda\,M(\lambda)-1\bigr|$, using the scalar multiplier. $E$ controls the KL divergence, $D_{\mathrm{KL}}\lesssim d\,E^2$ (\cref{app:compelxity-results-zoloflow}). Two findings:
\begin{itemize}
    \item \emph{Zoloflow wins by orders of magnitude.} The elliptic schedule with the elliptic finite-difference (FD) rule (\cref{def:zoloflow}) has error $2.4\times10^{-14}$ with $12$ score evaluations at $\kappa=100$, where the best tested standard solver--schedule pair has error $4.0\times10^{-2}$ (\cref{app:gauss-results-solvers}).
    \item \emph{Both parameters must match the spectrum.} On an unnormalized spectrum (\cref{app:spectrum_rescaling}), widening the schedule without rescaling, or rescaling without widening, stays orders of magnitude above the optimum (\cref{app:gauss-results-sweep}).
\end{itemize}

\subsubsection{Solvers and schedules on a normalized spectrum}
\label{app:gauss-results-solvers}

\begin{figure}[htbp]
    \centering
    \includegraphics[width=\textwidth,height=.73\textheight,keepaspectratio]{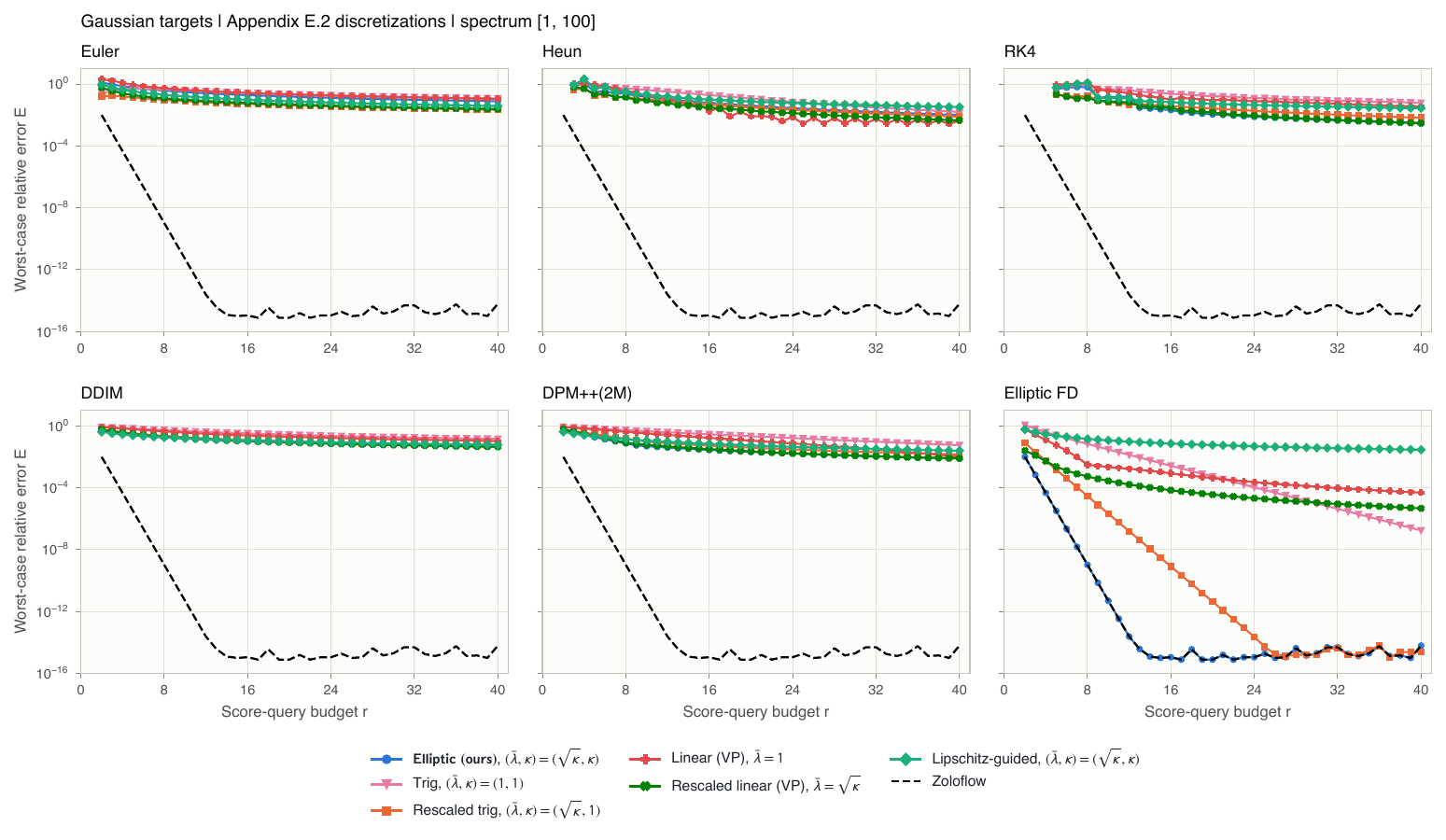}
    \caption{\textbf{Worst-case relative error $E$ against the score-query budget $r$ on the spectrum $[1,100]$.} Panels: solvers; colors: the six schedule settings. Every pair uses \cref{app:catalog-of-discretizations}, including its initialization and closing step. Heun and RK4 use $m\lfloor(r-1)/m\rfloor+1\leq r$ queries, with $m=2$ and $m=4$, respectively, and are shown only for $r>m$; the other solvers use exactly $r$. The black dashed line is the elliptic FD rule with the elliptic schedule ($\widehat E_{100,r}$).}
    \label{fig:gauss-r-app}
\end{figure}

\begin{figure}[htbp]
    \centering
    \includegraphics[width=\textwidth,height=.73\textheight,keepaspectratio]{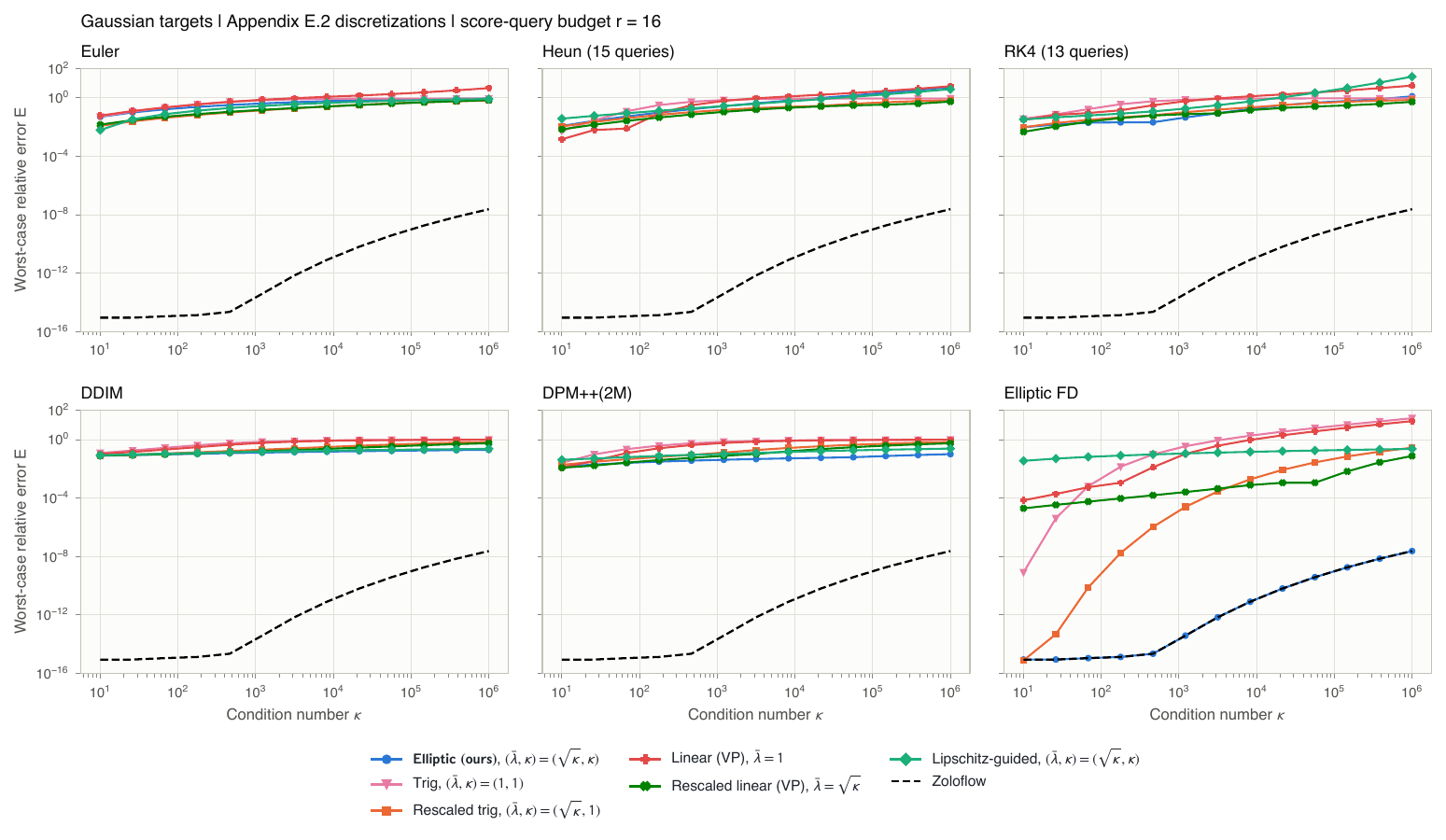}
    \caption{Worst-case relative error $E$ against $\kappa$ on the spectrum $[1,\kappa]$ at score-query budget $r=16$. Appendix~\ref{app:catalog-of-discretizations} uses $15$ queries for Heun, $13$ for RK4, and $16$ for the other solvers, for every schedule. Layout as in \cref{fig:gauss-r-app}.}
    \label{fig:gauss-kappa}
\end{figure}

\Cref{fig:gauss-r-app,fig:gauss-kappa} plot $E$ on the normalized spectrum $[1,\kappa]$ for five standard solvers and the elliptic FD rule (panels), with six schedule settings (colors). All use the discretizations of \cref{app:catalog-of-discretizations}. The black dashed line is the endpoint-normalized Zolotarev error $\widehat E_{\kappa,r}$ from the elliptic FD rule with the elliptic schedule. At $\kappa=100$ and budget $r=12$, its error is $2.4\times10^{-14}$, while the best tested standard solver--schedule pair has error $4.0\times10^{-2}$. FD with rescaled trig has error $1.6\times10^{-7}$, and FD with rescaled Linear (VP form) has error $1.6\times10^{-4}$. At this budget, rescaled trig improves on the elliptic schedule for Euler and Heun by factors $3.8$ and $1.5$, respectively. At budget $r=16$, Zoloflow stays below $2.4\times10^{-8}$ over the tested spectra up to $\kappa=10^6$; at $\kappa=10^6$, the best tested standard solver--schedule pair has error $0.102$, and FD with rescaled Linear has error $0.076$. The elliptic schedule with any of the five standard solvers stays above $10^{-4}$ for the tested budgets $r\leq40$, and FD with unscaled Trig still has error $1.3\times10^{-2}$ at $r=12$.

\subsubsection{Both center and width must match the spectrum}
\label{app:gauss-results-sweep}

\begin{figure}[htbp]
    \centering
    \includegraphics[width=\textwidth,height=.73\textheight,keepaspectratio]{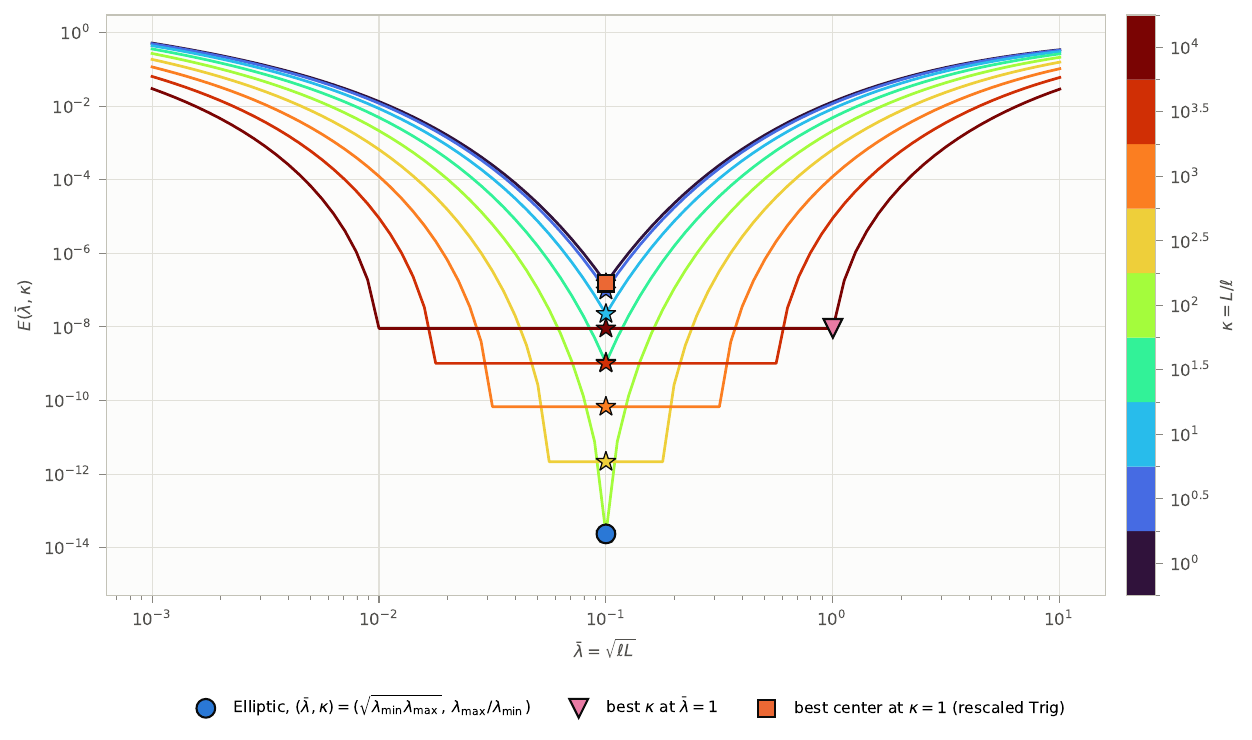}
    \caption{Gaussian target with spectrum $[0.01,1]$, $r=12$: $E$ against the center $\bar\lambda=\sqrt{\ell L}$, one curve per width $\kappa=L/\ell$; stars mark each curve's minimum over $\bar\lambda$, in the middle of its flat segment where there is one, on the grid $\log_{10}\bar\lambda\in[-3,1]$ (step $0.05$), $\log_{10}\kappa\in[0,4]$ (step $0.125$). Blue circle: $(\bar\lambda,\kappa)=(\sqrt{\lambda_{\min}\lambda_{\max}},\lambda_{\max}/\lambda_{\min})$; pink triangle: the best $\kappa$ at $\bar\lambda=1$; orange square: the best rescaled trig, $\kappa=1$.}
    \label{fig:gauss-sweep}
\end{figure}

\Cref{fig:gauss-sweep} takes a Gaussian whose spectrum $[\lambda_{\min},\lambda_{\max}]=[0.01,1]$ is not normalized. It runs Generalized Zoloflow (\cref{def:generalized-zoloflow}), the exponential-gauge rescaling of \cref{sec:exponential-rescaling}, with $r=12$ at every center and width $(\bar\lambda,\kappa)$ on a grid, that is, with the endpoints $(\ell,L)=(\bar\lambda/\sqrt{\kappa},\bar\lambda\sqrt{\kappa})$ in place of $(\lambda_{\min},\lambda_{\max})$. By \eqref{eq:generalized-zoloflow-output}, $M(\lambda)=\ell^{-1/2}\,\widehat q_{\kappa,r}(\lambda/\ell)$ per mode, so the spectrum's own center and width, $(\bar\lambda,\kappa)=(\sqrt{\lambda_{\min}\lambda_{\max}},\lambda_{\max}/\lambda_{\min})=(0.1,100)$, give $E=\widehat E_{100,12}$.

Each curve fixes $\kappa$, and its star marks the curve's minimum over $\bar\lambda$, placed in the middle of its flat segment where there is one. Every curve is centered on the spectral center $\bar\lambda=0.1$, and only the $\kappa=100$ curve reaches $\widehat E_{100,12}$ there (blue circle). Without widening ($\kappa=1$, rescaled trig), the best center is $\bar\lambda=0.1$ (orange square, $E=1.6\times10^{-7}$), about seven orders of magnitude above the matched setting ($2.4\times10^{-14}$). Without centering ($\bar\lambda=1$), even the widest schedule on the grid, $\kappa=10^4$, only reaches $E=9.1\times10^{-9}$ (pink triangle); Trig itself, $(\bar\lambda,\kappa)=(1,1)$, has $E=1.3\times10^{-2}$. The flat segments are exact: once $[\ell,L]$ contains the spectrum, $E=\widehat E_{\kappa,r}$ for every such center, so a schedule wider than the spectrum pays for its own, larger $\kappa$.

\subsubsection{Setup}
\label{app:gauss-setup}

\paragraph{Solvers and schedules.}
Euler, Heun, RK4, DDIM and DPM-Solver++(2M) (\cref{app:additional-discretizations}), and the elliptic FD rule of \cref{def:zoloflow} on its shared grid \eqref{eq:shared-query-grid}. Both figures include Elliptic \eqref{eq:appendix-elliptic-schedule} with $(\bar\lambda,\kappa)=(\sqrt{\kappa},\kappa)$ (endpoints $[1,\kappa]$), Trig with $(\bar\lambda,\kappa)=(1,1)$, rescaled Trig with $(\bar\lambda,\kappa)=(\sqrt\kappa,1)$, Linear and rescaled Linear with $\bar\lambda=1$ and $\bar\lambda=\sqrt\kappa$, and Lipschitz-guided \eqref{eq:drift-lipschitz-schedule} \citep{chen2025lipschitz}. Linear uses the normalized VP coefficients $(\alpha_t,\beta_t)=(1-t,t)/\sqrt{(1-t)^2+t^2}$, which preserve the linear interpolant's signal-to-noise ratio. Each rescaled Trig or Linear run integrates on the target rescaled by $\ell=\sqrt\kappa$ and divides the output by $\sqrt\ell$ (\cref{app:target-rescaling}). For FD, this specifies the implementation: the discrete update is not generally invariant under a time-dependent spatial rescaling.

\paragraph{Initialization and budget.}
All solvers use \cref{app:catalog-of-discretizations}: $x^0=z$ is the state at $t_1=1/(2r+1)$. One-query methods evaluate at $t_1,\ldots,t_r$ and use exactly $r$ score queries. Heun and RK4 use $M=\lfloor(r-1)/m\rfloor$ macro-steps from $t_1$ to $t_r$, followed by the Euler closing step \eqref{eq:si-closing-step}; they use $mM+1\leq r$ queries and never query at $t=1$. The horizontal variable $r$ is a query budget for these multistage methods.

\paragraph{Computing $E$.}
Each solver is run per mode on $4001$ log-spaced $\lambda$ values, with scalar refinement of the five largest local maxima and explicit inclusion of both endpoints. Values below $10^{-14}$ are treated as floating-point round-off.

\subsection{Log-concave experiments}
\label{app:logconcave-experiments}

\paragraph{Summary.}
The Gaussian theory sets the elliptic schedule parameters from the precision spectrum. For the log-concave targets here, we compare the covariance rule, which uses the endpoints of $\operatorname{spec}(\Sigma^{-1})$, with the curvature-endpoint rule and rescaled trig. These product targets have exact scores, and KL is evaluated by quadrature. The results depend on the solver: covariance-based widening improves on rescaled trig for DDIM and DPM-Solver++(2M) on targets A and B, but Euler and Heun have exceptions. On the 32-dimensional targets H1 and H2, the covariance rule is within factors $1.53$ and $3.47$ of the best displayed grid setting for DPM-Solver++(2M). Schedules, targets and the metric are defined in \cref{app:lc-schedules,app:lc-targets,app:lc-metric}.

\paragraph{Discretizations and budget.}
Both this subsection and \cref{app:Gaussian-experiments} use exactly \cref{app:catalog-of-discretizations}, including initialization at $t_1$. In particular, Heun takes $\lfloor(r-1)/2\rfloor$ macro-steps to $t_r$, followed by one Euler step to $1$; its plotted budgets $r=4,8,16,32,64$ use $3,7,15,31,63$ score queries. DDIM, DPM-Solver++(2M), and Euler use exactly $r$ queries.

\subsubsection{Two-dimensional targets: which rule to use}
\label{app:lc-results-2d}

\begin{figure}[p]
    \centering
    \includegraphics[width=\textwidth,height=.73\textheight,keepaspectratio]{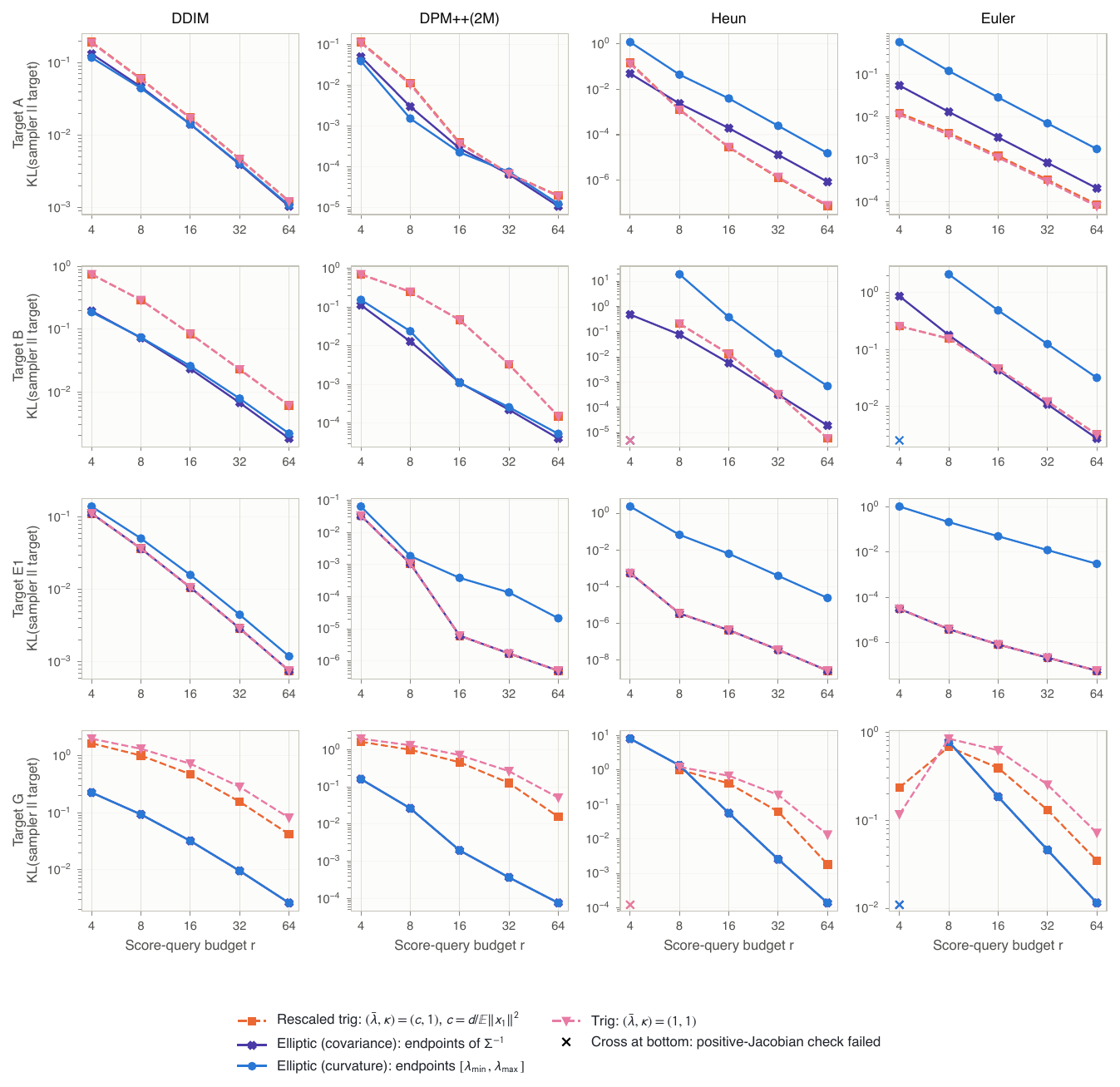}
    \caption{KL$(F_\#\gamma\,\|\,p_1)$ against score-query budget $r$, using \cref{app:catalog-of-discretizations}. Rows: targets A, B, E1, and G; columns: solvers. Heun uses $r-1$ queries at these even budgets. Purple is the covariance rule, orange rescaled trig, blue the curvature-endpoint rule, and pink unscaled trig. The scale is $c=c_{\mathrm{var}}=d/\mathbb E\|x_1\|^2$. Trig and rescaled trig are close on A and B and coincide on E1. A $\times$ marks failure of the positive-Jacobian check; the corresponding KL is not reported.}
    \label{fig:lc-2d}
\end{figure}

\Cref{fig:lc-2d} plots KL against the query budget on four two-dimensional targets (rows, \cref{tab:lc-targets}) for four solvers (columns). Purple denotes the covariance rule, orange rescaled trig, and blue the curvature-endpoint rule.
\begin{itemize}
    \item \emph{A, B: scales differ across directions.} The covariance rule improves on rescaled trig for DDIM and DPM-Solver++(2M) at every displayed budget. This does not hold for Euler or Heun. On A, rescaled trig is better for Euler at all five budgets and for Heun at $r=8,16,32,64$; the covariance rule's Heun KL is respectively $1.82$, $7.07$, $10.4$, and $12.0$ times larger.
    \item \emph{E1: $\Sigma=I$.} Covariance-based elliptic, rescaled trig, and unscaled trig coincide. The curvature rule instead uses $\kappa=100$ and has larger KL for every displayed solver and budget, with strongly solver-dependent ratios.
    \item \emph{G: Gaussian.} The covariance and curvature rules coincide because both recover the precision endpoints. Their improvement over rescaled trig depends on the solver and budget; it is not a pointwise dominance statement.
\end{itemize}

\begin{table}[htbp]
\centering
\small
\begin{tabular}{llrrrr}
\toprule
target & numerator / covariance & DDIM & DPM++(2M) & Heun & Euler \\
\midrule
A & rescaled trig & $1.23\ (5)$ & $1.84\ (5)$ & $0.142\ (5)$ & $0.368\ (5)$ \\
B & rescaled trig & $3.63\ (5)$ & $14.9\ (5)$ & $1.66\ (4)$ & $1.06\ (5)$ \\
E1 & curvature & $1.48\ (5)$ & $42.7\ (5)$ & $1.09\!\times\!10^4\ (5)$ & $5.48\!\times\!10^4\ (5)$ \\
G & rescaled trig & $14.8\ (5)$ & $204\ (5)$ & $10.2\ (4)$ & $2.48\ (4)$ \\
\bottomrule
\end{tabular}
\caption{Median KL ratios computed separately for each target and solver over the same budgets $r\in\{4,8,16,32,64\}$. Ratios exceed one when the covariance rule is better. Parentheses give the number of valid paired comparisons: a budget is omitted if either map fails the positive-Jacobian check, or the denominator is at the $10^{-14}$ numerical floor. No targets or solvers are pooled.}
\label{tab:lc-rule-ratios}
\end{table}

\subsubsection{Validation in $d=32$: distance to the best $(\bar\lambda,\kappa)$}
\label{app:lc-results-sweep}

\begin{figure}[htbp]
    \centering
    \includegraphics[width=\textwidth,height=.73\textheight,keepaspectratio]{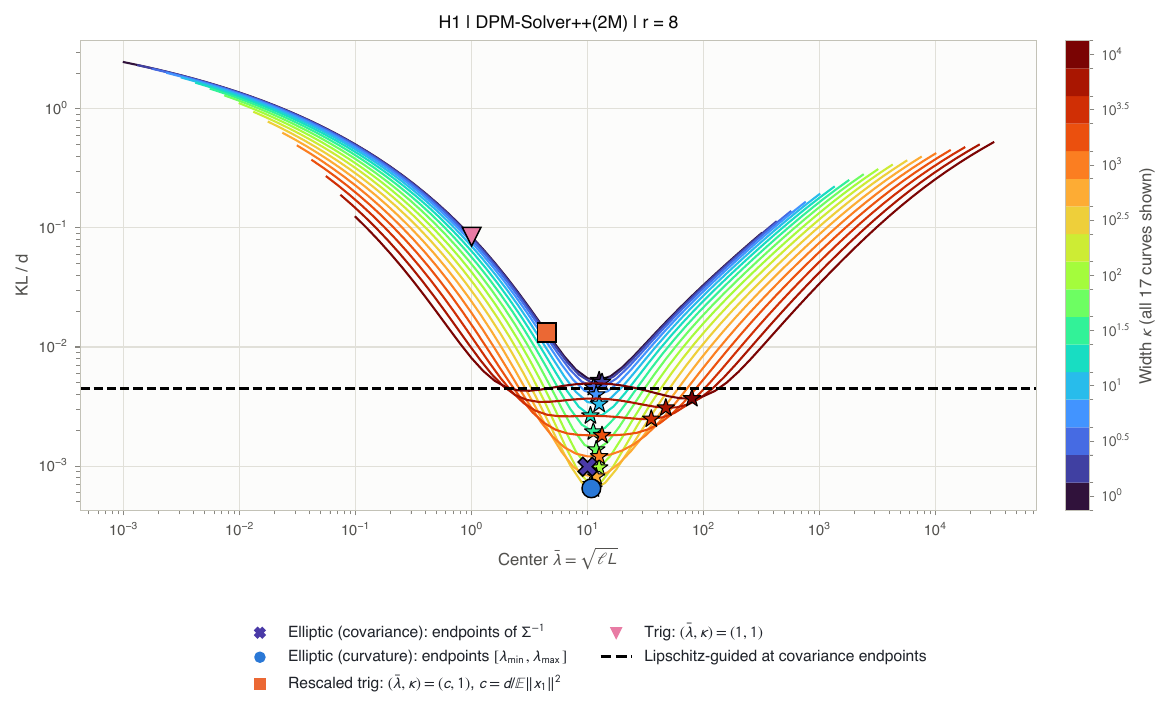}
    \includegraphics[width=\textwidth,height=.73\textheight,keepaspectratio]{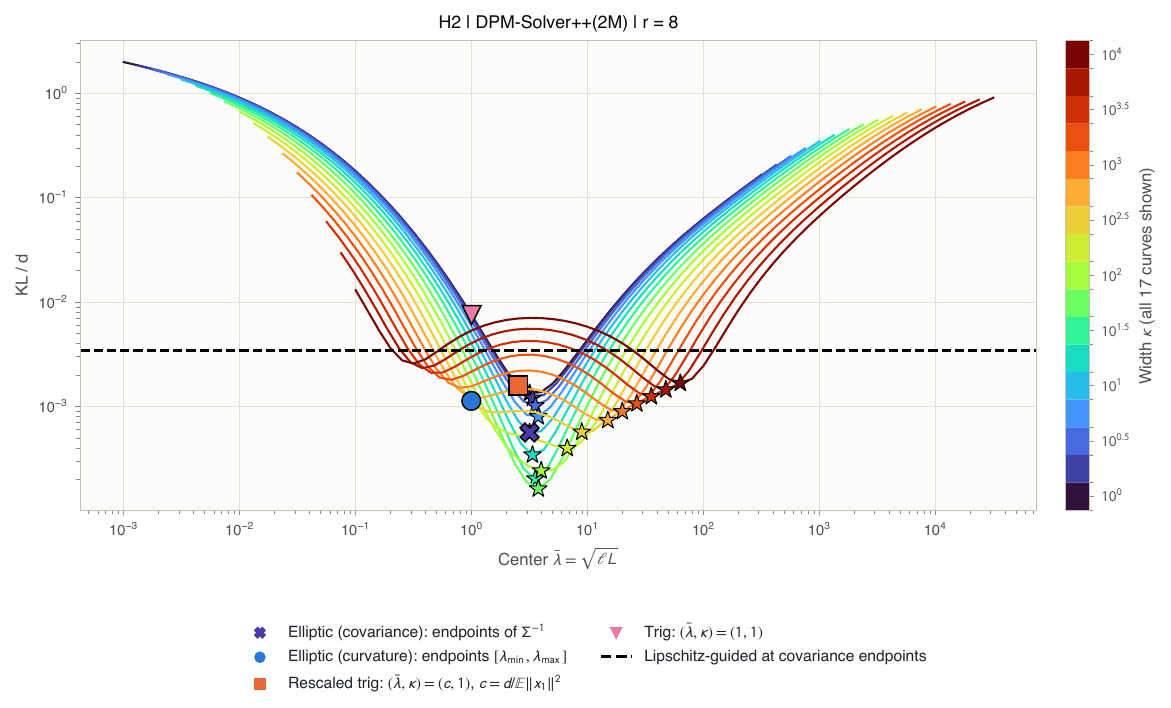}
    \caption{KL$/d$ of DPM-Solver++(2M), $r=8$, against $\bar\lambda=\ell\sqrt\kappa$; top H1, bottom H2. The grid has $\log_{10}\ell=-3+0.1j$, $j=0,\ldots,55$, and $\log_{10}\kappa=0.25k$, $k=0,\ldots,16$. All 17 width curves and their minima are plotted, without interpolation. The horizontal center coordinates therefore shift with the width. Ratios in the text use the minimum over exactly these 952 plotted settings. The purple cross is the covariance rule; the dashed line is Lipschitz-guided at the same endpoints.}
    \label{fig:lc-sweep}
\end{figure}

\Cref{fig:lc-sweep} evaluates DPM-Solver++(2M) with $r=8$ on targets H1 and H2. Every curve is an evaluated width slice, and its star marks its minimum over the tested lower endpoints. All 17 quarter-decade widths used in the numerical comparison are shown. Relative to the minimum over the displayed $56\times17$ grid, the covariance rule is worse by factors $1.53$ (H1) and $3.47$ (H2). The corresponding ratios are $20.6$ and $9.77$ for rescaled trig, $131$ and $46.2$ for unscaled trig, $6.94$ and $21.1$ for matched Lipschitz-guided, and $1.01$ and $6.96$ for the curvature rule. The grid minima are at $(\ell,\kappa)\approx(0.631,316)$ for H1 and $(0.501,56.2)$ for H2. These are minima over evaluated settings, not continuous optima.

\subsubsection{Schedules under comparison}
\label{app:lc-schedules}

All schedules are variance preserving, $\alpha_t^2+\beta_t^2=1$, with $(\alpha_0,\beta_0)=(1,0)$ and $(\alpha_1,\beta_1)=(0,1)$. The elliptic schedule with condition number $\kappa\geq 1$ is
\begin{equation}
    \alpha_t=\operatorname{cn}\!\big(K(m)\,t\mid m\big),\qquad
    \beta_t=\operatorname{sn}\!\big(K(m)\,t\mid m\big),\qquad
    m=1-\kappa^{-1},
    \label{eq:lc-elliptic}
\end{equation}
which reduces to $(\cos\tfrac{\pi t}{2},\sin\tfrac{\pi t}{2})$ at $\kappa=1$.

A \emph{schedule setting} is a center and width $(\bar\lambda,\kappa)$, equivalently the endpoints $(\ell,L)=(\bar\lambda/\sqrt{\kappa},\bar\lambda\sqrt{\kappa})$. It is realized by \emph{target rescaling} (\cref{app:target-rescaling}): the elliptic schedule \eqref{eq:lc-elliptic} with width $\kappa$ is run on the rescaled target $\sqrt{\ell}\,x_1$, and the output is divided by $\sqrt{\ell}$. We compare four settings of this family:
\begin{center}
\begin{tabular}{ll}
\toprule
schedule & setting $(\bar\lambda,\kappa)$ \\
\midrule
Trigonometric & $(1,1)$ \\
Rescaled trig & $(c,1)$, $c=c_{\mathrm{var}}=d/\mathbb E\|x_1\|^2$ \\
Elliptic (ours) & center and width of $\operatorname{spec}(\Sigma^{-1})$, \cref{app:cov-heuristic} \\
Elliptic, curvature endpoints & center and width of $[\lambda_{\min},\lambda_{\max}]$ \\
\bottomrule
\end{tabular}
\end{center}
Rescaled trig coincides with the closed-form schedule of Align Your Steps for isotropic Gaussian data with the target's average variance \citep[Theorem~3.1]{sabour_align_2024}. The sweep also shows the Lipschitz-guided schedule \eqref{eq:drift-lipschitz-schedule} of \citet{chen2025lipschitz} at the same $(\bar\lambda,\kappa)$ as Elliptic (ours).

\subsubsection{Target family: products of one-dimensional Gaussian mixtures}
\label{app:lc-targets}

\paragraph{Definition.}
The target is a product of $d$ independent, symmetric, two-component mixtures,
\begin{equation}
    p_1(y)=\prod_{i=1}^{d}p_i(y_i),\qquad
    p_i=\tfrac12\,\mathcal N(\mu_i,\sigma_i^2)+\tfrac12\,\mathcal N(-\mu_i,\sigma_i^2),
    \qquad 0\le\mu_i<\sigma_i .
    \label{eq:lc-target}
\end{equation}
Writing $p_i=e^{-V_i}$, a direct computation gives
\begin{equation}
    V_i(y)=\frac{y^2+\mu_i^2}{2\sigma_i^2}-\log\cosh\!\Big(\frac{\mu_i y}{\sigma_i^2}\Big)+\tfrac12\log(2\pi\sigma_i^2),
    \qquad
    V_i''(y)=\frac{1}{\sigma_i^2}-\frac{\mu_i^2}{\sigma_i^4}\operatorname{sech}^2\!\Big(\frac{\mu_i y}{\sigma_i^2}\Big).
    \label{eq:lc-V}
\end{equation}
The target has mean zero, covariance $\operatorname{Cov}(x_1)=\operatorname{diag}(\sigma_i^2+\mu_i^2)$, and second moment $\mathbb E\|x_1\|^2=\sum_i(\sigma_i^2+\mu_i^2)$.

\paragraph{Curvature.}
The Hessian $\nabla^2V(y)=\operatorname{diag}\big(V_1''(y_1),\dots,V_d''(y_d)\big)$ is diagonal at every point: its eigenbasis is fixed and its eigenvalues vary with position. With $\rho_i=\mu_i/\sigma_i$,
\begin{equation}
    \label{eq:lc-bounds}
\begin{aligned}
    V_i''(y)&\in\Big[\lambda_{\min,i},\,\lambda_{\max,i}\Big),\\
        \lambda_{\min,i}&=\frac{1-\rho_i^2}{\sigma_i^2}\ \text{(attained at }y=0), \\
        \lambda_{\max,i}&=\frac{1}{\sigma_i^2}\ \text{(approached as }|y|\to\infty),
    \end{aligned}
\end{equation}
so $p_1$ is $\lambda_{\min}$-strongly log-concave and $\lambda_{\max}$-smooth with $\lambda_{\min}=\min_i\lambda_{\min,i}$, $\lambda_{\max}=\max_i\lambda_{\max,i}$. Strong log-concavity requires $\rho_i<1$ strictly; at $\rho_i=1$ the mixture is exactly at the threshold where it would become bimodal and $\lambda_{\min,i}=0$. To prescribe a per-coordinate curvature range $[\lambda_{\min,i},\lambda_{\max,i}]$ we invert \eqref{eq:lc-bounds}:
\begin{equation}
    \sigma_i^2=\frac1{\lambda_{\max,i}},\qquad \mu_i^2=\frac{\lambda_{\max,i}-\lambda_{\min,i}}{\lambda_{\max,i}^2},\qquad
    \text{so that }\ \kappa_i\triangleq\frac{\lambda_{\max,i}}{\lambda_{\min,i}}=\frac{1}{1-\rho_i^2}.
    \label{eq:lc-constructor}
\end{equation}
The choice $\lambda_{\min,i}=\lambda_{\max,i}$ (i.e.\ $\mu_i=0$) makes coordinate $i$ Gaussian with precision $\lambda_{\max,i}$. Equivalently, a coordinate with variance $v_i$ and $\rho_i^2\in[0,1)$ has $\sigma_i^2=v_i/(1+\rho_i^2)$, $\mu_i^2=v_i\rho_i^2/(1+\rho_i^2)$ and curvature range $[(1-\rho_i^4)/v_i,\,(1+\rho_i^2)/v_i]$. We call coordinates with $\rho_i\approx1$ \emph{flat}: their curvature at the center, $\lambda_{\min,i}$, approaches $0$.

\paragraph{Targets.}
\Cref{tab:lc-targets} lists the targets. A and B have scales that differ across directions; E1 has $\Sigma=I$, so its curvature varies only within a direction; G is Gaussian. In $d=32$, H1 spreads its variances over two decades with mild flattening, and H2 is dominated by flattening.
\begin{table}[h]
\centering
\begin{tabular}{lll}
\toprule
target & $d$ & coordinates \\
\midrule
A & 2 & curvature ranges $[0.1,1]$, $[1,10]$ \\
B & 2 & curvature ranges $[0.01,1]$, $[1,100]$ \\
E1 & 2 & $v=(1,1)$, $\rho^2=(0,0.99)$ \\
G & 2 & Gaussian, $v=(1,10^{-3})$ \\
H1 & 32 & $v_k=10^{-2k/31}$, $\rho_k^2=0.8\,g_k$ \\
H2 & 32 & $v_k=10^{-k/31}$, $\rho_k^2=0.9+0.09\,g_k$ \\
\bottomrule
\end{tabular}
\caption{Log-concave targets. For H1 and H2, $k=0,\dots,31$ and $g_k=(k+1)\omega-\lfloor(k+1)\omega\rfloor$ with $\omega=(\sqrt5-1)/2$.}
\label{tab:lc-targets}
\end{table}

\paragraph{Closed-form interpolated law.}
Because the interpolant noise is isotropic, $x_t=\alpha_t z+\beta_t x_1$ remains a product, and each factor is again a two-component mixture:
\begin{equation}
    x_{t,i}\sim\tfrac12\,\mathcal N(\beta_t\mu_i,\,v_{t,i})+\tfrac12\,\mathcal N(-\beta_t\mu_i,\,v_{t,i}),
    \qquad v_{t,i}=\alpha_t^2+\beta_t^2\sigma_i^2 .
    \label{eq:lc-pt}
\end{equation}
Hence the score is available exactly for every $t$,
\begin{equation}
    s_{t,i}(x)=\partial_x\log p_{t,i}(x)=-\frac{x}{v_{t,i}}+\frac{\beta_t\mu_i}{v_{t,i}}\tanh\!\Big(\frac{\beta_t\mu_i\,x}{v_{t,i}}\Big),
    \label{eq:lc-score}
\end{equation}
and $\operatorname{Var}(x_{t,i})=\alpha_t^2+\beta_t^2(\sigma_i^2+\mu_i^2)$. Every solver is therefore run with the \emph{exact} score, so the measured error is discretization error alone and is not confounded with score estimation.

\subsubsection{Metrics}
\label{app:lc-metric}

Let $F:\mathbb R^d\to\mathbb R^d$ denote the output map implemented by a solver at query budget $r$, and let $\gamma=\mathcal N(0,I_d)$. The sampler's output law is the pushforward $F_\#\gamma$.

\paragraph{KL divergence by change of variables.}
If $F$ is a $C^1$ bijection, then $(F_\#\gamma)(F(z))=\gamma(z)/|\det\nabla F(z)|$, and substituting $x=F(z)$ in the definition of KL gives
\begin{equation}
    \mathrm{KL}\big(F_\#\gamma\,\big\|\,p_1\big)
    =\mathbb E_{z\sim\gamma}\Big[\log\gamma(z)-\log|\det\nabla F(z)|-\log p_1\big(F(z)\big)\Big].
    \label{eq:lc-kl}
\end{equation}
Every term is available exactly at every $z$: $\log p_1$ is the closed-form mixture density, and $\nabla F$ follows from the chain rule through the solver. For a one-query step $x\mapsto a_j x+b_j\,w_{t_j}(x)$ the derivative is $a_j+b_j\,\partial_x w_{t_j}(x)$, with $\partial_x w=1+\partial_x s$ closed form from \eqref{eq:lc-score}; multistage and multistep solvers propagate the running Jacobian through their stages. In the product setting $F$ is coordinatewise, $F(z)=(F_1(z_1),\dots,F_d(z_d))$, so \eqref{eq:lc-kl} splits into one-dimensional integrals against the standard normal,
\begin{equation}
    \mathrm{KL}=\sum_{i=1}^d\mathrm{KL}_i,\qquad
    \mathrm{KL}_i=\mathbb E_{\zeta\sim\mathcal N(0,1)}\Big[\log\varphi(\zeta)-\log F_i'(\zeta)-\log p_i\big(F_i(\zeta)\big)\Big],
    \label{eq:lc-kl-coord}
\end{equation}
with $\mathbb E\log\varphi=-\tfrac12(1+\log2\pi)$ known exactly. We use $512$-point Gauss--Hermite quadrature per coordinate. All 320 two-dimensional rule settings are re-evaluated with $1024$ nodes; the monotonicity flags agree, and every finite KL changes by less than $\max(0.01\,\mathrm{KL}_{1024},10^{-14})$. The eight marked elliptic/trig settings, two Lipschitz-guided settings, and two grid minima in the $d=32$ comparison are also checked with $1024$ nodes. The grid search itself uses 512 nodes. No reported KL uses Monte Carlo sampling.

\paragraph{Validity.}
We use \eqref{eq:lc-kl} only when $F_i'(\zeta)>0$ at every quadrature node and coordinate. A failed check invalidates this evaluation, so we omit its KL and mark a $\times$; it does not mean that KL divergence itself is undefined. Of the 320 displayed two-dimensional settings, eight fail this check, all with Euler or Heun at $r=4$, including some trigonometric settings with $\kappa=1$. No displayed DDIM or DPM-Solver++(2M) setting fails. Checking $40{,}001$ equally spaced initial values on $[-10,10]$ gives the same flags for all 320 settings. Positivity at finitely many nodes is a numerical check, not a proof of global invertibility.

\paragraph{Sanity check against the Gaussian theory.}
On a Gaussian coordinate with precision $\lambda$ every solver's map is linear, $F_i(\zeta)=q\,\zeta$, and \eqref{eq:lc-kl-coord} evaluates to $\tfrac12\big(\lambda q^2-1-\log\lambda q^2\big)$, which is the per-mode identity \eqref{eq:zoloflow-kl-relative-errors} with $1+e=\sqrt\lambda\,q$. For the elliptic finite-difference rule on Gaussian targets the pipeline reproduces the normalized Zolotarev multiplier $\hat q_{\kappa,r}(\lambda)$ to relative accuracy $10^{-14}$.

\clearpage
\subsection{2D Toy Experiments}
\label{app:toy2d}

\paragraph{Setup.}
The five targets are two-dimensional Gaussian mixtures tracing a fern frond, a radial fractal tree, the word ``ZOLO'', an ECG trace, and a double helix. Their interpolated mixtures have closed-form scores. We use DPM-Solver++(2M) (\cref{app:additional-discretizations}) with $r=8$ exact score evaluations on the grid \eqref{eq:shared-query-grid}, and $N=10^5$ samples per setting. All schedules share the same initial noise for each target.

\begin{figure}[p]
    \centering
    \includegraphics[width=\textwidth,height=.76\textheight,keepaspectratio]{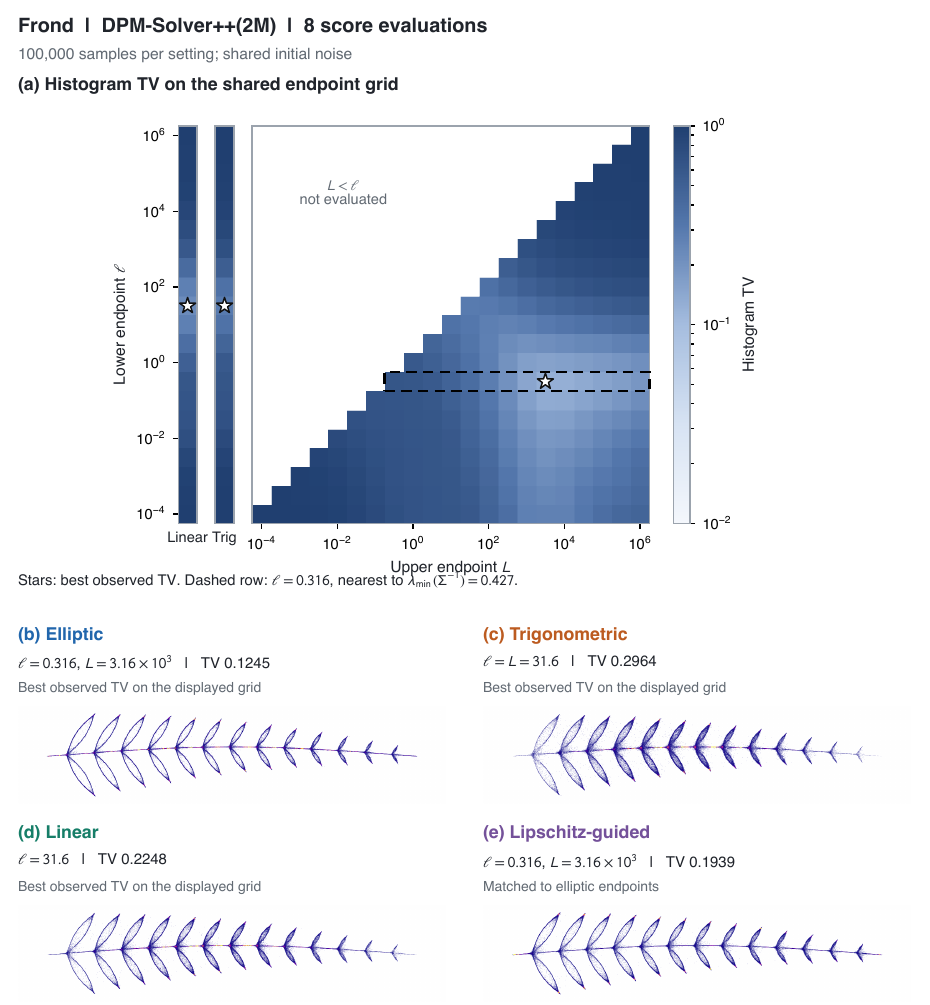}
    \caption{Frond target. DPM-Solver++(2M), eight exact score evaluations and $10^5$ samples per setting. (a) Histogram TV over the shared $(\ell,L)$ triangle, with the Linear and Trig scale sweeps to its left. Stars indicate the best observed TV in each family; the dashed rectangle marks the nearest covariance-precision row. (b)--(d) Samples at the selected elliptic, trigonometric, and linear settings. (e) Lipschitz-guided at the same endpoints as (b), without independent tuning. Each sample panel reports its parameters and histogram TV.}
    \label{fig:toy2d-frond}
\end{figure}

\begin{figure}[p]
    \centering
    \includegraphics[width=\textwidth,height=.76\textheight,keepaspectratio]{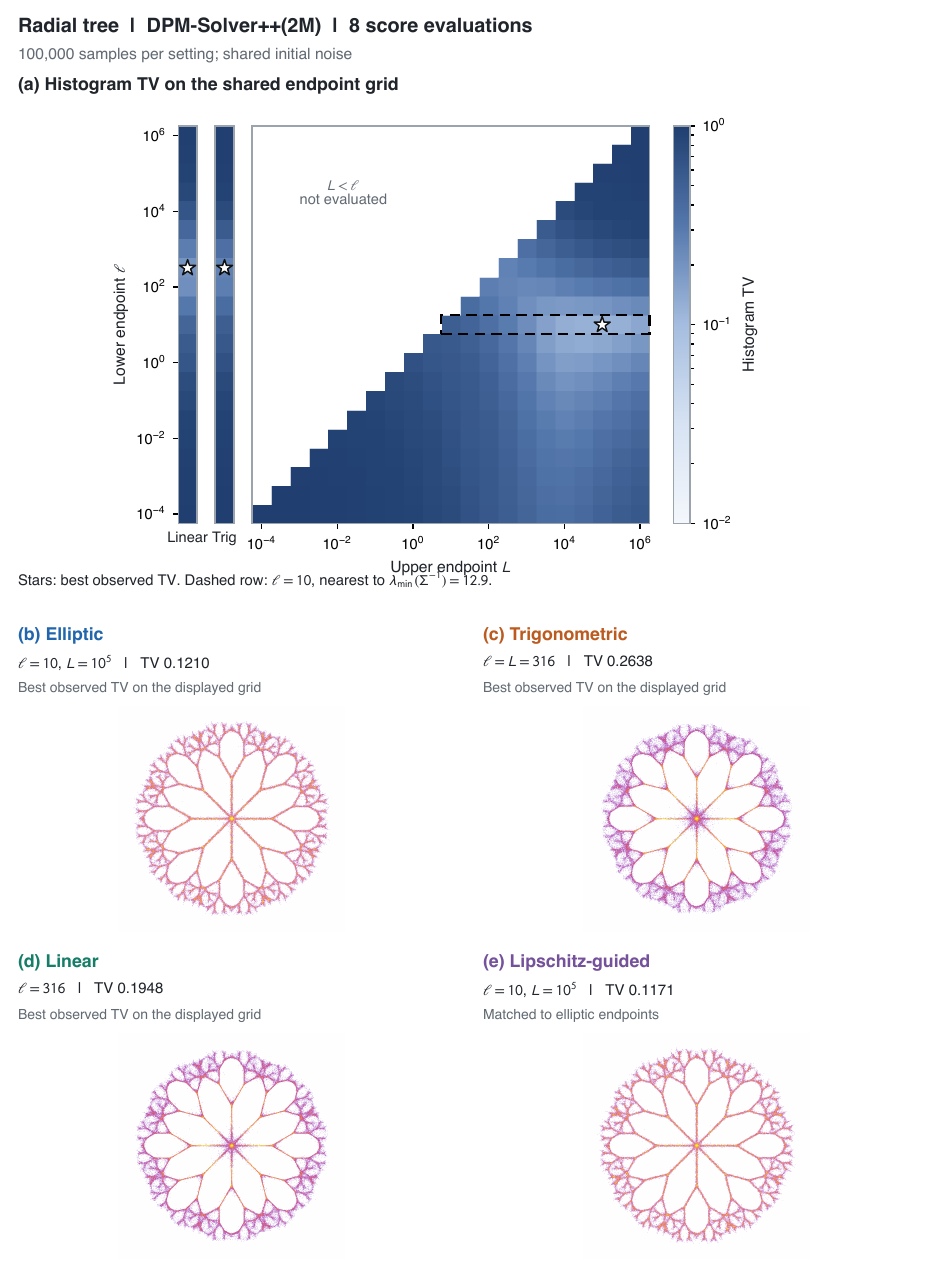}
    \caption{Radial tree target. DPM-Solver++(2M), eight exact score evaluations and $10^5$ samples per setting. (a) Histogram TV over the shared $(\ell,L)$ triangle, with the Linear and Trig scale sweeps to its left. Stars indicate the best observed TV in each family; the dashed rectangle marks the nearest covariance-precision row. (b)--(d) Samples at the selected elliptic, trigonometric, and linear settings. (e) Lipschitz-guided at the same endpoints as (b), without independent tuning. Each sample panel reports its parameters and histogram TV.}
    \label{fig:toy2d-tree}
\end{figure}

\begin{figure}[p]
    \centering
    \includegraphics[width=\textwidth,height=.76\textheight,keepaspectratio]{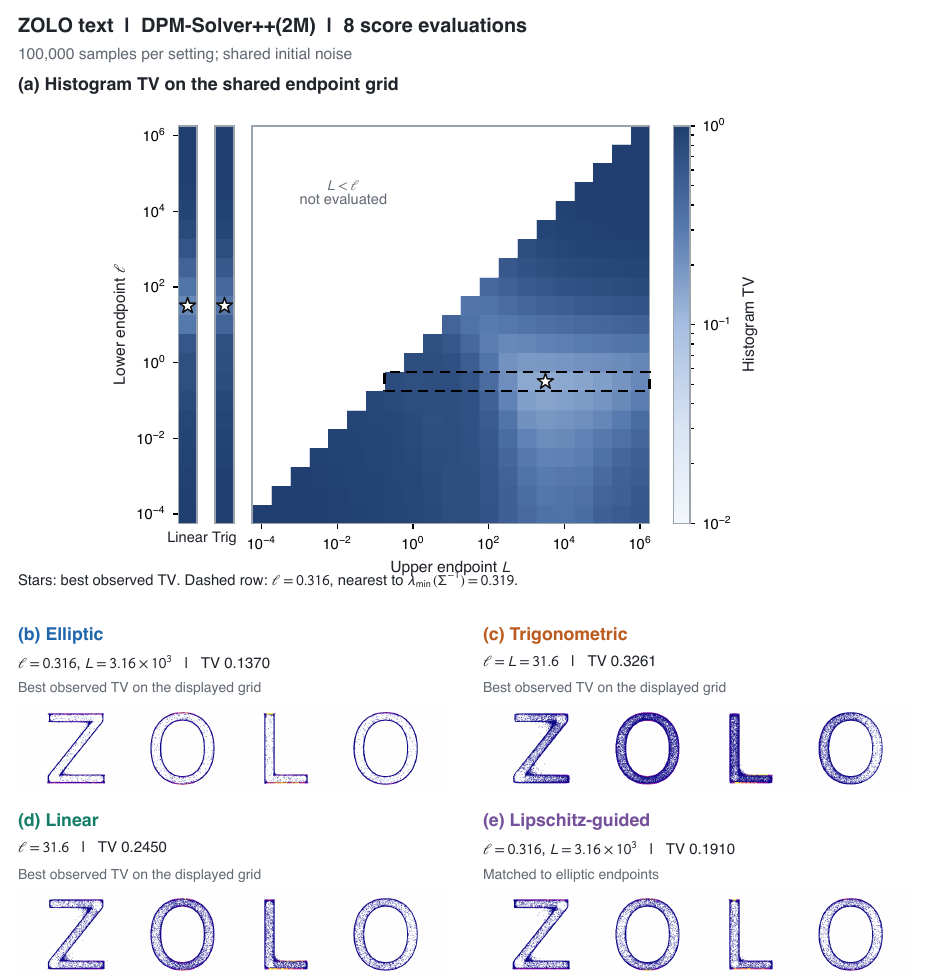}
    \caption{ZOLO text target. DPM-Solver++(2M), eight exact score evaluations and $10^5$ samples per setting. (a) Histogram TV over the shared $(\ell,L)$ triangle, with the Linear and Trig scale sweeps to its left. Stars indicate the best observed TV in each family; the dashed rectangle marks the nearest covariance-precision row. (b)--(d) Samples at the selected elliptic, trigonometric, and linear settings. (e) Lipschitz-guided at the same endpoints as (b), without independent tuning. Each sample panel reports its parameters and histogram TV.}
    \label{fig:toy2d-text}
\end{figure}

\begin{figure}[p]
    \centering
    \includegraphics[width=\textwidth,height=.76\textheight,keepaspectratio]{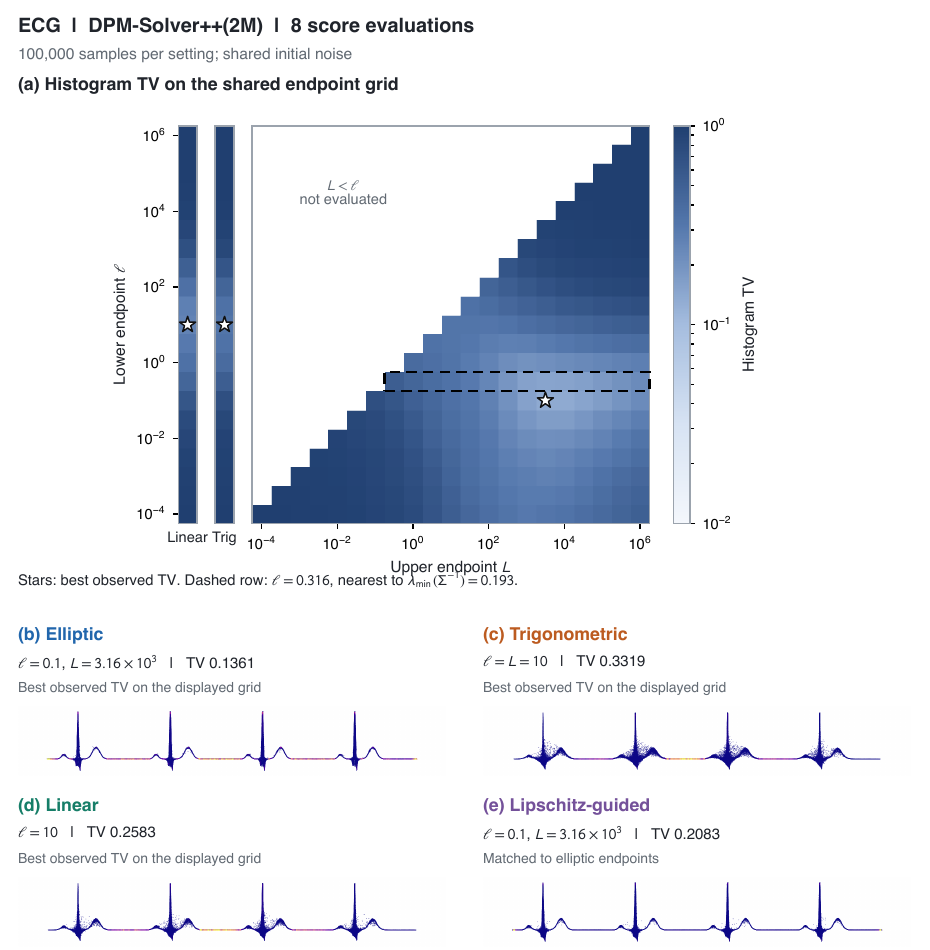}
    \caption{ECG target. DPM-Solver++(2M), eight exact score evaluations and $10^5$ samples per setting. (a) Histogram TV over the shared $(\ell,L)$ triangle, with the Linear and Trig scale sweeps to its left. Stars indicate the best observed TV in each family; the dashed rectangle marks the nearest covariance-precision row. (b)--(d) Samples at the selected elliptic, trigonometric, and linear settings. (e) Lipschitz-guided at the same endpoints as (b), without independent tuning. Each sample panel reports its parameters and histogram TV.}
    \label{fig:toy2d-ecg}
\end{figure}

\begin{figure}[p]
    \centering
    \includegraphics[width=\textwidth,height=.76\textheight,keepaspectratio]{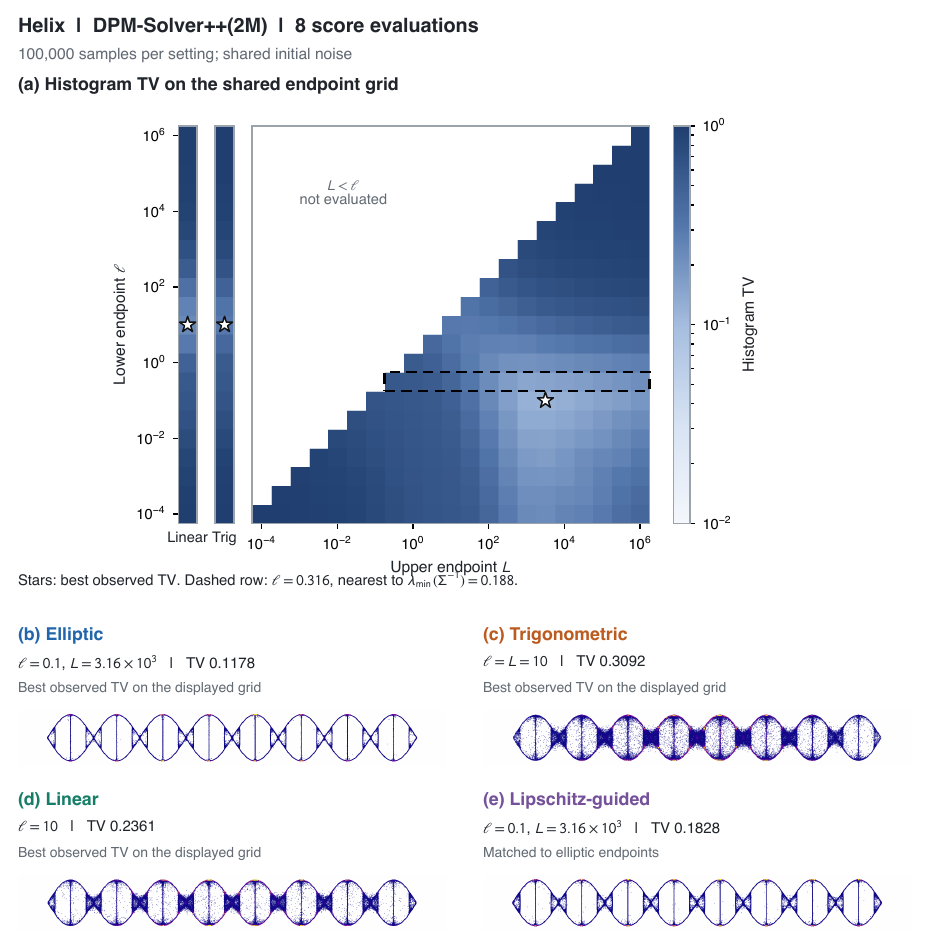}
    \caption{Helix target. DPM-Solver++(2M), eight exact score evaluations and $10^5$ samples per setting. (a) Histogram TV over the shared $(\ell,L)$ triangle, with the Linear and Trig scale sweeps to its left. Stars indicate the best observed TV in each family; the dashed rectangle marks the nearest covariance-precision row. (b)--(d) Samples at the selected elliptic, trigonometric, and linear settings. (e) Lipschitz-guided at the same endpoints as (b), without independent tuning. Each sample panel reports its parameters and histogram TV.}
    \label{fig:toy2d-helix}
\end{figure}

\paragraph{Shared endpoint sweep.}
\Cref{fig:toy2d-frond,fig:toy2d-tree,fig:toy2d-text,fig:toy2d-ecg,fig:toy2d-helix} use the same 21 values $10^{-4+j/2}$, $j=0,\ldots,20$, for both $\ell$ and $L$. We evaluate the 231 pairs with $L\geq\ell$ and select the elliptic pair with the lowest observed histogram TV. The trigonometric schedule is the diagonal $L=\ell$ ($\kappa=1$), with its scale selected independently over the same 21 values. The linear schedule is also selected independently over these 21 scales: its variance-preserving coefficients are $(1-t,t)/\sqrt{(1-t)^2+t^2}$, applied to the target rescaled by $\ell$ and followed by division by $\sqrt{\ell}$ (\cref{app:target-rescaling}). Linear has no upper-endpoint parameter $L$. Thus elliptic is selected from 231 candidates, compared with 21 each for linear and trig. These are minima over the displayed grid, not continuous optima.

\paragraph{Lipschitz-guided at the selected elliptic pair.}
After selecting the elliptic endpoints separately for each target, we run the Lipschitz-guided schedule of \citet[Proposition~3.9]{chen2025lipschitz}, given by \eqref{eq:drift-lipschitz-schedule}, at exactly that pair. It uses the same target rescaling, initial noise, solver, and eight-query budget. Its endpoints are matched to elliptic, not independently optimized. In these runs, elliptic has lower histogram TV than the tuned linear and trig schedules on all five targets. Matched Lipschitz-guided has slightly lower observed TV on Tree ($0.1171$ versus $0.1210$ for elliptic); elliptic has lower TV on the other four targets. The tree's isotropic covariance does not force the tuned endpoints to coincide: its selected elliptic pair is $(\ell,L)=(10,10^5)$.

\paragraph{Metric and figure conventions.}
Histogram TV is half the $\ell_1$ distance between the sample and target histograms on the same fixed $100\times100$ grid, including an additional outside bin; the target histogram uses $10^6$ exact draws. Tuning and the reported TV use the same samples. Each figure shows the endpoint TV heatmap with vertical Linear and Trig strips, followed by samples at the selected elliptic, trig, and linear settings and the matched Lipschitz-guided setting. All five heatmaps use the same axes and logarithmic color scale. Stars mark the selected settings. The black dashed rectangle encloses the row nearest on the log grid to $\ell=\lambda_{\min}(\Sigma^{-1})=1/\lambda_{\max}(\Sigma)$, where $\Sigma$ is the exact target covariance; it marks a reference value of $\ell$ across all tested $L$ in that row and does not constrain the tuning. Sample colors encode local density, normalized separately in each panel, with fixed viewing bounds within each target.

\clearpage

\clearpage

\subsection{Image Experiments}
\label{app:imgage-generation-experiments}

\paragraph{Oracle calls.}
One oracle call is one evaluation of the transferred velocity field $b_t(x)$. With classifier-free guidance, this is a single forward pass of the network (SiT-XL/2 or SD~3.5M) on a batch that stacks the conditional and unconditional inputs, and we count it as one call. 

\paragraph{Noise and prompts.}
For each model, hyperparameter sweeps and final FID-50K evaluation draw their initial Gaussian noise from different random seeds. For SD 3.5M, they also use prompts from disjoint image sets. Within each set, all schedules, solvers, and oracle budgets use identical noise and prompts. The class labels of SiT-XL/2 are drawn uniformly at random from the $1000$ ImageNet classes \citep{5206848,russakovsky2015imagenet}. Since the original training data of SD~3.5M are not publicly available, we use captions from MS-COCO 2014, with one caption per image \citep{lin2014microsoft,chen2015microsoft}. 
Some of the prompts in \Cref{fig:front-page-schedules,fig:sd3_image_comparison_nfe4,fig:sd3_image_comparison_nfe8} were written by the authors.

\paragraph{Encoding and decoding.}
Both models were trained in the latent space of a variational autoencoder (VAE). SiT-XL/2 uses \texttt{sd-vae-ft-ema}~\citep{stable-diffusion}, and SD~3.5M uses its own VAE, which follows the autoencoder design of Stable Diffusion~3~\citep{esser2024scaling}.\footnote{
SD~3.5M VAE configuration and weights: \url{https://huggingface.co/stabilityai/stable-diffusion-3.5-medium/tree/main/vae}.} To generate images, we sampled standard Gaussian noise in the latent space, solved the ODE using a given schedule and solver, and then we decoded the result with the model's VAE.

\paragraph{Metrics.}
We measure sample quality with the Fr\'echet Inception distance (FID) \citep{heusel2017gans}. We write FID-$N$K for an FID score that is computed using $N \times 10^3$ generated images. The hyperparameter sweeps calculate a cheaper FID-$N$K score because they serve only to find the relative performances of different configurations. Once the best hyperparameters are identified, we calculated FID over more samples to compare different schedules with less noise. For SiT-XL/2, the hyperparameter sweeps use 10,000 samples, and the FID-50K evaluation uses 50,000 samples. For SD~3.5M, the hyperparameter sweeps use captions of 5,000 MS-COCO val2014 images as prompts and measure FID against all 40,504 images in the split. The FID-50K evaluations use the captions from 50,000 other images (30,000 from the validation split and 20,000 from the training split) and measure FID against these images. A list of which FID scores we calculated can be found in \Cref{tab:fid-protocols}.

\begin{table}[h]
      \centering
      \small
      \caption{\textbf{FID protocols.} ADM is the TensorFlow evaluator of \citet{dhariwal2021diffusion}, and \texttt{pytorch-fid} is the implementation of \citet{Seitzer2020FID}. COCO is the MS-COCO 2014 dataset. ImageNet reference statistics are taken from the ADM evaluation suite.}
      \label{tab:fid-protocols}
      \begin{tabular}{@{}llcccl@{}}
      \toprule
      Model & Use & Samples & Resolution & Implementation & Reference statistics \\
      \midrule
      \multirow{2}{*}{SiT-XL/2} & \Cref{fig:sit-heatmap-nfe4,fig:sit-heatmap,fig:sit-2d-eq-budget-heatmap} & $10{,}000$ & $256\times256$ & \texttt{pytorch-fid} & ImageNet \\
       & \Cref{tab:fid50k,tab:sit-fid50k-equal-budget} & $50{,}000$ & $256\times256$ & ADM (TensorFlow) & ImageNet \\
      \midrule
      \multirow{2}{*}{SD~3.5M} & \Cref{fig:sd3-heatmap,fig:sd3-2d-eq-budget-heatmap} & $5{,}000$ & $512\times512$ & \texttt{pytorch-fid} & COCO val \\
       & \Cref{tab:fid50k,tab:sd3-fid50k-equal-budget} & $50{,}000$ & $512\times512$ & \texttt{pytorch-fid} & COCO train + val \\
      \bottomrule
      \end{tabular}
  \end{table}

\begin{figure}
    \centering
    \includegraphics[width=\linewidth,height=.73\textheight,keepaspectratio]{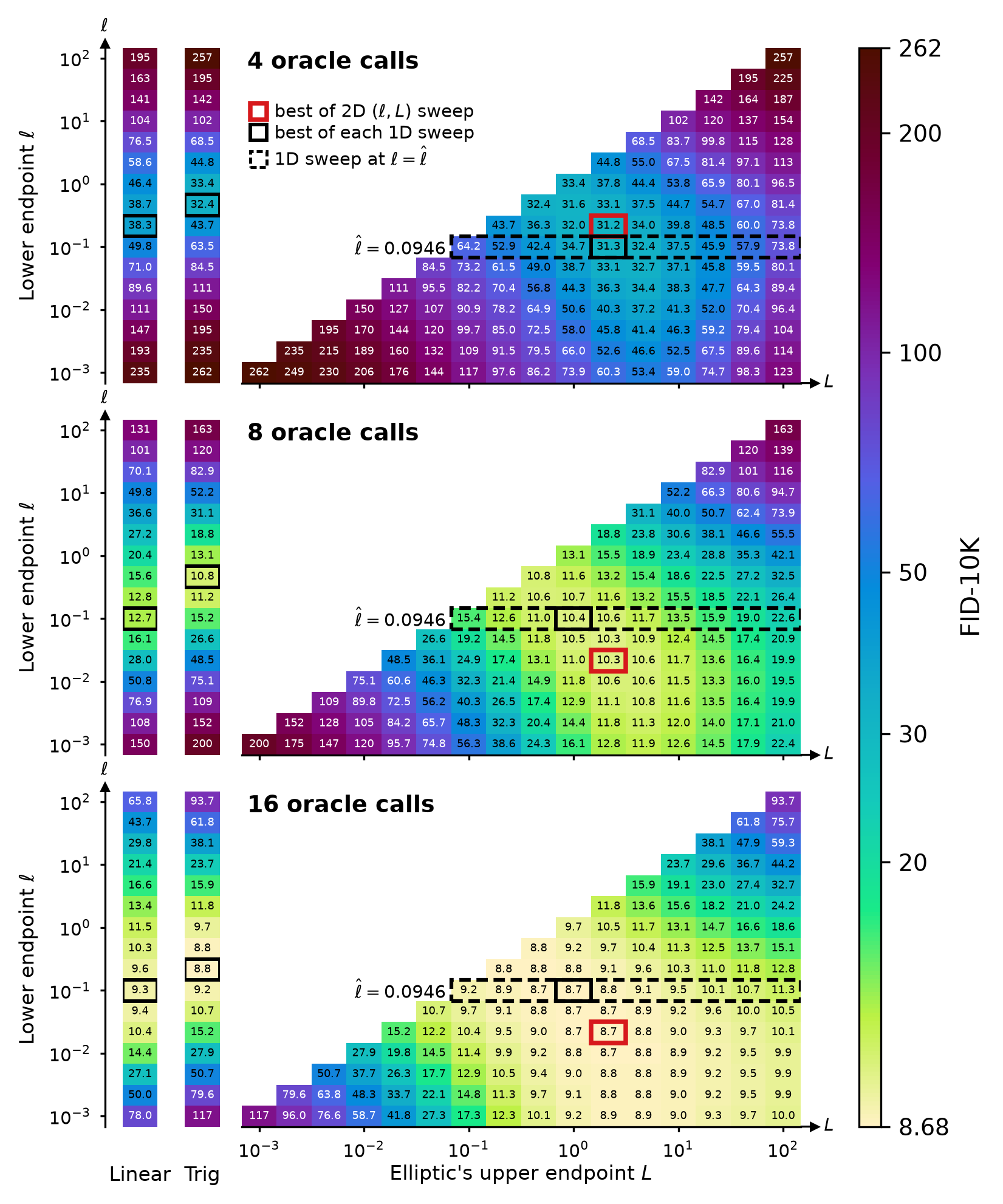}
    \caption{\textbf{FID-10K of hyperparameter sweeps of schedules with SiT-XL/2.} Panels show $4$, $8$, and $16$ oracle calls of SiT-XL/2, from top to bottom. Each cell is annotated with its FID-10K (lower is better). Colors use a single logarithmic scale shared across panels. \emph{Right:} the full 2D sweep over the endpoints $(\ell, L)$ of the target-rescaled elliptic schedule, with the best configuration boxed in red. \emph{Left:} 1D sweeps over $\ell$ for the linear and trigonometric schedules. For an equal-budget comparison, we also restrict the elliptic schedule to a 1D sweep over $L$ with $\ell$ fixed to our heuristic $\hat\ell = 0.0946$ (dashed box). The best configuration of each 1D sweep is boxed in black.}
    \label{fig:sit-heatmap}
\end{figure}

\begin{figure}
    \centering
    \includegraphics[width=\linewidth,height=.73\textheight,keepaspectratio]{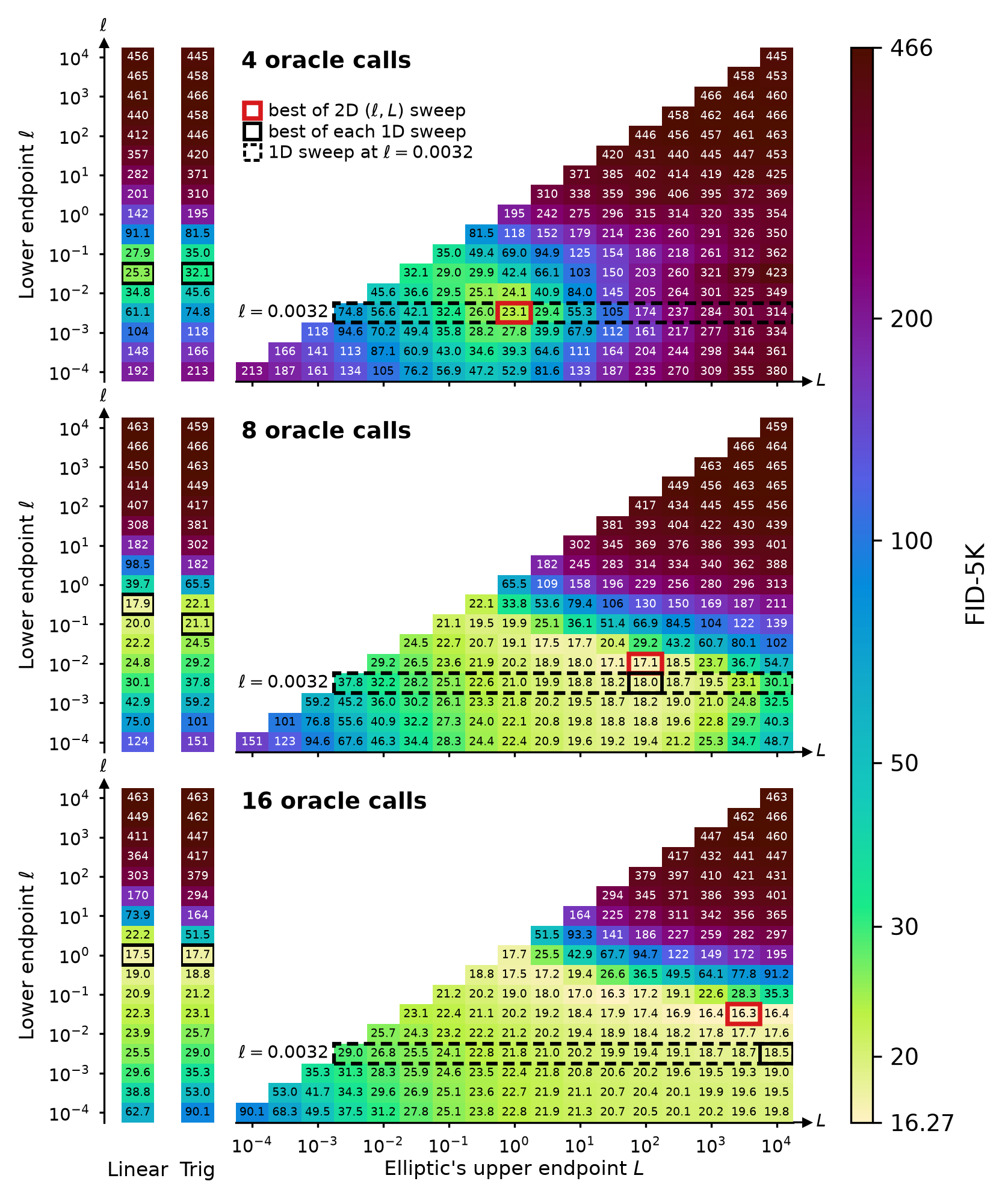}
    \caption{\textbf{FID-5K of hyperparameter sweeps of schedules with SD 3.5M.} Panels show $4$, $8$, and $16$ oracle calls of SD 3.5M, from top to bottom. Each cell is annotated with its FID-5K (lower is better). Colors use a single logarithmic scale shared across panels. \emph{Right:} the full 2D sweep over the endpoints $(\ell, L)$ of the target-rescaled elliptic schedule, with the best configuration boxed in red. \emph{Left:} 1D sweeps over $\ell$ for the linear and trigonometric schedules. 
    For an equal-budget comparison at $8$ and $16$ oracle calls, we also restrict the elliptic schedule to a 1D sweep over $L$ (dashed box), fixing $\ell=0.0032$ to its value in the best configuration from the full 2D sweep at $4$ oracle calls.
    The best configuration of each 1D sweep is boxed in black.}
    \label{fig:sd3-heatmap}
\end{figure}

\begin{figure}
    \centering
    \includegraphics[width=\linewidth,height=.73\textheight,keepaspectratio]{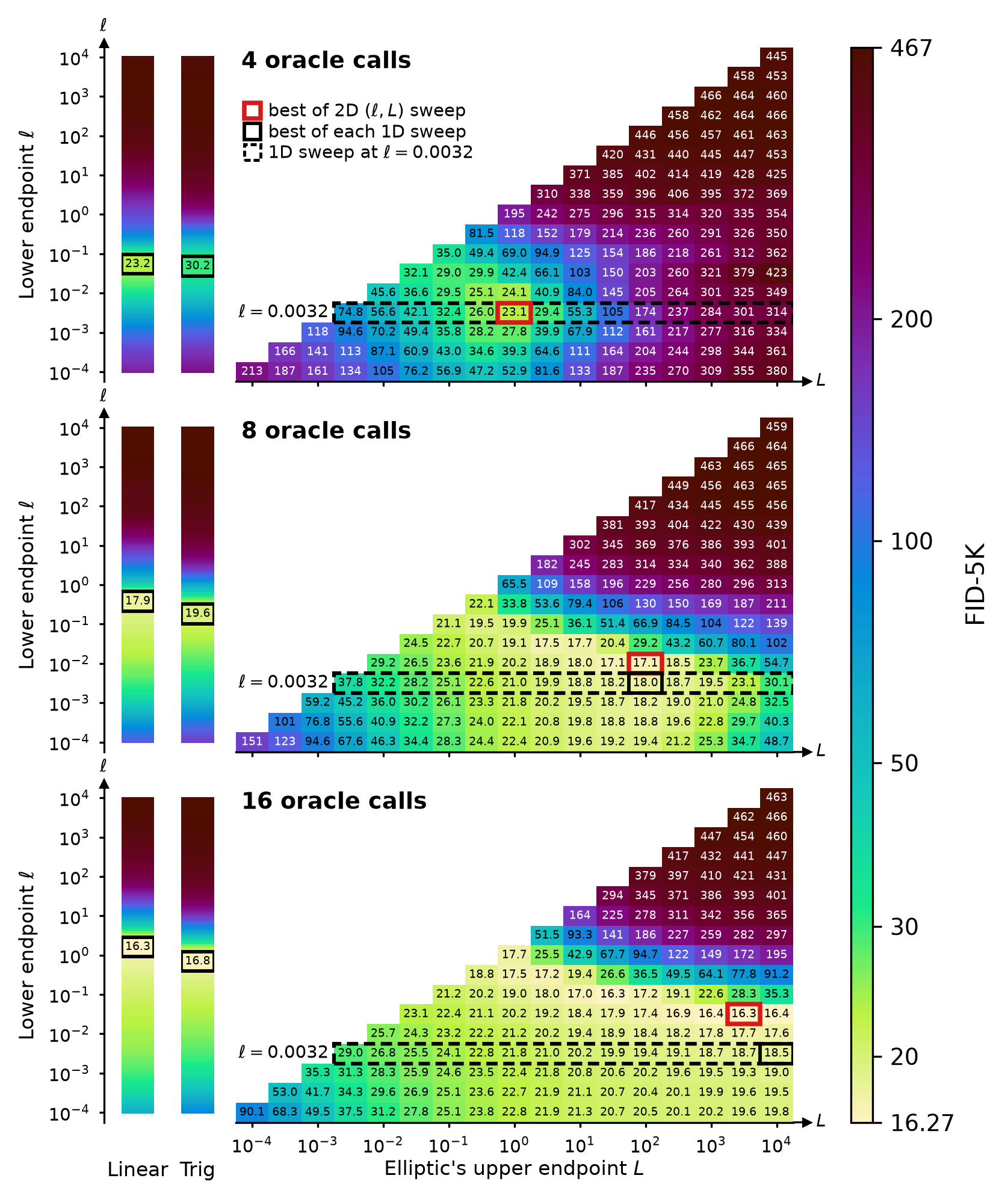}
    \caption{\textbf{FID-5K of hyperparameter sweeps of schedules with SD 3.5M, with equal tuning budgets.} Panels show $4$, $8$, and $16$ oracle calls of SD 3.5M, from top to bottom. Each cell is annotated with its FID-5K (lower is better). Colors use a single logarithmic scale shared across panels. \emph{Right:} the full 2D sweep over the endpoints $(\ell, L)$ of the target-rescaled elliptic schedule, with the best configuration boxed in red. \emph{Left:} 1D sweeps over $\ell$ for the linear and trigonometric schedules. Linear and trigonometric schedules are tuned over 153 values of $\ell$ in $[10^{-4}, 10^4]$ (19 per decade), the same number of configurations as the 2D elliptic sweep over the half-decade $(\ell, L)$ grid with $\ell \leq L$ on $[10^{-4}, 10^4]$. We also restrict the elliptic schedule to a 1D sweep over $L$ (dashed box), fixing $\ell=0.0032$ to its value in the best configuration from the full 2D sweep at $4$ oracle calls.
    The best configuration of each 1D sweep is boxed in black.}
    \label{fig:sd3-2d-eq-budget-heatmap}
\end{figure}

\begin{figure}
    \centering
    \includegraphics[width=\linewidth,height=.73\textheight,keepaspectratio]{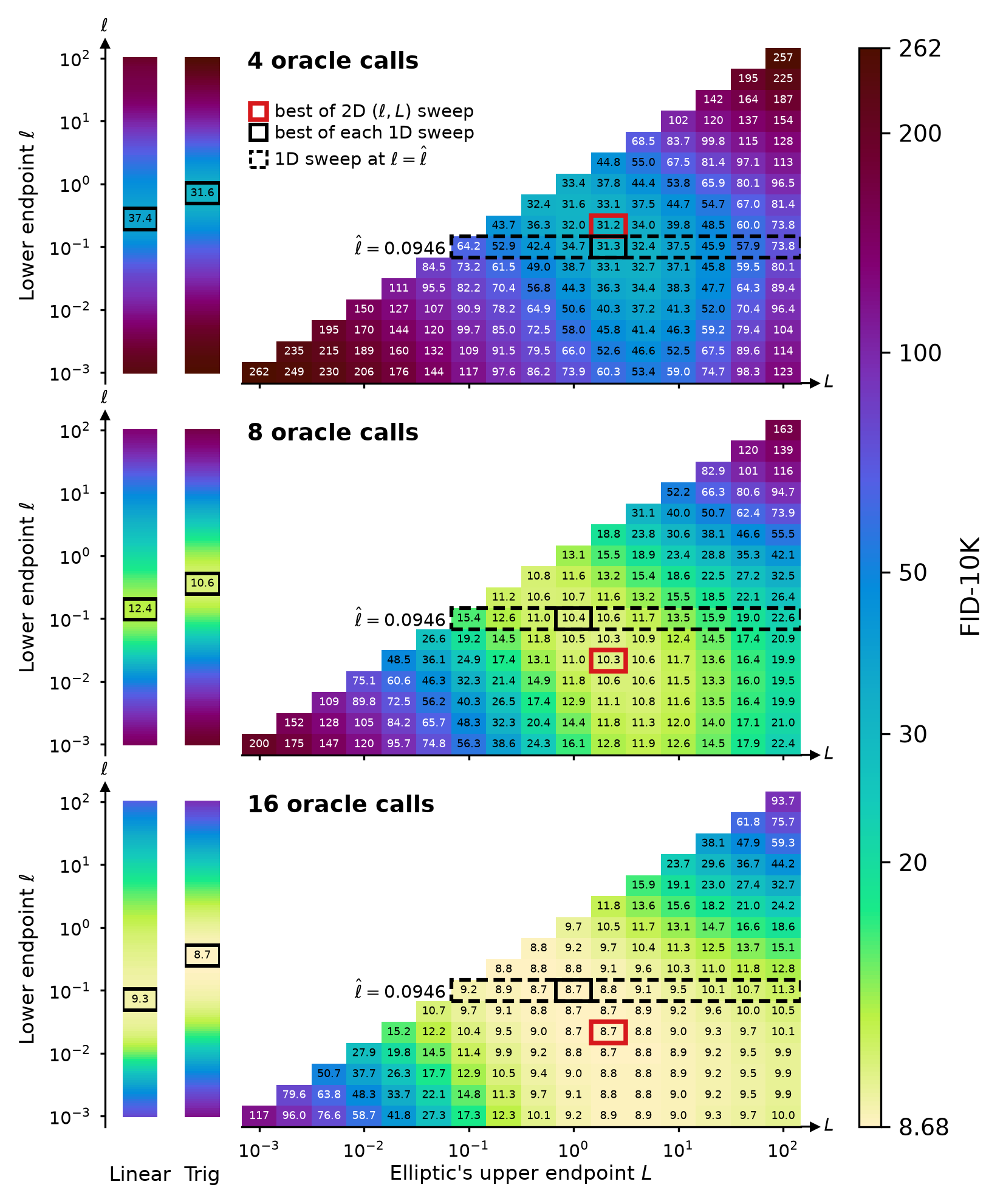}
    \caption{\textbf{FID-10K of hyperparameter sweeps of schedules with SiT-XL/2, with equal tuning budgets.} Panels show $4$, $8$, and $16$ oracle calls of SiT-XL/2, from top to bottom. Each cell is annotated with its FID-10K (lower is better). Colors use a single logarithmic scale shared across panels. \emph{Right:} the full 2D sweep over the endpoints $(\ell, L)$ of the target-rescaled elliptic schedule, with the best configuration boxed in red. \emph{Left:} 1D sweeps over $\ell$ for the linear and trigonometric schedules. Linear and trigonometric schedules are tuned over 136 values of $\ell$ in $[10^{-3}, 10^2]$ (27 per decade), the same number of configurations as the 2D elliptic sweep over the third-decade $(\ell, L)$ grid with $\ell \leq L$ on $[10^{-3}, 10^2]$. We also restrict the elliptic schedule to a 1D sweep over $L$ with $\ell$ fixed to our heuristic $\hat\ell = 0.0946$ (dashed box). The best configuration of each 1D sweep is boxed in black.}
    \label{fig:sit-2d-eq-budget-heatmap}
\end{figure}

\begin{figure}
    \centering
    \includegraphics[width=1\linewidth]{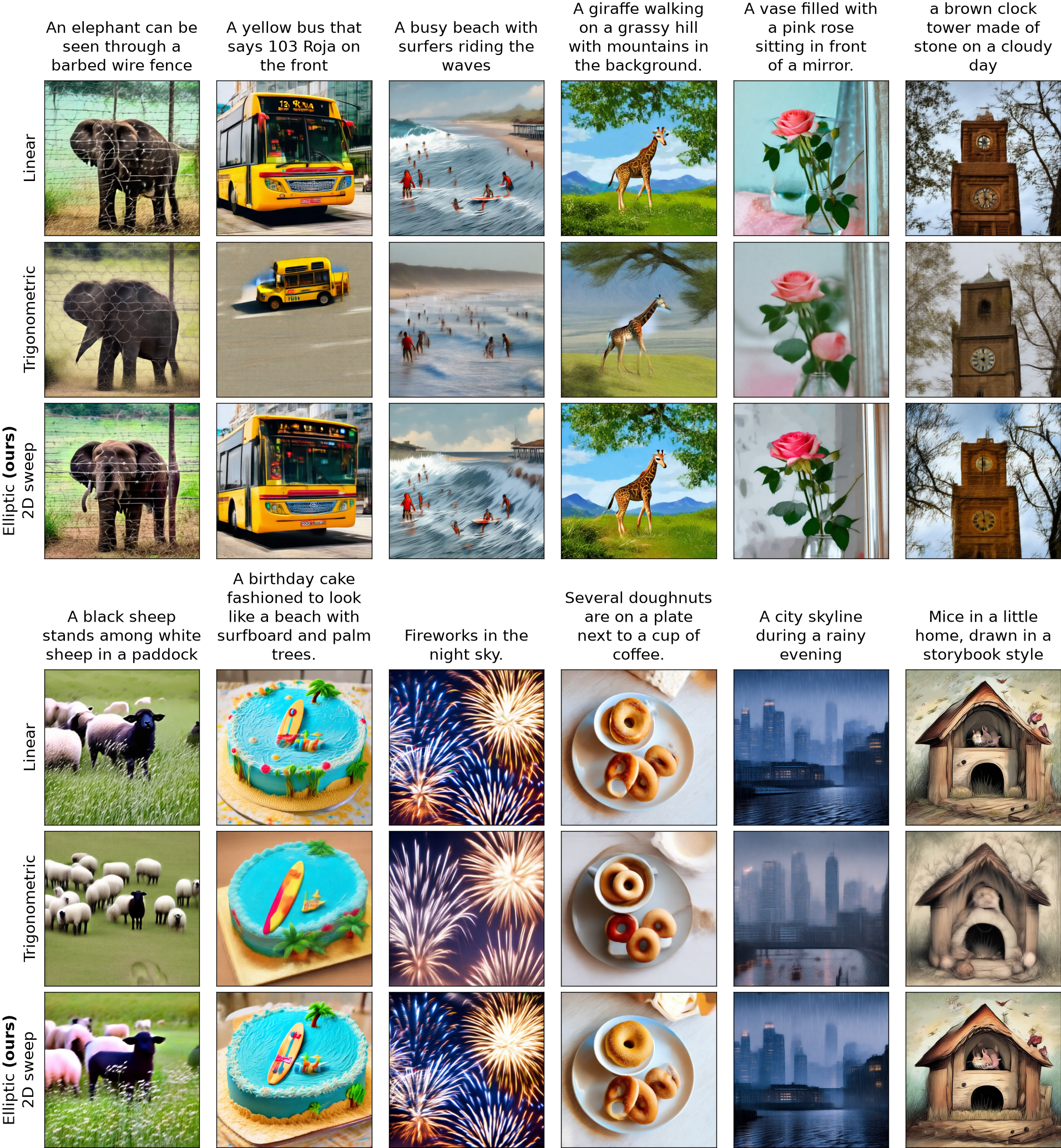}
    \caption{\textbf{Side-by-side comparison of schedules with DPM-Solver++(2M) and 4 oracle calls of SD~3.5M.} The hyperparameters of the schedules are those of the best configurations from our sweep (\cref{fig:sd3-2d-eq-budget-heatmap,tab:sd3-fid50k-equal-budget}). Here, the target-rescaled linear, trigonometric, and elliptic (2D sweep) schedules have the same hyperparameter-tuning budgets.}
    \label{fig:sd3_image_comparison_nfe4}
\end{figure}

\begin{figure}
    \centering
    \includegraphics[width=\linewidth,height=.73\textheight,keepaspectratio]{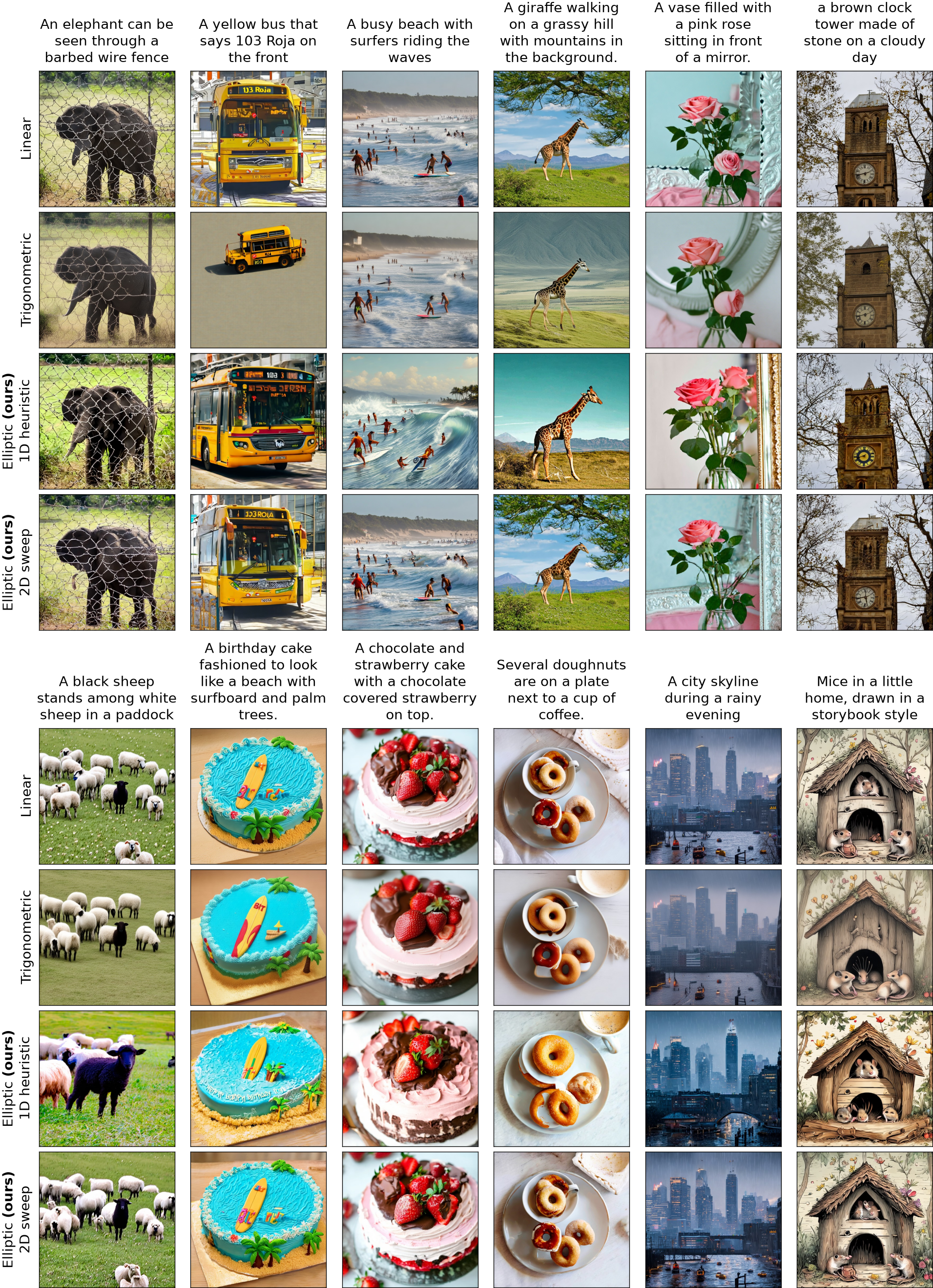}
    \caption{\textbf{Side-by-side comparison of schedules with DPM-Solver++(2M) and 8 oracle calls of SD~3.5M.} The hyperparameters of the schedules are those of the best configurations from our sweep (\cref{fig:sd3-2d-eq-budget-heatmap,tab:sd3-fid50k-equal-budget}). Here, the target-rescaled linear, trigonometric, and elliptic (2D sweep) schedules are swept over 153 hyperparameter configurations, while the 1D heuristic elliptic schedule is swept over only 14 configurations.}
    \label{fig:sd3_image_comparison_nfe8}
\end{figure}

\paragraph{Compute.}
On a single H200 GPU, generating samples and computing FID for one configuration at four oracle calls takes approximately seven minutes for SiT-XL/2 (FID-10K) and ten minutes for SD~3.5M (FID-5K). The full 2D sweep over $(\ell,L)$ at four oracle calls took 1.9 hours for SiT-XL/2 (\cref{fig:sit-heatmap,fig:sit-2d-eq-budget-heatmap}) and 3.1 hours for SD~3.5M (\cref{fig:sd3-heatmap,fig:sd3-2d-eq-budget-heatmap}), each using eight H200 GPUs. This cost can be reduced further by using fewer samples to estimate FID. When target samples are available, $\ell$ can be set heuristically to the reciprocal of the $99.5$th percentile of the empirical covariance spectrum. Otherwise, $\ell$ can be selected through a low-oracle-budget sweep over $(\ell,L)$ and reused at higher oracle budgets, because, as our experiments suggest, sensitivity to these hyperparameters decreases with the number of oracle calls (\cref{fig:sit-heatmap,fig:sd3-heatmap}). Fixing $\ell$ using either approach leaves only $L$ to be tuned, matching the number of tunable parameters in the target-rescaled linear and trigonometric schedules. \Cref{fig:sit-2d-eq-budget-heatmap,fig:sd3-2d-eq-budget-heatmap,tab:sit-fid50k-equal-budget,tab:sd3-fid50k-equal-budget} compare the linear and trigonometric schedules with the full 2D elliptic sweep under equal tuning budgets. Once the hyperparameters are selected, additional samples can be generated without retuning.

\begin{figure}
    \centering
    \includegraphics[width=1\linewidth]{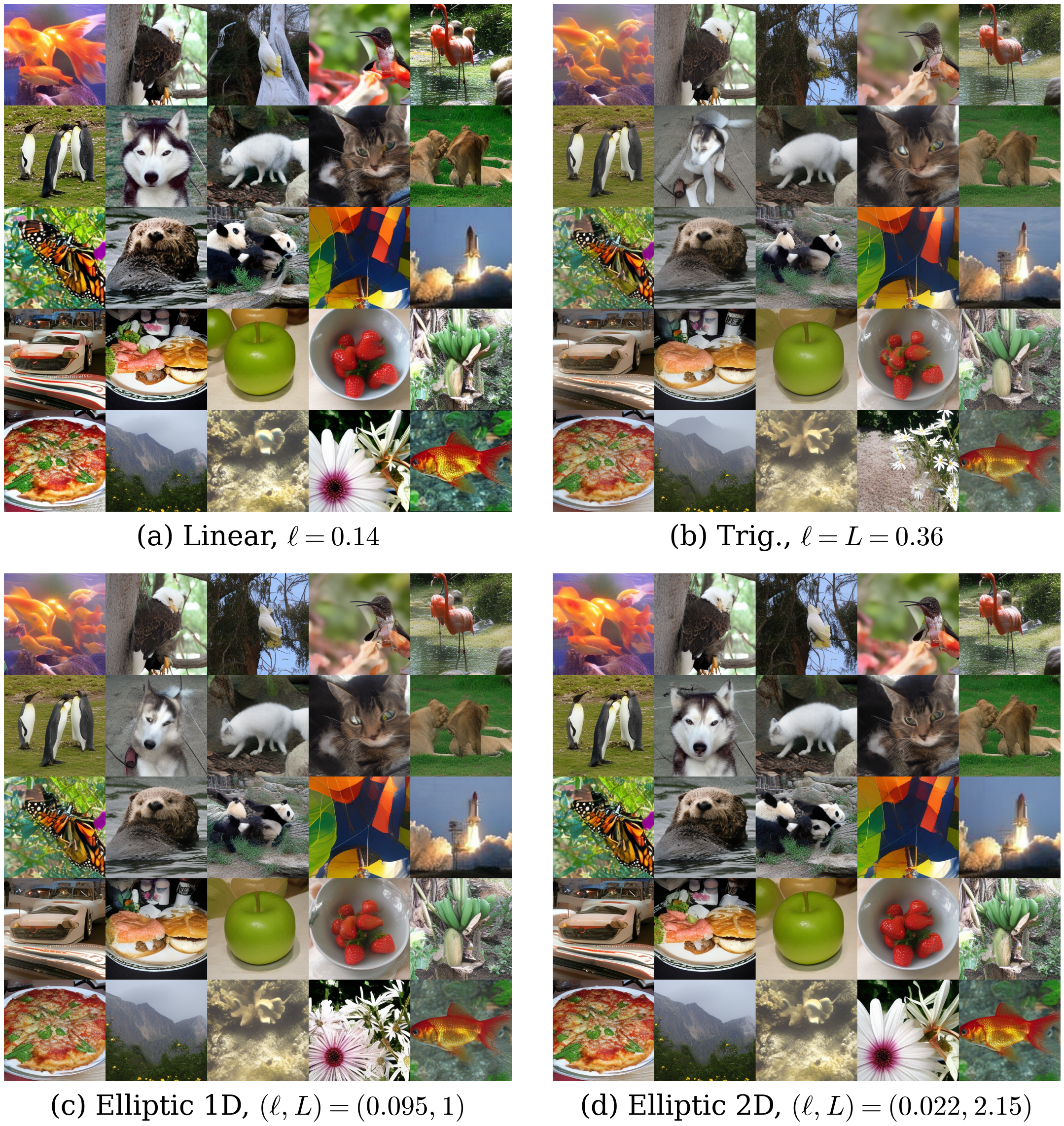}
    \caption{\textbf{Side-by-side comparison of schedules with 8 oracle calls of SiT-XL/2.} Images are sampled with DPM-Solver++(2M) from the same initial noise and class label in every panel. The hyperparameters of the schedules are those of the best configurations from our sweep (\cref{fig:sit-2d-eq-budget-heatmap,tab:sit-fid50k-equal-budget}). Here, the target-rescaled linear, trigonometric, and elliptic (2D sweep) schedules are swept over 136 hyperparameter configurations, while the 1D heuristic elliptic schedule is swept over only 10 configurations.}
    \label{fig:comparison-sit}
\end{figure}

\paragraph{Results.} The resulting FID-50K scores of the different schedules can be found in \Cref{tab:fid50k,tab:sit-fid50k-equal-budget,tab:sd3-fid50k-equal-budget}, and some of the sampled images are shown in \Cref{fig:sd3_image_comparison_nfe4,fig:sd3_image_comparison_nfe8,fig:comparison-sit}.

\begin{figure}
    \centering
    \includegraphics[width=0.75\linewidth,height=.73\textheight,keepaspectratio]{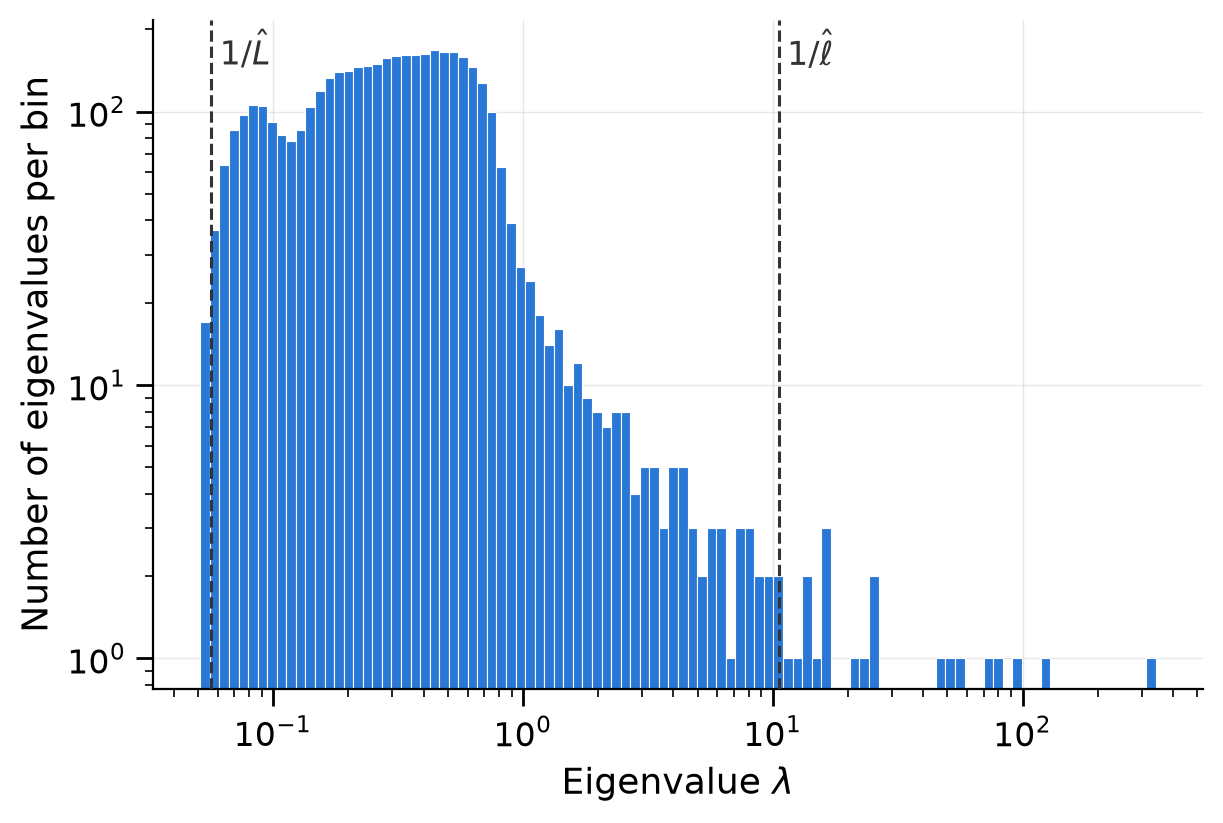}
    \caption{\textbf{Eigenvalue spectrum of the SiT-XL/2 training-data covariance.} Histogram of the eigenvalues $\lambda$ of the empirical covariance $\hat{\Sigma}$ of the latents on which SiT-XL/2 is trained, using 100 bins of equal width in $\log_{10}\lambda$. We estimate $\hat{\Sigma}$ (unbiased) from $N = 50{,}000$ ImageNet $256 \times 256$ training images passed through the exact SiT-XL/2 training pipeline, so each sample is identical in distribution to what the network sees during training. Dashed lines mark $1/\hat{L} = 0.0566$ and $1/\hat{\ell} = 10.57$, which are the 0.5th and 99.5th percentiles of the spectrum, respectively.}
  \label{fig:latent-cov-spectrum}
\end{figure}

\begin{table}[t]
    \centering
    \small
    \setlength{\tabcolsep}{4.5pt}
    \renewcommand{\arraystretch}{0.95}
    \caption{\textbf{SD 3.5M FID-50K with equal tuning budgets.} Linear and trigonometric schedules are tuned over 153 values of $\ell$ in $[10^{-4}, 10^{4}]$ (19 per decade), the same number of configurations as the 2D elliptic sweep over the half-decade $(\ell, L)$ grid with $\ell \le L$ on $[10^{-4}, 10^{4}]$. The 1D heuristic elliptic schedule fixes $\ell = 0.0032$ and sweeps only $L$ (14 configurations). All hyperparameters are selected by FID-5K on a held-out selection set (see \cref{fig:sd3-2d-eq-budget-heatmap}). Best per budget is bold, and second best is boxed.}
    \label{tab:sd3-fid50k-equal-budget}
    \vspace{0.4em}
    \setlength{\aboverulesep}{0pt}
    \setlength{\belowrulesep}{0pt}
    \setlength{\extrarowheight}{1.5pt}
    \setlength{\fboxsep}{1.5pt}
    \begin{tabular}{|c|l|cccc|}
    \toprule
    \rule{0pt}{2.5ex}\emph{\# Oracle Calls} & \emph{Schedule} & $\ell$ & $L$ & \# Configs & FID-50K$\downarrow$ \\
    \midrule
     \multirow{4}{*}{4} & Linear & 0.055 & --- & 153 & \fbox{17.28} \\
      & Trigonometric & 0.048 & 0.048 & 153 & 24.42 \\
      & Elliptic (\textbf{ours}) -- 1D heuristic & --- & --- & --- & --- \\
      & Elliptic (\textbf{ours}) -- 2D sweep & 0.0032 & 1 & 153 & \textbf{16.79} \\
    \midrule
     \multirow{4}{*}{8} & Linear & 0.38 & --- & 153 & 11.92 \\
      & Trigonometric & 0.18 & 0.18 & 153 & 13.74 \\
      & Elliptic (\textbf{ours}) -- 1D heuristic & 0.0032 & 100 & 14 & \fbox{11.56} \\
      & Elliptic (\textbf{ours}) -- 2D sweep & 0.01 & 100 & 153 & \textbf{10.71} \\
    \midrule
     \multirow{4}{*}{16} & Linear & 1.62 & --- & 153 & \fbox{10.16} \\
      & Trigonometric & 0.7 & 0.7 & 153 & 10.88 \\
      & Elliptic (\textbf{ours}) -- 1D heuristic & 0.0032 & 10000 & 14 & 12.05 \\
      & Elliptic (\textbf{ours}) -- 2D sweep & 0.032 & 3160 & 153 & \textbf{9.93} \\
    \bottomrule
    \end{tabular}
    \vspace{-1em}
\end{table}

\begin{table}[t]
    \centering
    \small
    \setlength{\tabcolsep}{4.5pt}
    \renewcommand{\arraystretch}{0.95}
    \caption{\textbf{SiT-XL/2 FID-50K with equal tuning budgets.} Linear and trigonometric schedules are tuned over 136 values of $\ell$ in $[10^{-3}, 10^{2}]$ (27 per decade), the same number of configurations as the 2D elliptic sweep over the third-decade $(\ell, L)$ grid with $\ell \le L$ on $[10^{-3}, 10^{2}]$. The 1D heuristic elliptic schedule fixes $\ell = \hat\ell = 0.0946$ and sweeps only $L$ (10 configurations). All hyperparameters are selected by FID-10K on a held-out selection set (see \cref{fig:sit-2d-eq-budget-heatmap}). Best per budget is bold, and second best is boxed.}
    \label{tab:sit-fid50k-equal-budget}
    \vspace{0.4em}
    \setlength{\aboverulesep}{0pt}
    \setlength{\belowrulesep}{0pt}
    \setlength{\extrarowheight}{1.5pt}
    \setlength{\fboxsep}{1.5pt}
    \begin{tabular}{|c|l|cccc|}
    \toprule
    \rule{0pt}{2.5ex}\emph{\# Oracle Calls} & \emph{Schedule} & $\ell$ & $L$ & \# Configs & FID-50K$\downarrow$ \\
    \midrule
     \multirow{4}{*}{4} & Linear & 0.28 & --- & 136 & 31.49 \\
      & Trigonometric & 0.71 & 0.71 & 136 & 26.23 \\
      & Elliptic (\textbf{ours}) -- 1D heuristic & 0.095 & 2.15 & 10 & \textbf{25.50} \\
      & Elliptic (\textbf{ours}) -- 2D sweep & 0.22 & 2.15 & 136 & \fbox{25.56} \\
    \midrule
     \multirow{4}{*}{8} & Linear & 0.14 & --- & 136 & 5.90 \\
      & Trigonometric & 0.36 & 0.36 & 136 & 4.35 \\
      & Elliptic (\textbf{ours}) -- 1D heuristic & 0.095 & 1 & 10 & \fbox{4.10} \\
      & Elliptic (\textbf{ours}) -- 2D sweep & 0.022 & 2.15 & 136 & \textbf{3.85} \\
    \midrule
     \multirow{4}{*}{16} & Linear & 0.071 & --- & 136 & 2.77 \\
      & Trigonometric & 0.36 & 0.36 & 136 & 2.41 \\
      & Elliptic (\textbf{ours}) -- 1D heuristic & 0.095 & 1 & 10 & \fbox{2.35} \\
      & Elliptic (\textbf{ours}) -- 2D sweep & 0.022 & 2.15 & 136 & \textbf{2.27} \\
    \bottomrule
    \end{tabular}
    \vspace{-1em}
\end{table}

\end{document}